\documentclass[twoside,11pt]{article}
\usepackage[preprint]{jmlr2e}
\IfFileExists{lmodern.sty}{\usepackage{lmodern}}{}
\usepackage{amsmath,amssymb}
\usepackage{booktabs}
\usepackage{enumitem}
\usepackage{graphicx}
\usepackage{url}
\usepackage{placeins}
\usepackage{array}
\usepackage{lastpage}

\jmlrheading{}{2026}{1-\pageref{LastPage}}{11/26}{}{}{Dmitry Ignatov}
\ShortHeadings{Quality--Resource Trade-offs: A Differential Theory}{Ignatov}

\makeatletter
\def\ps@jmlrtps{\let\@mkboth\@gobbletwo
  \def\@oddhead{}%
  \def\@oddfoot{\parbox[t]{\textwidth}{\raggedright\scriptsize
      \copyright\ 2026 Dmitry Ignatov.\hfill}}%
  \def\@evenhead{}\def\@evenfoot{}}
\makeatother

\firstpageno{1}

\newcommand{\eps}{\varepsilon}
\newcommand{\epsinf}{\eps_{\infty}}
\newcommand{\lamstar}{\lambda^{\star}}
\newcommand{\Hs}{\mathcal{H}}
\newcommand{\KL}{\mathrm{KL}}

\newcommand{\R}{\mathbb{R}}
\newcommand{\AM}{\mathrm{AM}}
\newcommand{\GM}{\mathrm{GM}}
\newcommand{\abar}{\bar a}
\newcommand{\Neff}{N_{\mathrm{eff}}}

\newcommand{\one}{\mathbf{1}}

\newcommand{\SM}[1]{the Appendix, \S\,``#1''}
\newcommand{\TR}[1]{the technical report in the archive, \S\,``#1''}
\newcommand{\SMonedim}{\TR{One dimension: self-duality and the realizable minimax}}

\newcommand{\SMchart}{Section~\ref{sm:chartzoo}}

\newtheorem{assumption}[theorem]{Assumption}
\newenvironment{proofof}[1]{\begin{proof}[Proof of #1]}{\end{proof}}

\begin{document}
\title{A Differential Theory of\\
  Quality--Resource Trade-offs for AI Model Selection:\\
  The Multiresource Frontier and Its Minimax Price}
\author{\name Dmitry Ignatov \email dmytro.ignatov@uni-wuerzburg.de \\
        \addr Computer Vision Lab, CAIDAS \& IFI\\ University of W\"urzburg\\
        John-Skilton-Stra\ss e 4a\\
        97074 W\"urzburg, Germany}
\editor{}
\maketitle

\begin{abstract}%
Comparing AI models across resource budgets yields quality–resource frontiers relating error to latency, parameter count, training compute, and energy. Offset power laws summarize such frontiers, but their fitted parameters reveal neither the price-independent geometry nor the manner in which trade-offs extend across several resource channels. This paper addresses both gaps by developing a differential theory in which this structure is made explicit. Each frontier admits a price-independent plane-curve representation relating the effective number of active resource channels, $\Neff=s^{2}/\kappa$ (squared slope over curvature), to the remaining log-error depth, $\Delta S=\ln(\eps/\epsinf)$. This representation separates intrinsic shape from the scale and origin of the resource axis. Complete monotonicity and divisibility yield differential inequalities and frontier bounds. When each of $N$ resources contributes one active channel, the theory gives the parameter-free operating rule $\Neff=1$ in delivered units and derives minimax resource prices $\lamstar_i=\alpha_i/(N+1)$ with an exact worst-case regret of $\ln(N+1)$ nats. The results make it possible to interpret measured frontiers and compare resource costs without hand-chosen exchange rates. The theory is evaluated on $117$ published frontiers and implemented in the open-source package \texttt{nn-rank}, which derives prices from measured quality and cost, ranks candidates, and reports supported choices and minimax regret: \url{https://github.com/ABrain-One/nn-rank}.
\end{abstract}

\section{Introduction: The Question and the Coordinates}
\label{sec:setup}
\begin{figure}[t]\centering
\includegraphics[width=\linewidth]{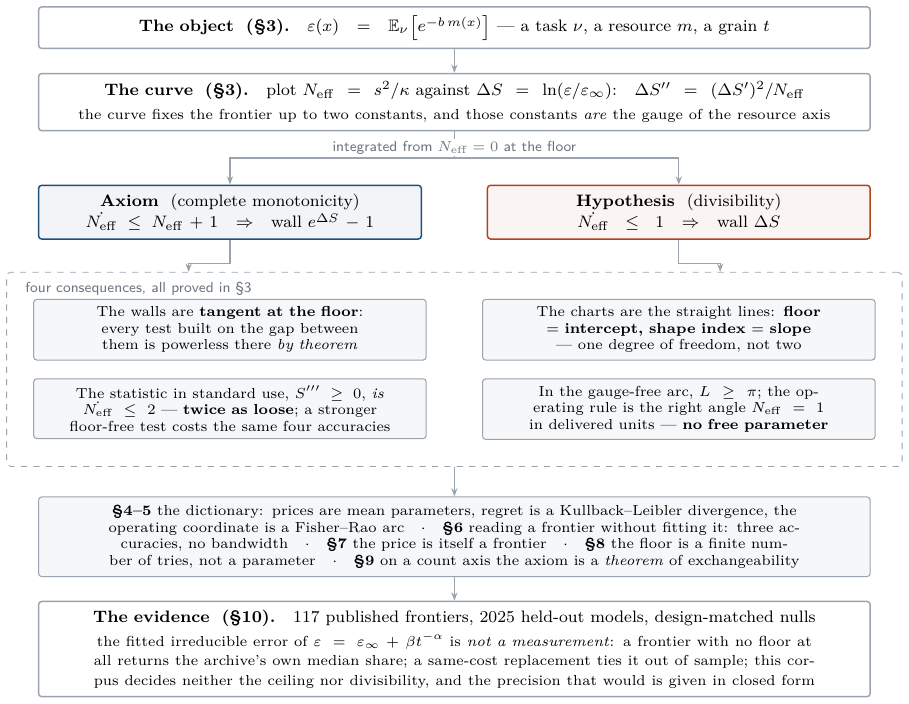}
\caption{The paper in one picture, and where each part is proved. The object
is a Laplace transform of a task's difficulty law evaluated at what the
resource buys; the theory is the plane curve it traces; the two structural
hypotheses are two differential inequalities on that curve, and the two walls
that bound every frontier are those inequalities integrated from the floor.
Everything in the dashed box is a consequence of the figure above it.}
\label{fig:roadmap}
\end{figure}
A candidate system has positive resources $t_1,\dots,t_n$---latency, parameter count, training
compute, and energy---and a quality $Q$ with chance level $q_0$. Throughout,
\begin{equation}
\eps=\frac{1-Q}{1-q_0}\in(0,1],\qquad x_i=\ln t_i,\qquad S=-\ln\eps,
\qquad s=S',\qquad \kappa=-S''.
\label{eq:defs}
\end{equation}
$S$ is log-error in nats; a price component $\lambda_i$ is nats per e-fold of resource $i$.
Selection means maximizing the scalarization
\begin{equation}
\sigma_\lambda(x)=S(x)-\lambda\cdot x,
\qquad
\sigma^*(\lambda)=\sup_x\sigma_\lambda(x).
\label{eq:score}
\end{equation}
The price $\lambda$ encodes the deployment: how much log-error a resource e-fold is worth. It is
normally not known when a checkpoint is published or a platform commits to one operating point for
many tenants. The question of this paper is what can be said in that situation. The answer separates three
objects that are usually conflated: the attainable frontier, the prices it can support, and the
rule used to choose among those prices.
\paragraph{Contributions.} The paper is organized around a single reduction and its consequences.
\begin{enumerate}[label=(C\arabic*),leftmargin=3.0em,itemsep=3pt]
\item \emph{One curve.} Everything a frontier decides before a price is declared is the plane curve
      $\Neff$ against $\Delta S$ (Section~\ref{sec:odefirst}). The frontier is that curve plus the
      two integration constants of an initial-value problem, and those constants are the scale and
      origin of the resource axis---they carry no information about the trade-off.
\item \emph{Two hypotheses, two differential inequalities, two walls.} Complete monotonicity and
      divisibility are $\dot\Neff\le\Neff+1$ and $\dot\Neff\le1$ on that curve. Integrating each
      from the floor bounds every frontier (Corollary~\ref{cor:wallivp}). The four
      previously separate propositions each reduce to a single line.
\item \emph{A dictionary, not a new formalism.} Prices are mean parameters, regret is a
      Kullback--Leibler divergence, and the operating coordinate is a Fisher--Rao arc length
      (Sections~\ref{sec:lawII}--\ref{sec:lawIII}). This establishes a frontier-specific
      dictionary between established geometric and information-theoretic objects.
\item \emph{A parameter-free operating rule, read from three accuracies.} The rule is the right
      angle $\Neff=1$ read in delivered units, equivalently $t\Neff=1$ in per-item units at
      scoring grain $t$ (Section~\ref{sec:lawfree}). It requires no bandwidth,
      kernel, or fitted floor. Theorem~\ref{thm:canonical} gives the corresponding price and its
      exact regret.
\item \emph{Four consequences for the saturating power law}, stated in Section~\ref{sec:related}
      and evaluated in Section~\ref{sec:evidence} on $117$ published frontiers containing $2\,259$
      model measurements (a median of $11$ per frontier, ranging from $8$ to $87$), together with
      $2\,025$ leave-one-out prediction events on those measurements. The central finding is that
      the fitted irreducible term is not a direct measurement of irreducible error.
\item \emph{A measurement protocol the paper applies to itself.} Each of the nine plug-in
      statistics is evaluated against a null simulated using the same measurement design, window,
      and residual scale. Results that cannot be distinguished from their design-matched nulls are
      reported as inconclusive.
\end{enumerate}
\paragraph{The finished object.} A task presents items; an item has a per-attempt failure
probability $g\in(0,1]$ drawn from a law $\nu$, and we write $b=-\ln g$, with mass at $g=1$ corresponding to
items that no attempt ever solves. A resource buys $m(x)$
independent attempts per item. A delivered unit consists of $t$ independently drawn items, and it fails only
if all of them fail. Then
\begin{equation}
\eps(x)=\mathbb{E}_\nu\bigl[e^{-b\,m(x)}\bigr]=P\bigl(m(x)\bigr),
\qquad
P(k)=\int_0^1 g^{k}\,\mathrm{d}\nu(g),
\qquad
\eps_{\text{unit}}=\eps^{\,t},
\label{eq:oneline}
\end{equation}
\emph{the error is the Laplace transform of the task's difficulty law, evaluated at the attempt
count purchased by the resource.} The three objects are the task law $\nu$, the resource map $m$, and the scoring grain $t$.
Exchangeability of attempts forces complete monotonicity on the count axis
(Proposition~\ref{prop:countaxis}). The theory also uses two assumptions of the same form: an
e-fold of resource and an item's difficulty can each be subdivided. The final input, which no
frontier can supply, is the scoring grain (Proposition~\ref{prop:retry}). The three numbers fitted
per axis are features of the same unknown function $m$: its rate, its index, and its total.
Equation~\ref{eq:oneline} carries two counts, and they are not interchangeable. Attempts
condition on one item, so the task's heterogeneity persists across them. Items are redrawn, so the
heterogeneity averages out across items. A Laplace transform is log-convex, so $-\ln P$ is concave through the origin, giving
\begin{equation*}
S(tk)\le t\,S(k)\quad\text{for every }t\ge1,
\end{equation*}
with equality if and only if $\nu$ is a single atom. In words: \emph{spending $t$ attempts on one
item is worth at most what scoring $t$ fresh items is worth.}
The discount $D(t)=S(t)/\{t\,S(1)\}$ measures the shortfall. It lies in $[1/t,1]$ and is
non-increasing. At $D=1$ the task is homogeneous and a retry is worth a fresh item; at $D=1/t$ the
attempts nest perfectly and a retry is worthless. The per-item panel of this archive measures
$D(2)=0.547$ and $0.522$ against the bound $\tfrac12$, so the second attempt yields only four to nine percent of the first attempt's
benefit. A retry budget $t_r$ therefore delivers grain $S(t_r)/S(1)<t_r$,
and the shortfall compounds: grain $2.2$ costs about $1.3\times10^{3}$ retries at $D(2)=0.52$,
against $2.6$ retries at $D(2)=0.90$. The exact statement, the local exchange rate it defines, and
the panel's numbers are in \SM{Proofs: The two counts}.
\paragraph{How to read this paper.} Section~\ref{sec:odefirst} presents the reduction and is the only
section on which the rest of the paper depends; it contains the core theory. Figure~\ref{fig:roadmap}
provides a one-page map of the argument and indicates where each statement is proved.
Sections~\ref{sec:lawII}--\ref{sec:lawIII} are the dictionary, Section~\ref{sec:lawfree} the estimator,
Section~\ref{sec:compose} the two hypotheses and what makes them one statement,
Section~\ref{sec:floor} the floor, Section~\ref{sec:count} the count axis, and
Section~\ref{sec:evidence} the measurement. Section~\ref{sec:scope} states the structural and
empirical scope of the results. Proofs are in the Appendix; Table~\ref{tab:provenance}
maps each numbered statement and reported number to the script that produces it.
\section{Related Work and Contributions}
\label{sec:related}
\paragraph{Neural scaling laws.} The empirical literature fits a frontier and extrapolates it
\citep{hestness2017,rosenfeld2020,kaplan2020,hoffmann2022,bahri2024}. Writing the error against a
resource $t$, the forms in use are a pure power law $\eps=\beta t^{c}$; a power law with an
irreducible term,
\begin{equation}
\boxed{\ \eps=\epsinf+\beta\,t^{-\alpha}\ }
\label{eq:M2}
\end{equation}
which \citet{alabdulmohsin2022} call the most commonly used model and which is the single-resource
case of the Chinchilla law $L=E+A/N^{\alpha}+B/D^{\beta}$ \citep{hoffmann2022}; a reparameterized
variant; and, more recently, forms that add break points or further shape parameters
\citep{alabdulmohsin2022,caballero2023}. Where that literature asks what \emph{generates} an exponent---for example, the dimension of the data
manifold \citep{sharma2022} or a vocabulary law carried into a loss \citep{debowski2026}---this
paper asks what any frontier must satisfy, regardless of its generating mechanism. The two questions
are independent: the results below hold for every mechanism in that list. The same functional form
continues to inform both empirical fitting and theoretical interpretation. For example,
\citet{han2026invariance} estimate its irreducible term from four to six model sizes and study the
transformations that leave that term fixed, while \citet{sun2026roofline} fit it against latency and
memory to guide on-device architecture selection. The results below clarify which properties of \eqref{eq:M2} are structurally determined and
which cannot be identified from frontiers measured at that resolution. In the coordinates of \eqref{eq:defs}, \eqref{eq:M2} is
exactly the two-atom chart of this paper, $\eps=\epsinf+c\,e^{-\alpha x}$. The chance level $\eps_0$ used by \citet{alabdulmohsin2022} for normalization corresponds to
our $q_0$. This correspondence connects the present analysis directly to \eqref{eq:M2} and
yields four results, each evaluated on the archive's $117$ published frontiers.
\begin{enumerate}[label=(\roman*),leftmargin=2.2em,itemsep=4pt]
\item \emph{The irreducible term and the shape index are one degree of freedom, not two.} In the
      plane of Section~\ref{sec:odefirst} they are the intercept and the slope of a single straight
      line (Corollary~\ref{cor:odepencil}). A line in a plane is determined by two parameters, not three. Thus, a fitted floor can only be
      a monotone readout of the fitted shape index. The consequence is
      measurable in a controlled simulation, with no plug-in estimator anywhere in it. Fit
      \eqref{eq:M2} to frontiers whose floor is \emph{exactly zero}, on the archive's own median
      design, and the fitted floor share runs monotonically with the shape index across
      effectively the whole of $(0,1)$: $0.98$ at index $0.05$, $0.01$ at $0.99$
      (Proposition~\ref{prop:nofloor}). The curve passes through $0.96$ at index $0.10$; this is
      also the archive's median fitted share. \emph{A fitted irreducible error is therefore not a
      measurement of the irreducible error.} It is a reading of the shape of the spectrum near the
      origin, taken at a window. This bears directly on work whose conclusions are about the floor
      itself \citep{han2026invariance,sun2026roofline}.
\item \emph{The axiom puts a ceiling on the irreducible term---and that ceiling is not testable on
      published windows, at any precision.} The axiom \eqref{eq:axiom} forces $\epsinf\le\eps/(1+\Neff)$ at every
      point of every frontier (Corollary~\ref{cor:wallivp}), and the bound needs three accuracies,
      no bandwidth and no fit of one's own. On the published windows, this bound is not empirically discriminating; the theory identifies
      the structural reason. Form \eqref{eq:M2} \emph{saturates} that wall identically, so the ceiling
      test is a goodness-of-fit test of \eqref{eq:M2} whose null is \eqref{eq:M2} itself; and by
      Corollary~\ref{cor:wallivp}(iii) the walls are tangent at the floor, where the published
      windows sit. Section~\ref{sec:evidence} reads the test against a design-matched null and
      reports a nominal level and no power---at any precision, because the obstruction is depth
      and not noise.
      The natural plug-in is not a test of it: minimizing $\eps/(1+\Neff)$ over a frontier's
      triples reports violations on $89$ of $117$ frontiers, but the same procedure applied to
      simulated frontiers that satisfy the ceiling \emph{by construction} reports a mean of
      $100.3$ apparent violations per $117$-frontier replicate. Such a count measures the
      estimator rather than the corpus. The ceiling therefore remains a theoretical guarantee, rather than an empirically resolved
      statistic on these windows.
\item \emph{The standard form does not satisfy the divisibility condition.} Section~\ref{sec:compose}
      asks whether the price of error is itself a frontier; under that hypothesis the admissible region is
      $\Neff\le\Delta S$. Form \eqref{eq:M2} has $\Neff=\rho-1>\ln\rho=\Delta S$ at every
      $\rho=\eps/\epsinf>1$. This is an analytic statement about the form and not a measurement, so
      no null is involved. The positive gap places the standard form outside the divisible class
      at second order rather than third. It is strictly positive at every point of the curve and
      closes only as $\rho\to1$ at the floor,
      where Corollary~\ref{cor:wallivp}(iii) shows that a second-order test has no power.
\item \emph{A same-cost replacement matches its out-of-sample performance and carries exact
      structure.} The cure chart \eqref{eq:cure} uses the same three parameters and is the
      extremal of the divisibility wall. On held-out data from the same corpus, the two forms have nearly indistinguishable
      accuracy: the cure chart has a pooled median relative error of $2.57\%$ across the $2\,013$
      prediction events for which it returns a fit, compared with $2.60\%$ for \eqref{eq:M2}
      across all $2\,025$ events. The cure chart satisfies six exact identities using
      those same three parameters. It also provides an operating point one nat above the floor and an
      admissible arc fixed by the floor (Proposition~\ref{prop:cure}).
\end{enumerate}
\noindent \textbf{What the theory delivers.} The contribution is a structural account of the admissible frontier class and of the quantities
that can be read from a frontier without fitting a parametric chart, rather than a claim of uniformly
improved interpolation accuracy. Section~\ref{sec:evidence} therefore evaluates prediction in
three held-out regimes against the floorless power law and two model-free baselines, while
Section~\ref{sec:scope} states the scope of those results. Two practical consequences follow.
\emph{First, current selection rules can produce materially different operating choices.} This
spread can be measured directly from the frontier, without specifying an objective function. Take the eight selection rules that a
practitioner can apply to a discrete frontier without fitting a model or specifying an additional
hyperparameter. On the $96$
single-resource frontiers of the archive, the eight rules select a median of \emph{five distinct
models} per frontier. The extreme selected points differ by a median resource ratio of
$19\times$ (interquartile range $4\times$ to $236\times$); this resource difference
corresponds to a median error difference of $1.72$ bits. On $52$ of the $96$
frontiers the spread exceeds $10\times$; on $29$ frontiers, the difference corresponds to less
than one bit. The median exchange
rate across the selection disagreement is $12$ resource units per bit. One of the three rules currently
in use is not scale-free, and a change of unit---GFLOP to MFLOP, seconds to milliseconds---moves its
pick on $94$ of those $96$ frontiers. A rule whose selection changes under a change of units encodes a price, not
an intrinsic measurement.

A second loss-free measurement identifies which Pareto candidates can be optimal for any
nonnegative price. Selection compares candidates through $\sigma_i(\lambda)=-\ln\eps_i-\lambda x_i$, which is
\emph{linear} in the price, so the candidates optimal at some nonnegative price are exactly the
vertices of the upper envelope of $K$ lines---one sort and one stack pass, with no fit, no floor, and
no price. On the archive's already Pareto-filtered frontiers, the median number of reachable Pareto points
is $5$ out of $11$. Conversely, the median share of Pareto-efficient models that are \emph{never
optimal at any nonnegative price} is $62.5\%$. Pareto efficiency alone therefore does not establish whether a candidate can be selected at any
nonnegative price; the upper-envelope calculation answers that question directly.

None of these numbers uses this paper's loss, its price, or its floor; they are properties of the
published frontiers. Section~\ref{sec:evidence} then scores the rules against the derived price,
where the derived rule's own row is nearly an identity and is interpreted as a consistency
reference.
\emph{Empirical scope of the structural tests.} The two structural hypotheses distinguish subclasses of admissible frontiers, but the published
compute-axis windows do not empirically resolve divisibility, as predicted by
Corollary~\ref{cor:wallivp}(iii). For divisibility, \eqref{eq:powerreq} quantifies
the required per-model noise: $0.73$ milli-nats, approximately $1/27.5$ of the published scale; on
a count axis, the corresponding item counts are reported in Section~\ref{sec:evidence}. The
ceiling requires a different measurement design: because its discriminating signal is linear in
the remaining depth and the published windows lie a median of $0.045$ nats above their fitted
floor, a shallower window is required rather than a finer instrument. The analysis thus specifies
how these hypotheses can be tested while keeping the empirical conclusions within the resolution
of the available data.
\paragraph{Multi-objective and resource-aware model selection.} Pareto frontiers over accuracy and
cost are a standard representation in architecture search and deployment
\citep{elsken2019,tan2019mnasnet,cai2020ofa,zhao2024lamoo}, and the standard output is the
frontier itself, leaving the trade-off to the user. In practice, users typically apply one of three
declared rules: a hard budget, which identifies the feasible endpoint but does not select among the
remaining points \citep{sun2026roofline}; a quality-to-cost ratio, whose arbitrariness is documented by
\citet{dehghani2022efficiency} but which is nevertheless used to rank open models
\citep{prucs2025pareto}; or a sweep of scalarization weights, which addresses how to space the weights
without specifying which weight should be selected \citep{jiang2026surf}. Each is a declared price that is rarely examined; Table~\ref{tab:namedrules} quantifies its consequences on the archive's frontiers.
The complementary question addressed here is: given a frontier,
which point should be selected, and what is the cost of choosing incorrectly? The price is derived from the frontier rather
than elicited (Theorem~\ref{thm:canonical}), the cost of a wrong price is exact rather than bounded
(Theorem~\ref{thm:regret}), and the one input the frontier cannot supply is not a weight but the
grain of the delivered unit (Theorem~\ref{thm:noinvariant}, Proposition~\ref{prop:retry}).
\paragraph{Relation to the single-resource theory.} The coordinates of \eqref{eq:defs}, the
scalarization \eqref{eq:score}, the reading of a price as the frontier's marginal slope, and the
two-atom chart's logistic yield are those of \citet{ABrain.Rank}, which treats one resource and one
reducible channel. There the dual is a binary entropy, the regret of a wrong price is a binary
divergence, and the minimax declaration is $\lamstar=\alpha/2$ with a worst case of one bit. Each
of those three is the $N=1$ member of a statement proved here for $N$ resources and a general
spectrum: Theorem~\ref{thm:duality} replaces the binary entropy by the Shannon entropy of the error
shares, Theorem~\ref{thm:regret} replaces the binary divergence by $\KL$ on $\Delta^{N}$, and
Theorem~\ref{thm:canonical} gives $\lamstar_i=\alpha_i/(N+1)$ with a worst case of $\ln(N+1)$
nats, of which one bit is the case $N=1$. The generalization does not simply extend the same proof. The binary statements concern a scalar
in $[0,1]$, whereas their $N$-resource counterparts concern the simplex and the rank of the
exponent matrix. The minimax value also acquires an interpretation---the capacity of deployment
uncertainty---that is absent when $N=1$.
Nothing else is inherited: the selection rules from that paper appear here only as baselines to be
scored, not as premises. The two works differ in one important respect.
\citet{ABrain.Rank} \emph{assumes} \eqref{eq:M2}. This paper instead derives an admissible class
from a hypothesis about subdivision, places \eqref{eq:M2} outside it, and reports the four
consequences listed above.

The present paper contributes the autonomous reduction, the two frontier bounds and their
tangency, the sharpened differential statistic, the arc representation and its covering law, the
finite-window floor results, and the count-axis analysis. It also establishes $\Neff=1$ in delivered units as a
coordinate-free operating rule for a general spectrum. On the two-atom chart at $N=1$, this rule
selects the same point, $\eps=2\epsinf$, that \citet{ABrain.Rank} obtains from the minimax
argument. The empirical analysis also introduces new research questions, estimands, and measurements.
\paragraph{Test-time compute.} Repeated sampling, self-consistency and best-of-$n$ trade accuracy
against inference cost \citep{wang2023selfconsistency,brown2024,snell2025}.
Section~\ref{sec:count} shows that, for any resource that counts independent attempts, the axiom
follows as a \emph{theorem} from exchangeability and infinite extendibility rather than as an
assumption about the technology. All three laws therefore apply on that axis, and an attempt can be
priced exactly against an e-fold of compute because both are measured in nats.
\paragraph{What is classical.} Section~\ref{sec:law1} distinguishes established results from the contributions developed
here:
the identification of a saturating frontier with a Laplace transform is Bernstein's theorem
\citep{widder1941}, prices as mean parameters and regret as a Bregman divergence are
exponential-family duality \citep{bregman1967,banerjee2005}, with the
divergence--risk correspondence for a binary experiment---the single-channel case of
Theorem~\ref{thm:duality}---set out by \citet{reid2011}, the arc is Fisher--Rao length
\citep{rao1945,amari2000}, the minimax value is the redundancy--capacity theorem
\citep{davisson1973,haussler1997}, and the one-sidedness of every bound is Markov--Krein
\citep{karlin1966}. Two correspondences relocate standing problems into settled literature:
$-\ln\eps$ is a cumulative hazard and $\kappa$ a frailty variance, so diminishing returns is
decreasing failure rate \citep{proschan1963}; and $\epsinf$ is a \emph{cure fraction}, whose
non-identifiability under finite follow-up is a classical theorem
\citep{farewell1982,li2001,yakovlev1996,chen1999} rather than a feature of compute. What is not
classical is the reduction of the whole theory to one curve and two differential inequalities on
it.
\paragraph{What is new, statement by statement.} The list below is the detailed form of
(C1)--(C6) of Section~\ref{sec:setup}, with the numbered statement that carries each one.
\begin{enumerate}[label=(\arabic*),leftmargin=2.2em,itemsep=3pt]
\item A frontier \emph{is} the curve $\Neff$ against $\Delta S$, through an autonomous
      second-order equation, up to two integration constants that are the scale and origin of its
      resource axis: the gauge of the theory is the pair of initial values of an initial-value
      problem (Proposition~\ref{prop:odefirst}). Each hypothesis is a Hankel ladder on that
      curve whose first rung is a single differential inequality, the two walls that bound every
      frontier are those two rungs integrated, and their tangency at the floor makes every test
      built on the gap powerless there by theorem (Proposition~\ref{prop:rungs},
      Corollary~\ref{cor:wallivp}).
\item The operating rule is $\Neff=1$ in delivered units---no floor, no fit, no units, no
      origin---where the canonical price is the price the frontier
      itself quotes, $\lamstar=s$ (Propositions~\ref{prop:neff}, \ref{prop:retry},
      \ref{prop:planeprice}). It is Theorem~\ref{thm:canonical}'s minimax point exactly when the
      frontier has one reducible channel, where the regret is zero; with $N$ reducible channels
      that point is $\Delta S=\ln(N+1)$ instead, and $\Neff=1$ is the gauge-free surrogate that
      stays well posed when the channel count is not identified. Three accuracy measurements
      suffice to recover the curve without estimating a bandwidth; the spacing cancels identically.
\item Four results against the standard form \eqref{eq:M2}, listed above, and a same-cost
      alternative that matches its out-of-sample accuracy while providing exact structure.
\item The commonly used third-order statistic is twice as permissive as the divisibility condition
      it is intended to test; the sharper test uses the same four accuracy measurements
      (Proposition~\ref{prop:strongrung}). Against nulls matched to each frontier's own design, the
      corpus does not resolve divisibility at the published precision. The required precision is
      given in closed form (Section~\ref{sec:evidence}).
\end{enumerate}
\section{The Frontier Is One Curve}
\label{sec:odefirst}\label{sec:planemain}
\begin{figure}[tbp]
\centering
\includegraphics[width=0.82\linewidth]{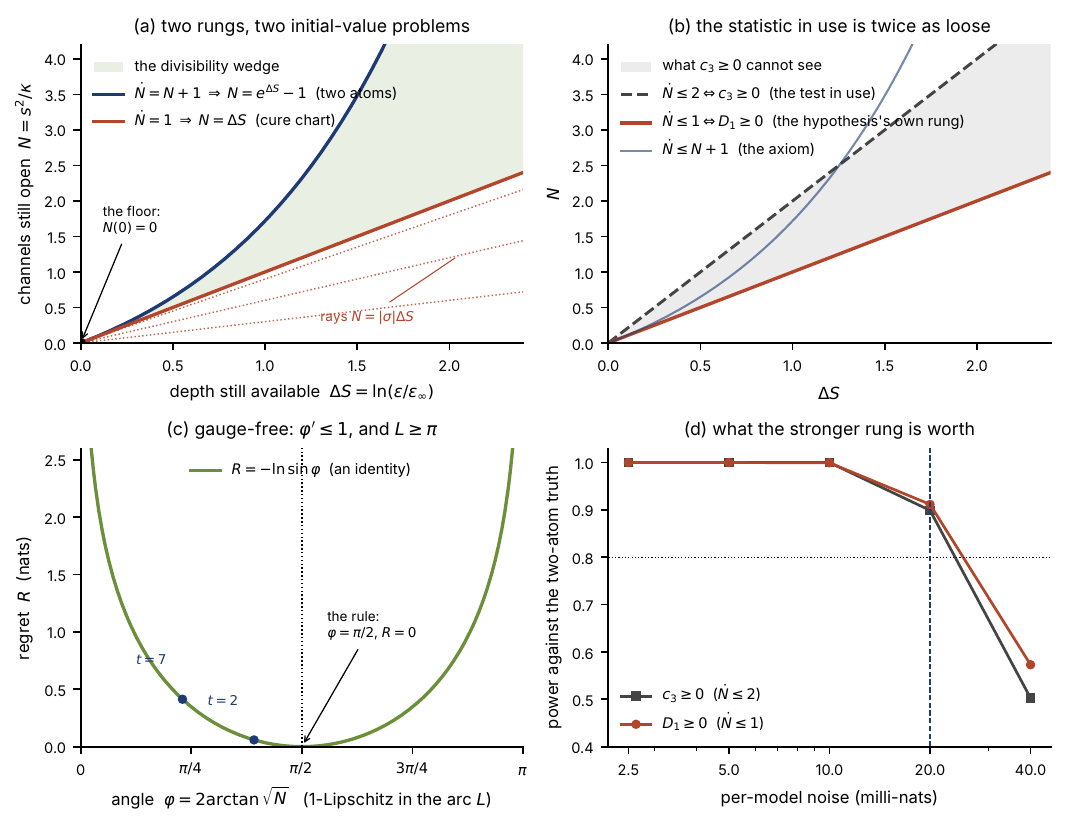}
\caption{\textbf{The theory as one curve.} \textbf{(a)} The primitive is
$\Neff$ against the depth still available, with $\Neff(0)=0$ at the floor. The
axiom bounds its growth by $\dot\Neff\le\Neff+1$ and the hypothesis by
$\dot\Neff\le1$; integrated from the same initial condition those two
inequalities \emph{are} the two charts, so the tangency at the origin is a
property of one initial-value problem. The shaded wedge is the whole
divisibility question, and the dotted rays represent the admissible constant-slope chart family.
\textbf{(b)} The working statistic $c_3\ge0$ is $\dot\Neff\le2$: it cannot see
the band between the two slopes, which $\widehat{D_1}\ge0$ closes at no extra
measurement cost. \textbf{(c)} In the gauge-free coordinates of
Proposition~\ref{prop:anglearc} the single-channel regret is
$-\ln\sin\varphi$ identically, the rule is the right angle, and the axiom states
that $\varphi$ is $1$-Lipschitz with respect to arc length---whence $L\ge\pi$ over the
complete sweep. \textbf{(d)} What the stronger rung is worth on clean
charts at matched level; on this corpus both statistics
stay at their nominal level.}
\label{fig:odefirst}
\end{figure}

Everything in this paper except Section~\ref{sec:count} is a property of a
single function of a single variable. Write the depth still available and the
channels still open as
\begin{equation*}
\Delta S=\ln(\eps/\epsinf)=S_\infty-S,\qquad
\Neff=\frac{s^{2}}{\kappa},\qquad
\dot{\ }=\frac{\mathrm{d}}{\mathrm{d}\Delta S},
\qquad S_\infty=-\ln\epsinf,
\end{equation*}
and take as the primitive not the representation
$\eps=\mathbb{E}_\nu[e^{-b\,m(x)}]$ but the curve $\Neff$ against $\Delta S$.
The representation serves one purpose: Section~\ref{sec:lawI} uses it to establish the positivity
of this curve and the alternating signs of its derivatives. Here $s$ and
$\kappa$ are the mean and the variance of the exponent under the tilted measure
of Section~\ref{sec:law1}, so $\Neff$ is an effective count of active
difficulty channels, which Proposition~\ref{prop:neff} makes exact. What is
assumed about the curve is two differential inequalities, one per hypothesis,
and they are the only content of the two assumptions.
\begin{proposition}[The curve is the frontier]
\label{prop:odefirst}
Along any twice-differentiable frontier with $s>0$ and $\kappa>0$:
\begin{enumerate}[label=(\roman*),leftmargin=2em,itemsep=1pt]
\item \begin{equation}
      \Delta S''=\frac{(\Delta S')^{2}}{\Neff(\Delta S)},
      \label{eq:odeivp}
      \end{equation}
      an autonomous second-order equation in which the curve enters as the
      only coefficient;
\item separating it,
      \begin{equation}
      s(\Delta S)=s_{0}\exp\Bigl\{\int_{\Delta S_{0}}^{\Delta S}\frac{\mathrm{d}u}{\Neff(u)}\Bigr\},
      \qquad
      x=x_{0}-\int_{\Delta S_{0}}^{\Delta S}\frac{\mathrm{d}u}{s(u)},
      \label{eq:reconstruct}
      \end{equation}
      so a frontier is determined by $\Neff$ up to the two integration
      constants $(s_{0},x_{0})$, which are exactly the scale and origin of the
      resource axis. Consequently \emph{the gauge of the theory is the pair of
      constants of an initial-value problem};
\item the shape is the curve, and an inequality between frontier functionals
      that is saturated pointwise determines the frontier, so each such
      inequality has a \emph{unique} extremal chart;
\item the grain acts as a dilation about the origin, since
      $S,\Delta S,s,\kappa$ and $\Neff$ are all extensive in it.
\end{enumerate}
\emph{(Verified by quadrature from the curve alone on four charts, $159/159$
checks at $40$ digits with maximum error $1.5\times10^{-39}$ for the
reconstruction and $9.2\times10^{-41}$ for the axis gauge; script
\texttt{134}, sections C and D.)}
\end{proposition}
\paragraph{The role of the reduction.} Part~(i) is an identity: substituting
$\Neff=s^{2}/\kappa$ into $\Delta S''=\kappa$ gives \eqref{eq:odeivp}, while part~(ii)
follows by separating $\mathrm{d}p/\mathrm{d}u=p/N(u)$ with $p=\Delta S'$ viewed as a
function of the monotone depth $\Delta S$. The significance is the direction of the
representation. Conventional analyses treat $(\epsinf,\alpha,\beta)$ as fitted parameters
and $\Neff$ as a diagnostic; Proposition~\ref{prop:odefirst} reverses this reading. The
function $\Neff(\Delta S)$ is the invariant, and the frontier is recovered from it through an
initial-value problem. The two initial values set only the scale and origin of the resource axis,
not the trade-off itself. The reduction makes three consequences explicit.
\begin{enumerate}[label=(\alph*),leftmargin=2.2em,itemsep=2pt]
\item \emph{The hypotheses become comparable.} Complete monotonicity and divisibility are infinite
      families of determinant conditions on different objects---the reducible spectrum and the
      L\'evy measure. Proposition~\ref{prop:rungs} shows that their first rungs are two bounds on
      one scalar derivative, $\dot\Neff\le\Neff+1$ and $\dot\Neff\le1$, so the two can be ordered
      and the gap between them computed. That is what makes the tangency result of
      Corollary~\ref{cor:wallivp}(iii) statable, and the tangency is what decides what this paper's
      measurements can and cannot establish.
\item \emph{What is a property and what is a unit become decidable.} Because the gauge is exactly
      two constants of an initial-value problem, any reported quantity can be checked against it.
      Theorem~\ref{thm:noinvariant} and the exponent's status as the unit of its axis are both
      consequences; so is the fact that a change of resource unit cannot move a correct operating
      rule, which Section~\ref{sec:evidence} finds that rules in current use do not satisfy.
\item \emph{The admissible charts reduce to two canonical forms.} Each inequality has exactly
      one extremal, identified
      before any algebra (Corollary~\ref{cor:odepencil}, \SM{Proofs: Reading a frontier without
      fitting it}), which is why the same two charts recur throughout: two atoms, and the cure
      chart \eqref{eq:cure}.
\end{enumerate}
Because the curve is the invariant, a frontier quantity that changes under a rescaling or a shift
of the resource axis reflects the choice of gauge rather than the underlying trade-off. This
provides a practical criterion for interpreting reported frontier quantities.
\begin{proposition}[Both ladders are differential inequalities on the curve]
\label{prop:rungs}
Let $c_k=(-1)^{k}\Delta S^{(k)}$ and $E_k=(-1)^{k}\eps^{(k)}$ be the depth
cumulants and the error derivatives, and let
$D_k=c_kc_{k+2}/c_{k+1}^{2}-1$ and $\tilde E_k=E_kE_{k+2}/E_{k+1}^{2}-1$ be the
rungs of the two Hankel ladders of Section~\ref{sec:compose}. A dot denotes
differentiation with respect to the \emph{depth remaining}, $\Delta S$, so
$\dot\Neff=-\mathrm{d}\Neff/\mathrm{d}S=-\sigma$ of \eqref{eq:slope}; read as
$\mathrm{d}/\mathrm{d}S$ instead, every inequality below reverses. Then every rung is
a statement about $\Neff$ alone:
\begin{equation}
\boxed{\ D_0=\frac{\Delta S}{\Neff}-1,\qquad
D_1=1-\dot\Neff,\qquad
\tilde E_0=\frac{1}{\Neff},\qquad
\tilde E_1=\frac{\Neff+1-\dot\Neff}{(\Neff+1)^{2}}\ }
\label{eq:rungs}
\end{equation}
\begin{equation}
\boxed{\ D_2=1-\frac{1}{2-\dot\Neff}
-\frac{\Neff\ddot\Neff}{(2-\dot\Neff)^{2}}\ }
\label{eq:rungtwo}
\end{equation}
and in particular the first rung of each ladder is a bound on how fast the
channel count may grow with depth:
\begin{equation}
\boxed{\ \text{\emph{axiom}}:\ \dot\Neff\le\Neff+1,
\qquad
\text{\emph{hypothesis}}:\ \dot\Neff\le1\ }
\label{eq:tworungs}
\end{equation}
Both are floor-free: $\mathrm{d}/\mathrm{d}\Delta S=-s^{-1}\mathrm{d}/\mathrm{d}x$
does not see $\epsinf$. \emph{(Verified on four charts at two points each, $40$
digits.)}
\end{proposition}
\begin{corollary}[The two walls are the two rungs integrated]
\label{cor:wallivp}
A frontier with a floor has $\Neff\to0$ as $\Delta S\to0$. Integrating
\eqref{eq:tworungs} from that initial condition gives
\[
\Neff\le e^{\Delta S}-1\quad\text{(axiom)},
\qquad
\Neff\le\Delta S\quad\text{(hypothesis)},
\]
equivalently $\epsinf\le\eps/(1+\Neff)$ and $\epsinf\le\eps\,e^{-\Neff}$.
Equality at a single point forces the reducible spectrum, respectively the
L\'evy measure of $S$, to be one atom; equality at every point holds exactly on
the two-atom chart and on the cure chart, and on nothing else up to gauge.
Consequently
\begin{enumerate}[label=(\roman*),leftmargin=2em,itemsep=1pt]
\item the second wall is the tangent to the first at the origin,
      $e^{u}-1-u=\tfrac12u^{2}+O(u^{3})$, and the fraction of the admissible
      range the hypothesis excludes is $1-u/(e^{u}-1)=\tfrac12u+O(u^{2})$;
\item the two walls are one initial-value problem read twice, and their
      tangency at the origin is the statement that $\dot\Neff\le\Neff+1$ and
      $\dot\Neff\le1$ agree at $\Neff=0$;
\item the vanishing power of any test built on the gap between them, at the
      deep end, is that tangency rather than a property of the data;
\item the two closed-form floor bounds $\eps/(1+\Neff)$ and $\eps e^{-\Neff}$
      are the same two walls solved for $\epsinf$, and the two operating
      depths $\ln(1+1/t)$ and $1/t$ are the same two walls read at the height
      $\Neff=1/t$.
\end{enumerate}
\end{corollary}
\begin{corollary}[The admissible charts form a pencil of constant-slope solutions]
\label{cor:odepencil}
$\dot\Neff=|\sigma|$ constant integrates to the ray
\begin{equation}
\boxed{\ \Neff=|\sigma|\,\Delta S\ },
\qquad
\text{slope in }\Delta S=|\sigma|=-\sigma,
\qquad
S\text{-intercept}=-\ln\epsinf,
\label{eq:planeline}
\end{equation}
where $\sigma:=\mathrm{d}\Neff/\mathrm{d}S=-\dot\Neff$ of \eqref{eq:slope}. The slope in the plane
actually plotted is therefore $|\sigma|$, and the floored members have $\sigma\le0$. Three
consequences follow.
\begin{enumerate}[label=(\roman*),leftmargin=2.2em,itemsep=2pt]
\item \emph{The charts of Section~\ref{sec:floor} are this one-parameter family.} On a ray every
      rung of \eqref{eq:rungs} collapses to the single condition $\dot\Neff\le1$, that is
      $\sigma\ge-1$, which is the trichotomy. The floored power charts (\SMchart, where the exponent
      is $\beta<0$) are the members with $|\sigma|=\beta/(\beta-1)\in(0,1)$; the cure chart is
      $|\sigma|=1$; the floorless charts are the members read in $S$, where $\epsinf=0$ and the ray
      $\Neff=\sigma S$ has $\sigma=\beta/(1-\beta)\ge0$, unbounded above. A $\Gamma(k)$ spectrum is
      the degenerate member at the separator $\sigma=0$: it has $\Neff\equiv k$ and a free
      intercept rather than the family's zero one, so it is the horizontal line of the pencil and
      not a ray of it.
\item \emph{The floor is the line's own intercept and the shape index is its slope.}
      Proposition~\ref{prop:nofloor} is simply the statement that a line in a plane is determined
      by two parameters, not three. Fit a floored chart to a floorless frontier of index $\beta$ and it returns, at
      depth $S$, the \emph{local} floor share $\hat p_0=(1-\beta)/\{(1-\beta)+\beta S\}$ of
      Proposition~\ref{prop:neff}(i), the true share $\epsinf/\eps$ being zero there. That
      expression is strictly decreasing in $\beta$ at every fixed depth. A fitted floor can
      therefore only be a monotone readout of $\beta$, and the two are one degree of freedom.
\item \emph{Two additional structural consequences follow.} A bounded non-degenerate spectrum is never divisible,
      because the two-atom chart has $\dot\Neff=\Neff+1>1$ at every point; and a divisible frontier
      has no elbow, because $\kappa'=-s^{3}(2-\dot\Neff)/\Neff^{2}$.
\end{enumerate}
\end{corollary}
The tangency result has a direct implication for empirical testing. \emph{Because the walls touch at the
origin the divisibility gap vanishes quadratically at the floor, so any
procedure that separates the hypothesis from the axiom by the position of
$\Neff$ between them has a discriminating margin that vanishes with the depth
and therefore no power there, whatever the measurement.}
Section~\ref{sec:evidence} reports that the published windows sit deep; that is
not bad luck in the sampling of corpora but a theorem.
\begin{proposition}[The statistic in use is twice as loose, and the repair is free]
\label{prop:strongrung}
The working third-order test of this paper, $c_3=S'''\ge0$, is by Proposition~\ref{prop:rungs}
exactly $\dot\Neff\le2$. The hypothesis's own first rung is $\dot\Neff\le1$. Both are available
from the same four accuracies: with divided differences on a quadruple,
$s=\mathrm{dd}_1$, $\kappa=-2\,\mathrm{dd}_2$ and $S'''=6\,\mathrm{dd}_3$, so
\begin{equation}
\boxed{\ \widehat{D_1}
=\frac{3}{2}\,\frac{\mathrm{dd}_1\,\mathrm{dd}_3}{\mathrm{dd}_2^{2}}-1\ \ge0\ }
\label{eq:d1hat}
\end{equation}
is floor-free, bandwidth-free, and strictly stronger than $\mathrm{dd}_3\ge0$.
\end{proposition}
\noindent The statement is exact; finite-span estimation introduces approximation error, which this
subsection characterizes. Three aspects should be distinguished: the limit of \eqref{eq:d1hat},
its behavior at a finite span, and its behavior on irregular nodes.
\paragraph{It is a plug-in, and plug-ins have a span.} Expression \eqref{eq:d1hat} is the exact
invariant $D_1=sS'''/\kappa^{2}-1$ with each derivative replaced by its divided difference. Since
$\mathrm{dd}_k=S^{(k)}(\xi_k)/k!$ at a different mean-value node $\xi_k$ per order, it equals
$D_1$ only as the span vanishes. The size of the discrepancy is measurable on known truths. On a
$\Gamma(2)$ truth, whose exact $D_1$ is $1$, an equally spaced quadruple of span $3$ returns
$0.42$ and one of span $9$ returns $0.27$. On the cure chart at equal spacing it returns $0$
identically, for every span and every $(\Lambda,c)$---which is the saturation discussed below.
\paragraph{The correction improves calibration but does not add new evidence.} The
archive's median residual scale on the frontiers carrying a usable quadruple is $47$ milli-nats.
On the archive's own quadruples at that scale the test remains at its nominal level, with median power $0.056$. On a clean five-e-fold design,
it separates the alternatives: power is $0.912$ versus $0.899$ at a per-model noise of $20$
milli-nats, and $0.573$ versus $0.502$ at $40$ milli-nats. Those are two points of a swept design grid, not measurements of this
corpus. Thus, \eqref{eq:d1hat} improves the test statistic without changing the conclusions supported by
the available evidence.
\paragraph{Applicability and calibration.} The statistic \eqref{eq:d1hat} requires $s>0$ and
$\kappa>0$ on the quadruple, and is therefore available on $78$ of the $111$ frontiers on which
$\mathrm{dd}_3$ can be evaluated. Its sharper boundary also calls for calibrated inference: the
cure chart saturates $D_1$, so a noisy plug-in sign alone does not distinguish it reliably from a
nearby divisible truth. By contrast, the looser $\mathrm{dd}_3$ statistic leaves divisible charts
strictly inside the admissible region. The design-matched comparisons below account for this
boundary behavior.
Node geometry accounts for a second component of the rejection rate and, in this archive, is the larger component. The
quadruple is the widest four-point subset at index thirds of a measured grid of \emph{log}-resource,
and the resulting resource values are not generally geometrically spaced. The
nodes are therefore irregular: not one of the $117$ frontiers yields an equally spaced quadruple,
and the median ratio of widest to narrowest gap is $3.0$. To isolate this effect, we remove
measurement noise and evaluate a strictly interior divisible truth---a compound-Poisson mixture
with median exact $D_1=0.22$---on each frontier's four nodes. The sign test rejects this truth on
$54\%$ of $3\,183$ readable draws, compared with $0\%$ on equally spaced nodes spanning the same
interval. At zero noise, node geometry alone accounts for this rejection rate. Reading the rung on an equally
spaced quadruple removes it wherever one is available. Near the boundary, the remaining uncertainty makes the result sensitive to measurement noise.
The sharper rung must therefore be evaluated against a null simulated on the same design. This
is the methodological principle introduced in Section~\ref{sec:lawfree}, applied here to the new
statistic. Section~\ref{sec:evidence} uses the same principle for the empirical tests.
\begin{proposition}[In gauge-free coordinates both rungs are Lipschitz bounds]
\label{prop:anglearc}
Put
\begin{equation}
\varphi=2\arctan\sqrt{\Neff}\in(0,\pi),
\qquad
\mathrm{d}L=\frac{|\mathrm{d}\Delta S|}{\sqrt{\Neff}},
\label{eq:angle}
\end{equation}
the second being the Fisher--Rao arc of Section~\ref{sec:lawIII}, so that
$L=\int_0^{\lambda}\mathrm{d}v/\sqrt{N(v)}$ with $\lambda=-\ln\epsinf$ is a
functional of the curve alone, as Proposition~\ref{prop:odefirst}(ii) requires.
\emph{This $\varphi$ is the channel angle, measured from the floor:
$\varphi\to0$ as $\Neff\to0$ and $\varphi\to\pi$ as $\Neff\to\infty$. The
Fisher--Rao coordinate of the binary simplex oriented the other way, by
increasing floor share, is the mirror $\pi-\varphi=2\arcsin\sqrt{\hat p_0}$ with
$\hat p_0=1/(1+\Neff)$ the local floor share of Proposition~\ref{prop:neff}(i),
which is the true share $\epsinf/\eps$ on the one reducible channel of
Proposition~\ref{prop:binary}; $\sin\varphi$---hence the regret and the arc
length---is the same in either, while $\cos\varphi$ and the direction of travel
are not.} Then $\sin\varphi=\operatorname{sech}(\tfrac12\ln\Neff)$ and
$\cos\varphi=(1-\Neff)/(1+\Neff)$, so
\[
R_{1}:=\ln\cosh\bigl(\tfrac12\ln\Neff\bigr)=-\ln\sin\varphi
\quad\text{identically},
\qquad
\frac{\mathrm{d}\varphi}{\mathrm{d}L}=\frac{\dot\Neff}{\Neff+1},
\]
with $R_1$ the single-channel regret \eqref{eq:logcosh}, which is the regret
$\KL(u\Vert p)$ of Theorem~\ref{thm:master} exactly when the spectrum is two
atoms,
and the two rungs \eqref{eq:tworungs} read
\[
\text{\emph{axiom}}:\ \varphi'\le1,
\qquad
\text{\emph{hypothesis}}:\ \varphi'\le\cos^{2}(\varphi/2).
\]
\emph{The axiom says the angle advances no faster than the arc.} Three
consequences follow directly. Law~II and Law~III become definitions:
the single-channel regret is $-\ln\sin\varphi$ and the rule $\Neff=1$ is
$\varphi=\pi/2$, the right angle, where it is exactly zero; at grain $t$ the
rule \eqref{eq:retry} places the operating point at $\varphi=2\arctan(1/\sqrt
t\,)$ instead, where it is the regret read in \emph{delivered} units that
vanishes. Law~III's constant is a Lipschitz bound, and both bounds below are the two walls
of Corollary~\ref{cor:wallivp} substituted into $L=\int_0^{\lambda}\mathrm{d}v/\sqrt{N(v)}$: the
axiom's $N\le e^{v}-1$ gives
$L\ge\int_0^{\lambda}(e^{v}-1)^{-1/2}\mathrm{d}v=2\arccos\sqrt{\epsinf}$ for the \emph{admissible}
arc, the part of the frontier at or below chance level, with equality iff $\varphi'\equiv1$, that is,
iff two atoms; letting $\lambda\to\infty$ gives $L\ge\pi$ for the complete sweep, $p_0$ over all of
$(0,1)$, which a single-channel chart traverses as $\Delta\varphi=\pi$. The
divisible bound $L\ge2\sqrt{-\ln\epsinf}$ is the hypothesis wall $N\le v$ integrated the
same way:
\begin{equation}
\boxed{\ L\ \ge\ 2\arccos\sqrt{\epsinf}\ }\quad\text{(axiom)},
\qquad\qquad
\boxed{\ L\ \ge\ 2\sqrt{-\ln\epsinf}\ }\quad\text{(hypothesis)},
\label{eq:planecover}
\end{equation}
both for the admissible arc, and each with equality on its own extremal: the
axiom's on the two-atom chart, the hypothesis's on the cure chart. (The
two-atom chart is not divisible, so it does not satisfy the second bound, and
nothing here claims it does.) The hypothesis bound is the binding one throughout,
$2\sqrt\lambda>2\arccos\sqrt{e^{-\lambda}}$ for every $\lambda>0$ with the two
agreeing to second order at the origin, and it passes the axiom's limiting
value $\pi$ at $\lambda=\pi^{2}/4$, that is at $\epsinf=e^{-\pi^{2}/4}=0.0848$:
below that floor a divisible frontier is forced to carry more than one channel,
since $N\ge L/\pi$.
\emph{(Verified on four charts, $102/102$ checks at $40$
digits.)}
\end{proposition}
The constant of Law~III is therefore not an independent fact:
$\int_{0}^{\infty}(e^{u}-1)^{-1/2}\mathrm{d}u$ is a Beta integral, the same $\pi$ as the
Fisher--Rao diameter of Proposition~\ref{prop:binary}, so \emph{Law~III is a corollary}---$\pi$ is
the smallest arc a frontier with a vanishing floor can traverse, attained by the chart that
saturates the axiom. Read in the opposite direction, the same inequality shows that a window arc cannot certify the
channel count: the \emph{admissible} arc of a floored frontier is bounded below by
$2\arccos\sqrt{\epsinf}<\pi$ and, on the two-atom chart, equals it, so a
\emph{window} arc below $\pi$ is satisfied by every frontier whatever its
channel count and carries no information about it. The surviving implication is
the bound $N\ge L/\pi$ for the complete sweep, vacuous below $\pi$; what
$L/\pi$ measures on a window is window completeness, and
Section~\ref{sec:evidence} reads it that way.
Figure~\ref{fig:odefirst} collects the three readings of the reduction used
below: the primitive and the two walls integrated from the same initial
condition, the gap the working third-order statistic cannot see, and the
gauge-free coordinates in which the operating rule is a right angle.
\subsection*{What the Curve Costs to Read, and the Price at the Operating Point}
\begin{figure}[tbp]
\centering
\includegraphics[width=0.82\linewidth]{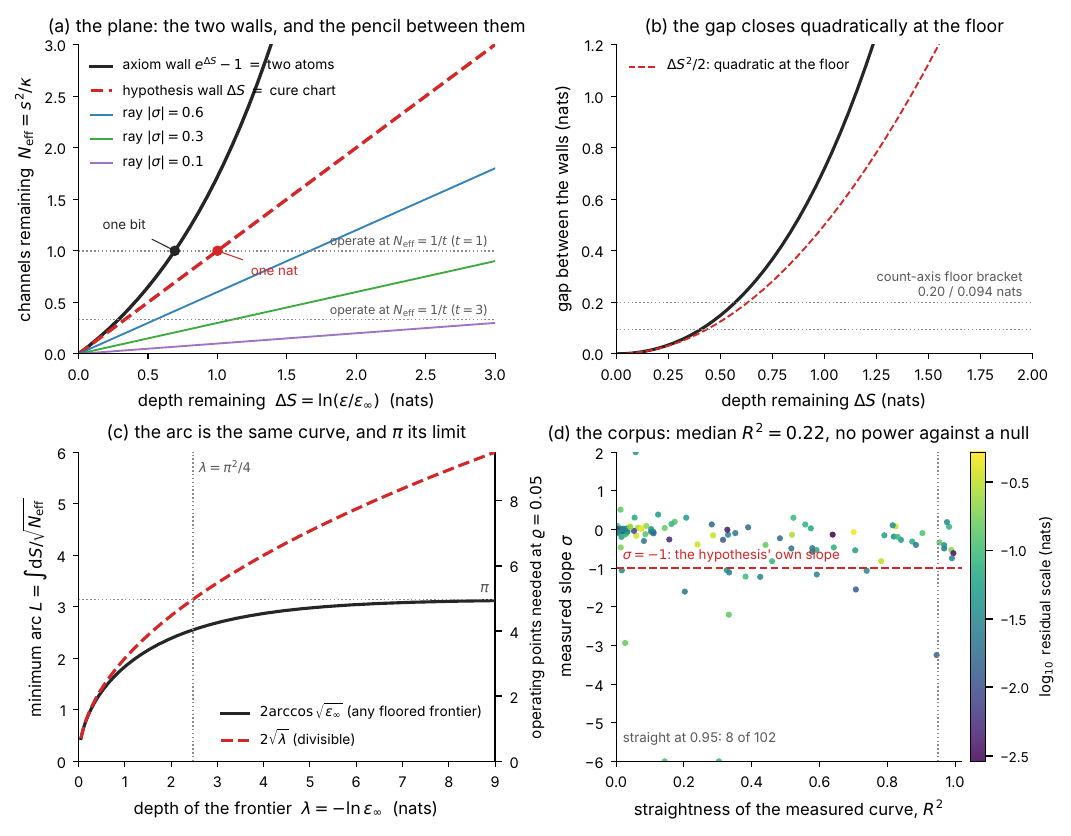}
\caption{\textbf{The theory is one plane curve.}
(a) Channels remaining against depth remaining. The axiom confines every
completely monotone frontier below $e^{\Delta S}-1$ and the hypothesis below
that curve's own tangent at the origin, $\Delta S$; the two bounding solutions
are charts, the two-atom chart and the cure chart, and each is the unique
frontier saturating its wall (Proposition~\ref{prop:odefirst}(iii)). Every
constant-$\sigma$ chart is a ray whose slope is $|\sigma|$ and whose intercept
is the floor, so the admissible charts form a pencil (Corollary~\ref{cor:odepencil}).
The operating point is the height $N_{\mathrm{eff}}=1/t$; where it meets a wall
is the recommended depth, one bit on the axiom's extremal against one nat on the
hypothesis's.
(b) The gap between the walls closes quadratically at the origin, and the origin
is the floor: a second-order test is powerless at the deep end by theorem, not by
accident of this corpus.
(c) The admissible arc is a functional of the same curve, so the walls bound it
from below; with the covering law the right-hand axis reads the number of
operating points a benchmark needs, and the hypothesis bound passes the axiom's
limiting value $\pi$ at $\lambda=\pi^{2}/4$.
(d) The archive's $117$ frontiers, read with three accuracies per point and no
bandwidth. The curve is noise-dominated: the median straightness is $R^{2}=0.22$
and against a design-matched null no statistic read from it has power
(Section~\ref{sec:evidence}).}
\label{fig:plane}
\end{figure}

The curve can be read from three accuracy measurements without estimating a bandwidth. With
$\ell=\ln\eps$ on an equally spaced triple,
\begin{equation}
\boxed{\ \widehat{N}_{\mathrm{eff}}
=\frac{(\ell_+-\ell_-)^{2}}{4\,(\ell_--2\ell_0+\ell_+)}\ }
\label{eq:nhat}
\end{equation}
because $\Neff$ is a square over a second difference and the $h$ in
$s\approx(\ell_--\ell_+)/2h$ cancels identically against the $h^{2}$ in
$\kappa\approx(\ell_--2\ell_0+\ell_+)/h^{2}$, both positive by the
log-convexity of $\eps$. Everything above is then
arithmetic in $(\eps,\Neff,t)$. The estimator \eqref{eq:nhat} is a ratio with a second
difference in its denominator, and its empirical precision is limited on this corpus: median
straightness is $R^{2}=0.2233$ on the $102$ of $117$ frontiers that return an estimate, and the
statistic has no power against a design-matched null (Section~\ref{sec:evidence}). The curve
therefore provides a reporting standard for new measurements; the operating price, by contrast,
can be obtained without estimating curvature.
\begin{proposition}[The canonical price is the local price]
\label{prop:planeprice}
With $t$ the grain and $\lamstar=\sqrt{\kappa/t}$ the canonical price of
Section~\ref{sec:lawfree},
\begin{equation}
\boxed{\ \lamstar=\frac{s}{\sqrt{t\,N_{\mathrm{eff}}}}\ }
\qquad\Longrightarrow\qquad
\boxed{\ \lamstar=s\ \text{ exactly where }N_{\mathrm{eff}}=1/t\ },
\label{eq:planeprice}
\end{equation}
with $\lamstar>s$ deeper, where $t\Neff<1$, and $\lamstar<s$ shallower. In
delivered units the count is $N^{\mathrm{unit}}=t\,N_{\mathrm{eff}}$---every
one of $S,\Delta S,s,\kappa,\Neff$ being extensive in the grain---and the rule
\eqref{eq:retry} reads $N^{\mathrm{unit}}=1$, a statement with no parameter in
it, at which \eqref{eq:logcosh} read in those units,
$R_1=\ln\cosh(\tfrac12\ln N^{\mathrm{unit}})$, vanishes and vanishes only; and
that point is the unique fixed point of the involution
$N^{\mathrm{unit}}\mapsto1/N^{\mathrm{unit}}$.
\end{proposition}
\emph{The canonical price is the frontier's own local price at the operating point selected by
the theory.} No curvature estimate, Prony root, or fitted chart is needed to price this
recommendation; such quantities are needed only to price a point away from it, where the
shading factor is $(t\Neff)^{-1/2}$. Together with Proposition~\ref{prop:odefirst}(iv), this
completes the picture: the grain dilates the curve about its origin, and the operating rule selects
the dilation for which the operating point has height one. The qualifications in Section~\ref{sec:scope} follow from this picture: the curve is
invariant, the two integration constants are not, and an operating rule must specify the grain.
\paragraph{Reconstruction and empirical estimation.} The reconstruction
\eqref{eq:reconstruct} determines a frontier from an \emph{exact} curve. With noisy measurements,
the inverse problem is ill-posed in Hadamard's sense, and Markov--Krein extremality makes the
resulting tests one-sided. The empirical curve is also noise-dominated on the present corpus.
Accordingly, the exact representation identifies the structural object to measure, while
finite-sample use requires uncertainty calibration and the initial values that set the resource
axis's scale and origin. Figure~\ref{fig:plane} shows the resulting plane: the
two walls, the pencil of admissible charts between them, and the height at
which the operating point meets each wall.
\section{The Dictionary: Laws I and II Are Exponential-Family Duality}
\label{sec:law1}\label{sec:lawI}\label{sec:law}\label{sec:duality}
The curve of Section~\ref{sec:odefirst} is positive and its derivatives alternate because of one
property of a frontier, statable before any model of what a frontier is:
\begin{equation*}
(-1)^k\,\frac{\mathrm{d}^k\eps}{\mathrm{d}x^k}\ \ge\ 0
\qquad\text{for every }k\ge0 :
\tag{A}
\label{eq:axiom}
\end{equation*}
\emph{returns diminish, and so does every rate at which they diminish.} Three routes reach it. The
modeling route---error splitting into positive mode contributions, each continuous, non-increasing
and scale-free---forces $\eps(x)=\sum_jc_je^{-a_j\cdot x}$ with $c_j>0$ and $a_j\ge0$, the modes
with $a_j=0$ combining into one irreducible floor (\SM{Proofs: Law~I}). The second is a proof from
a \emph{symmetry}: exchangeable resource increments give \eqref{eq:axiom} by de Finetti alone
(Proposition~\ref{prop:countaxis}). The third subsumes both and is the one this paper rests on:
Proposition~\ref{prop:compose} shows that \emph{the price curve being itself a frontier} implies
\eqref{eq:axiom}, so that single hypothesis replaces the axiom and the Bernstein assumption
together.
\begin{proposition}[Spectral form]
\label{prop:spectral}
If the error is an average of scale-free modes against a finite positive
measure $\mu$ on $\R^n_{\ge0}$, then
\begin{equation}
\eps(x)=\int e^{-a\cdot x}\,\mathrm{d}\mu(a),
\qquad
\epsinf=\mu(\{0\}).
\label{eq:spectral}
\end{equation}
The finite law is the atomic case, and the floor is the mass at the origin.
Along any direction, $\eps$ is completely monotone, and by Bernstein's
theorem every completely monotone function arises this way from a unique
$\mu$: \eqref{eq:spectral} characterizes the family rather than merely
generating it.
\end{proposition}
Equation~\ref{eq:spectral} is the whole of Law~I, because it says that
\begin{equation}
\ln\eps(x)=\Lambda(-x),
\qquad
\Lambda(\theta)=\ln\int e^{\theta\cdot a}\,\mathrm{d}\mu(a),
\label{eq:cgf}
\end{equation}
is the cumulant generating function of the spectrum at minus the log-resource, and the
\emph{tilted measure}
\begin{equation}
\mathrm{d}p_x(a)=\frac{e^{-a\cdot x}\,\mathrm{d}\mu(a)}{\eps(x)}
\label{eq:tilt}
\end{equation}
---the distribution of error over modes at the operating point---is its Gibbs measure, an
exponential family with natural parameter $-x$ and sufficient statistic $a$, whose derivative
identity gives $s=\mathbb{E}_p[a_v]$ and $\kappa=\operatorname{Var}_p(a_v)$.
\emph{Diminishing returns are not a separate postulate about technology; they are the Fisher
information of the frontier} (\SM{Proofs: Law~I}).
\subsection*{Prices Are Error Shares, and the Dual Value Is a Rate Function}
Let $A$ have rows $a_j^\top$, let $\mathcal{C}=\operatorname{conv}\{a_0,\dots,a_N\}$, let
$C=\sum_jc_j$ and $\bar c_j=c_j/C$, and let $p_j(x)=c_je^{-a_j\cdot x}/\eps(x)$ be the error
shares.
\begin{theorem}[Price--error-share duality]
\label{thm:duality}
For $\lambda\in\mathcal{C}$,
\begin{equation}
\sigma^*(\lambda)
=\min_{p\in\Delta^N,\ A^\top p=\lambda}\KL(p\Vert\bar c)-\ln C
=-\sum_jp_j\ln c_j-\Hs(p),
\label{eq:dual}
\end{equation}
and $\sigma^*(\lambda)=+\infty$ otherwise. For $\lambda$ in the relative
interior the supremum is attained, up to directions in $\ker A$, at the
unique point where
\begin{equation}
A^\top p(x^*)=\lambda.
\label{eq:moment}
\end{equation}
\end{theorem}
By \eqref{eq:cgf}, $\sigma^*(\lambda)=\Lambda^*(\lambda)$: \emph{the dual value of a declared price
is the Cram\'er rate function of the failure-rate spectrum}, and \eqref{eq:dual} is its Sanov form
\citep{dembo1998}. A declared price is a declared \emph{mean failure rate}, so the admissible
prices are the interior of the convex hull of the spectrum's support. The gradient identity
$\nabla S=A^\top p$ is Shephard's lemma with error shares in place of cost shares. If
each resource has its own channel, $a_i=\alpha_ie_i$, then
\begin{equation}
p_i=\frac{\lambda_i}{\alpha_i},
\qquad
p_0=1-\sum_i\frac{\lambda_i}{\alpha_i},
\label{eq:diagshare}
\end{equation}
so a price is literally a share vector; and since
$\mathcal{C}\subseteq\operatorname{row}(A)$, the supported price space has affine dimension at most
$\operatorname{rank}(A)$---a frontier can be measured along many resource dimensions yet support only one
independent price direction. On the $82$-frontier panel the recovered
channel count is strictly below the number of measured resources on $51$ frontiers and equals one
on $76\%$ of them. When this condition fails, the one-dimensional correction is given in \SMonedim.
\begin{theorem}[Exact regret of a wrong price]
\label{thm:regret}
Let $\lambda$ and $\hat\lambda$ lie in the relative interior, with share
vectors $p$ and $\hat p$. Committing to the point chosen at $\hat\lambda$ when
the realized price is $\lambda$ costs
\begin{equation}
R(\lambda\Vert\hat\lambda)=\KL(p\Vert\hat p)\quad\text{nats}.
\label{eq:regret}
\end{equation}
\end{theorem}
Regret is the Bregman divergence of the dual value function \citep{bregman1967,banerjee2005}, which
for a rate function is the relative entropy between the tiltings it indexes, so disagreement about
exchange rates has an exact cost and not merely a bound. With one resource and one reducible
channel $p$ is a scalar and \eqref{eq:regret} is the binary divergence of \citet{ABrain.Rank};
what is proved here is the statement on $\Delta^{N}$. The same quantity can also be computed without specifying a spectrum:
$R=\ln(\hat\eps/\eps)+\lambda\cdot(\hat x-x)$, using two measured points and the
realized price.
\subsection*{Law~II: The Canonical Price Is the Mean Exponent, and Regret Is an
arithmetic--geometric mean gap}
\label{sec:lawII}\label{sec:canonical}\label{sec:acconly}
Let $p_0(x)=\epsinf/\eps(x)$ be the share carried by the floor. Then
$\ln(\eps/\epsinf)=-\ln p_0$ identically, for every spectral measure with $\epsinf>0$ and every
number of modes: \emph{the remaining reducible log-error under unlimited resource use is the
surprisal of the floor mode}, so resources buy nats of evidence against the proposition that the error is
irreducible (\SM{Proofs: Law~II}). Write $e_j(x)=c_je^{-a_j\cdot x}$ for mode $j$'s
contribution, so $\eps=\sum_je_j$ and $p_j=e_j/\eps$; let
\begin{equation*}
\abar=\frac{1}{N+1}\sum_{j=0}^{N}a_j
\end{equation*}
be the \emph{unweighted} mean of the exponents, with the floor entering as the exponent $0$; and
let $\AM,\GM$ be the arithmetic and geometric means of the $N+1$ numbers $e_j$.
\begin{theorem}[Master identity]
\label{thm:master}
For every spectrum and every operating point,
\begin{equation}
\boxed{\ \sigma_{\abar}(x)\;=\;\mathrm{const}-\ln\frac{\AM(e(x))}{\GM(e(x))}\ },
\qquad
\KL\bigl(u\Vert p(x)\bigr)=\ln\frac{\AM(e(x))}{\GM(e(x))},
\label{eq:master2}
\end{equation}
with $u$ the uniform share vector and the constant
$-\ln(N+1)-\frac1{N+1}\sum_j\ln c_j$ independent of $x$. Consequently the
scalarization at the price $\abar$ is maximal exactly where all mode errors
are equal, and the maximum is unique.
\end{theorem}
The price $\abar$ is the canonical price, $\abar_i=\alpha_i/(N+1)$ with one
channel per resource. The equality case is the canonical operating point
$\eps^\star=(N+1)\epsinf$---\emph{every reducible error component equals the irreducible one}.
Writing $R(x)=\KL(u\Vert p(x))$, the three laws are three properties of one function: it
is convex in the log-resource (Law~I), it vanishes at equipartition of the error across modes
(Law~II), and---\emph{for one reducible channel}---in the arc coordinate of
Section~\ref{sec:lawIII} it is $-\ln\cos\theta$ on an interval of length $\pi$ (Law~III). Off
two atoms the third reading is the regret of the \emph{tangent} two-atom chart, written $R_1$
below, and not $R$ itself.
Consider a provider committing to one operating point before deployment conditions are known. Four
requirements on a pricing rule are natural: the declared price is supported by the frontier; the
selected point is unchanged when a resource axis is reparameterized $t_i\mapsto t_i^{\beta_i}$; the
rule is unchanged by a rescaling of the error or of the resource units; and the rule minimizes
worst-case regret. The first three alone force the rule to declare the same error shares everywhere.
The fourth fixes which shares.
\begin{theorem}[Canonical reference]
\label{thm:canonical}
Assume one channel per resource, so that $\operatorname{rank}A=N$ and every
share vector in $\Delta^N$ is realizable. Then the frontier-induced minimax
problem has a unique solution. It declares the uniform share vector
$\hat p_j=1/(N+1)$, that is, $\lamstar=\abar$ with $\lamstar_i=\alpha_i/(N+1)$,
and
\begin{enumerate}[label=(\roman*),leftmargin=2em,itemsep=1pt]
\item its worst-case regret is exactly $\ln(N+1)$ nats, and no rule,
      admissible or not, attains a smaller worst case;
\item since $\KL(p\Vert u)=\ln(N+1)-\Hs(p)$, the minimax declaration is the
      maximum-entropy share vector and the minimax value is the maximal
      entropy of the shares; equivalently $\ln(N+1)$ is the Hartley entropy of
      $\operatorname{supp}\mu$, while the dual value \eqref{eq:dual} carries
      its Shannon entropy;
\item for any prior on the realized shares with mean $\bar p$ the Bayes
      declaration is $\bar p$; the uniform prior on the vertices is least
      favorable;
\item the selected point satisfies
      \begin{equation}
      \lamstar_i=\frac{\alpha_i}{N+1},
      \qquad
      \eps^\star=(N+1)\,\epsinf,
      \qquad
      \Delta S=\ln(N+1),
      \label{eq:canon}
      \end{equation}
      which is the equality case of Theorem~\ref{thm:master}.
\end{enumerate}
\end{theorem}
\paragraph{The canonical price for interacting resources.}
``One channel per resource'' is $\theta=1$ in the power-mean family of \SM{The empirical study}, and it
is not a modeling convenience: the archive's $21$ factorial language-model grids fit $\theta$ with a
profile interval, and $\theta=1$ falls inside it on only $\mathbf{6}$ of them. The theorem above is
therefore the \emph{special case} on those grids and not the general one. What replaces it is not a
repair but the same statement with the shares nested: writing $m=n^{1/\theta}$, the minimax declaration
has reducible share $m/(1+m)$, worst case $\ln(1+m)$ and selected depth
$\Delta S=\ln\bigl(1+n^{1/\theta}\bigr)$, which is Theorem~\ref{thm:canonical} at $\theta=1$ and is
proved, with the nested entropy that carries it, in \TR{Interacting resources: the nested dual}
(verified to $40$ digits over $285$ checks). The correction is worth having: across
those $21$ grids, assuming additivity misplaces the canonical depth by a median $0.340$ nats, about
seven times the tolerance inside which the recommendation does not change.
\emph{For a single resource, as in every frontier in Table~\ref{tab:namedrules}, $n=1$ and the two
coincide for every value of $\theta$}; the nesting matters exactly when resources are read together.
For any closed set of share
vectors the deployer admits, the minimax value is the \emph{capacity} of the deployment uncertainty
and the price to declare is the capacity-achieving mixture---the redundancy--capacity theorem
\citep{davisson1973,haussler1997} in the frontier's coordinates, with
Theorem~\ref{thm:canonical} the unrestricted case (\SM{Proofs: Law~II}). At $N=1$ the
declaration is $\lamstar=\alpha/2$ and the worst case is $\ln2$, one bit, which is the
single-resource rule of \citet{ABrain.Rank}; the entropy reading of (ii) and the capacity reading
above are empty there and are what the general case adds. The reference is
feasible only if $\epsinf\le1/(N+1)$; and with a single reducible
channel the frontier depends on the resources through one dimensionless coordinate,
\begin{equation}
\zeta=a\cdot x-\ln(c/\epsinf),
\qquad
\frac{\eps}{\epsinf}=1+e^{-\zeta},
\label{eq:universal}
\end{equation}
with canonical rule $\lamstar=a/2$, $\eps=2\epsinf$, $\Delta S=\ln2$---\emph{increase resources
until the reducible error equals the irreducible error}.
Specializing Theorem~\ref{thm:master} to one reducible channel gives every scalar statement at
once. With $r=\eps-\epsinf$ and $\psi=r/\epsinf$,
\begin{equation}
R=\ln\frac{\AM(\epsinf,r)}{\GM(\epsinf,r)}
 =\ln\frac{\epsinf+r}{2\sqrt{\epsinf r}}
 =\ln\cosh\tfrac{\zeta}{2}
 =\ln\cosh\bigl(\tfrac12\ln\psi\bigr),
\label{eq:lncosh}
\end{equation}
so $R\le\varrho$ exactly when $|\zeta|\le2\,\mathrm{arccosh}\,e^{\varrho}$; at $\varrho=0.05$ nats
that is $\eps/\epsinf\in[1.53,2.89]$, so \emph{a canonical recommendation should be reported as a
band, not a point.} Define the \emph{canonical value}
$J(\eps)=\tfrac12\ln(\eps-\epsinf)-\ln\eps$; then $J=-R+\mathrm{const}$, so the canonical ranking
of a candidate set depends only on the error list and $\epsinf$; the fitted $c$ and
$\alpha$ cancel (\SM{Proofs: Law~II}). The key implication is that the two quantities a
fit is least able to identify on a short frontier are precisely the two that cancel.
\begin{corollary}[Two candidates]
\label{cor:two}
For $\eps_1>\eps_2$ the more accurate candidate carries the larger canonical
value if and only if
\begin{equation}
\frac{1}{\epsinf}>\frac{1}{\eps_1}+\frac{1}{\eps_2},
\label{eq:harmonic}
\end{equation}
that is, if and only if the harmonic mean of the two errors exceeds
$2\epsinf$. In the floor shares $f_i=\epsinf/\eps_i$ the test is $f_1+f_2<1$;
in the saturation coordinate $m_i=1-2\epsinf/\eps_i$ it is
\begin{equation}
m_1+m_2>0,
\label{eq:mmid}
\end{equation}
and where both candidates lie strictly above the floor it is
$\zeta_1+\zeta_2<0$.
\end{corollary}
The four forms are algebraically identical and agree on all $20\,206$ archived Pareto pairs.
Equation~\ref{eq:mmid} is the preferred form because $\zeta$ is undefined for the $9.2\%$ of
pairs in which the more accurate candidate lies at or below the fitted floor, whereas $m$ remains
well defined and selects the less expensive candidate.
\paragraph{The blind spot.} That $R$ depends on $\eps$ and $\epsinf$ alone is also a blind spot.
Scoring each candidate at its \emph{measured} resource instead, the two objectives differ
by exactly the candidate's resource excess; on the
panel they select the same candidate on only $40.7\%$ of cells, the accuracy-only choice being
usually close (median regret $0.029$ nats) but with a ninetieth percentile of $0.65$ nats, and the
gap is law misfit and nothing else. \emph{Resources are needed exactly to the extent that the
frontier disobeys the law, and both numbers should be reported} (\SM{Proofs: Law~II}).
\paragraph{Interacting resources.}
\label{sec:nested}
With $\eps=\epsinf+M_\theta(g)$ a power mean, the moment map is unchanged, the entropy nests, the
regret splits into ``how much to leave reducible'' plus ``how to divide it,'' and the canonical
point is $\eps^\star=(1+n^{1/\theta})\epsinf$ with worst case $\ln(1+n^{1/\theta})$, falling to one
bit in the bottleneck limit. On the curve of Section~\ref{sec:odefirst} an interaction changes the
spectrum, not the primitive; the statements, their proofs, and the factorial evidence are in the
technical report in the archive.
\section{Law III: One Channel Is an Arc of Length \texorpdfstring{$\pi$}{pi}}
\label{sec:lawIII}
Laws~I and~II say what to do at a point, not how much there is to decide. That question has an
absolute answer: $x\mapsto p_x$ is a curve, and a curve has a length. The Fisher--Rao length
\citep{rao1945,amari2000} between two operating points is
\begin{equation*}
L(x_1,x_2)=\int_{x_1}^{x_2}\sqrt{\kappa(u)}\,\mathrm{d}u
          =\int_{x_1}^{x_2}\sqrt{-S''(u)}\,\mathrm{d}u,
\end{equation*}
free of the floor because only $S''$ enters, and invariant under $t_i\mapsto t_i^{\beta_i}$ and
under a constant rescaling of the error, $\eps\mapsto C\eps$
\citep{salamon1983,crooks2007}---but \emph{not} under the change of scoring
grain $\eps\mapsto\eps^{\,t}$, which multiplies it by $\sqrt t$
(Section~\ref{sec:compose}). By Proposition~\ref{prop:anglearc} it
is a functional of the curve of Section~\ref{sec:odefirst} and $L\ge\pi$ is the axiom integrated
over the complete sweep.
\paragraph{The complete sweep and the admissible part of it.} For two atoms the integral of
$\mathrm{d}p_0/\sqrt{p_0(1-p_0)}$ over $p_0\in(0,1)$ is $\pi$ whatever the exponent, the floor, and
the scale. That range is the chart's \emph{complete} sweep, $x$ over all of $\R$; because $\eps=1$
is chance level the part a measurement can reach is $p_0\ge\epsinf$, whose arc is
$2\arccos\sqrt{\epsinf}<\pi$, rising to $\pi$ only as $\epsinf\to0$
(Proposition~\ref{prop:anglearc}). \emph{Every constant in this section is a constant of the
complete sweep, and is an upper bound on what a floored frontier delivers.} With that said, the
dichotomy is sharp: $L\equiv0$ for a pure power
law with no floor, $L=\pi$ for one channel and a floor, $L=\infty$ for a spectrum accumulating at
the origin; and for $N+1$ atoms with $N\ge1$, $\pi\le L\le N\pi$ with $L=\pi$ iff $N=1$ (\SM{Proofs: Law~III}).
Write $\varphi=2\arccos\sqrt{\hat p_0}\in(0,\pi)$ for the position on the arc, the $\varphi$ of
\eqref{eq:angle} with $\hat p_0=1/(1+\Neff)$ the local floor share, so that $\varphi=0$ is the floor
and $\varphi=\pi$ the pure power law; the mirror
$\pi-\varphi=2\arcsin\sqrt{\hat p_0}$ runs the other way and gives the same $\sin\varphi$. On the
one reducible channel of this section $\hat p_0$ \emph{is} the true share $\epsinf/\eps$, and the
scalar statements below are trigonometric functions of $\varphi$ in that case:
$m=-\cos\varphi$, $\zeta=-2\ln\tan(\varphi/2)$, $\sqrt\kappa=\tfrac\alpha2\sin\varphi$, and
\begin{equation}
\boxed{\ R=-\ln\sin\varphi\ },
\qquad
e^{-R_1}=\sin\varphi=2\sqrt{p_0p_1}=\frac{\GM}{\AM},
\label{eq:lnsin}
\end{equation}
with the canonical point at $\varphi=\pi/2$ (\SM{Proofs: Law~III}); for one reducible channel
$R_1=R=\KL(u\Vert p)$, and \eqref{eq:lnsin} is a statement about that case. On the two-atom chart $-S''$
is maximal exactly at $\varphi=\pi/2$, so there the ``knee'' coincides with the minimax rule rather
than approximating it; off that chart Theorem~\ref{thm:monofloor}(iii) orders the two, and on the
cure chart $-S''$ is strictly decreasing and there is no knee at all.
\subsection*{The Regret Law, and How Many Operating Points a Frontier Needs}
Write $\theta=\varphi-\pi/2$, so $R=-\ln\cos\theta$. With respect to arc length---the Jeffreys measure of this family---the regret has a
parameter-free distribution.
\begin{proposition}[The regret law]
\label{prop:regretlaw}
Let the operating point be placed uniformly in arc length on the complete
sweep of a single-channel frontier. Then
\begin{equation}
\boxed{\ \Pr(R\le\varrho)=\frac{2}{\pi}\arccos e^{-\varrho}
       \;=\;\frac{2\sqrt2}{\pi}\sqrt{\varrho}\,\bigl(1+O(\varrho)\bigr)\ },
\label{eq:regretlaw}
\end{equation}
and
\begin{equation}
\mathbb{E}[R]=\ln2,\qquad \mathrm{med}(R)=\tfrac12\ln2.
\label{eq:onebit}
\end{equation}
For a spectrum with $N+1$ modes and the same measure on the share simplex---%
the Dirichlet$(\tfrac12,\dots,\tfrac12)$ law---%
\begin{equation}
\mathbb{E}[R]=\psi\Bigl(\tfrac{N+1}{2}\Bigr)-\psi\Bigl(\tfrac12\Bigr)-\ln(N+1)
\ \nearrow\ \gamma+\ln2=1.27036\ \text{nats},
\label{eq:meanregretN}
\end{equation}
equal to $\ln2$ at $N=1$ and to exactly one nat at $N=3$.
 Placed instead on the
admissible part alone, $\varphi$ uniform on $(0,\varphi_{\max})$ with
$\varphi_{\max}=2\arccos\sqrt{\epsinf}$, the law is
\[
\Pr(R\le\varrho)=\frac{\bigl\{\min(\varphi_{\max},\,\pi-a)-a\bigr\}^{+}}{\varphi_{\max}},
\qquad a=\arcsin e^{-\varrho},
\]
\[
\mathbb{E}[R]=-\frac{1}{\varphi_{\max}}\int_{0}^{\varphi_{\max}}\!\!\ln\sin\varphi\,\mathrm{d}\varphi,
\]
both proved in \SM{Proofs: Law~III},
in closed form $\mathbb{E}[R]=\ln2+\mathrm{Cl}_{2}(2\varphi_{\max})/(2\varphi_{\max})$
with $\mathrm{Cl}_2$ the Clausen function. On $\epsinf\le\tfrac12$ that mean never exceeds $\ln2$:
it equals $\ln2$ at both ends of that interval, as $\epsinf\to0$ and at $\epsinf=\tfrac12$, and dips
to a minimum of $0.4945$ nats at $\epsinf=0.1037$. Above $\tfrac12$ it rises
without bound---$0.79$ nats at $\epsinf=0.6$ and $2.61$ at $0.99$---because a
frontier whose floor is most of its chance-level error has only a short arc, all
of it far from the right angle. Thus, \eqref{eq:onebit} is the value of the
complete sweep and the ceiling of the admissible case \emph{for any frontier
whose floor is below half its chance-level error}---which by
Theorem~\ref{thm:budget}(iv) holds, more than comfortably, for every
compositional frontier that \emph{has} a canonical point, those needing
$\epsinf\le1/e=0.368$.
\end{proposition}

\begin{figure}[t]\centering
\includegraphics[width=\linewidth]{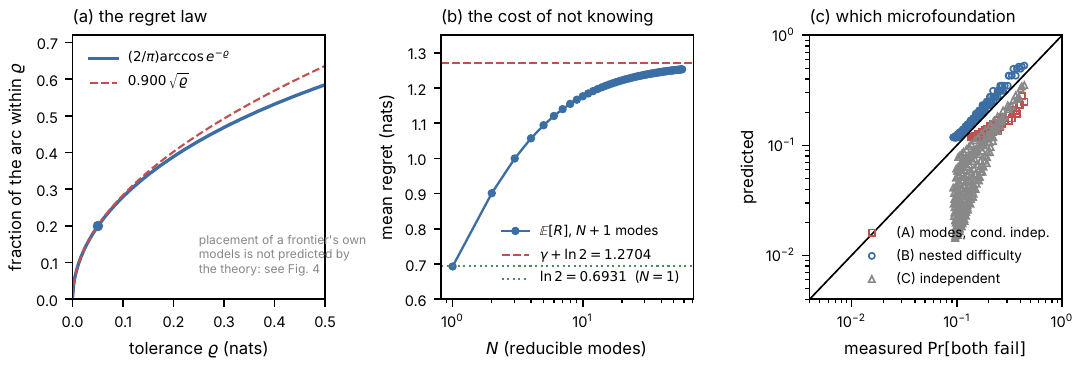}
\caption{(a) The regret law \eqref{eq:regretlaw}: the fraction of a complete
frontier within $\varrho$ nats of the canonical point, with the
$0.900\sqrt\varrho$ form, and (inset) the $35$ panel cells' observed in-band
fractions against the prediction conditional on each cell's own arc. (b) The
mean regret \eqref{eq:meanregretN} of an operating point placed uniformly in
arc length: exactly one bit for a single reducible channel, rising to
$\gamma+\ln2$ and never beyond. \emph{This is the cost of not knowing where to
operate, and it is bounded.} (c) The microfoundation comparison: measured pairwise
co-failure rates for the three mechanisms on the per-item panel of
\TR{What the item level says}.}
\label{fig:regretmicro}
\end{figure}

Three readings. The square-root form says why the optimum is broad: the $\varrho$-optimal fraction
of a frontier grows as the \emph{square root} of the tolerance, with the universal constant
$2\sqrt2/\pi=0.900$. Equation~\ref{eq:onebit} says a model placed at random on the complete sweep
of a single-channel frontier is on average exactly one bit of log-error from the canonical point, and
\eqref{eq:meanregretN} bounds the same quantity by $1.83$ bits on every frontier that exists:
\emph{the entire value of knowing where to operate is one universal interval, of at most one bit
for a single channel and at most $1.83$ bits for any spectrum.}
\begin{theorem}[Covering law]
\label{thm:covering}
Let the realized price range over the frontier, so that the arc position of
the optimum ranges over an interval of length $L_{\mathrm{rng}}\le\pi$, and let a
checkpoint at $\hat\varphi$ cost $\KL(p_\varphi\Vert p_{\hat\varphi})$ when
the optimum is at $\varphi$. Then the number of published operating points
needed so that some checkpoint is within $\varrho$ nats of optimal for every
realized price is
\begin{equation}
K\;\simeq\;\frac{L_{\mathrm{rng}}}{2\sqrt{2\varrho}},
\label{eq:covering}
\end{equation}
The optimal checkpoints are equally spaced in $\varphi$, not in log-resource. However, as
Section~\ref{sec:scope} reports, the plug-in arc is too noisy on the published windows for this
difference to be measured. For $L_{\mathrm{rng}}=\pi$, the exact covering numbers are given in
\SM{Proofs: Law~III}. Since $L_{\mathrm{rng}}\le\pi$ for a single channel, with
$L_{\mathrm{rng}}=2\arccos\sqrt{\epsinf}$ on the admissible part of a two-atom frontier,
$K$ at $L_{\mathrm{rng}}=\pi$ is an upper bound: \emph{five operating points cover any
single-channel frontier to $0.05$ nats}, whatever its exponent, floor,
resource or units.
\end{theorem}
\noindent\emph{Where the tolerance comes from.} On a log-resource axis the natural unit is a factor
of two, and one doubling of $\psi=(\eps-\epsinf)/\epsinf$ costs, by
\eqref{eq:lncosh},
\begin{equation}
\varrho_{2}=\ln\cosh\bigl(\tfrac12\ln2\bigr)=0.05889\ \text{nats},
\qquad
K=\frac{\pi}{2\sqrt{2\varrho_2}}=4.577,
\label{eq:onedoubling}
\end{equation}
so the $0.05$-nat tolerance requires approximately five checkpoints per resource doubling; the
count follows from the choice of a twofold resource unit.
Equation~\ref{eq:covering} is the second-order expansion of a Fisher--Rao metric
\citep{salamon1983} (\SM{Proofs: Law~III}).
The measured arcs are short. Across the $18$ archived ImageNet single-resource frontiers, the
median arc returned by a three-accuracy chord is $0.31\pi$, while the median frontier contains
$28.5$ Pareto points. For the $12$ frontiers on which the chord returns a positive arc, the median
is $0.39\pi$; the remaining six yield no measurable arc. The resulting coverage statistics are:
\begin{quote}
\emph{The median frontier carries pricing information equivalent to $1.74$ checkpoints when the
six no-arc cases count as zero, or $2.14$ checkpoints among the $12$ frontiers with a measurable
arc. The corresponding redundancy is $13\times$, and $13.8$ consecutive models fit within a
$0.05$-nat indifference band.}
\end{quote}
\noindent Those are the vision panel's numbers, not the whole corpus's; the corpus-wide arc, read
using plug-in curvature rather than a chord, is reported in
Section~\ref{sec:scope} and is longer.
\subsection*{Why $\pi$, and Why Laws~II and~III Are One Law}
\begin{figure}[tbp]\centering
\includegraphics[width=0.82\linewidth]{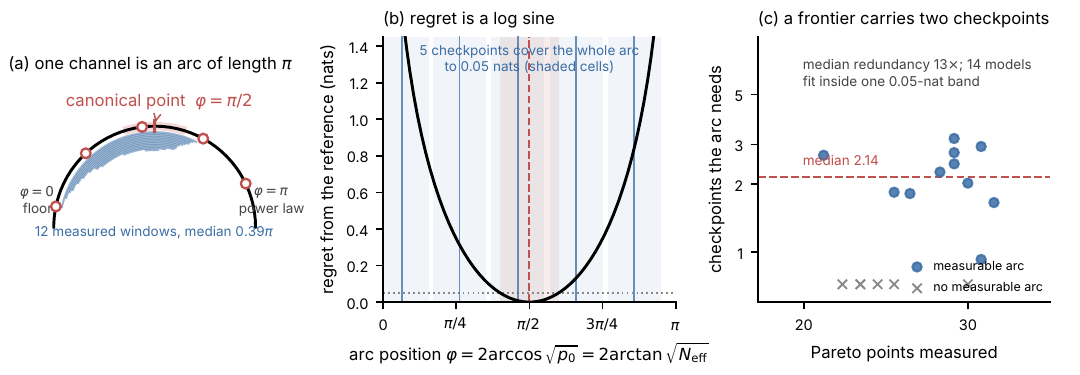}
\caption{(a) The complete sweep of a single-channel frontier is the same arc of
Fisher--Rao length $\pi$, from the pure power law at $\varphi=\pi$ to the floor
at $\varphi=0$; a frontier with a floor reaches only $2\arccos\sqrt{\epsinf}$
of it at or below chance level. The shaded segment is the $0.05$-nat band, the
open circles are the five checkpoints of Theorem~\ref{thm:covering}, and the
inner curves are the $12$ of the $18$ archived ImageNet single-resource windows
that have a measurable chord arc, median $0.39\pi$, drawn on the same absolute
scale and centered on the right angle so that their \emph{lengths} can be
compared---the chord returns a length and not a position, and the positions the
jet does return sit deep, at a median $\Neff=0.465$ and hence
$\varphi=1.20<\pi/2$. (b) Regret from the reference is minus
the log sine of arc position, \eqref{eq:lnsin}; the shaded cells are the five
checkpoints' covering neighborhoods. (c) Checkpoints the arc needs against
Pareto points measured, one point per frontier; crosses are the six frontiers
whose log-convexity defect is non-positive, on which the chord returns no arc
and the law fails.}
\label{fig:arc}
\end{figure}

Section~\ref{sec:odefirst} derives $\pi$ as a Lipschitz bound; a shorter argument identifies the
constants.
\begin{proposition}[The binary reading]
\label{prop:binary}
With one reducible channel the share vector is the two-point distribution
$p=(p_0,1-p_0)$ with
\[
p_0=\frac{\epsinf}{\eps(x)}
  =\Pr\bigl[\text{an item still being failed at }x\text{ is irreducible}\bigr],
\]
$2\arcsin\sqrt{p_0}=\pi-\varphi$ is its Fisher--Rao arc-length coordinate
measured from the pure power law, so the $\varphi$ of \eqref{eq:angle} is
$2\arccos\sqrt{p_0}$, measured from the floor; and
\begin{equation}
e^{-R}=\sin\varphi
      =\bigl\langle\sqrt{p},\sqrt{p^{\,\mathrm{swap}}}\bigr\rangle
      =\cos\Bigl(\tfrac12 d_{\mathrm{FR}}(p,p^{\,\mathrm{swap}})\Bigr),
\label{eq:binaryR}
\end{equation}
the Hellinger affinity between the share vector and its own mirror image, the
same in either orientation.
\end{proposition}
Thus, $L=\pi$ is the \emph{Fisher--Rao diameter of the binary simplex}: the complete sweep of a
single-channel frontier is a full geodesic corresponding to one binary decision. A frontier with a
floor traverses the fraction $\arccos\sqrt{\epsinf}/(\pi/2)$ of this diameter at or below
chance level. The remaining constants in this section therefore have the standard geometric
interpretation associated with one bit. The canonical operating point has a direct interpretation:
\emph{operate where the reducible and irreducible components of the remaining error are equal}. Laws~II and~III are therefore not two laws but the two canonical geometries of one
object, the share path $x\mapsto p_x$: its Bregman geometry gives the price, and its Fisher--Rao
geometry gives the arc, the covering, and the regret law.
Figure~\ref{fig:arc} draws that arc with the covering checkpoints of
Theorem~\ref{thm:covering} on it, and places the archived ImageNet windows on
the same absolute scale, where their median length is a fraction of the
complete sweep.
\section{Reading a Frontier Without Fitting It}
\label{sec:lawfree}\label{sec:onedim}
Two quantities follow from $(s,\kappa)$ that need no fitted law. Because the floor is the atom at
$a=0$ it enters neither moment, so $\hat\alpha=\mathbb{E}[a^2]/\mathbb{E}[a]=s+\kappa/s$ is
identified without estimating $\epsinf$; and $g=\kappa/s^{2}$ recovers the floor pointwise on a
two-atom window, $\epsinf=\eps\,\kappa/(s^2+\kappa)$
(\SM{Proofs: Reading a frontier without fitting it}). For two atoms the width is the collapse
coordinate itself,
\begin{equation}
\zeta=\ln g=\ln(\kappa/s^2),
\qquad
\zeta'=\alpha,
\qquad
\zeta=0\iff\epsinf=\tfrac12\eps,
\label{eq:zeta}
\end{equation}
so one straight-line regression of $\zeta$ on $x$ returns the branch of the dichotomy, the
exponent, and---at its zero crossing---the floor; and two points suffice for the exponent even when
the window never reaches that crossing, since
\begin{equation}
\alpha=\frac{2\,\Delta s}{\Delta m}
\label{eq:alphatwopoint}
\end{equation}
exactly, for any two points of a two-atom frontier, any floor, and any scale.
\subsection*{What Is a Property of the Frontier and What Is a Choice of Clock}
A resource axis is defined only up to its unit and its origin. The error is not like that: $\eps=1$
is the chance level, so $S$ is absolute, in nats. Under $x\mapsto\lambda x+c$ one has $S\mapsto S$,
$s\mapsto s/\lambda$, $\kappa\mapsto\kappa/\lambda^2$, so the invariant local data of a frontier is
the jet $(S;w,w_2,\dots)$ with $w=\kappa/s^2$ and $w_2=\mathrm{d}\ln\kappa/\mathrm{d}S$; and
\emph{$\alpha$ is the unit of the axis and is not invariant, so it cannot appear in any invariant
statement and no derivation of it from \eqref{eq:axiom} can exist}
(\SM{Proofs: The gauge of a resource axis, and the invariant jet}). This turns a known non-identifiability result into a theorem: a
gauge-invariant axiom system cannot determine the unit. It also says where the one remaining number can come
from, namely an axis with no gauge freedom. A count has none, and
Section~\ref{sec:count} derives its price, $-\ln(1-q_0)$ nats per attempt, from the
chance level alone.
The complete set of axis changes preserving
\eqref{eq:axiom} is the class of \emph{Bernstein} functions: if $\eps$ satisfies \eqref{eq:axiom} so does
$\eps\circ\chi$ for Bernstein $\chi$, and the new spectrum is the law of a subordinator evaluated at
the old---\emph{a change of clock is a subordination of the difficulty spectrum}. They compose but do
not invert, so the gauge is a \emph{semigroup} and the axiom is a joint property of a frontier
\emph{and} its clock (\SM{Proofs: The gauge of a resource axis, and the invariant jet}). Complete monotonicity is a theorem
on a count axis (Proposition~\ref{prop:countaxis}), so if a resource buys $m(x)$ effective
exchangeable attempts and $m$ is Bernstein then the axiom holds on that resource axis too:
Thus, the assumption imposed by the theory on a compute axis becomes a corresponding condition on
the resource map: diminishing returns at every order, a measurable property of the curve. The parameter
$\beta$ is the \emph{index at the origin of the clock's L\'evy measure} \eqref{eq:dict}. Because
the clock is a $\beta$-stable subordinator by Proposition~\ref{prop:compose}(b), the same index
also governs the resulting difficulty spectrum. Thus, $\beta$ is a property of the subdivision
rather than a scale, which helps explain why it is weakly identified and corpus-specific.
\subsection*{The Invariant Is an Inverse Channel Count, and That Is the Rule}
\begin{proposition}[Effective channels]
\label{prop:neff}
Write $N_{\mathrm{eff}}:=1/w=s^{2}/\kappa$. Then
\begin{enumerate}[label=(\roman*),leftmargin=2em,itemsep=1pt]
\item for a $\Gamma(k)$ spectrum---the difficulty of a task built from $k$
      independent exponential subtasks---$N_{\mathrm{eff}}=k$ at every point,
      exactly; in general it is the effective number of independent difficulty
      channels still active. Write $\hat p_0:=1/(1+N_{\mathrm{eff}})$ for the
      \emph{local} floor share, the share the two-atom chart tangent to the
      frontier at that point would carry. It equals the true floor share
      $p_0=\epsinf/\eps$ exactly on the two-atom chart and nowhere else up to
      gauge (Corollary~\ref{cor:wallivp}), and $\hat p_0\ge p_0$ always, the
      axiom's wall being precisely that statement. The arc element of
      \eqref{eq:angle} is $\mathrm{d}L=|\mathrm{d}\Delta
      S|/\sqrt{N_{\mathrm{eff}}}=|\mathrm{d}S|/\sqrt{N_{\mathrm{eff}}}$, a
      length and so unsigned, and the
      angle advances by $\mathrm{d}\varphi=\dot
      N_{\mathrm{eff}}\,\mathrm{d}L/(N_{\mathrm{eff}}+1)$, which is
      $\mathrm{d}L$ exactly on the two-atom chart;
\item \eqref{eq:lncosh} becomes
      \begin{equation}
      \boxed{\ R_1=\ln\cosh\bigl(\tfrac12\ln N_{\mathrm{eff}}\bigr)\ }
      \label{eq:logcosh}
      \end{equation}
      --- the same identity read as a channel count, since $\psi=1/w$ on the
      two-atom chart. Here $R_1$ is the regret of the two-atom chart tangent to
      the frontier at that point. It equals the spectrum's own
      $R=\KL(u\Vert p)$ exactly when the spectrum \emph{is} two atoms, and
      otherwise differs from it in \emph{either} direction and by an unbounded
      amount---on the three-atom spectrum
      $0.05\delta_0+0.6\delta_{0.4}+0.35\delta_{1.3}$ at $\Neff=1.92$ it reads
      $0.05$ nats against $R=0.38$, and over random spectra the gap $R-R_1$ takes
      both signs. The operating rule below minimizes $R_1$ rather than $R$, because $R_1$ is a
      function of the curve alone. The rule is therefore a gauge-free surrogate for, rather than the same object as, Law~II's
      minimax point. They coincide when $N=1$, the one-reducible-channel case of
      Theorem~\ref{thm:canonical}. The surrogate depends only on $|\ln N_{\mathrm{eff}}|$, so
      the operating rule is to \emph{operate where $|\ln N_{\mathrm{eff}}|$ is smallest}. This
      corresponds to $\kappa=s^{2}$ wherever $w=1$ is attained and remains well posed when it is
      not; the tolerance in \eqref{eq:onedoubling} is one doubling of $N_{\mathrm{eff}}$;
\item the floorless chart conserves $S\,w$, so
      $N_{\mathrm{eff}}=S/S^{\star}$ with $S^{\star}=(1-\beta)/\beta$:
      \emph{one new effective channel per $S^{\star}$ nats of log-error}, and
      $S^{\star}$---the depth at which the frontier carries one effective
      channel, which is its canonical point and not a knee, that chart's
      curvature being strictly decreasing---is the reportable form of the shape
      index;
\item \textup{(Karamata.)} If $\mu[0,a]\sim Ca^{\theta}$ then
      $\eps\sim C\Gamma(\theta+1)x^{-\theta}$ and
      $N_{\mathrm{eff}}\to\theta$. Every frontier obeying \eqref{eq:axiom}
      therefore ends in exactly one of three ways: an atom at the origin
      ($N_{\mathrm{eff}}\to0$, a floor), an edge index $\theta$
      ($N_{\mathrm{eff}}\to\theta$), or an isolated slowest mode
      ($N_{\mathrm{eff}}\to\infty$, a pure power law). The scaling exponent of
      a power-law frontier \emph{is} an effective channel count, and the floor
      question is the $\theta=0$ case of an edge-exponent question.
\end{enumerate}
\end{proposition}
Part (ii) gives the operating rule recommended in this paper. It is the first formulation that
requires no floor estimate, fitted model, resource units, or resource origin. The rule is also
incomplete without the grain specified below.
\begin{theorem}[No gauge-invariant operating rule]
\label{thm:noinvariant}
Every local gauge invariant is a horizon ratio $D_k$, and on a constant-$\sigma$
chart every $D_k$ is \emph{constant along the frontier}---equal to $1/(k-\beta)$
on the power members and to $0$ on the cure chart, which saturates every rung
(\SMchart). Consequently, no condition on invariants alone can select an
operating point there. An operating rule is therefore necessarily
gauge-dependent: it must take one input from outside the frontier.
\end{theorem}
Equation~\ref{eq:oneline} identifies the required external input: the scoring grain $t$.
The effective channel count $\Neff$ is extensive, transforming as $\Neff\mapsto t\Neff$
under $\eps\mapsto\eps^{t}$. Here, $t$ is the \emph{grain}: the number of independently drawn
items that must all fail for the delivered unit to fail. It is not a retry budget, which acts on the
same item and is worth less because of the discount described in Section~\ref{sec:setup}.
\begin{proposition}[The rule with its input]
\label{prop:retry}
Let $t$ be the grain of the delivered unit. Then
\begin{equation}
\boxed{\ \text{operate where }N_{\mathrm{eff}}=1/t\ },
\quad\text{equivalently}\quad
\boxed{\ 1/N_{\mathrm{eff}}\ \text{is the grain the model is sized for}\ }
\label{eq:retry}
\end{equation}
Part (ii) above is the $t=1$ case. The canonical resource also moves with the
grain at a rate that is gauge-invariant:
\begin{equation*}
\frac{\mathrm{d}x^{\star}}{\mathrm{d}\ln t}=-\frac{h_1}{\sigma}
\ =\ h_0=\frac{\Delta S}{s}\quad\text{on a floored similarity chart.}
\end{equation*}
This rate is exact and integrable on the cure chart, where
$x^{\star}(t)-x^{\star}=h_0\ln t$.
\end{proposition}
Measured with the archive's own jet on the $104$ of the $117$ frontiers long enough to carry one, the median $\Neff$
over a window is $0.465$, so the median published frontier is sized for grain $t=2.2$ with
interquartile range $[1.0,7.3]$, and within a frontier the implied grain rises from about $1$ to
about $11$: \emph{a frontier is a sequence of models each canonical at a coarser criterion.} If only
$t\in[t_1,t_2]$ is known, the minimax point is $\Neff=1/\sqrt{t_1t_2}$ at exact worst-case cost
$\ln\cosh\{\tfrac14\ln(t_2/t_1)\}$: the archive's own interquartile range of implied grains costs
$0.119$ nats at $\Neff=1/\sqrt{7.3}=0.370$, more than twice the band inside which the
recommendation is constant (\SM{Proofs: The tests without a bandwidth: the grid lemma, the floor bracket, the operating depths, and the grain}). \emph{The grain has to be
declared; defaulting it to one is itself a priced decision.}
A separate qualification concerns reading $\Neff$ from a compute axis. With
$s_m=-\mathrm{d}\ln m'/\mathrm{d}x$ the clock's own price, $1/\Neff^{x}=1/\Neff^{k}+s_m/s$
identically, so $\Neff^{x}\le\Neff^{k}$ with equality iff the axis \emph{is} a count
(\SM{Proofs: Reading a frontier without fitting it}). A measured compute-axis channel count is
therefore a \emph{lower bound} on the task's own, and, where the compute-axis rule applies, it
\emph{undersizes} relative to the task-level criterion by $s_m/s$ in inverse units.
\paragraph{The identity is exact; its plug-in estimator is not.} Every plug-in reading of
\eqref{eq:zeta} needs $s$ and $\kappa$, which carry a bandwidth. \emph{A statistic must be evaluated against a null simulated on the same design, rather than
against its analytic value.} This paper evaluates the rule using eight statistics (\TR{Two warnings about fitting the shape index}). \emph{Where a choice
exists, prefer the statistic with no bandwidth in it.} A correction for reducible modes sharing a
single resource direction---by the rank evidence of Section~\ref{sec:lawI} the normal case---is in
\SMonedim.
\section{The Price Is a Frontier}
\label{sec:compose}
The theory so far has two statements of unequal standing: an axiom about $\eps$, and an assumption
about the map that produces attempts. They are the same statement.
\begin{proposition}[The composition law]
\label{prop:compose}
For a frontier along one direction the following are equivalent.
\begin{enumerate}[label=(\alph*),leftmargin=2em,itemsep=1pt]
\item \textup{(Composition.)} The difficulty spectrum $\mu$ is infinitely
      divisible: an item's log-difficulty is a sum of independent
      contributions, in arbitrarily fine subdivision.
\item \textup{(Clock.)} $S=-\ln\eps$ is a Bernstein function, so $\eps=e^{-x}$
      in some clock: the orbit of the pure exponential under the semigroup of
      Section~\ref{sec:lawfree} is \emph{exactly} the infinitely divisible
      frontiers.
\item \textup{(Stages.)} $\eps^{t}$ is completely monotone for every $t>0$.
\item \textup{(Self-reference.)} The price curve $s=-\mathrm{d}\ln\eps/
      \mathrm{d}x$ is itself a completely monotone frontier.
\item \textup{(Depth.)} The remaining depth $\Delta S=\ln(\eps/\epsinf)$ is
      itself completely monotone---\emph{the axiom, applied to the budget
      instead of to the error} (for $\epsinf=0$ the statement is (d), the
      zeroth cumulant being infinite).
\end{enumerate}
Any of them implies \eqref{eq:axiom}, with $\mu$ the law of $\sigma_1$ for the
subordinator whose Laplace exponent is $S$; the converse fails.
\end{proposition}
Form (d) is the most direct statement: \emph{diminishing returns applies not only to the error
but also to the price of error.} It contains the axiom, because $e^{-S}$ is completely monotone whenever $S$ is
Bernstein, so the theory has one hypothesis where it had two and the weaker is now a corollary.
Form (b) settles the chart question: the stretched-exponential chart \emph{is} the orbit of $e^{-u}$
and the two-atom chart \eqref{eq:M2} lies outside it, because no non-degenerate compactly supported
law is infinitely divisible \citep{sato1999}. \emph{Model selection between the two is therefore not
a choice of clock but the physical question of whether difficulty is divisible}, and divisibility is
preserved by every clock.
Writing the clock in L\'evy--Khintchine form, which (b) guarantees exists,
\begin{equation}
S(x)=b\,x+\int_0^{\infty}\bigl(1-e^{-cx}\bigr)\,\nu(\mathrm{d}c),
\qquad
\nu_x(\mathrm{d}c):=e^{-cx}\nu(\mathrm{d}c),
\qquad
m_k:=\int c^k\,\nu_x(\mathrm{d}c),
\label{eq:levy}
\end{equation}
and comparing with $S^{(k)}=(-1)^{k+1}c_k$ gives the dictionary on which everything below runs:
\begin{equation}
\boxed{\ c_k=m_k(\nu_x)\quad(k\ge2),
\qquad s=b+m_1(\nu_x),
\qquad \Delta S=m_0(\nu_x)\ }
\label{eq:dict}
\end{equation}
the last for $b=0$. A floor exists iff the clock is compound Poisson, and then
\begin{equation}
\begin{gathered}
\epsinf=e^{-\nu(\text{total})}=\Pr[\text{the clock never ticks}],\\
-\ln\epsinf=\mathbb{E}[\text{improvements the resource can ever buy}],
\end{gathered}
\label{eq:poissonzero}
\end{equation}
so \emph{depth in nats is a count}. The three numbers fitted per axis are three functionals of one
positive measure: $\nu$'s scale is $\alpha$, its index at the origin is $\beta$, its mass is
$-\ln\epsinf$.
Write the frontier as the composition assumed by the clock argument,
$x\xrightarrow{m}\text{attempts}\xrightarrow{P}\eps$. Complete monotonicity is required at each
arrow, and the assumption made at both stages is closure under \emph{refinement}
(\TR{The Attempt Chain}). The chain gives the floor a finite-count interpretation: the error at
unbounded resource equals the error after $\Lambda$ attempts,
\begin{equation}
\boxed{\ \epsinf=P(\Lambda)\ },
\qquad\text{hence}\qquad
\boxed{\ P(\lceil\Lambda\rceil)\ \le\ \epsinf\ \le\ P(\lfloor\Lambda\rfloor)\ }
\label{eq:attemptfloor}
\end{equation}
so a floor exists \emph{iff} the attempt budget saturates. \emph{On a count axis with a finite
budget there is no irreducible error; there is a finite number of tries}---and the atom at $a=0$ of
\eqref{eq:spectral} is what that looks like once the budget has saturated, not a rival account of
it. In practice, \eqref{eq:attemptfloor} requires exchangeability and a
finite budget and \emph{not} the composition law, so $P$ can be measured at integer values using any pass-at-$k$ curve. The floor is then
bracketed by two measured values, to within one attempt's price---$0.20$ and $0.094$ nats on the
two panel families---rather than being extrapolated to $x=\infty$.
\subsection*{One Cauchy--Schwarz, One Level Up}
\begin{theorem}[The channel--budget inequality]
\label{thm:budget}
Let $\eps$ satisfy the composition law. Then $m_1^2\le m_0m_2$, which is to say
\begin{equation}
\boxed{\ N_{\mathrm{eff}}\ \le\ \Delta S=\ln\!\bigl(\eps/\epsinf\bigr)\ }
\qquad\Longleftrightarrow\qquad
\boxed{\ \epsinf\ \le\ \eps\,e^{-N_{\mathrm{eff}}}\ }
\label{eq:budget}
\end{equation}
with equality exactly when $\nu_x$ is a single atom. \emph{No frontier carries
more effective channels than it has nats remaining.} Consequently
\begin{enumerate}[label=(\roman*),leftmargin=2em,itemsep=1pt]
\item the bound is tighter than Theorem~\ref{thm:monofloor}(i) wherever
      $N_{\mathrm{eff}}>0$, by the ratio $e^{N}/(1+N)$---$1.4\times$ at
      $N_{\mathrm{eff}}=1$, $25\times$ at $5$, $2\,000\times$ at $10$---and the
      ratio tends to $1$ as $N_{\mathrm{eff}}\to0$, which is the
      tangency of Corollary~\ref{cor:wallivp}(ii): \emph{approaching the floor
      the two walls coincide and the gain vanishes};
\item it is again the infimum of a measured monotone sequence, because
      $e^{-G}$ with $G$ of \eqref{eq:slope} is non-increasing, so every window
      returns a valid bound and the bound improves by construction;
\item the canonical point satisfies $\Delta S^{\star}\ge N_{\mathrm{eff}}=1$,
      that is, $\eps^{\star}\ge e\,\epsinf$: \emph{never operate within one nat of
      the floor}. The rule $\eps=2\epsinf$ of \eqref{eq:universal} is therefore
      unavailable to a compositional frontier, by $\ln(e/2)=0.307$ nats and in a
      definite direction---not a contradiction but a separation of classes,
      since $\eps=2\epsinf$ is the canonical point of the two-atom chart, which
      is not compositional, while $\eps^{\star}=e\,\epsinf$ is the canonical
      point of the cure chart, which is compositional. The invariant rule $\Neff=1$ is the
      same in both; only its reading as a floor multiple differs, and that
      reading is chart-bound;
\item a compositional frontier has a canonical point only if $S_{\infty}\ge1$,
      that is, only if $\epsinf\le1/e=0.368$---a separation threshold that needs no
      spectral ratio.
\end{enumerate}
\end{theorem}
Without the composition law the axiom gives only $\Neff\le e^{\Delta S}-1$, tight on the two-atom
chart. The pair states the additional content of the composition hypothesis directly: it tightens the
axiom's exponential bound, $\Neff\le e^{\Delta S}-1$, to the linear bound,
$\Neff\le\Delta S$. Both are read at a \emph{depth}, not at a resource span, so the comparison
depends on where on the frontier it is taken. Over the whole frontier, at the median fitted total
depth $-\ln\epsinf=2.78$ nats on the $115$ frontiers the cure chart fits, the two walls give $15.1$
channels against $2.78$, a factor of $5.4$. At a window's deep edge, where the median depth still
available is $0.056$ nats on the $91$ frontiers that report one, they give $0.057$ against $0.056$,
a factor of $1.03$. \emph{The hypothesis tightens the bound by a factor of five at the deep end of the
frontier, but the published windows do not provide power to distinguish the two bounds.} This
empirical pattern is the numerical counterpart of Corollary~\ref{cor:wallivp}(iii). One rung up, Cauchy--Schwarz on $\nu_x$ bounds the channel slope
$\sigma:=\mathrm{d}\Neff/\mathrm{d}S$ by
\begin{equation}
\boxed{\ \sigma\ \ge\ -1\ }
\qquad\Longleftrightarrow\qquad
\boxed{\ G:=\Neff+S\ \text{ is non-decreasing}\ }
\label{eq:slope}
\end{equation}
where the axiom alone gives only $\sigma\ge-(1+\Neff)$, tight on the two-atom chart: \emph{a
compositional frontier loses at most one effective channel per nat of log-error}
(\SM{Proofs: The composition law: the dictionary and the second Cauchy--Schwarz}).
\subsection*{The Chart Missing from the Dichotomy}
Proposition~\ref{prop:neff}(iv) separates the floor question from the chart question, while
Proposition~\ref{prop:compose} identifies the relevant axis. Together, they leave one cell empty
in the $2\times2$ classification: frontiers that are divisible \emph{and} have a floor. The cure
chart fills this cell and yields six exact identities without additional fitting.
\begin{proposition}[The cure chart]
\label{prop:cure}
The frontier
\begin{equation}
\eps(x)=\exp\bigl\{-\lambda\bigl(1-e^{-c(x-x_0)}\bigr)\bigr\},
\qquad \epsinf=e^{-\lambda},
\label{eq:cure}
\end{equation}
is the unique chart with $\kappa/s$ constant. It is compound Poisson with one
jump size, hence compositional, and it is the promotion-time cure model of
survival analysis \citep{yakovlev1996,chen1999}, whose cure fraction is
$\epsinf$. With the same three parameters used for the competing charts, the cure chart satisfies
the following identities exactly, without additional fitting:
\begin{equation}
N_{\mathrm{eff}}=\Delta S,\quad
\sigma=-1,\quad
G\equiv-\ln\epsinf,\quad
\eps^{\star}=e\,\epsinf,\quad
L_{\mathrm{adm}}=2\sqrt{-\ln\epsinf},\quad
\frac{\mathrm{d}w}{\mathrm{d}S}=w^{2},
\label{eq:cureids}
\end{equation}
and it saturates both rungs of the depth ladder, $D_1=D_2=0$ at every point,
being the extremal of the hypothesis.
\end{proposition}
\subsection*{What Is Testable, and at What Precision}
The hypothesis imposes a sharper shape restriction: \emph{the axiom requires the price to
decrease, whereas the hypothesis requires it to decrease convexly}. This restriction follows from
the equivalence $\Delta S$ completely monotone $\iff S$ is Bernstein $\iff s$ is completely
monotone; its first differential condition is $s''=c_3\ge0$. Table~\ref{tab:tests} summarizes the resulting tests. Each row is proved in
\SM{Proofs: The composition law: the dictionary and the second Cauchy--Schwarz} or
\SM{Proofs: The tests without a bandwidth: the grid lemma, the floor bracket, the operating depths, and the grain}; all of the tests are one-sided.
\begin{table}[tbp]\centering
\small
\begin{tabular}{llll}
\toprule
test & costs & order & where it can refute\\
\midrule
bounded non-degenerate spectrum & a certificate on $a_{\max}$ & --- & anywhere\\
curvature rise ($c_3<0$) & a jet, or 4 accuracies & 3 & left of the elbow only\\
divided difference \eqref{eq:dd3} & 4 accuracies, no spacing & 3 & left of the elbow only\\
sharpened rung \eqref{eq:d1hat} & the same 4 accuracies & 3 & where $s,\kappa>0$\\
wall gap $\Neff\le\Delta S$ & 3 accuracies $+$ a floor & 2 & shallow edge\\
floor bracket $\epsinf^{W}\le\epsinf^{A}$ & 3 accuracies & 2 & at or left of the elbow\\
sampled-depth minor, rung 2 & 5 accuracies & 2 & shallow edge\\
count-axis four-point test & $P(1\ldots4)$, exact & --- & any census\\
count-axis rung 2, offset $0$ & $P(1\ldots5)$, exact & --- & any census\\
bundle rescoring & rescore at grain $k$ & --- & any frontier\\
\bottomrule
\end{tabular}
\caption{What refutes the divisibility hypothesis, what it costs, and where it can be read. ``Order''
is the lowest derivative of $S$ the test involves. In the derivatives of $S$ \emph{alone} the axiom
and the hypothesis coincide at every order $\le2$, so without a floor the discriminating content
begins at order three; the order-$2$ rows buy their discrimination with a floor instead, which is
the second of the two currencies named below.}
\label{tab:tests}
\end{table}
Four implications of the table merit emphasis. \emph{The elbow determines where the test applies.}
A positive,
strictly decreasing, convex price on the whole axis diverges at $-\infty$, so the composition law
forces $a_{\max}=\infty$: \emph{a bounded non-degenerate spectrum is never divisible}. (The
degenerate case is an exception: a single atom $\delta_a$ gives the pure
exponential, whose price is the constant $a$, and that frontier \emph{is} divisible.) The spectral width is the only
non-vanishing charge of the resource flow, so $c_3$ must change sign and the frontier is refutable
exactly to the left of its elbow. This gives the admissibility condition observed in the archive
without requiring a chart or a floor.
\emph{The empirical gain is expressed by the margin between the bounds.} The wall-gap width $\Neff-\ln(1+\Neff)$ equals minus the log of the
tightening factor $(1+\Neff)e^{-\Neff}\le1$ reported for \eqref{eq:budget}, so the median tightening
factor of $0.9988$ measured at the deep edge of the archive's windows
\emph{is} a width of $0.0012$ nats: the gain was measured where the test has no power. The
width increases in $\Neff$, so the test is strongest at the shallow edge---the only side on which
refutation is possible---where the median width is $0.72$ nats, and three accuracies bound the floor
twice, once per hypothesis, against a count-axis bracket of $0.094$ and $0.20$ nats
(\SM{Proofs: The tests without a bandwidth: the grid lemma, the floor bracket, the operating depths, and the grain}).
\emph{The precision required to resolve this question also has a closed-form expression.} The
discriminating content is
reachable without estimating a third derivative, by a divided difference over any four models at
distinct log-resources,
\begin{equation}
\boxed{\
[x_0,x_1,x_2,x_3]\,S\ :=\ \sum_{i=0}^{3}
\frac{S(x_i)}{\prod_{j\ne i}(x_i-x_j)}\ \ge\ 0\ }
\label{eq:dd3}
\end{equation}
which equals $S'''(\xi)/6$ for some $\xi$ in the hull \citep{widder1941}; being linear in the
measured log-errors with weights $u_i=1/\prod_{j\ne i}(x_i-x_j)$, its power and the precision it
requires are
\begin{equation}
\text{power}=\Phi\Bigl\{\frac{\mathrm{dd}_3^{\mathrm{div}}
-\mathrm{dd}_3^{\mathrm{alt}}}{\sigma_\ell\|u\|_2}-z_{.95}\Bigr\},
\qquad
\sigma_\ell^\star=\frac{\mathrm{dd}_3^{\mathrm{div}}-\mathrm{dd}_3^{\mathrm{alt}}}
{\|u\|_2\,(z_{.95}+z_{1-\delta})},
\label{eq:powerreq}
\end{equation}
These expressions require no simulation. Here, $\sigma_\ell$ denotes the per-model log-error
noise, not the channel slope $\sigma$ in \eqref{eq:slope}; $\|u\|_2$ is the norm of the
weight vector, not the scalar invariant $w=\kappa/s^2$ or $w_2$; and $1-\delta$ is the target
power. On the archive's quadruples, with the cure chart as the divisible truth and \eqref{eq:M2}
as the alternative, the widest quadruple is available on $89$ frontiers. It requires a median
per-model log-error noise $\sigma_\ell$ of $\mathbf{0.73}$ milli-nats to attain $80\%$ power---
approximately $1/27.5$ of the nominal $20$ milli-nats and $1/47.7$ of the archive's
median residual scale on those same $89$ frontiers, $47$ milli-nats
(\texttt{table\_power\_summary.csv}). Two measurement requirements follow: the informative
quadruples are \emph{wide}, as expected when resolving a third derivative, and placing the
quadruple where the theorem predicts the signal costs a further factor $4.9$. \emph{A divisibility test can gain discrimination in two ways: by measuring one additional
derivative or by obtaining a floor estimate. The present compute-axis analysis relies on the former;
item-level pass@$k$ measurements can provide the latter through a floor bracket.}
\emph{The error also has a scale symmetry, and the divisibility hypothesis specifies when the
model class is closed under that symmetry.} The error has an absolute \emph{zero} at chance level
but no absolute \emph{scale}: scoring items individually or in bundles changes $\eps$ by a power.
The transformation $\eps\mapsto\eps^{\,t}$ preserves the class precisely when the hypothesis
holds \citep{steutel2004}. The axiom corresponds to $t=1$, whereas the composition law is the limit $t\to0$. The parameter therefore connects the two hypotheses. Closure under this transformation is
equivalent to the stronger condition: \emph{the hypothesis is a consistency requirement, not an
extra assumption}. A class that is not closed under a change of error scale describes the
benchmark's scoring convention rather than the task. Under the transformation, $S$, $\Delta S$,
$s$, $\kappa$, $\Neff$, and $-\ln\epsinf$ are \textbf{extensive}, whereas every $D_k$---and
therefore $\sigma$, $\beta$, and every horizon---is \textbf{invariant}. This also shows why the
floor cannot be derived for the same reason as $\alpha$: two of the theory's three fitted numbers
are the gauge coordinates of a single decomposition (\SM{Proofs: The gauge of a resource axis, and the invariant jet}). The cleanest
test of the hypothesis therefore requires no fitting: \textbf{rescore the same models in
$k$-item bundles}. Under this rescaling of the scoring grain, $S$, $\Delta S$ and $\Neff$ must multiply by exactly $k$, the arc by
$\sqrt k$, and every $D_k$, $\sigma$ and $\beta$ must be unchanged. Any measured departure refutes
the composition law directly, on models already evaluated.
\section{The Floor Is Not a Parameter}
\label{sec:floor}
Two identifiability facts bound what any finite experiment can settle. Both are theorems rather
than sample-size problems (\SM{Proofs: The floor}).

First, finitely many observations cannot distinguish a finite spectrum from a continuous one. For any spectral measure and any $m$
log-resource points there is an atomic measure with at most $m$ atoms matching $\eps$ at every one
of them. No test of ``finitely many modes'' against ``a continuum'' therefore has power exceeding
its level. At the origin, the same result means that a floor cannot be distinguished from spectral mass just
above it. Thus, \emph{the presence of an irreducible error and the finiteness of a benchmark's spectrum
cannot be identified separately, for the same reason.} Second, a local jet known to
order $2n$ determines exactly $n$ quadrature nodes and the floor, but no further spectral detail
\citep{karlin1966,gautschi2004}---equivalently, by the constructive statement below, a law of order
$n$ needs $2n+1$ accuracy measurements. Published windows resolve two derivatives reliably and a
third only marginally, so \textbf{one shape parameter is identifiable, whereas identifying a second
requires the fourth derivative}.
On a geometric grid, the result is constructive: a law of order $k$ is determined exactly by
$2k+1$ accuracy measurements (\SM{Proofs: Law~I}). The case $k=1$ is Aitken's $\Delta^2$: for any equally spaced triple
$x_0-h,x_0,x_0+h$ on the frontier,
\begin{equation}
  \epsinf=\eps_0-\frac{(\eps_0-\eps_-)(\eps_+-\eps_0)}
                      {\eps_--2\eps_0+\eps_+},
  \label{eq:aitken}
\end{equation}
so three accuracy measurements supply the floor without requiring a separate resource measurement. Interpreted correctly,
\eqref{eq:aitken} is a sharp \emph{upper} bound: it gives the largest floor compatible with the
axiom.
Everything above except the floor is a derivative of $S$ at an interior point. The floor is a
boundary quantity: observed on a window $[x_0,x_0+X]$ with log-error noise $\sigma_\ell$, every exponent
with $a\lesssim\sigma_\ell/(x_0+X)$ is observationally equivalent to the floor, and the identified set
given noiseless observations and monotonicity alone is $[0,\min_k\eps(x_k)]$. The axiom then
shrinks this interval to $[0,\,\eqref{eq:aitken}]$ for a mirror triple and to $[0,F(x)]$ at every point by
Theorem~\ref{thm:monofloor}(i). The bound is on the curve and not on the
measurements, so under noise a fitted floor above the best observed error is an ordinary residual
rather than an infeasibility (\SM{Proofs: The floor}). \emph{A fitted floor is window-dependent}:
it is defined for each measurement window and should be reported together with that window.
\begin{figure}[t]
\centering
\includegraphics[width=\linewidth]{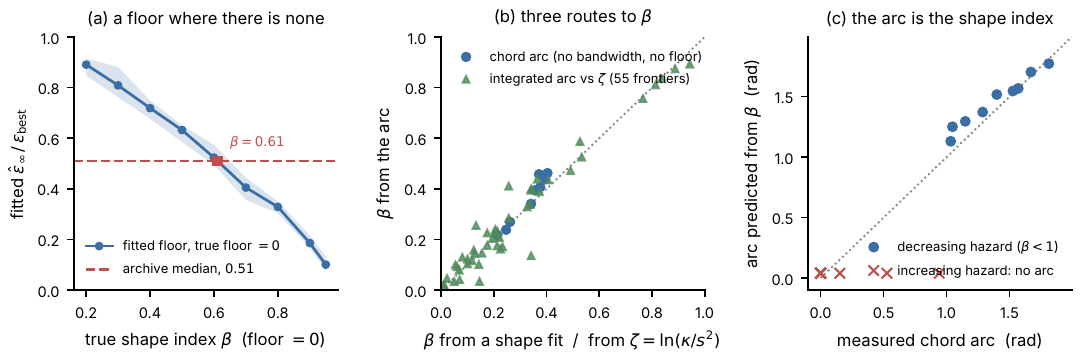}
\caption{The floor is not a parameter. (a) A two-atom fit to a frontier whose
floor is \emph{exactly zero} returns a floor that is a large fraction of the
best observed error, monotone in the shape index, and the curve sweeps the
whole of $(0,1)$ as the index does; the archive's own median fitted share,
$0.96$ of the best observed error, is what a floorless frontier with
$\beta=0.10$ produces. (b) The shape index from three independent
routes---a direct floorless fit, the bandwidth-free three-point chord arc, and
the collapse coordinate $\zeta$---on the frontiers where each is defined.
(c) The arc predicted from $\beta$ against the measured chord arc; the
increasing-hazard frontiers, which have no arc at all, are the three
distribution-shift evaluations. Statements, proofs and scripts in
\SM{Proofs: The floor}.}
\label{fig:nofloor}
\end{figure}
\subsection*{One Inequality Bounds Every Frontier}
\begin{theorem}[The local floor can only fall]
\label{thm:monofloor}
Let $\eps$ satisfy \eqref{eq:axiom}. Then $\kappa'=-\mu_3$, the third central
moment of the rate posterior, and because
$\mathbb{E}[a]\,\mathbb{E}[a^{3}]\ge\mathbb{E}[a^{2}]^{2}$ by Cauchy--Schwarz,
every completely monotone frontier obeys
\begin{equation}
\boxed{\ \frac{\mathrm{d}\ln\kappa}{\mathrm{d}S}\ \le\ 1-\frac{\kappa}{s^{2}}
\quad\Longleftrightarrow\quad
\frac{\mathrm{d}w}{\mathrm{d}S}\le w(1+w)
\quad\Longleftrightarrow\quad
\frac{\mathrm{d}}{\mathrm{d}x}\,\frac{\kappa\,\eps}{s^{2}+\kappa}\ \le\ 0\ }
\label{eq:monofloor}
\end{equation}
with equality at every point if and only if the frontier is a floor plus
exactly one reducible channel. Consequently
\begin{enumerate}[label=(\roman*),leftmargin=2em,itemsep=1pt]
\item $F(x):=\kappa\eps/(s^{2}+\kappa)=\hat p_0\,\eps$, with $\hat
      p_0=1/(1+\Neff)$ the \emph{local} floor share of
      Proposition~\ref{prop:neff}(i), is a non-increasing upper
      bound on $\epsinf$ at every point, so the floor is the \emph{infimum of
      a measured monotone sequence} rather than a parameter: every finite
      window returns a valid bound, and the bound improves by construction.
      It is tighter than $\eps(x_0+X)$ by exactly $1/\hat p_0$. (With the true
      floor share $p_0=\epsinf/\eps$ one would have $F=\epsinf$ identically,
      which is the two-atom case and the equality case of the inequality
      above.)
\item The two-atom chart is the \emph{envelope} of the admissible region, so a fitted floor can err
      in only one direction. This one-sided behavior of the window-dependent floor estimate is
      therefore a geometric property, not merely an estimator artifact.
\item At the canonical point $w=1$ the inequality reads $\kappa'\le0$: the
      point of maximal curvature never comes \emph{after} the canonical point,
      and the two coincide exactly when the frontier is two-atom. The three
      routes to the canonical point are therefore ordered, not merely
      discrepant.
\item \eqref{eq:monofloor} is a one-sided condition on a measurable statistic, so it can be
      tested: a window on which $F$ rises beyond its design-matched null provides evidence
      against \eqref{eq:axiom} on that window. This is the first falsification test the
      axiom has had that is not a ratio with a two-sided null.
\end{enumerate}
\end{theorem}
\noindent Clause~(i) is a statement about the exact derivatives. With plug-in
derivatives it is clause~(iv) that one measures. Call a frontier
\emph{second-order-readable} when all its measured errors are positive, it
carries at least seven points, and the local-quadratic jet returns $s>0$ and
$\kappa>0$ at five or more points, so that $\Neff=s^2/\kappa$ is defined
across a window rather than at a single point. This criterion is met by $92$ of the $117$
frontiers; all results reported for second-order-readable frontiers use this population. Among those $92$, $F$ is nonincreasing over the full window for only $5$;
the median trend is $\mathrm{d}\ln F/\mathrm{d}S=+0.284$. The new bound beats the best
observed error on $77$ of the $92$. The bound remains valid wherever the jet is defined. What fails is the monotonicity of the
plug-in estimate, which is why the running infimum is reported alongside, rather than in place of,
the right-edge value.
Theorem~\ref{thm:budget} supersedes $F$ wherever the composition law is granted.
\paragraph{Why every test in this paper is one-sided.} By Markov--Krein the extremal measures of
every moment-constrained bound are \emph{atomic} \citep{karlin1966}, while no non-degenerate
compactly supported law is infinitely divisible, so the divisible measures lie strictly in the
\emph{interior} of every bound the theory can write and its complement contains the boundary: a
violation refutes, and satisfaction cannot confirm, because the equality cases all belong to the
alternative.
\subsection*{The Floor Is the Shape Index Read at a Window}
\begin{proposition}[The floor is the shape index read at a window]
\label{prop:nofloor}
Let $\eps=\exp\{-[\alpha(x-x_0)]^{\beta}\}$ with $0<\beta\le1$. This function is the Laplace transform of a one-sided stable density of index $\beta$; it is
therefore completely monotone and has \emph{no atom at the origin}. Write $S=-\ln\eps$.
Then, pointwise and without specifying
a resource origin, a floor, or a fitted model,
\begin{equation}
\boxed{\ \beta=\frac{s^2}{s^2+\kappa S}=\frac{1}{1+S\,e^{\zeta}},
\qquad
\hat p_0=\frac{1-\beta}{(1-\beta)+\beta S}\ }
\label{eq:nofloor}
\end{equation}
and consequently
\begin{enumerate}[label=(\roman*),leftmargin=2em,itemsep=1pt]
\item here $\hat p_0=1/(1+\Neff)$ is the \emph{local} floor share of
      Proposition~\ref{prop:neff}(i)---the share the tangent two-atom chart
      carries---whereas the true floor share $\epsinf/\eps$ is identically zero.
      The limiting cases make this relation explicit: $\beta=1$ gives $\hat p_0\equiv0$ and
      $\beta\to0$ gives $\hat p_0\to1$, so a window
      sees a floor exactly to the extent that the spectrum accumulates at the
      origin, and the two charts---an atom plus one channel, or no atom and a
      shape index---are one chart;
\item $\hat p_0\sim(1-\beta)/(\beta S)$, so there is no asymptotic floor on any
      frontier with $\beta>0$, and the window-dependent floor estimate defined earlier in
      this section has a closed-form decay rate: its apparent share decreases
      inversely with the log-error;
\item the canonical point is a \emph{depth in nats},
      \begin{equation}
      S^\star=\frac{1-\beta}{\beta},
      \qquad
      \eps^\star=e^{\,1-1/\beta},
      \label{eq:kneedepth}
      \end{equation}
      with no $\alpha$, no floor, no units and no resource measurement in it;
\item integrating the line element of Law~III over a window,
      \begin{equation}
      L=2\sqrt{\frac{1-\beta}{\beta}}
        \Bigl(\sqrt{S_{\max}}-\sqrt{S_{\min}}\Bigr),
      \qquad
      \beta=\Bigl[1+\bigl(L/2\Delta\!\sqrt S\bigr)^{2}\Bigr]^{-1},
      \label{eq:arcbeta}
      \end{equation}
      so the arc measures the shape index with an integral and a one-sided
      bound rather than with the pointwise ratio that
      Section~\ref{sec:lawfree} warns about;
\item read as an estimator, \eqref{eq:nofloor} leaves its own range
      $0<\beta\le1$ exactly when $\kappa<0$: an \emph{increasing} hazard,
      forbidden by the axiom, and by \eqref{eq:arcbeta} the same event as a
      window with no measurable arc. Which way it leaves depends on the
      denominator, so the diagnostic is the exit and not its direction:
      $\hat\beta>1$ while $s^2+\kappa S>0$, and $\hat\beta<0$ once
      $S>s^2/|\kappa|$ carries the denominator through zero.
\end{enumerate}
\end{proposition}
\paragraph{A fitted floor on a frontier that has none.} Simulate from the chart of
Proposition~\ref{prop:nofloor} with the floor set to \emph{exactly zero}, add the corpora's own
$0.02$-nat log-error noise, and fit \eqref{eq:M2}. On the archive's own median
design---$5.1$ nats of log-resource and $11$ points---the fitted floor returns a share of the best
observed error that falls monotonically from $0.98$ at $\beta=0.05$ through $0.45$ at $\beta=0.6$ to
$0.01$ at $\beta=0.99$; on a longer window ($8$ nats, $22$ points) the same curve reads $0.98$ to
$0.02$, through $0.52$ at $\beta=0.6$. \emph{The zero-floor null therefore spans $[0.01,0.98]$ as
the shape index runs over $(0,1)$, so no measured floor share is evidence of a floor.} The archive's own median fitted share
is $0.96$ of the best observed error over the $117$ frontiers ($0.99$ over the $92$
second-order-readable ones), and the null reproduces that at $\beta=0.10$ with no floor anywhere in
it. \emph{A measured floor share is a readout of how the spectrum approaches the origin, not
evidence that the origin carries mass.} By Corollary~\ref{cor:odepencil} this is not a measured
coincidence but a corollary of the pencil: the floor is the line's intercept and $\beta$ its slope,
so a fitted floor can only be a monotone function of $\beta$.
\paragraph{The fitted floor carries its axis.} If the fitted floor were an estimate of the Bayes
error, fitting it along different resource axes on the same evaluation set would return a common
value. It does not:
on ImageNet-1k the three single-resource fits give $0.112$, $0.124$ and $0.085$ with no single
value inside all three $90\%$ intervals---the parameter fit is disjoint from the two latency fits,
which overlap each other---while the window bound is \emph{identical} across the three axes, and $\epsinf$ exceeds
the label error of \citet{northcutt2021} by factors of $1.4$ to $2.1$, so it cannot be interpreted as a
label-noise ceiling. By \eqref{eq:attemptfloor} that is what a floor must do: $\Lambda$
is a property of the axis, so a fitted floor is one too
(\SM{Proofs: Reading a frontier without fitting it}).
\paragraph{What this rules out.} The predecessor of this work \citep{ABrain.Rank} estimates the
floor from the co-failure of two models alone. Section~\ref{sec:evidence} rejects this mechanism
by nineteen standard errors; accordingly, the present analysis does not use an estimator of that form.
\section{The Count Axis}
\label{sec:count}
The item-level assumption tested in Section~\ref{sec:evidence} places the exponential link
\emph{per item in log-resource}: it assigns one scalar $a$ to each item and assumes that models
fail independently conditional on that rate. On the per-item panel, observed co-failure exceeds the
prediction under conditional independence by a median of $57\%$ ($19$ standard errors). The data
instead support a nested-difficulty structure. The exponential link remains meaningful, however,
on a different axis.
Fix a resource that \emph{counts independent attempts}: members of an ensemble, retries of a
stochastic decoder, samples in a best-of-$n$ or self-consistency scheme, votes of a committee
\citep{wang2023selfconsistency,brown2024,snell2025}. The procedure succeeds if any attempt
succeeds. Let $g$ be the probability that one attempt fails a given item and $\nu$ the law of $g$
over items. The weakest hypothesis that yields all of Law~I at once is not independence but
a symmetry.
\begin{proposition}[Law~I on a count axis]
\label{prop:countaxis}
Suppose the attempt sequence is \emph{exchangeable}---the joint law of the
outcomes is invariant under permuting the attempts---and infinitely
extendible, so that it is the restriction of an infinite exchangeable sequence.
(Both are needed: a finite exchangeable sequence is not in general a mixture of
i.i.d.\ ones.) Then by de Finetti's theorem
$P(m):=\Pr[\text{all }m\text{ attempts fail}]=\int_0^1 g^m\,\mathrm{d}\nu(g)$.
Hence $P$ is a completely monotone sequence, $P(m)=\int e^{-bm}\mathrm{d}\nu$
with $b=-\ln g$, and by Hausdorff's theorem $\nu$ is determined by $P$. The
floor is $\nu(\{1\})$, the items that no attempt can solve.
\end{proposition}
\paragraph{The hypothesis is a symmetry, and the independence is a conclusion.} Exchangeability
says only that the order in which attempts are made does not matter, and that the sequence extends;
de Finetti then \emph{produces} the latent $g$ and the conditional independence given it. At the item level on a count axis, the proposition holds trivially---$\nu=\delta_{g_i}$ and
$P_i(m)=g_i^{\,m}$. The substantive content emerges at the population level, where $\nu$ is
the distribution of item difficulties. The corresponding \emph{compute}-axis statement cannot be established item by item: an
individual item's error as a function of resource is a step function, which is not completely
monotone. This is the distinction enforced in Section~\ref{sec:evidence}; the refutation reported
there concerns independence across models conditional on an item, not exchangeability across
attempts. Thus, the count-axis result requires no additivity, scale-freeness, or separate Bernstein
assumption. Exchangeability and infinite extendibility provide the structure that
Section~\ref{sec:law1} must assume on the compute axis. Every consequence of complete
monotonicity therefore transfers directly, including Laws~II and~III, the regret identity
\eqref{eq:regret}, the capacity form, and the covering law. This brings test-time compute within
the same theoretical framework under a single symmetry assumption. Equation~\ref{eq:regret}
also prices an attempt against an e-fold of compute exactly, because both are measured in nats.
\section{What the Evidence Establishes}
\label{sec:evidence}
The empirical analysis evaluates the theory on $117$ single-resource frontiers spanning the
archive's eight corpus categories---\texttt{vision}, \texttt{vision-shift}, \texttt{vision-size},
\texttt{nas}, \texttt{tabular}, \texttt{hpo}, \texttt{materials},
\texttt{medicine}---using $2\,025$ leave-one-out prediction events and a
per-item panel constructed for this study
\citep{wightman2019timm,dong2020nasbench201,dong2022natsbench,erickson2025tabarena,pfisterer2022yahpo,riebesell2025,doerrich2025medmnistplus}.
\emph{This section uses two related but distinct sample definitions.} The set of $117$ includes
every frontier in the archive. Table~\ref{tab:namedrules} is measured on a stricter
build of the same records---at least eight Pareto-efficient points, and one frontier per (corpus,
benchmark, resource) rather than one per variant---which leaves $96$ frontiers, of which $54$ have a usable floor
and $26$ contain the canonical point within the measured curve. Each count below identifies the
population to which it refers. Five findings follow; their supporting
tables are provided in \SM{The empirical study}.
\paragraph{What the evidence reveals about the standard form.} Three complementary findings bear
on \eqref{eq:M2}: a controlled zero-floor simulation, a design-calibrated assessment of the
admissible ceiling, and an out-of-sample comparison with a same-cost alternative.
\subparagraph{Finding 1: the fitted floor is a shape-dependent readout.}
By Corollary~\ref{cor:odepencil} the floor
and the shape index are the intercept and the slope of one line, so they are one degree of freedom.
The consequence is then demonstrated without touching the corpus: fit \eqref{eq:M2} to simulated
frontiers whose floor is exactly zero, on the archive's own median design of $5.1$ nats and $11$
points. The fitted floor share sweeps effectively the whole of $(0,1)$ as the shape index runs over
it---$0.978$ at $\beta=0.05$, $0.529$ at $0.50$, $0.007$ at $0.99$---and it passes through $0.957$
at $\beta=0.10$. The archive's own median fitted share is $0.956$ over the $117$ frontiers and
$0.991$ over the $92$ second-order-readable ones. \emph{The entire measured range lies inside the
zero-floor null.} No plug-in estimator, no bandwidth and no corpus noise enter this statement; it is
a property of the three-parameter fit itself, so it transfers to any published fit of that form.
\subparagraph{Finding 2: the admissible ceiling is a structural guarantee with limited finite-window power.}
The axiom gives
$\epsinf\le\eps/(1+\Neff)$ at every point of every frontier, which invites a count: on $89$ of
$117$ frontiers the archive's fitted floor exceeds a plug-in evaluation of the right-hand side.
That count is not evidence, and Table~\ref{tab:ceilnull} is the design-matched null it has to be
read against.
The construction is forced by the theory. Form \eqref{eq:M2} is the two-atom chart, which
\emph{saturates} the axiom wall---$\Neff=\rho-1$ and $\epsinf=\eps/(1+\Neff)$ identically, for
every parameter value. A frontier of that form therefore can never violate its own ceiling, which
makes the ceiling test a goodness-of-fit test of \eqref{eq:M2} whose null hypothesis is
\eqref{eq:M2} itself. Accordingly, the null is constructed separately for each frontier: fit
\eqref{eq:M2} to that frontier, take
its own residual scale as the per-model noise, draw replicates on its own nodes, and re-run the
identical pipeline---refit, re-estimate, re-decide---on each replicate.
Three observations follow, beginning with the calibration of the original count. The published statistic minimized $\eps/(1+\Neff)$ over all triples on each frontier
(median: $75$ triples; maximum: $16\,970$). Because $\Neff=s^{2}/\kappa$ diverges as the
measured curvature $\kappa$ approaches zero from above, this minimum is an extreme value selected
from thousands of heavy-tailed estimates, not a valid bound. On $16$ of the $117$ frontiers, the
companion wall returned exactly zero, which is incompatible with a frontier having a positive floor.
Under the null, the same procedure has an expected $100.3$ declared violations per replicate,
out of $117$---\emph{more} than the $89$ observed. When calibrated against its own null, the count is
$10$ of $117$.
Second, the calibrated readings carry no signal either. Fixing the location in advance instead of
searching over triples---the median over consecutive triples, or the single widest triple, both
chosen before the data are seen---brings the rejection rate to the nominal level: the two readings produce $0$ and
$5$ rejections at the $5\%$ level, compared with $5.6$ and $4.4$ expected, respectively. No
rejections survive a Benjamini--Hochberg correction.
Third, the calibrated test has limited power on the published windows because the available depth,
rather than measurement noise alone, constrains discrimination. An alternative must
leave the two-atom shape, since \eqref{eq:M2} cannot violate its own ceiling, so we deform the wall
itself: $\Neff=\theta(e^{\Delta S}-1)$ with $\theta\ge1$, which Proposition~\ref{prop:odefirst}(ii)
integrates in closed form to $s=\alpha(1-e^{-\Delta S})^{1/\theta}$. Its floor exceeds its ceiling
by $\ln[(1+\theta(\rho-1))/\rho]$, which is $\ln\theta$ at depth but, expanded at the floor,
\begin{equation*}
\ln\frac{1+\theta(\rho-1)}{\rho}=(\theta-1)\,\Delta S\Bigl(1-\tfrac{\theta}{2}\Delta S\Bigr)+O(\Delta S^{3}),
\end{equation*}
\emph{linear in the depth remaining}. This is Corollary~\ref{cor:wallivp}(iii) with an explicit coefficient and explains the limited
power near the floor. The archive's frontiers sit a median of $0.045$ nats
above their own fitted floor, and $65$ of the $117$ sit within $0.10$ nats of it. Thus, even a truth
whose floor is \emph{twice} its admissible value offers a median signal of $0.0435$ nats at the best
measured point, against a median per-model residual of $50.5$ milli-nats across all $117$. Two further effects then
work in the same direction. The plug-in estimator also depends on $\kappa$, which vanishes with depth; consequently, its
relative error increases as the signal shrinks. In addition, the three-parameter fit absorbs part of
the deformation: on a $\theta=2$ truth, it returns a floor equal to $0.782$ times the true value,
shifting the estimate by $0.245$ nats in the direction that masks the violation. This effect
compounds the loss of power described above.

At $\theta=2$, power is $0.010$ for the median-triple reading and $0.096$ for the widest-triple
reading. Table~\ref{tab:ceilnull} shows that neither achieves substantial power even when the
per-model noise is reduced to $2$ milli-nats. \emph{The published compute frontiers neither refute nor confirm the ceiling: the limiting factor
is the depth covered by the measurements, not simply their precision.} Resolving the question therefore requires a \emph{shallower window}, not merely a finer
instrument: the frontier should be measured where $\Delta S$ is of order one rather than $0.04$,
that is, across a region where a substantial fraction of the error remains reducible. Such a window
is accessible on a count axis, as discussed in Section~\ref{sec:count}. The published compute frontiers do not cover that region, so replication within the same windows
cannot resolve the question.
\subparagraph{Finding 3: a same-cost alternative matches held-out accuracy.}
The cure chart \eqref{eq:cure} uses the same three parameters and achieves a pooled median
relative error of $2.57\%$ on its $2\,013$ scored points, compared with $2.60\%$ for \eqref{eq:M2}
across all $2\,025$ prediction events. The cure chart also satisfies \eqref{eq:cureids} on the $115$
frontiers it fits, whereas \eqref{eq:M2} does not.
\begin{table}[tbp]\centering
\small\setlength{\tabcolsep}{4.5pt}
\begin{tabular}{lccc}
\toprule
per-model noise & incumbent, false & median triple, power & widest triple, power \\
\midrule
$\phantom{0}2$ milli-nats  & $1.000$ & $0.020$ & $0.040$ \\
$\phantom{0}5$ milli-nats  & $1.000$ & $0.013$ & $0.076$ \\
$10$ milli-nats            & $1.000$ & $0.013$ & $0.062$ \\
$20$ milli-nats            & $1.000$ & $0.020$ & $0.063$ \\
\midrule
$50.5$ milli-nats (the archive's own) & $1.000$ & $0.010$ & $0.096$ \\
\bottomrule
\end{tabular}
\caption{The admissible-ceiling test of Corollary~\ref{cor:wallivp}, read against a null in which
each frontier \emph{is} the three-parameter form \eqref{eq:M2} fitted to it, drawn on its own nodes
at its own residual scale. ``False'' is the per-frontier rate at which the procedure declares a
violation when none exists, and power is measured against the deformed wall
$\Neff=\theta(e^{\Delta S}-1)$ at $\theta=2$, a floor at twice its admissible value. The incumbent
statistic minimizes $\eps/(1+\Neff)$ over every triple and is not calibrated under the null at
any precision reported here. The two pre-registered readings are at their nominal level and have no power on these
windows, which is what Corollary~\ref{cor:wallivp}(iii) predicts. \emph{Precision is not the
binding constraint}: the incumbent's false-violation rate is $1.000$ at every precision down to $2$
milli-nats, a factor $25$ below the archive's own residual scale, and the two pre-registered
readings never leave $[0.01,0.10]$ in power. The constraint is depth, not noise---see the text.}
\label{tab:ceilnull}
\end{table}
\emph{Predictive evaluation and comparison baselines.} The protocol is leave-one-out, and the number of prediction events warrants clarification. For
each interior point on each frontier, the law is refitted using the other $n-1$ measurements from
the same frontier and used to predict the withheld measurement; the two endpoints cannot be
interpolated and are never predicted. Across the $117$ frontiers, the procedure produces $2\,025$ prediction events from $2\,259$
measurements. These events are \emph{not} $2\,025$ independent tests: each fit uses a median of $91\%$ of its frontier's measurements, and each
measurement is reused as training data by neighboring fits.
Uncertainty is therefore bootstrapped over the $117$ frontiers and never over the $2\,025$ events.
Pooled over the events, the law returns $R^{2}=0.9897$ on the log-error and a median relative error
of $2.6\%$, against $0.9752$ and $4.4\%$ for the floorless power law, and its median absolute
error, $0.026$ nats, is below the $0.05$-nat band inside which its own recommendation does not
change. Per frontier, which is the unit the bootstrap respects, Table~\ref{tab:predbase} scores it
against that power law and against two predictors that assume no model at all.
\begin{table}[tbp]\centering
\small
\begin{tabular}{lccc}
\toprule
 & \multicolumn{3}{c}{per-frontier median error (nats), with $95\%$ CI} \\
\cmidrule(lr){2-4}
predictor & interior & forward & backward \\
\midrule
\eqref{eq:M2}, three parameters   & $0.037\,[0.030,0.046]$ & $0.074\,[0.052,0.116]$ & $0.159\,[0.116,0.257]$ \\
floorless power law               & $0.062\,[0.047,0.083]$ & $0.205\,[0.147,0.256]$ & $0.142\,[0.120,0.206]$ \\
log-linear interpolation          & $0.018\,[0.014,0.028]$ & $0.073\,[0.063,0.092]$ & $0.162\,[0.124,0.217]$ \\
nearest measured point            & $0.033\,[0.029,0.040]$ & $0.076\,[0.063,0.110]$ & $0.161\,[0.124,0.213]$ \\
\midrule
 & \multicolumn{3}{c}{frontiers on which \eqref{eq:M2} is the better of the pair} \\
\cmidrule(lr){2-4}
versus floorless power law        & $\mathbf{73.5\%}$ & $\mathbf{73.5\%}$ & $\mathbf{61.5\%}$ \\
versus log-linear interpolation   & $25.6\%$          & $50.4\%$          & $\mathbf{67.5\%}$ \\
versus nearest measured point     & $44.4\%$          & $53.8\%$          & $\mathbf{65.0\%}$ \\
\bottomrule
\end{tabular}
\caption{Held-out prediction, scored per frontier. \emph{Interior} is leave-one-out on each interior
measurement; \emph{forward} trains on the least expensive two-thirds and predicts the most expensive third;
\emph{backward} trains on the most expensive two-thirds and predicts the least expensive third.
The error is $|\ln\eps_{\mathrm{pred}}-\ln\eps_{\mathrm{obs}}|$, summarized per frontier and then across
frontiers, with intervals bootstrapped over the $117$ frontiers---never over the $2\,025$ dependent prediction events. The lower panel is the paired comparison on the same frontiers.
\emph{The law is better than the floorless power law on a majority of frontiers in all three regimes;
it loses the interior to a model-free interpolator and wins the backward extrapolation against
every baseline.} Script
\texttt{169}, from script \texttt{68}'s per-frontier output.}
\label{tab:predbase}
\end{table}
The table supports three complementary interpretations of the law's predictive performance.
\emph{Against the floorless power law, the differential law performs better on a majority of
frontiers in each regime.} The paired median improvement is $0.011$ nats interior and $0.040$
nats forward, both with intervals clear of zero, and the law is the better of the pair on $73.5\%$
of frontiers in each of those regimes and on $61.5\%$ in backward extrapolation.
\emph{For interior prediction, local interpolation is more accurate.} Log-linear interpolation
through the two neighboring observations is more accurate on $74.4\%$ of frontiers, with a paired
median advantage of $0.013$ nats. This comparison reflects the different objectives: the interpolator uses both neighbors of its
target, whereas the frontier law also provides a floor, a price and an operating point. It therefore
clarifies the predictive scope of the pooled $R^{2}$ reported above.
\emph{The clearest advantage over model-free prediction is in backward extrapolation.}
For forward extrapolation, the two are statistically indistinguishable: the law is better on
$50.4\%$ of frontiers, with an interval that includes no difference. For backward extrapolation,
where the low-resource end must be predicted from higher-resource observations, the law is better
on $67.5\%$ of frontiers, with a paired median improvement of $0.026$ nats, and it is the only predictor in the table that beats all three
others there. \emph{On these windows, the structural model's clearest predictive advantage is backward
extrapolation.}
The prediction error plateaus at about two and a half times the measurement-noise scale. That scale
also has a useful interpretation: $0.015$ nats is the leave-one-out residual of a local quadratic,
which provides both an upper bound on the per-model noise and a benchmark for a model-free smoother.
Interpreted as a noise estimate, it suggests that the three-parameter law is misspecified; interpreted
as a predictive baseline, it indicates that the smoother also outperforms the law for interior
prediction. These interpretations describe the same empirical pattern.
Fitting the whole completely monotone class nonparametrically does no
better at the training sizes where the law is normally used---$m\le6$ anchor points, where the two
stay within $9\%$ of each other in either direction---but it does overtake the law when more are
available, by $8\%$ at $m=8$ on the $66$ frontiers that carry eight anchor points and by $21\%$ at
$m=16$ on the $36$ that carry sixteen, and a stable index beats the law by $43\%$ at the
same $m$, so the broader spectrum is worth more than the nonparametric freedom that finds it.
\emph{The residual is consistent with a broader spectrum than two atoms, which becomes more
visible as the measurement window supplies additional anchor points.}
\emph{What is measured and new.} The pointwise arc--depth identity is testable on these windows and
holds: fitting the cumulative arc as an affine function of $\sqrt{\Neff}$ beats $\Neff$ and $S$ on
$76$ of the $90$ frontiers that carry a cumulative arc at all, median $R^{2}=0.9994$. The smallest claim the theory makes,
Corollary~\ref{cor:two}, is an instance at every Pareto pair: on $20\,206$ pairs the harmonic rule
\eqref{eq:harmonic} agrees with the resource-aware choice on $90.9\%$ against $77.4\%$ for
``always take the cheaper.'' This extends the two-system test of \citet{ABrain.Rank} from $5\,459$
pairs to $20\,206$, with the floor-share form \eqref{eq:harmonic} in place of a fitted weight. On the $4\,229$ near-ties, defined by a decision margin below $0.05$ nats, agreement
remains $70.9\%$ (skill $0.24$); because these pairs have small decision margins by construction,
disagreements carry relatively little regret. The weakest performance occurs on the $334$ pairs
whose own frontier is furthest from the law, with a gap above $0.30$ nats: agreement is $52.4\%$
with skill $-0.07$. This pattern is consistent with the model-misfit interpretation of
Section~\ref{sec:lawII}.
\begin{figure}[t]\centering
\includegraphics[width=0.82\linewidth]{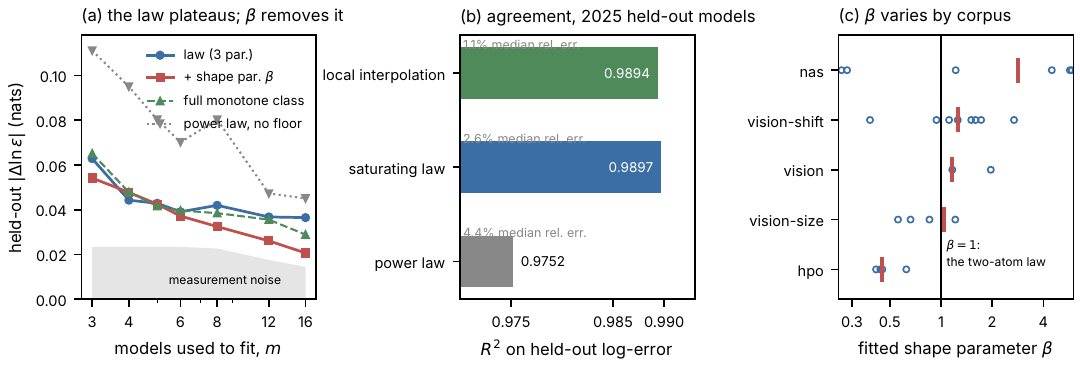}
\caption{(a) Held-out error against the number of models used to fit, on $117$
frontiers carrying the archive's eight corpus labels. The three-parameter law
plateaus well above the measurement noise (shaded); the whole completely
monotone class, fitted nonparametrically, plateaus with it; the one shape
parameter $\beta$ of the stretched-exponential form removes most of the gap.
(b) Agreement on the $2\,025$ pooled interior leave-one-out predictions.
(c) The fitted $\beta$, with one point per frontier and a bar at the corpus median:
corpus-specific and, on these windows, weakly identified. Details are provided in
\SM{The empirical study}.}
\label{fig:prediction}
\end{figure}
\paragraph{How large the decision is, measured without an objective function.} Before scoring
rules against this paper's loss, it is useful to establish the magnitude of the choice in units that
do not depend on a particular objective function.
Take the eight rules a practitioner can apply to a discrete frontier without fitting a model or
specifying an additional hyperparameter: twice-the-best-error, the knee, half-slope, most accurate, cheapest, accuracy per unit cost,
accuracy per $\log_{10}$ cost, and a mid-range budget. Apply them to the $96$ single-resource
frontiers---no floor is needed, so this is the widest population in the paper---and read off only
what the measured frontier says.
These eight rules select a median of \emph{five distinct models} per frontier. The extreme picks
are separated by a median resource ratio of $19\times$ (interquartile range $4\times$ to
$236\times$), or $2.95$ e-folds. This resource difference corresponds to a median of $1.72$ bits
of error (interquartile range $[0.86,3.05]$). The spread exceeds $10\times$ on $52$ of the $96$
frontiers; on $29$, it corresponds to less than one bit. The median exchange rate across the
selection disagreement is $12$ resource units per bit.
One of the three rules in current use is not scale-free, with a substantial practical consequence:
multiplying the resource by $10^{3}$---GFLOP to MFLOP, seconds to milliseconds---moves its pick on
$\mathbf{94}$ of those $\mathbf{96}$ frontiers. \emph{A rule whose selection changes when the resource unit changes encodes a price rather than an
intrinsic measurement.}

\begin{figure}[t]\centering
\includegraphics[width=\linewidth]{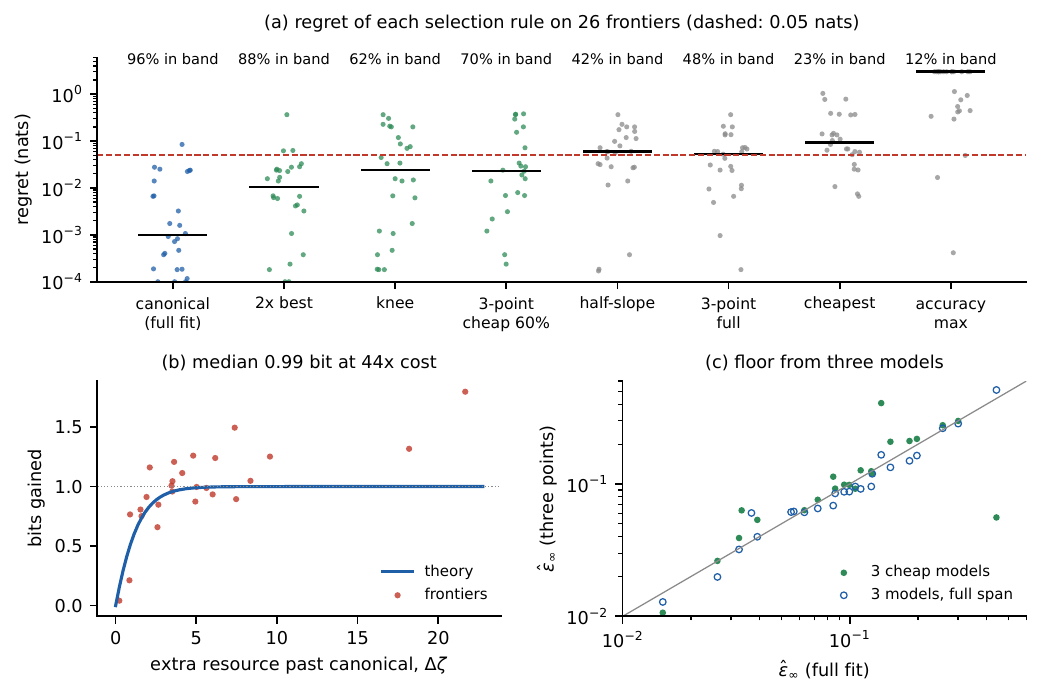}
\caption{The rules actually in use, scored against the derived price on the
same frontiers. (a) Regret of each selection rule, one point per frontier,
bars at the median; the dashed line is the $0.05$-nat band inside which the
recommendation does not change. The canonical rule lies inside the band on almost every frontier, while the
comparison quantifies the cost of alternative choices. (b) The bits gained against
the extra resource spent past the canonical point: the median frontier buys
$0.99$ bits at $44\times$ the cost, which is the price of not knowing where to
operate. (c) The floor read from three models against the floor from a full
fit. Detail in \SM{The empirical study}.}
\label{fig:rules}
\end{figure}

The same analysis shows that many Pareto-efficient points on a published frontier are never
optimal for any nonnegative price. Because
$\sigma_i(\lambda)=-\ln\eps_i-\lambda x_i$ is linear in $\lambda$, the candidates optimal at some
nonnegative price are the vertices of the upper envelope of $K$ lines, obtained with one sort and
one stack pass. Across the $117$ frontiers, the median number of reachable points is $5$ out of $11$. Conversely,
the median share of Pareto-efficient models that are \emph{never optimal at any
nonnegative price} is $\mathbf{62.5\%}$; their absence from the upper envelope establishes this
directly. None of these numbers uses this paper's loss, price or floor.
\emph{Against the rules actually in use.} Every rule below returns one model from the same
frontier; the regret is \eqref{eq:regret} against the derived price---a different quantity from
the excess over the minimax-optimal candidate reported in Section~\ref{sec:scope}, which
scores a band of prices rather than one---and the \emph{scale-free} column indicates whether the answer is unchanged under a change in
the resource unit. Each rule is implemented directly on a discrete measured frontier, which is why the knee row is not at zero although the knee and the canonical
point coincide on the chart. The comparison is across
frontiers rather than within a single frontier, following the setting described by
\citet{demsar2006}, and the
corpora are of the kind a benchmark suite assembles \citep{gijsbers2024amlb}; the quantity
compared is a decision and its cost in nats, not an accuracy.
\paragraph{Interpreting the selection comparison.} Four qualifications help interpret
Table~\ref{tab:namedrules}.
First, \emph{the scoring uses this paper's own loss and the derived price is that loss's analytic
minimizer}, so the first row is nearly an identity and serves as a consistency reference rather than an
independent comparative result.
The table therefore measures the \emph{spread} among rules currently in use, a quantity
already established in the previous paragraph without reference to the paper's loss.
Second, the ranking is evaluated statistically. Over the nine rules defined on every one of
the $26$ frontiers, Friedman gives $\chi^{2}=114.0$ ($p=6\times10^{-21}$; Iman--Davenport
$F=30.3$). At the Nemenyi critical difference of $2.36$ the derived price has the best mean rank,
$1.62$, and is separated from six of the other eight---but \emph{not} from twice-the-best-error or
from the knee. The ranking therefore does not statistically distinguish these three rules. The derived rule's in-band fraction
carries a bootstrap interval of $[88\%,100\%]$ and its median regret $[0.0004,0.0067]$ nats.
Third, the result is not specific to the floor estimator used to determine attainability. Because
that determination depends on a fitted floor, we repeat the selection procedure using the
bandwidth-free three-point estimate. Of the $96$ single-resource
frontiers, $54$ have a usable fitted floor and $26$ are attainable. Under the three-point floor,
$48$ have a usable estimate and $32$ are attainable; the two sets agree on $24$. The derived-price rule then falls within the tolerance band on $94\%$
of the attainable frontiers, with a median regret of $0.0019$ nats and the ranking unchanged.
Fourth, two checks in the table do \emph{not} use this loss. The first measures what additional
resource buys: a median gain of $0.99$ bits (interquartile range $[0.85,1.20]$) at a median cost equal to $44\times$ the canonical point's cost, and fewer than one bit on $14$ of the $26$. The other is the closed-form price
of misquoting the floor, which predicts the realized plug-in regret with a median absolute error
of $0.000$ nats on these same frontiers. Figure~\ref{fig:rules} shows the same
comparison per frontier, together with the bits bought by the extra resource
spent past the canonical point and the three-model floor read against a full
fit.
\begin{table}[tbp]\centering
\small
\setlength{\tabcolsep}{4pt}
\begin{tabular}{llccc}
\toprule
rule & needs & scale-free & median regret & within $0.05$ \\
\midrule
derived price (Theorem~\ref{thm:canonical}) & a fit & yes & $\mathbf{0.0010}$ & $\mathbf{96\%}$ \\
twice the best error                        & nothing      & yes & $0.0104$ & $88\%$ \\
knee of the curve                           & nothing      & yes & $0.0243$ & $62\%$ \\
fixed exponent, $w=-0.07$ & an exponent  & yes & $0.0520$ & $50\%$ \\
accuracy $/\log_{10}$ cost                  & a unit       & \textbf{no} & $0.0657$ & $42\%$ \\
accuracy per unit cost                      & nothing      & yes & $0.0677$ & $27\%$ \\
cheapest model                              & nothing      & yes & $0.0936$ & $23\%$ \\
hard budget, mid-range                      & a budget     & yes & $0.1698$ & $19\%$ \\
most accurate model                         & nothing      & yes & $\infty$ & $12\%$ \\
\bottomrule
\end{tabular}
\caption{Selection rules on the $26$ attainable single-resource frontiers of the archive, scored
against the derived price. Regret is in nats; the tolerance is $0.05$ nats inside which the
recommendation does not change. Each rule is implemented directly on a discrete measured frontier. This explains why the knee
row is not zero even though the knee and the canonical point coincide on the chart. The ratio of accuracy to $\log_{10}$ cost is the ``efficiency score''
of \citet{prucs2025pareto}, whose limitations are discussed by
\citet{dehghani2022efficiency}; the knee is
\citet{das1999knee,satopaa2011kneedle}; the budget is the rule of \citet{sun2026roofline} at the
geometric midpoint of each frontier's own range. The fixed-exponent row is the soft objective of
\citet{tan2019mnasnet}, $\mathrm{ACC}\cdot(\mathrm{LAT}/T)^{w}$, whose target enters only through
the constant $T^{-w}$ and cancels from the argmax: that rule is target-free and scale-free on all
$26$ frontiers, leaving the exponent as its single declared number, and moving $w$ over
$[-0.20,-0.01]$ moves the pick by a median $0.488$ nats, ten times the tolerance.
\emph{The first row is a consistency reference rather than an independent comparative result};
see the qualifications above and the loss-free measurement of rule dispersion.}
\label{tab:namedrules}
\end{table}
\emph{Two readings that do not use the paper's own loss.} The rule requires one estimated number, and the
cost of misestimating it has a closed-form expression: quoting $\phi\,\epsinf$ costs $\ln\cosh(\tfrac12\ln(2\phi-1))$
on the law, with a pole at $\phi=\tfrac12$ and only logarithmic growth above, so overestimating the floor is less costly than underestimating it (\SM{The floor}). Take the three-point floor as the plug-in on the $26$ attainable
frontiers---three accuracies, no bandwidth. The formula then predicts the \emph{realized} regret of the selected point with a median
absolute error of $0.000$ nats. The estimator underestimates the floor on $16$ of $25$ frontiers, and the selected point remains
within the band on $80\%$ of them, compared with $92\%$ under the law-only benchmark.
\emph{This closed-form relationship remains informative even for frontiers that do not obey the
assumed law.} The nested-resource correction also matters: on the $21$ language-model grids the additive
reference $\theta=1$ lies within the $95\%$ profile interval for only $6$ of the $21$ grids, and assuming it misplaces the
canonical depth by a median $0.340$ nats---seven times the tolerance---against the nested
$\Delta S=\ln(1+n^{1/\theta})$ of \TR{Interacting resources: the nested dual}.
\paragraph{Per-item mechanism and empirical scope of divisibility.} The per-item panel also tests
the proposed interpretation in which models fail independently conditional on an item's failure
rate. Across $21$ classifiers, observed co-failure exceeds that prediction by a median of $57\%$
($19$ standard errors). Instead, the measurements support a nested-difficulty relation,
$\Pr[\text{both fail}]=\kappa\min(h_i,h_j)$, with $\kappa=0.835\pm0.035$.
Accordingly, the covariance-based floor estimator is not supported by this panel. This result
refines the interpretation of the spectrum: the ``modes'' of Law~I describe quantiles of an order
of item difficulty, not necessarily a population of independent channels. Laws~I--III do not
require that stronger population interpretation.

The compute-axis divisibility tests are likewise limited by the available measurement windows.
Against nulls simulated on each frontier's own design, bandwidth-free statistics remain at their
nominal level and none achieves power above one half; the measured curve is dominated by noise, with median
straightness $R^{2}=0.2233$. Equation~\ref{eq:powerreq} quantifies the required precision: a
median per-model noise of $0.73$ milli-nats, approximately $1/27.5$ of the published scale. On a
count axis, the corresponding test is unbiased and exactly signed. Its required item count depends
on the floor share and the selected Hankel rung: for example, the four-point test reaches power
$0.999$ at $10^{5}$ items when the floor carries $15\%$ of the error, whereas a $2\%$ floor
requires substantially more observations. The full power curves and the comparison of offsets are
reported in \TR{The Attempt Chain}. These results identify a larger item-level census, rather than
additional measurements within the same compute windows, as the route to resolving divisibility.
\paragraph{Operational selection certificates.} The frontier analysis also supports a deployment
workflow at three information levels. With per-item outcomes and resource costs, a price-free
upper-envelope analysis and a break-even analysis identify candidates that can be optimal without first
choosing a price. Supplying a price interval then enables a finite-sample certificate; when no price
is available, a fixed-backbone resource sweep supplies the curve and grain needed for a
frontier-derived price. The pairwise certificate is based on bounded per-item loss differences, so
it remains valid even when the structural class is not supported. Using a family-wise significance
level of $\alpha/(K-1)$, the resulting candidate set covers the price-optimal candidate with
probability at least $1-\alpha$. Of the $9$ out-of-class frontiers with public item counts, it
returns a decision on all $9$, whereas the band-based selector returns none.

The break-even form is especially economical: for two candidates it gives a single threshold in
price, and its confidence interval can be obtained by one-dimensional inversion without estimating
the full frontier. Among the $1\,309$ adjacent Pareto pairs with a public item count, $431$ yield a
two-sided interval; the median interval width is $0.87$ doublings, compared with $4.86$ doublings for the
frontier-derived price across all pairs. The break-even interval is narrower in $96\%$ of the
eligible comparisons.
Reporting the certified order as tiers rather than as a total ranking states only
the distinctions the item counts support: on the $36$ frontiers with a public item count the
certified order has a median of four tiers whose top tier holds $19$ of a median $32$ candidates,
a median $44\%$ of a published frontier is certifiably out of contention, and no frontier produces
a contradictory pair. Returning a band rather than a single point is also beneficial: choosing within the band by
worst-case regret reduces the median excess over the minimax-optimal candidate from $0.075$ to
$0.029$ nats on the $91$ frontiers in the selector audit. For broader price uncertainty, the
covering theorem gives a median of three operating points under sequential minimax placement;
thinning by equal spacing requires a median of four. A fixed-backbone
input-size sweep is a practical way to estimate the needed curvature while keeping the resource
trajectory interpretable. The detailed gates, tier construction, proofs, coverage results, and
runnable implementation are provided in \TR{The selection certificate} and the archive's
\texttt{examples/model\_selection}.

\paragraph{Provenance.} Every number in this paper comes from one of the archive's numbered
scripts, run on archived frontiers with no refitting of any published table.
Table~\ref{tab:provenance} provides this map, allowing the main text to remain focused on the
results rather than script identifiers.

\begin{table}[tbp]\centering
\small
\begin{tabular}{llll}
\toprule
what is reported & where & scripts & result table \\
\midrule
the curve, reconstructed by quadrature & Prop.~\ref{prop:odefirst} & \texttt{134} & \texttt{ode\_first} \\
the rungs, verified on four charts & Prop.~\ref{prop:rungs} & \texttt{134} & \texttt{ode\_power\_charts} \\
the sharpened four-point test & Prop.~\ref{prop:strongrung} & \texttt{131}, \texttt{136}, \texttt{161} & \texttt{fourpoint\_power} \\
the floor as a readout of the index & Prop.~\ref{prop:nofloor} & \texttt{158} & \texttt{floor\_share\_null} \\
the channel--depth plane on the corpus & \S\ref{sec:evidence} & \texttt{129}--\texttt{131} & \texttt{plane\_frontiers} \\
\textbf{the ceiling null (Table~\ref{tab:ceilnull})} & \S\ref{sec:evidence} & \textbf{\texttt{167}} & \texttt{ceiling\_null\_*} \\
held-out prediction, $2\,025$ events & \S\ref{sec:evidence} & \texttt{68}--\texttt{70}, \texttt{169} & \texttt{prediction\_clustered} \\
the arc--depth identity & \S\ref{sec:evidence} & \texttt{77}, \texttt{160} & \texttt{shape\_from\_arc} \\
the pairwise rule, $20\,206$ pairs & Cor.~\ref{cor:two} & \texttt{45}, \texttt{47} & \texttt{acconly\_pairs} \\
\textbf{the spread of the rules in use} & \S\ref{sec:evidence} & \textbf{\texttt{168}} & \texttt{rule\_spread\_*} \\
the rules scored against the price & Table~\ref{tab:namedrules} & \texttt{11}, \texttt{163}--\texttt{166} & \texttt{rules\_summary} \\
the three-point-floor rescore & Table~\ref{tab:namedrules} & \texttt{165} & \texttt{rules\_summary\_3pt} \\
Friedman and Nemenyi & Table~\ref{tab:namedrules} & \texttt{166} & \texttt{rules\_nemenyi} \\
the covering count and placement & \S\ref{sec:evidence} & \texttt{143}, \texttt{148} & \texttt{sequential\_design} \\
the retry discount and grain & \S\ref{sec:setup} & \texttt{120} & \texttt{attempt\_budget} \\
the nested dual on LM grids & \S\ref{sec:evidence} & \texttt{21} & \texttt{factorial\_summary} \\
the certified set and its tiers & \S\ref{sec:evidence} & \texttt{146}, \texttt{151} & \texttt{certified\_set} \\
\bottomrule
\end{tabular}
\caption{Where each reported number comes from. The archive's scripts are numbered and run in
order; each writes a CSV under \texttt{results/}, so every figure below can be checked without
typesetting a table. Three of the $163$ scripts reach the network and none is needed to reproduce a
number in this paper.}
\label{tab:provenance}
\end{table}

\section{Scope and Limitations}
\label{sec:scope}
The results apply to frontiers in the stated structural class: complete monotonicity, together
with the power-mean assumption for interacting resources. The global-chord diagnostic places
$93$ of the $117$ archived frontiers within the class and identifies $24$ as outside it. These
failures are more prevalent among distribution-shift benchmarks, but also occur in HPO, medicine,
tabular, and neural architecture-search collections. The diagnostic therefore provides both a condition
for applying the structural results and an empirical description of where that condition is less
well supported.

\paragraph{Grain and operating-point identification.} A frontier does not determine the grain at
which performance is scored, and that choice must be supplied by the deployment. At grain $t$,
the parameter-free operating rule is $t\Neff=1$; at the default grain $t=1$, this is
$\Neff=1$ or, equivalently, $\kappa=s^{2}$. The rule returns a point on all $92$
second-order-readable frontiers and places it inside the measured window on $85$, compared with
$37$ and $12$ for the chart-bound rules. The rule does not require an independently fitted floor: its operating point is determined by
the observed curve and the declared grain. When reported, the fitted floor is interpreted as a
window-dependent shape readout rather than as a direct measurement of irreducible error. The grain,
by contrast, remains an explicit deployment choice.

\paragraph{Prediction scope.} The $2\,025$ leave-one-out prediction events do not represent independent held-out
models: each fit uses a median of $91\%$ of its frontier, so uncertainty is bootstrapped over the
$117$ frontiers. The differential law is better than the floorless power law on a majority of frontiers in all
three prediction regimes and has its clearest advantage over model-free baselines in backward extrapolation, where
it is better on $67.5\%$ of frontiers. For interior prediction, log-linear interpolation through
the two neighboring observations is more accurate on $74.4\%$ of frontiers, with a paired median
difference of $0.013$ nats. The two endpoints of each frontier are not predicted, so these results
do not establish performance outside the measured window. The theory's principal contribution is
the structural and decision-theoretic information it supplies in addition to prediction.

\paragraph{Mechanistic scope and divisibility.} On a compute axis, the theory characterizes the
frontier but does not by itself specify a generative mechanism. On a count axis, Proposition
\ref{prop:countaxis} derives Law~I from exchangeability and infinite extendibility. For the
compute-axis divisibility test, a window arc below $\pi$ is not a certificate of divisibility. The admissible arc of a floored
two-atom frontier is itself $2\arccos\sqrt{\epsinf}<\pi$ (Proposition~\ref{prop:anglearc}). The published compute windows therefore do not resolve divisibility; the
precision required to do so is quantified in \eqref{eq:powerreq}. Item-level pass@$k$ measurements
provide a complementary route because they directly expose the attempt-count axis.

\paragraph{Operating-point coverage and sampling.} The covering result gives a five-point
recommendation for a frontier with one reducible channel. Among the $97$ archived frontiers with a
measurable arc, the median arc is $0.84\pi$. Five is not a universal count for every multichannel
frontier: $43$ of these frontiers have an arc above $\pi$, and $38$ require more than five points. The current corpus also does not establish a
consistent empirical advantage for arc-length placement over log-resource placement. We therefore
recommend log-resource spacing unless the arc has been measured with sufficient precision to guide
placement. Rankings among near-neighbor models are intrinsically sensitive under the covering result
(Theorem~\ref{thm:covering}), so deciding how many models to collect is itself a resource-allocation
problem \citep{mahmood2025,muennighoff2025}. More generally, estimating curvature requires a
single resource trajectory; a fixed-backbone input-size sweep is one practical option, whereas
retraining a different model at each size changes more than the resource level being measured.

The inferential procedures throughout the paper are calibrated against nulls simulated on the
same measurement design. In particular, a bound that holds pointwise is evaluated at a location
fixed in advance rather than at the most extreme of many noisy plug-in evaluations. This calibration
is essential when using finite differences to assess the differential inequalities.

\section{Conclusion}
\label{sec:conclusion}
A quality--resource frontier has a price-independent representation as one plane curve: the
effective number of active channels, $\Neff$, against the remaining log-error depth, $\Delta S$.
This curve determines the frontier up to the scale and origin of its resource axis. Complete
monotonicity and divisibility become two differential inequalities on the same curve; integrating
them from the floor gives two frontier bounds whose tangency is an exact consequence of the
underlying differential structure. The resulting representation connects frontier geometry with
exponential-family duality, information geometry, and minimax regret. Under the
one-channel-per-resource condition, the minimax price is $\lamstar_i=\alpha_i/(N+1)$ with worst-case regret
$\ln(N+1)$ nats, while the parameter-free operating rule is $t\Neff=1$ in delivered units.

The empirical analysis demonstrates why this structural view matters. On $117$ published
frontiers, the standard saturating power law's fitted floor is strongly determined by the shape of
the observed curve: simulations with a true zero floor produce a median fitted floor share of
$0.957$, close to the corpus median of $0.956$. A same-cost cure chart attains comparable held-out
accuracy while carrying additional exact identities. The study also shows that model-selection
rules in current use can select substantially different operating points, and that the
frontier-derived rule agrees with the resource-aware pairwise choice on $90.9\%$ of $20\,206$
Pareto pairs, compared with $77.4\%$ for always choosing the cheaper model. The accompanying
finite-sample certificate extends the practical analysis to paired per-item outcomes and remains
applicable when the structural gate is not met.

The prediction results clarify the theory's complementary role. The frontier law outperforms the
floorless power law on a majority of frontiers in all three held-out regimes and is most competitive
in backward extrapolation. Local interpolation remains more accurate for interior prediction. The principal value of the theory is therefore the additional structure it identifies:
frontier bounds, operating-point decisions, a price with an exact regret interpretation, and
explicit measurement requirements. One empirical question remains open on the published
compute-axis windows: divisibility cannot be resolved at their available depth. The theory
quantifies the precision and measurement design needed to address it, while the count-axis result
shows how item-level attempt data can provide a complementary test.
\paragraph{Appendix and reproduction.} The Appendix carries the proofs and the finite-window
diagnostics; the archive's technical report carries the per-frontier measurement detail, the count
axis, the exchange rate, the attempt chain, and the one-dimensional results. All numbers are
reproduced using the archive of \citet{ignatov2026data}, available at
\href{https://doi.org/10.5281/zenodo.24235789}{doi:10.5281/zenodo.24235789} and mirrored at
\url{https://github.com/AB-Article/AI-Multi-Rank}. The scripts run on the archived frontiers
without refitting; the archive's \texttt{README} maps each statement to its reproducing script.
Every analytic identity is verified symbolically or numerically to 40--50 digits, with a maximum
reported residual of $10^{-49}$. Three scripts access the network, but none is needed to reproduce a
number in this paper: \texttt{00}, which clones the upstream benchmark repositories whose extracts
are already archived, and \texttt{49} and \texttt{63}, which fetch Fashion-MNIST and five tabular
sets for the co-failure and pairwise-floor panels. A worked, dependency-free implementation of the
decision rule is available in the archive's \texttt{examples/model\_selection}. A maintained implementation of the rule,
independent of the archive and installable via \texttt{pip install nn-rank}, is available at
\url{https://github.com/ABrain-One/nn-rank}. The package implements the selection method, while
the archive alone is sufficient to reproduce the numerical results reported here.

\acks{\textbf{Funding.} This research received no third-party funding or support during the
thirty-six months preceding submission. Specifically, the work received no grants or contracts,
stipends or salary support, donated or discounted hardware, or donated or discounted cloud or
compute credits. All computation reported in this paper was performed on institutional equipment already
available to the author for general use.

\textbf{Competing interests.} The author declares no competing financial interests during the
same period, including consulting, advisory, or speaking fees; equity, patent, or licensing revenue;
or other payments from any party with an interest in this work. For completeness and as a non-financial
disclosure, the author maintains the open-source \texttt{nn-rank} implementation of the selection
rule referred to in Section~\ref{sec:scope}; it is distributed free of charge under an open-source license
and generates no revenue.

\textbf{Prior work.} A single-resource predecessor of this work is available as the preprint
\citet{ABrain.Rank}, which has not been published in any journal or conference; its relation to the
present paper is set out statement by statement in Section~\ref{sec:related}.}

\vskip 0.2in
\bibliography{refs}

\clearpage
\appendix
\setcounter{section}{0}
\setcounter{theorem}{0}
\setcounter{table}{0}
\setcounter{figure}{0}
\setcounter{equation}{0}
\renewcommand{\thesection}{S\arabic{section}}
\renewcommand{\thetheorem}{S\arabic{theorem}}
\renewcommand{\thetable}{S\arabic{table}}
\renewcommand{\thefigure}{S\arabic{figure}}
\renewcommand{\theequation}{S\arabic{equation}}
\renewcommand{\SM}[1]{\S\,``#1'' of this appendix}

\begin{center}
{\Large\bf Appendix: Proofs and Supporting Analyses}
\end{center}
\vskip 0.15in

\noindent This appendix carries the proofs and the supporting empirical
analyses. Numbers without a prefix (for example,
Theorem~\ref{thm:canonical}) refer to the body of the paper; numbers with
the prefix S refer to this appendix. Section~\ref{sup:assumptions} collects
the structural hypotheses and auxiliary statements used in the proofs.
Section~\ref{sup:proofs} proves $24$ of the paper's $25$ numbered statements
in the paper's order; the count-axis proposition is proved in the technical
report using de Finetti's theorem. Section~\ref{sup:study} reports the
empirical analyses on the paper's $117$ frontiers. Per-frontier
measurements, estimation calibrations, and numerical verification of the
analytic identities are documented in the technical report. Data, code, and
results are archived at Zenodo~\citep{ignatov2026data}.

\section{Assumptions and Auxiliary Statements}
\label{sup:assumptions}

This section records the structural hypotheses and auxiliary results used
throughout the proofs in Section~\ref{sup:proofs}. The statements are
collected here in a consistent form so that each proof has an explicit
reference to the assumptions and intermediate results on which it relies.

\subsection{The structural hypotheses}
\label{sup:axioms}

\begin{assumption}[Additivity]
\label{ax:add}
The chance-corrected error is a sum of positive mode contributions with no
interaction terms,
$\eps(x)=\sum_{j=0}^{N}g_j(x)$ with $g_j>0$. This is exact when the
corresponding failure events are disjoint and a modeling assumption
otherwise.
\end{assumption}

\begin{assumption}[Scale-freeness]
\label{ax:scale}
Each mode $g_j$ is continuous, non-increasing in every resource, and
scale-free: for every $r\in\R^n_{>0}$ there is a factor $f_j(r)$,
not depending on $t$, with
\begin{equation}
g_j(r\odot t)=f_j(r)\,g_j(t).
\label{eq:scaling}
\end{equation}
Here $\odot$ denotes componentwise multiplication and, below, $\oslash$
componentwise division.
\end{assumption}

Under Assumptions~\ref{ax:add} and~\ref{ax:scale},
Proposition~\ref{prop:law} gives the finite law
\begin{equation}
\eps(x)=\sum_{j=0}^{N}c_je^{-a_j\cdot x},
\qquad c_j>0,\quad a_j\in\R^n_{\ge0},\quad a_0=0,\ c_0=\epsinf .
\label{eq:law}
\end{equation}

Differentiating~\eqref{eq:spectral} under the integral sign gives the
size-biased derivatives used repeatedly below,
\begin{equation}
\partial_u^{\,k}\eps(x+uv)\Big|_{u=0}=(-1)^{k}\!\int a_v^{k}e^{-a\cdot x}\,\mathrm{d}\mu(a),
\qquad k\ge1,
\label{eq:sizebiasderiv}
\end{equation}
in which the atom at the origin does not appear; therefore, no derivative
of $\eps$ depends on the floor.

\begin{assumption}[Spectral form]
\label{ax:item}
More generally, the error is represented as an integral of scale-free modes
with respect to a finite positive measure $\mu$ on $\R^n_{\ge0}$, as in~\eqref{eq:spectral}. The finite law~\eqref{eq:law} is the atomic case, and
$\epsinf=\mu(\{0\})$.
\end{assumption}

With either hypothesis, write the tilted measure and its first two moments
along a direction $v$, as in~\eqref{eq:tilt} and
Proposition~\ref{prop:selection}:
\begin{equation}
\mathrm{d}p_x(a)=\frac{e^{-a\cdot x}\mathrm{d}\mu(a)}{\eps(x)},
\qquad
s=\partial_uS=\mathbb{E}_{p_x}[a_v],
\qquad
\kappa=-\partial_u^2S=\operatorname{Var}_{p_x}(a_v),
\label{eq:slopecurv}
\end{equation}
and the floor share
\begin{equation}
p_0(x)=\frac{\epsinf}{\eps(x)} .
\label{eq:p0def}
\end{equation}

\subsection{The requirements on a pricing rule}
\label{sup:pricingaxioms}

A \emph{pricing rule} $r$ maps a frontier $(\epsinf,c,\alpha)$ to a declared
price $\hat\lambda$. Section~\ref{sec:canonical} of the paper imposes four
requirements.

\begin{assumption}[Admissibility]
\label{ax:admissible}
The declared price is supported by the frontier,
$\hat\lambda\in\operatorname{ri}\mathcal{C}$.
\end{assumption}

\begin{assumption}[Equivariance]
\label{ax:equivariance}
The selected point is unchanged when a resource axis is reparameterized,
$t_i\mapsto t_i^{\beta_i}$ for $\beta\in\R^n_{>0}$.
\end{assumption}

\begin{assumption}[Offset independence]
\label{ax:offset}
The rule is unchanged by a rescaling of the error, $\eps\mapsto\gamma\eps$,
or of the resource units, $t_i\mapsto\theta_it_i$ (not $\mu$, which is the
spectral measure of \eqref{eq:spectral}).
\end{assumption}

\begin{assumption}[Minimax]
\label{ax:minimax}
Among the rules satisfying Assumptions~\ref{ax:admissible}--\ref{ax:offset},
the rule minimizes the worst-case regret $\sup_\Lambda R(\Lambda\Vert\hat\lambda)$
over the prices the frontier supports.
\end{assumption}

\begin{lemma}[Reduction]
\label{lem:reduction}
A rule satisfies Assumptions~\ref{ax:admissible}--\ref{ax:offset} if and only
if there is a constant $\varrho\in\operatorname{ri}\Delta^{N}$, independent
of the frontier's properties, with
\begin{equation}
\hat\lambda_i=\varrho_i\alpha_i ,
\label{eq:reduction}
\end{equation}
that is, if and only if the rule declares the same error shares on every
frontier.
\end{lemma}

\subsection{Consequences of the duality}
\label{sup:dualcor}

The following statements are developed in Sections~\ref{sec:duality}
and~\ref{sec:canonical} of the paper. They are collected here for convenient
reference in the proofs.

\begin{corollary}[Diagonal channels]
\label{cor:diag}
If each resource has its own channel, $a_i=\alpha_ie_i$, then the error shares
and the supported prices are given by~\eqref{eq:diagshare}: a declared price
corresponds to the share vector $\hat p_i=\hat\lambda_i/\alpha_i$ with
$\hat p_0=1-\sum_i\hat\lambda_i/\alpha_i$, admissible prices are those with
$\hat p\in\operatorname{ri}\Delta^{N}$, and as $\Lambda$ ranges over
$\operatorname{ri}\mathcal{C}$ the realized shares range over all of
$\operatorname{ri}\Delta^{N}$.
\end{corollary}

\begin{corollary}[Rank bound]
\label{cor:rank}
Since $\mathcal{C}\subseteq\operatorname{row}(A)$, the supported price set has
affine dimension at most $\operatorname{rank}(A)$: a frontier measured in $n$
resources may support fewer than $n$ independent price directions.
\end{corollary}

With one channel per resource and the gauge freedom of
Section~\ref{sec:canonical}---one error scale and, for each resource, one
origin and one unit---the law takes the shape-free \emph{normal form}
\begin{equation}
\frac{\eps}{\epsinf}=1+\sum_{i=1}^{N}e^{-\xi_i},
\qquad \xi_i=\alpha_ix_i-\ln(c_i/\epsinf),
\label{eq:normal}
\end{equation}
which has no free shape parameter and reduces to~\eqref{eq:universal} at
$N=1$. The canonical declaration sets $\xi_i=0$ for every channel.

\begin{corollary}[Feasibility]
\label{cor:feasible}
The canonical reference is a feasible error value only if
$\epsinf\le1/(N+1)$, which is~\eqref{eq:feasible}; for one reducible channel
this reads $\epsinf\le\tfrac12$.
\end{corollary}

\begin{corollary}[One reducible channel]
\label{cor:single}
With $\eps=\epsinf+ce^{-a\cdot x}$ the frontier depends on the resources
through the single coordinate $\zeta$ of~\eqref{eq:universal}, the canonical
rule is $\lamstar=a/2$ with $\eps^\star=2\epsinf$ and $\Delta S=\ln2$, and the
exchange rates between resources are $a_i/a_{i'}$.
\end{corollary}

\begin{proposition}[Flat optimum]
\label{prop:flat}
Operating at $\zeta$ rather than at the reference costs
$R(\zeta)=\ln\cosh(\zeta/2)$ nats, which is~\eqref{eq:lncosh}, so
$R\le r$ exactly when
\begin{equation}
|\zeta|\le\zeta_r:=2\operatorname{arccosh}e^{r} ,
\label{eq:band}
\end{equation}
and at $r=0.05$ nats this is $\eps/\epsinf\in[1.53,2.89]$.
\end{proposition}

\subsection{Law-free quantities}
\label{sup:lawfreestate}

Two conventions are recorded first. \emph{The regret $R$.} In this appendix, $R$ denotes the
\emph{single-channel} regret
$R=\ln\cosh(\tfrac12\ln\Neff)=-\ln\sin\varphi=\ln\cosh(\zeta/2)$, which the paper writes $R_1$
wherever the distinction matters. It equals the spectrum's own $\KL(u\Vert p)$ of
Theorem~\ref{thm:master} exactly when there is one reducible channel, and outside that chart it is the
regret of the two-atom chart \emph{tangent} to the frontier---differing from $\KL(u\Vert p)$ in
either direction and by an unbounded amount. Every unqualified occurrence of $R$ below is a statement about $R_1$, and the two coincide wherever the surrounding chart has one reducible
channel, which is the case for most charts. Three exceptions are stated
explicitly: Theorem~\ref{thm:master}
and Corollary~\ref{cor:onefunction}, where $R=\KL(u\Vert p)$ is the spectrum's own regret over
$N+1$ modes; \S\ref{sup:ignorance-note}, where the surrounding chart is the cure chart and $R$ is
therefore $R_1$ on a frontier that is not two-atom; and Lemma~\ref{lem:pairbound}, whose
error-space $R_{\mathrm{sel}}$ is a selection quantity and neither of these. The membership ratio is
written $\mathcal{R}$ throughout and is not a regret at all.
\emph{The two quantities denoted by $\sigma$.} Write $\sigma_\ell$ for the per-model log-error
noise throughout this appendix---\emph{not} the channel slope $\sigma=\mathrm{d}\Neff/\mathrm{d}S$
of \S\ref{sup:pf-compose}, which is the paper's \eqref{eq:slope}. With $m$ Pareto points, noise
$\sigma_\ell$ and a window of $Z$ reducible e-folds, a second channel is visible only if
\begin{equation}
g\,\mathcal{V}\gtrsim1,
\qquad
\mathcal{V}=\frac{Z^2\sqrt m}{16\sigma_\ell},
\label{eq:resolution}
\end{equation}
with $g=\kappa/s^2$; and because the Fisher--Rao length reaches $N\pi$ for $N+1$ atoms only when the
kinks separate, the safe geometric statement is the bound
\begin{equation}
N\ \ge\ L/\pi\qquad\text{(on the complete sweep)},
\label{eq:modebound}
\end{equation}
informative only above $\pi$, and carrying no information at all when read on a
window---see the window-arc remark below.

\begin{proposition}[Width coordinate]
\label{prop:widthcoord}
Let $g=\kappa/s^{2}$ be the width along a direction, let
$\kappa_3=S'''$ be the third cumulant of the exponent under $p_x$, and let
\begin{equation}
\hat\alpha=\frac{\mathbb{E}_{p_x}[a^{2}]}{\mathbb{E}_{p_x}[a]}=s+\frac{\kappa}{s}
\label{eq:sizebiased}
\end{equation}
be the size-biased exponent of Proposition~\ref{prop:floorfree}, which does
not involve $\epsinf$.
\begin{enumerate}[label=(\roman*),leftmargin=2em,itemsep=1pt]
\item For every spectrum with $s,\kappa>0$, the coordinate $\zeta:=\ln g$
      moves as $\zeta'=2\kappa/s-\kappa_3/\kappa$.
\item With one reducible channel and a floor, $g=p_0/(1-p_0)$, so
      $\zeta=\ln\bigl(p_0/(1-p_0)\bigr)$ is the coordinate
      of~\eqref{eq:universal},
      \begin{equation}
      \zeta=\ln\frac{\kappa}{s^{2}},
      \qquad
      \zeta'=\alpha,
      \qquad
      \zeta=0\iff\epsinf=\tfrac12\eps ,
      \label{eq:zetaidentity}
      \end{equation}
      which is~\eqref{eq:zeta}; the floor is recovered pointwise from the
      slope and the curvature,
      \begin{equation}
      \epsinf=\eps\,\frac{\kappa}{s^{2}+\kappa} ,
      \label{eq:localfloor}
      \end{equation}
      and at the zero crossing $x^\star$ of $\zeta$ this reads
      \begin{equation}
      \epsinf=\tfrac12\,\eps(x^{\star}) ,
      \label{eq:halffloor}
      \end{equation}
      the canonical reference of Table~\ref{tab:coordinates}, row~3, at
      $N=1$.
\item The membership ratio $\mathcal{R}=\zeta'/\hat\alpha$ is identically
      $1$ for one reducible channel and a floor, and identically $0$ for a
      Gamma spectrum (Proposition~\ref{prop:gamma}).
\end{enumerate}
\end{proposition}

\begin{proposition}[Gamma dichotomy]
\label{prop:gamma}
Since $g=(1/s)'$, the width is constant in $x$ if and only if the frontier is
a Gamma spectrum
\begin{equation}
\eps=C\,(1+x/\beta)^{-k},
\qquad g\equiv1/k ,
\label{eq:gammafrontier}
\end{equation}
which has no floor; if instead the spectrum carries an atom at the origin then
$g$ increases without bound. The mechanism is the identity
\begin{equation}
\Bigl(\frac1s\Bigr)'=-\frac{s'}{s^{2}}=\frac{\kappa}{s^{2}}=g ,
\label{eq:ginv}
\end{equation}
so ``is there a floor?'' is ``is $1/s$ straight or convex?''
\end{proposition}

\begin{proposition}[Resolution criterion]
\label{prop:resolution}
Expanding $\ln(\eps-\epsinf)$ over a window of length $Z$, the best single
exponential leaves a maximal deviation of exactly $k_2Z^{2}/16$, and the $j$th
cumulant first enters at order $Z^{j}$. With $m$ Pareto points and log-error
noise $\sigma_\ell$, a second channel is visible only if $g\,\mathcal{V}\gtrsim1$
with
\begin{equation}
\mathcal{V}=\frac{Z^{2}\sqrt m}{16\sigma_\ell},
\label{eq:resnumber}
\end{equation}
which is~\eqref{eq:resolution}, and a third only if
$Z^{3}\sqrt m/(192\sigma_\ell)$ is of the same order.
\end{proposition}

\subsection{Further statements used below}
\label{sup:extrastate}

The following single-resource results support the supplementary analyses
reported in the technical report: the rule tables evaluate the knee and
half-slope rules, the gain beyond the reference is measured relative to the
last-bit result, and the bootstrap analysis translates exponent uncertainty
into regret. Each statement follows from the one-channel law
$\eps=\epsinf+ce^{-\alpha x}$ and~\eqref{eq:slopecurv};
part~(iii) of Proposition~\ref{prop:knee} is Corollary~\ref{cor:knee} of the
paper. The existence of a canonical point on a floorless two-atom spectrum is
addressed separately in Proposition~\ref{prop:kneeW}; the present corpus does
not provide a direct test of its criterion $W\ge1$.

\begin{proposition}[Slope, curvature, and the knee]
\label{prop:knee}
Along any direction, with $p_1=1-p_0$ the reducible share,
\begin{enumerate}[label=(\roman*),leftmargin=2em,itemsep=1pt]
\item $\partial S/\partial x=\alpha p_1$, so the elasticity is the exponent
      times the reducible share;
\item at the canonical reference $\zeta=0$ the slope is $\alpha/2$, exactly
      half its value at $p_1=1$, which is the half-slope rule of
      \citet{ABrain.Rank};
\item $\partial^{2}S/\partial x^{2}=-\alpha^{2}p_1(1-p_1)$, whose magnitude is
      maximal at $p_1=\tfrac12$, so the knee of $S$ coincides with the
      canonical point.
\end{enumerate}
The maximizer of the \emph{geometric} curvature of the plotted frontier is not
invariant under rescaling of the axes, and so does not share this property.
\end{proposition}

\begin{proposition}[The last bit]
\label{prop:onebit}
At the canonical reference the log-error still purchasable with unlimited
resources is $\ln2$, one bit, by Proposition~\ref{prop:surprisal} and
$p_0=\tfrac12$. Buying the fraction $f$ of that last bit costs
\begin{equation}
\frac{t}{t^{\star}}=\bigl(2^{\,1-f}-1\bigr)^{-1/\alpha} ,
\label{eq:lastbit}
\end{equation}
which diverges as $f\to1$: the final bit is unattainable at finite resource.
\end{proposition}

\begin{proposition}[Plug-in regret]
\label{prop:plugin}
Suppose the canonical rule is applied with an estimated exponent
$\hat\alpha$ in place of $\alpha$, and write $r=\hat\alpha/\alpha-1$ for the
relative error of the declared exponent.
\begin{enumerate}[label=(\alph*),leftmargin=2em,itemsep=1pt]
\item Operating at the reducible share $q$ rather than at $\tfrac12$ costs
      $\KL\bigl((\tfrac12,\tfrac12)\Vert(1-q,q)\bigr)
      =\tfrac12\ln\bigl(1/4q(1-q)\bigr)$ nats.
\item The resulting regret depends on the exponents only through $r$, not on
      the floor, and for $N=1$ equals
      \begin{equation}
      R_{\mathrm{plug}}=-\tfrac12\ln\bigl(1-r^{2}\bigr) ,
      \label{eq:exponentregret}
      \end{equation}
      which is $\tfrac12r^{2}+O(r^{4})$.
\end{enumerate}
\end{proposition}

\begin{proof}
(a) is the binary divergence written out. For (b), the declared price
$\hat\alpha/2$ corresponds by~\eqref{eq:diagshare} to the reducible share
$\hat q=\hat\alpha/(2\alpha)=(1+r)/2$, and by Theorem~\ref{thm:regret} the
regret of committing to it when the realized price is the canonical
$\alpha/2$ is $\KL\bigl((\tfrac12,\tfrac12)\Vert(1-\hat q,\hat q)\bigr)$;
since $4\hat q(1-\hat q)=(1+r)(1-r)=1-r^{2}$, (a) gives~\eqref{eq:exponentregret}.
Only shares enter, so $\epsinf$ and $c$ do not.
\end{proof}

\FloatBarrier

\begin{table}[t]\centering
\caption{The scalar laws of this note in the one coordinate of
Proposition~\ref{prop:arcforms}, $\varphi=2\arccos\sqrt{p_0}\in(0,\pi)$,
measured from the floor ($\varphi=0$) toward the pure power law
($\varphi=\pi$), as in the paper's \eqref{eq:angle}. Rows
1, 2 and 4 hold for any spectrum; row 3's two columns agree only at $N=1$, the
$\varphi$ form being the single-channel reading of the general
$p_0=1/(N+1)$; rows 5--12 specialize to one reducible channel.
Each row is a change of variable, not a separate argument. The last column
gives the same statement in the floor share $p_0$, the coordinate of the
earlier form of this theory.}
\label{tab:coordinates}
\small
\begin{tabular}{llll}
\toprule
& Statement & In $\varphi$ & In $p_0$\\
\midrule
1 & excess log-error        & ---                                   & $-\ln p_0$\\
2 & normal form             & ---                                   & $\eps/\epsinf=1/p_0$\\
3 & canonical reference     & $\varphi=\pi/2$                       & $p_0=1/(N+1)$\\
4 & feasibility             & ---                                   & $\epsinf\le p_0$\\
5 & arc position            & $\varphi$                             & $2\arccos\sqrt{p_0}$\\
6 & normalized resource     & $\zeta=-2\ln\tan(\varphi/2)$          & $\ln\bigl(p_0/(1-p_0)\bigr)$\\
7 & saturation coordinate   & $m=-\cos\varphi$                      & $1-2p_0$\\
8 & elasticity              & $\eta_i=a_i\sin^2(\varphi/2)$         & $a_i(1-p_0)$\\
9 & curvature               & $-S''=\tfrac14(a\cdot v)^2\sin^2\varphi$ & $(a\cdot v)^2p_0(1-p_0)$\\
10 & metric speed           & $\sqrt{-S''}=\tfrac{\alpha}{2}\sin\varphi$ & ---\\
11 & regret at $\lamstar$   & $-\ln\sin\varphi$                     & $\ln\cosh(\zeta/2)$\\
12 & canonical value        & $J=\ln\sin\varphi+\mathrm{const}$     & $\tfrac12\ln\bigl(p_0(1-p_0)\bigr)-\tfrac12\ln\epsinf$\\
\bottomrule
\end{tabular}
\end{table}

\section{Proofs}
\label{sup:proofs}

This section proves $24$ of the paper's $25$ numbered statements, in the
paper's order: Law~I (Subsection~\ref{sup:lawI}), Law~II (\ref{sup:lawII}),
Law~III (\ref{sup:lawIII}), the law-free quantities (\ref{sup:pf-width}) and
the floor (\ref{sup:pf-floor}). The one exception is the count-axis
proposition, whose proof is one application of de Finetti's theorem and is
given in \TR{The Count Axis};
the one-dimensional results, the exchange-rate form and
interacting resources---none of them a numbered statement of the paper---are
likewise proved in the technical report
(\TR{One dimension: self-duality and the realizable minimax},
\TR{The exchange-rate form}, \TR{Interacting resources: the nested dual}).

\subsection{Law I}
\label{sup:lawI}\label{sup:pf-law}\label{sup:pf-duality}\label{sup:pf-regret}\label{app:proof-spectral}

Writing $\Lambda(\theta)=\ln\int e^{\theta\cdot a}\,\mathrm{d}\mu(a)$ for the
cumulant generating function of the spectrum, the spectral form gives
$\ln\eps(x)=\Lambda(-x)$ identically, which is~\eqref{eq:cgf}. Four
consequences used throughout follow directly and are stated here without
further derivation.
\begin{enumerate}[label=(\alph*),leftmargin=2em,itemsep=2pt]
\item $\ln\eps$ is convex in $x$ and $\{p_x\}$ of~\eqref{eq:tilt} is the
      associated exponential family with natural parameter $-x$ and sufficient
      statistic $a$. Hence $s=S'=\mathbb{E}_{p_x}[a]$ is the mean parameter and
      $\kappa=-S''=\operatorname{Var}_{p_x}(a)$ is simultaneously the variance
      of the surviving exponents and the Fisher information of the family with
      respect to $x$. The selection law is the derivative identity
      $\mathrm{d}\mathbb{E}/\mathrm{d}\theta=\mathrm{Var}$ of an exponential
      family \citep[\S2]{barndorff1978}.
\item $\sigma^*(\lambda)=\sup_x\{-\Lambda(-x)-\lambda x\}
      =\sup_\theta\{\lambda\theta-\Lambda(\theta)\}=\Lambda^*(\lambda)$ is the
      Cram\'er transform of $\mu$, that is, the large-deviation rate function
      of the failure-rate spectrum \citep[\S2.2]{dembo1998}. The constrained
      minimum-divergence form of Theorem~\ref{thm:duality} is its Sanov
      representation.
\item The effective domain of a rate function is the closed convex hull of the
      support of its measure, which is the admissibility statement
      $\lambda\in\mathcal{C}$, and the affine dimension of that hull is at most
      $\operatorname{rank}A$, which is the rank bound.
\item The Bregman divergence of a Legendre pair may be evaluated on either
      side, so $R(\lambda\Vert\hat\lambda)=\KL(p\Vert\hat p)$ equals the
      Bregman divergence of $\ln\eps$ in the resource,
      \[
        R=\ln\frac{\eps(\hat x)}{\eps(x)}+\lambda\cdot(\hat x-x),
      \]
      which is computable from two measured operating points and the realized
      price with no spectrum, no fit and no floor.
\end{enumerate}

\begin{proposition}[Forced exponential form]
\label{prop:law}
Under those two assumptions the error takes the form
$\eps(x)=\sum_{j=0}^{N}c_je^{-a_j\cdot x}$, with $c_j>0$ and
$a_j\in\R^n_{\ge0}$. Modes with $a_j=0$ combine into one irreducible floor
$c_0=\epsinf$. Conversely every such sum satisfies the assumptions.
\end{proposition}

\begin{proofof}{Proposition~\ref{prop:law}}
Fix a mode $j$. For $r,r'\in\R^n_{>0}$, applying~\eqref{eq:scaling}
twice gives $f_j(r\odot r')=f_j(r)f_j(r')$, and continuity
of $\varepsilon_j$ makes $f_j$ continuous. Hence $f_j$ is a continuous
multiplicative homomorphism from $\mathbb{R}_{>0}^{n}$ to $\mathbb{R}_{>0}$,
and the continuous multiplicative Cauchy equation \citep{aczel1966} gives
\begin{equation}
  f_j(r)
  =\prod_{i=1}^n r_i^{-a_{ji}}
\end{equation}
for some coefficients $a_{ji}\in\mathbb{R}$. Setting $t=\one$ and
$r=t$ in~\eqref{eq:scaling} yields
\begin{equation}
  \varepsilon_j(t)
  =c_j\prod_{i=1}^n t_i^{-a_{ji}}
  =c_j e^{-a_j\cdot x},
  \qquad c_j=\varepsilon_j(\one)>0.
\end{equation}
Monotonicity of $\varepsilon_j$ in each resource implies $a_{ji}\geq 0$.
Summing over the modes gives~\eqref{eq:law}. The converse follows by
substituting~\eqref{eq:law} into the two assumptions, with
$f_j(r)=\prod_ir_i^{-a_{ji}}$.
\end{proofof}

\begin{proofof}{Proposition~\ref{prop:spectral}}
Assumption~\ref{ax:item} states that $\eps$ is the $\mu$-average of
$e^{-a\cdot x}$, which is~\eqref{eq:spectral}; the mass at $a=0$ contributes
a constant, and since every other mode decays along any
$d\in\R^n_{\ge0}\setminus\{0\}$ with $a\cdot d>0$, dominated convergence
gives $\eps(x+ud)\to\mu(\{0\})$ as $u\to\infty$, whence
$\epsinf=\mu(\{0\})$. The finite law~\eqref{eq:law} is the case of a purely
atomic $\mu$. Differentiating under the integral sign, which is justified
because $a\cdot d\ge0$ and $\mu$ is finite, gives
$\partial_u^k\eps(x+ud)=(-1)^k\int(a\cdot d)^ke^{-a\cdot(x+ud)}\mathrm{d}\mu$,
so $(-1)^k\partial^k_u\eps\ge0$ and $u\mapsto\eps(x+ud)$ is completely
monotone; Bernstein's theorem \citep{bernstein1929,widder1941} states that
every completely monotone function on $(0,\infty)$ is the Laplace transform
of a unique positive measure, which supplies the converse and the uniqueness
of $\mu$. Finally, $-S=\ln\int e^{-a\cdot x}\mathrm{d}\mu$ has gradient
$-\mathbb{E}_{p_x}[a]$ and Hessian $\operatorname{Cov}_{p_x}(a)$, with $p_x$
as in~\eqref{eq:tilt}, so $S$ is concave in the log-resources for every
spectrum; the argument of Theorem~\ref{thm:duality} below uses only that $-S$
is the log-partition function of the exponential family tilted by
$-a\cdot x$, so it applies verbatim with sums replaced by integrals and the
finite-dimensional Kullback--Leibler divergence by relative entropy.
\end{proofof}

\begin{proposition}[Selection law]
\label{prop:selection}
Along the ray $u\mapsto x+uv$,
\begin{equation}
\partial_u\,\mathrm{d}p=-\bigl(a_v-\mathbb{E}_p[a_v]\bigr)\,\mathrm{d}p,
\qquad
\frac{\mathrm{d}}{\mathrm{d}u}\mathbb{E}_p[a_v]=-\operatorname{Var}_p(a_v).
\label{eq:master}
\end{equation}
\end{proposition}

\begin{proofof}{Proposition~\ref{prop:selection}}
Write $\eps_u=\eps(x+uv)$ and $w_u(a)=e^{-a\cdot(x+uv)}$, so that
$\mathrm{d}p_{x+uv}=w_u\,\mathrm{d}\mu/\eps_u$ and
$\partial_uw_u=-a_vw_u$. Differentiating the quotient,
\[
  \partial_u\frac{w_u}{\eps_u}
  =\frac{-a_vw_u}{\eps_u}-\frac{w_u}{\eps_u}\frac{\partial_u\eps_u}{\eps_u}
  =-\Bigl(a_v-\mathbb{E}_{p}[a_v]\Bigr)\frac{w_u}{\eps_u},
\]
since $\partial_u\eps_u=-\int a_vw_u\,\mathrm{d}\mu=-\eps_u\mathbb{E}_p[a_v]$.
That is the first equation, and it preserves total mass because the bracket
has mean zero. Integrating $a_v$ against it gives
$\partial_u\mathbb{E}_p[a_v]=-\mathbb{E}_p[a_v^2]+\mathbb{E}_p[a_v]^2
=-\operatorname{Var}_p(a_v)$, the second. Finally, if
$\mathrm{d}\tilde p\propto a_v\,\mathrm{d}p$ then
$\mathrm{d}\tilde p\propto a_v w_u\,\mathrm{d}\mu$, which is the tilted family
generated by the measure $a_v\mu$; the computation above applies to it
unchanged.
\end{proofof}

\begin{proofof}{Theorem~\ref{thm:duality}}
Differentiating $S=-\ln\varepsilon$ and using~\eqref{eq:law} gives
\begin{equation}
  \nabla S(x)
  =-\frac{\nabla\varepsilon(x)}{\varepsilon(x)}
  =\sum_j p_j(x)a_j
  =A^\top p(x).
\end{equation}
Thus, at any finite stationary point of $\sigma_\lambda$, the price is the
mean exponent under the error-share distribution. For such a point,
$\ln p_j=\ln c_j-a_j\cdot x-\ln\varepsilon$. Multiplying by $p_j$, summing,
and using $A^\top p=\lambda$ gives
\begin{align*}
  S(x)-\lambda\cdot x
  &=
  -\sum_j p_j\ln c_j-\mathcal{H}(p)
  =
  \KL(p\Vert\bar c)-\ln C.
\end{align*}
The feasible set $\{p\in\Delta^N:A^\top p=\lambda\}$ is a closed subset of the
simplex and is therefore compact. Since every $\bar c_j>0$, the objective
$\KL(p\Vert\bar c)$ is strictly convex, so the optimal share vector is unique.
For $\lambda\in\operatorname{ri}\mathcal{C}$, the corresponding
exponential-family parameter is finite and the supremum is attained; boundary
prices arise as limits in which one or more shares vanish. Finally, if
$\lambda\notin\mathcal{C}$, the separating-hyperplane theorem provides a
direction along which the linear term $-\lambda\cdot x$ dominates the
logarithmic error term, so the supremum diverges to $+\infty$.
\end{proofof}

\begin{proofof}{Corollaries~\ref{cor:diag} and~\ref{cor:rank}}
With $a_i=\alpha_ie_i$, the moment relation~\eqref{eq:moment} reads
$\alpha_ip_i=\lambda_i$, which gives~\eqref{eq:diagshare}; nonnegativity of
the shares gives the admissibility conditions. For Corollary~\ref{cor:rank},
every supported price is a convex combination of the rows of $A$
by~\eqref{eq:moment}, and hence lies in $\mathcal{C}\subseteq\operatorname{row}(A)$.
\end{proofof}

\begin{proofof}{Theorem~\ref{thm:regret}}
The value function $\sigma^*$ is differentiable on
$\operatorname{ri}\mathcal{C}$ modulo the null directions of $A$, with
$\nabla\sigma^*(\lambda)=-x(\lambda)$. The regret is therefore the Bregman
divergence associated with the dual value function. By
Theorem~\ref{thm:duality}, the dual representation is the entropy projection
onto the affine constraint $A^\top p=\lambda$, and the Bregman divergence
between two members of this exponential family is the Kullback--Leibler
divergence between their share vectors \citep{banerjee2005}. Hence
$R(\lambda\Vert\hat{\lambda})=\KL(p\Vert\hat p)$, with the true price in the
first argument and the deployed price in the second.
\end{proofof}

\begin{proposition}[Linear normal form]
\label{prop:ode}
A frontier of order $k$ along a direction---a floor and $k$ reducible
modes---is exactly a solution of a constant-coefficient linear ordinary
differential equation of order $k+1$ in the log-resource,
$\prod_{j=0}^{k}(\mathrm{d}/\mathrm{d}x+a_j)\,\eps=0$, one of whose roots is
$a_0=0$. The Prony and Hankel statements of
Corollary~\ref{cor:prony} are its discrete shadow, and the minimal order is
the rank of the Hankel matrix of the first differences.
\end{proposition}

\begin{proofof}{Proposition~\ref{prop:ode}}
If $\eps=\sum_{j=0}^{k}c_je^{-a_jx}$ then each summand is annihilated by
$(\mathrm{d}/\mathrm{d}x+a_j)$ and the operators commute, so
$\prod_j(\mathrm{d}/\mathrm{d}x+a_j)\eps=0$. Conversely the general solution
of a constant-coefficient linear equation of order $k+1$ with distinct real
roots $-a_j$ is that sum; positivity of the $c_j$ and non-negativity of the
$a_j$ are the content of Proposition~\ref{prop:law}, and $a_0=0$ is the
presence of a floor, which makes $\mathrm{d}/\mathrm{d}x$ a factor and leaves
an equation of order $k$ in $\eps'$. Sampling that equation on a geometric
grid and eliminating the floor by first differences is
Corollary~\ref{cor:prony}; the minimal order is the rank of the Hankel matrix
of $\Delta\eps$ because a sum of $k$ geometric sequences has Hankel rank $k$.
\end{proofof}

\begin{corollary}[Exact recovery on a geometric grid]
\label{cor:prony}
Let the law have order $k$ along one direction: a floor and $k$ reducible
modes, $2k+1$ unknowns. On a geometric resource grid $x_j=x_0+j\Delta$,
$j=0,\dots,2k$, the $2k+1$ accuracies determine all of them exactly. The first
differences annihilate the floor and form a sum of $k$ geometric sequences, so
the ratios $e^{-\alpha_i\Delta}$ are the roots of the Prony polynomial of
$\Delta\eps$, after which the weights and $\epsinf$ follow linearly; the
minimal order is the rank of the Hankel matrix of $\Delta\eps$. This is
Proposition~\ref{prop:ode} read on a grid.
\end{corollary}

\begin{proofof}{Corollary~\ref{cor:prony}}
On the grid $x_j=x_0+j\Delta$ write $\eps_j=\epsinf+\sum_{i=1}^{k}w_i\rho_i^{\,j}$
with $w_i=c_ie^{-\alpha_ix_0}>0$ and $\rho_i=e^{-\alpha_i\Delta}\in(0,1)$
distinct. The first differences $\Delta\eps_j=\eps_{j+1}-\eps_j
=\sum_i w_i(\rho_i-1)\rho_i^{\,j}$, $j=0,\dots,2k-1$, no longer contain the
floor and form a sum of $k$ geometric sequences. Such a sequence is
annihilated by the Prony polynomial $\prod_i(z-\rho_i)=z^k+\sum_{l<k}q_lz^l$,
that is, $\Delta\eps_{j+k}+\sum_{l<k}q_l\Delta\eps_{j+l}=0$ for
$j=0,\dots,k-1$: $k$ linear equations in the $k$ coefficients $q_l$, whose
Hankel matrix $(\Delta\eps_{j+l})_{j,l<k}$ is nonsingular because the $\rho_i$
are distinct and the weights $w_i(\rho_i-1)$ are nonzero. The roots give the
$\rho_i$, hence $\alpha_i=-\ln\rho_i/\Delta$; the weights then follow from the
first $k$ differences by a Vandermonde system, and $\epsinf=\eps_0-\sum_iw_i$.
Thus $2k+1$ accuracies determine the $2k+1$ unknowns. The Hankel matrix of
$(\Delta\eps_j)$ has rank exactly $k$ for a sum of $k$ geometric sequences, so
the minimal order is read off its rank. For $k=1$ the Prony root is
$\rho=\Delta\eps_1/\Delta\eps_0$ and $\epsinf=\eps_0-\Delta\eps_0/(1-\rho)$,
which simplifies to~\eqref{eq:aitken}, Aitken's $\Delta^2$ \citep{aitken1926};
the three observations identify a finite positive floor exactly when
$\Delta\eps_0<\Delta\eps_1<0$ and $\eps_-\eps_+>\eps_0^2$, the latter being
strict log-convexity of $\eps$ (Subsection~\ref{sup:pf-width}). For unequally
spaced candidates $\alpha$ solves a scalar equation, obtained by bracketing.
\end{proofof}

\begin{proposition}[Logistic normal form]
\label{prop:logistic}
With one reducible channel of exponent $\alpha$ along a direction, the
marginal return obeys the logistic equation and the cumulant hierarchy closes
at second order:
\begin{equation}
s'=-s(\alpha-s),
\qquad
\kappa=s(\alpha-s),
\qquad
S'''=-\alpha\,\kappa\,m,
\label{eq:logistic}
\end{equation}
with $m=1-2p_0$. Equivalently $S$ is a softplus and $p_0$ a logistic in the
normalized resource $\zeta$ of~\eqref{eq:universal}; the frontier is a
two-level system and the canonical point of Section~\ref{sec:lawII} is its
half-saturation.
\end{proposition}

\begin{proofof}{Proposition~\ref{prop:logistic}}
With two atoms $\{0,\alpha\}$ and $p_1=1-p_0$ the moments are $s=\alpha p_1$
and $\kappa=\alpha^2p_0p_1$, so
$\kappa=\alpha^2(s/\alpha)(1-s/\alpha)=s(\alpha-s)$; combined with
$s'=-\kappa$ from Proposition~\ref{prop:selection} this is the logistic
equation, whose solution is $s=\alpha p_1$ with $p_1$ a logistic in
$\zeta=\alpha x-\ln(c/\epsinf)$, hence $S=-\ln\epsinf-\ln(1+e^{-\zeta})$, a
softplus. The identity $\kappa=s(\alpha-s)$ rearranges to $s+\kappa/s=\alpha$,
which is Proposition~\ref{prop:floorfree}(ii). For the third equality,
$\mathrm{d}\kappa_n/\mathrm{d}x=-\kappa_{n+1}$ for the cumulants of the tilted
measure, so $S'''=\kappa_3$; for a two-point distribution on $\{0,\alpha\}$,
$\kappa_3=\alpha^3p_0p_1(p_0-p_1)=-\alpha\kappa\,m$ with $m=1-2p_0$. The
closure is exact for two atoms and fails for three, which is why it
characterizes the class.
\end{proofof}

\subsection{Law II}
\label{sup:lawII}\label{sup:pf-canonical}\label{sup:pf-surprisal}

\begin{proposition}[Surprisal identity]
\label{prop:surprisal}
For every spectral measure with $\epsinf>0$, every $n$, and every number of
modes,
\begin{equation}
\boxed{\ \ln\frac{\eps(x)}{\epsinf}=-\ln p_0(x)\ }
\label{eq:surprisal}
\end{equation}
nats; equivalently $S(\infty)-S(x)=-\ln p_0(x)$.
\end{proposition}

\begin{proofof}{Proposition~\ref{prop:surprisal}}
By Proposition~\ref{prop:spectral}, $\epsinf=\mu(\{0\})$ and the tilted
measure~\eqref{eq:tilt} assigns the origin the mass
$p_0(x)=\epsinf e^{-0\cdot x}/\eps(x)=\epsinf/\eps(x)$, since the atom is
unaffected by $x$. Hence $\eps/\epsinf=1/p_0$ and
$\ln(\eps/\epsinf)=-\ln p_0$. Because $S=-\ln\eps$ and
$S(\infty)=-\ln\epsinf$ along any direction $d$ with $a\cdot d>0$
$\mu$-almost everywhere off the origin, the same quantity equals
$S(\infty)-S(x)$. No property of $\mu$ beyond $\mu(\{0\})>0$ was used, so the
identity holds for any $n$, any number of modes, and any spectrum.

The rows of Table~\ref{tab:coordinates} follow by substitution. Rows~1
and~2 are the identity and its exponential. Row~3: the minimax rule of
Theorem~\ref{thm:canonical} declares $\hat p_j=1/(N+1)$ for every $j$,
including $j=0$, and row~1 then gives $\Delta S=\ln(N+1)$, which is
\eqref{eq:canon}. Row~4: $\eps\le1$ and $p_0=\epsinf/\eps$ give
$\epsinf=p_0\eps\le p_0$, and at the canonical point this is
$\epsinf\le1/(N+1)$, that is, Corollary~\ref{cor:feasible}. For one
reducible channel, $\eps/\epsinf=1+e^{-\zeta}=1/p_0$ gives
$e^{-\zeta}=(1-p_0)/p_0$, hence $\zeta=\ln\bigl(p_0/(1-p_0)\bigr)$, the
logit of the floor share (row~6); $m=1-2p_0=-\tanh(\zeta/2)$ (row~7), so $m$
is the odd image of $\zeta$ and the mode-exchange involution
$p_0\leftrightarrow p_1$ is $m\mapsto-m$. Rows~8 and~9 are
Proposition~\ref{prop:knee} with $p_1=1-p_0$, the second form of row~9
following from $p_0(1-p_0)=(1-m^2)/4$. Row~11 is
Proposition~\ref{prop:flat}, in which $\ln\cosh(\zeta/2)$ is the binary
divergence between the declared share $\tfrac12$ and the realized $p_0$, by
row~6. Rows~5, 10 and~12, and the $\varphi$ column throughout, are the arc
forms of Proposition~\ref{prop:arcforms}, proved in
Subsection~\ref{sup:lawIII}.
\end{proofof}

\begin{proofof}{Theorem~\ref{thm:master}}
Write $M=N+1$ and $e_j(x)=c_je^{-a_j\cdot x}$, so $\eps=\sum_je_j$ and
$p_j=e_j/\eps$. For a share vector $\AM(p)=\frac1M\sum_jp_j=\frac1M$, hence
\[
  \KL(u\Vert p)=\sum_j\frac1M\ln\frac{1/M}{p_j}
  =-\ln M-\frac1M\sum_j\ln p_j
  =\ln\frac{1/M}{\GM(p)}
  =\ln\frac{\AM(p)}{\GM(p)} .
\]
Both $\AM$ and $\GM$ are positively homogeneous of degree one, so their ratio
is scale-invariant and $\AM(p)/\GM(p)=\AM(e)/\GM(e)$: the shares may be
replaced by the raw mode errors. Next,
\[
  \ln\GM(e)=\frac1M\sum_j\bigl(\ln c_j-a_j\cdot x\bigr)
  =\frac1M\sum_j\ln c_j-\abar\cdot x
\]
is affine in $x$ with gradient $-\abar$, while $\ln\AM(e)=\ln\eps-\ln M$.
Therefore
\[
  -\ln\frac{\AM(e)}{\GM(e)}
  =-\ln\eps+\ln M+\frac1M\sum_j\ln c_j-\abar\cdot x
  =\sigma_{\abar}(x)+\ln M+\frac1M\sum_j\ln c_j ,
\]
which is~\eqref{eq:master2}. By the arithmetic--geometric mean inequality
\citep[Theorem~9]{hardy1952} the ratio is at least $1$, with equality if and
only if $e_0=\dots=e_N$; strict concavity of $\ln$ makes the equality case
unique among share vectors, and on a frontier of full rank it is attained at
the unique $x$ solving $c_je^{-a_j\cdot x}=\epsinf$ for every $j$, where
$\eps=(N+1)\epsinf$.
\end{proofof}

Three remarks clarify the scope of the theorem.

\paragraph{The price is the counting barycenter.} The gradient of the master
identity is $\abar$, the mean of the exponents under the \emph{counting}
measure on $\operatorname{supp}\mu$, not under $\mu$ itself; the weights $c_j$
enter only the additive constant. This is the precise sense in which the
canonical rule is Laplace's principle of insufficient reason applied to the
spectrum's modes rather than to the deployment. With one channel per resource
it reproduces $\lamstar_i=\alpha_i/(N+1)$.

\paragraph{Two entropies.} The value $\ln(N+1)$ of
Theorem~\ref{thm:canonical} is $\ln|\operatorname{supp}\mu|$, the Hartley
(order-zero R\'enyi) entropy of the spectrum's support, whereas the dual value
\eqref{eq:dual} carries the Shannon entropy of the shares. The declaration
problem and the valuation problem therefore sit at two different orders of the
same ladder, and the nested reference of \TR{Interacting resources: the nested dual} is the
Hartley entropy of an effective support of fractional size $1+n^{1/\theta}$.

\paragraph{Two divergences.} The operating-point regret used throughout the
selection sections is $\KL(u\Vert p)$---the cost of operating at $p$ when the
canonical price is right---and it is this that equals $\ln(\AM/\GM)$. The
declaration regret of Theorem~\ref{thm:canonical} is
$\sup_p\KL(p\Vert\hat p)$, which is a vertex evaluation. They coincide in
pointing at the uniform share vector and differ in everything else; confusing
them is the one error the compactness of the master identity invites.

\begin{proofof}{Lemma~\ref{lem:reduction}}
Let $r$ satisfy Assumptions~\ref{ax:admissible}--\ref{ax:offset}.

\emph{Equivariance.} Under $t_i\mapsto t_i^{\beta_i}$ the coordinate becomes
$x_i'=\beta_ix_i$, and the diagonal law becomes
$\eps=\epsinf+\sum_ic_ie^{-(\alpha_i/\beta_i)x_i'}$: the floor and the
coefficients are unchanged, while $\alpha_i\mapsto\alpha_i/\beta_i$. A price
$\lambda'$ in the new coordinates scores
$S-\lambda'\!\cdot x'=S-\sum_i\lambda_i'\beta_ix_i$. Suppose
$\lambda_i'\beta_i\neq\lambda_i$ for some $i$. Both values lie in
$(0,\alpha_i)$ by admissibility, and along coordinate $i$ the marginal yield
$\partial S/\partial x_i=\alpha_ip_i$ takes every value in $(0,\alpha_i)$;
by continuity there are two frontier points differing only in $x_i$ whose
break-even rate lies strictly between $\lambda_i'\beta_i$ and $\lambda_i$,
and the two scores rank them oppositely. Hence equivariance holds if and only
if
\[
  r_i(\epsinf,c,\alpha\oslash\beta)=r_i(\epsinf,c,\alpha)/\beta_i
  \qquad\text{for all }\beta\in\R^N_{>0},
\]
with $\oslash$ the componentwise division of \eqref{eq:scaling}.

\emph{Offset independence.} The maps $\eps\mapsto\gamma\eps$ and
$t_i\mapsto\theta_it_i$ send $(\epsinf,c_1,\ldots,c_N)$ to
$(\gamma\epsinf,\gamma c_1\theta_1^{\alpha_1},\ldots,\gamma c_N\theta_N^{\alpha_N})$
and leave $\alpha$ unchanged. For fixed $\alpha$ they act transitively on
$\R^{N+1}_{>0}$, so $r(\epsinf,c,\alpha)=r(1,\one,\alpha)=:\rho(\alpha)$.

\emph{Reduction.} Taking $\beta=\alpha$ in the equivariance identity gives
$\rho_i(\one)=\rho_i(\alpha)/\alpha_i$, that is, $\rho_i(\alpha)=\varrho_i\alpha_i$
with $\varrho_i:=\rho_i(\one)$. By Corollary~\ref{cor:diag}, the declared
shares are $\hat p_i=\hat\lambda_i/\alpha_i=\varrho_i$ and
$\hat p_0=1-\sum_i\varrho_i=:\varrho_0$, and admissibility is equivalent to
$\varrho\in\operatorname{ri}\Delta^N$. Conversely, a rule of the
form~\eqref{eq:reduction} is admissible, is unaffected by rescalings of the
error and of resource units, and maps to $\hat\lambda_i/\beta_i$ under
$t_i\mapsto t_i^{\beta_i}$, which preserves every ranking.
\end{proofof}

\begin{proofof}{Theorem~\ref{thm:canonical}}
Fix a frontier. A declared price $\hat\lambda\notin\mathcal{C}$ has
$\sigma^*(\hat\lambda)=+\infty$ by Theorem~\ref{thm:duality}, so no finite
operating point is selected, and we assign it infinite regret. A declared
price $\hat\lambda\in\mathcal{C}\setminus\operatorname{ri}\mathcal{C}$ has,
by Corollary~\ref{cor:diag}, a share vector $\varrho$ with $\varrho_j=0$ for
some $j$; defining the regret by continuity along declarations
$\hat\lambda_k\to\hat\lambda$ in $\operatorname{ri}\mathcal{C}$, it equals
$\KL(p\Vert\varrho)=+\infty$ for every realized $p$ with $p_j>0$. Hence a
finite worst case requires $\hat\lambda\in\operatorname{ri}\mathcal{C}$:
admissibility is implied by the minimax requirement rather than imposed. For
$\hat\lambda\in\operatorname{ri}\mathcal{C}$, let
$\varrho\in\operatorname{ri}\Delta^N$ be its share vector. By
Corollary~\ref{cor:diag}, the realized shares range over all of
$\operatorname{ri}\Delta^N$ as $\Lambda$ ranges over
$\operatorname{ri}\mathcal{C}$, and by Theorem~\ref{thm:regret} the regret
is $\KL(p\Vert\varrho)$. The worst case is
\[
  \Psi(\varrho)=\sup_{p\in\operatorname{ri}\Delta^N}\KL(p\Vert\varrho).
\]
The map $p\mapsto\KL(p\Vert\varrho)$ is convex and continuous on the closed
simplex (with $0\ln0=0$), so its supremum over the relative interior equals
its maximum over $\Delta^N$, which is attained at a vertex $e_j$, where
$\KL(e_j\Vert\varrho)=-\ln\varrho_j$. Hence
\[
  \Psi(\varrho)=\max_j\bigl(-\ln\varrho_j\bigr)=-\ln\min_j\varrho_j .
\]
Since $\sum_j\varrho_j=1$, $\min_j\varrho_j\le1/(N+1)$, with equality if and
only if $\varrho$ is uniform. Therefore $\Psi(\varrho)\ge\ln(N+1)$, with
equality exactly at $\varrho=u$. On every frontier, the unique minimax
declaration is therefore $\hat\lambda_i=\alpha_i/(N+1)$. The rule so defined
is of the form~\eqref{eq:reduction} with $\varrho=u$, and by the converse
part of Lemma~\ref{lem:reduction} it satisfies
Assumptions~\ref{ax:admissible}--\ref{ax:offset}. This proves uniqueness
and~(i); Lemma~\ref{lem:reduction} is used only for the converse. Note that
Assumptions~\ref{ax:admissible}--\ref{ax:offset} alone do not single out the
uniform shares: every constant $\varrho\in\operatorname{ri}\Delta^N$ defines a
rule satisfying them, with worst case $-\ln\min_j\varrho_j$, unbounded over
$\varrho$. Note also that $\KL(Pp\Vert P\varrho)=\KL(p\Vert\varrho)$ for every
permutation matrix $P$, so $\Psi(P\varrho)=\Psi(\varrho)$ and the unique
minimizer is fixed by every permutation of the modes; this is the Hunt--Stein
reading given in Section~\ref{sec:onedim} of the paper, and it was not used
above.

(ii) Expanding,
$\KL(p\Vert u)=\sum_jp_j\ln p_j+\ln(N+1)=\ln(N+1)-\Hs(p)$. Since
$\Hs\le\ln(N+1)$, with equality only at $p=u$, the uniform vector is the
maximum-entropy share, and the supremum of $\KL(p\Vert u)$ is $\ln(N+1)$,
approached at the vertices.

(iii) For a prior $\pi$ with mean $\bar p$,
$\mathbb{E}_\pi[\KL(p\Vert\hat p)]=\mathbb{E}_\pi[\sum_jp_j\ln p_j]
-\sum_j\bar p_j\ln\hat p_j$. The first term is finite, lying in
$[-\ln(N+1),0]$, and does not involve $\hat p$; by Gibbs' inequality, the
second is minimized uniquely at $\hat p=\bar p$. A permutation-invariant
prior has $\bar p=u$. The uniform prior on the vertices of $\Delta^N$, the
limit of the admissible set, has Bayes act $u$ and risk
$\KL(e_j\Vert u)=\ln(N+1)$ at every point of its support; a Bayes rule with
constant risk on the support of its prior is minimax, and that prior is least
favorable \citep{berger1985}, which re-derives (i).

(iv) With $\hat p=u$, the normal form~\eqref{eq:normal} gives $e^{-\xi_i}=1$,
hence $\xi_i=0$ for every channel, so that every reducible component
$c_ie^{-\alpha_ix_i}=\epsinf e^{-\xi_i}$ equals $\epsinf$.
Equation~\ref{eq:normal} yields $\eps^\star=(N+1)\epsinf$,
the normalized price yields $\lambda_i^\star=\alpha_i/(N+1)$, and
$\Delta S=\ln(\eps^\star/\epsinf)=\ln(N+1)$, which is the equality case
$e_0=\dots=e_N$ of Theorem~\ref{thm:master}.
\end{proofof}

\begin{theorem}[Capacity form]
\label{thm:capacity}
Let $\Lambda\subseteq\Delta^N$ be any closed set of share vectors the deployer
admits, and let $\mathcal{P}(\Lambda)$ be the probability measures on it.
Then
\begin{equation}
\min_{\hat p\in\Delta^N}\ \sup_{p\in\Lambda}\KL(p\Vert\hat p)
\;=\;\max_{\pi\in\mathcal{P}(\Lambda)}\int\KL\bigl(p\Vert\bar p_\pi\bigr)\,
\mathrm{d}\pi(p)\;=:\;C(\Lambda),
\qquad
\bar p_\pi=\int p\,\mathrm{d}\pi .
\label{eq:capacity}
\end{equation}
The minimizer is unique and equals $\bar p_{\pi^\star}$ for any maximizing
$\pi^\star$, and $\KL(p\Vert\bar p_{\pi^\star})\le C(\Lambda)$ on $\Lambda$
with equality $\pi^\star$-almost surely. Theorem~\ref{thm:canonical} is the
case $\Lambda=\Delta^N$, where $C=\ln(N+1)$, the maximizing $\pi^\star$ is
uniform on the vertices and $\bar p_{\pi^\star}$ is uniform.
\end{theorem}

\begin{proofof}{Theorem~\ref{thm:capacity}}
Write $f(\hat p,\pi)=\int\KL(p\Vert\hat p)\,\mathrm{d}\pi(p)$ for
$\hat p\in\Delta^N$ and $\pi\in\mathcal{P}(\Lambda)$. Three facts about $f$.
First, for fixed $\pi$ with mean $\bar p_\pi$ the compensation identity
\[
  \int\KL(p\Vert\hat p)\,\mathrm{d}\pi
  =\int\KL(p\Vert\bar p_\pi)\,\mathrm{d}\pi+\KL(\bar p_\pi\Vert\hat p)
\]
holds, because $\int\sum_jp_j\ln(1/\hat p_j)\,\mathrm{d}\pi
=\sum_j\bar p_{\pi,j}\ln(1/\hat p_j)$ is linear in $p$; hence
$\min_{\hat p}f(\hat p,\pi)=f(\bar p_\pi,\pi)$, attained only at
$\hat p=\bar p_\pi$. This is the Bregman identity that makes the Bayes
declaration the prior mean in Theorem~\ref{thm:canonical}(iii). Second, for
fixed $\hat p$, $f$ is linear in $\pi$, so
$\sup_\pi f(\hat p,\pi)=\sup_{p\in\Lambda}\KL(p\Vert\hat p)$, the supremum
over Dirac masses. Third, $\hat p\mapsto f(\hat p,\pi)$ is convex and lower
semicontinuous on the compact convex set $\Delta^N$, and
$\pi\mapsto f(\hat p,\pi)$ is linear and weak$^*$-continuous on the compact
convex set $\mathcal{P}(\Lambda)$, $\Lambda$ being closed in the simplex. By
Sion's minimax theorem,
\[
  \min_{\hat p}\sup_{\pi}f=\sup_{\pi}\min_{\hat p}f ,
  \qquad\text{that is}\qquad
  \min_{\hat p}\ \sup_{p\in\Lambda}\KL(p\Vert\hat p)
  =\sup_{\pi}\int\KL(p\Vert\bar p_\pi)\,\mathrm{d}\pi=C(\Lambda),
\]
which is~\eqref{eq:capacity}; the right-hand supremum is attained because it
is an upper semicontinuous function of $\pi$ on a compact set. For
uniqueness, $\hat p\mapsto\sup_{p\in\Lambda}\KL(p\Vert\hat p)$ is a supremum
of functions $-\sum_jp_j\ln\hat p_j+\mathrm{const}$ each of which is strictly
convex in $\hat p$ on the simplex, so it is strictly convex and has one
minimizer $\hat p^\star$. If $\pi^\star$ maximizes the right-hand side then,
by the compensation identity and the saddle-point property,
$\sup_p\KL(p\Vert\bar p_{\pi^\star})\le C(\Lambda)
=f(\bar p_{\pi^\star},\pi^\star)\le\sup_p\KL(p\Vert\bar p_{\pi^\star})$,
so $\bar p_{\pi^\star}$ attains the minimum and equals $\hat p^\star$;
moreover $\KL(p\Vert\bar p_{\pi^\star})\le C$ on $\Lambda$ while its
$\pi^\star$-average equals $C$, so equality holds $\pi^\star$-almost surely.
For $\Lambda=\Delta^N$, the vertex argument of Theorem~\ref{thm:canonical}
gives $\sup_p\KL(p\Vert u)=\ln(N+1)$, and the uniform prior on the vertices
has $\bar p_\pi=u$ and $\int\KL(e_j\Vert u)\,\mathrm{d}\pi=\ln(N+1)$, so
$C=\ln(N+1)$ with $\pi^\star$ uniform on the vertices and
$\bar p_{\pi^\star}=u$. Monotonicity of $C$ in $\Lambda$ is immediate from
the left-hand side of~\eqref{eq:capacity}. The right-hand side is the mutual
information between a price drawn from $\pi$ and the shares it induces,
which identifies $C(\Lambda)$ as a capacity
\citep{davisson1973,haussler1997}.
\end{proofof}

\begin{corollary}[Entropy tangency]
\label{cor:tangency}
With one reducible channel and the reducible share known only to lie in
$[a,b]\subseteq[0,1]$, the minimax declaration is the unique $\xi\in(a,b)$ at
which the binary entropy is tangent to its own chord,
\begin{equation}
\Hs_b'(\xi)=\frac{\Hs_b(b)-\Hs_b(a)}{b-a},
\qquad\text{that is}\qquad
\xi=\Bigl(1+e^{(\Hs_b(b)-\Hs_b(a))/(b-a)}\Bigr)^{-1},
\label{eq:tangency}
\end{equation}
the operating point is $\eps^\star=\epsinf/(1-\xi)$, and the worst-case regret
is the common gap between the entropy and that tangent at the two endpoints.
At $[a,b]=[0,1]$ this is $\xi=\tfrac12$, $\eps^\star=2\epsinf$, $C=\ln2$.
\end{corollary}

\begin{proofof}{Corollary~\ref{cor:tangency}}
With one reducible channel a share vector is $(1-s,s)$ with $s$ the reducible
share, and $\Lambda=\{s\in[a,b]\}$. For a declaration $\xi\in(0,1)$ the
regret $k(s):=\KL\bigl((1-s,s)\Vert(1-\xi,\xi)\bigr)$ is convex in $s$, so
its supremum over $[a,b]$ sits at an endpoint. As a function of $\xi$,
$k_a(\xi)=\KL(a\Vert\xi)$ is increasing on $(a,1)$ and $k_b(\xi)=\KL(b\Vert\xi)$
is decreasing on $(0,b)$, so on $(a,b)$ the worst case
$\max\{k_a(\xi),k_b(\xi)\}$ is minimized at the unique point where the two
are equal, and a declaration outside $(a,b)$ is dominated. Writing
$\KL(s\Vert\xi)=-\Hs_b(s)-s\ln\xi-(1-s)\ln(1-\xi)$, the equality
$k_a(\xi)=k_b(\xi)$ reads
\[
  \Hs_b(b)-\Hs_b(a)+(b-a)\ln\frac{\xi}{1-\xi}=0
  \qquad\Longleftrightarrow\qquad
  \ln\frac{1-\xi}{\xi}=\frac{\Hs_b(b)-\Hs_b(a)}{b-a},
\]
and since $\Hs_b'(\xi)=\ln\{(1-\xi)/\xi\}$ this is the tangency condition and
the closed form~\eqref{eq:tangency}. The tangent to $\Hs_b$ at $\xi$ is
$T(s)=\Hs_b(\xi)+\Hs_b'(\xi)(s-\xi)=-s\ln\xi-(1-s)\ln(1-\xi)$, so
$T(s)-\Hs_b(s)=\KL(s\Vert\xi)$ for every $s$: the worst-case regret is the
common gap between the entropy and its tangent at the two endpoints. The
operating point has reducible share $\xi$, hence $\eps^\star=\epsinf/(1-\xi)$
by~\eqref{eq:p0def}. At $[a,b]=[0,1]$ the chord slope is $0$, so
$\xi=\tfrac12$, $\eps^\star=2\epsinf$ and $C=\KL(0\Vert\tfrac12)=\ln2$. This
is Theorem~\ref{thm:capacity} at $N=1$ with $\pi^\star$ supported on the two
endpoints, with weights chosen so that $\bar p_{\pi^\star}=\xi$.
\end{proofof}

\noindent Since $\eps\le1$ and $p_0=\epsinf/\eps$, the canonical reference is a feasible
error value only if
\begin{equation}
\epsinf\le\frac{1}{N+1},
\label{eq:feasible}
\end{equation}
so a benchmark whose floor is too high admits no canonical point.

\begin{proofof}{Corollary~\ref{cor:feasible}}
The error condition follows from $\varepsilon^\star=(N+1)\varepsilon_\infty$
and $\varepsilon\leq1$. Empirical attainability additionally requires that
the corresponding resource coordinates belong to the achievable candidate
region.
\end{proofof}

\begin{proofof}{Corollary~\ref{cor:single}}
With $\zeta$ as defined, $c\,e^{-a\cdot x}=\epsinf e^{-\zeta}$, which
gives~\eqref{eq:universal}. The share vector is $(1-p_1,p_1)$ with
$p_1=(1+e^{\zeta})^{-1}$, and $\nabla S=a\,p_1$ by Theorem~\ref{thm:duality},
so the supported prices are $\{s\,a:s\in(0,1)\}$ and $s=p_1$ is the declared
share. The argument of Lemma~\ref{lem:reduction} applies with one channel:
admissibility gives $\hat\lambda=s\,a$, offset independence removes the
dependence of $s$ on $(\epsinf,c)$, and equivariance under
$t_i\mapsto t_i^{\beta_i}$, which maps $a\mapsto a\oslash\beta$, requires
$s(a\oslash\beta)=s(a)$ for all $\beta$, so $s$ is a constant. The proof of
Theorem~\ref{thm:canonical} with $N=1$, which does not use this reduction,
gives $s=\tfrac12$ on every frontier, worst-case
regret $\ln2$, $\zeta=0$, $\eps=2\epsinf$, $\lambda^\star=a/2$, and
$\Delta S=\ln2$. Feasibility follows from Corollary~\ref{cor:feasible} with
$N=1$, with strict inequality required for the reference to lie below
chance-level error.
\end{proofof}

\begin{corollary}[Canonical value]
\label{cor:accvalue}
Let $\eps=\epsinf+ce^{-\alpha x}$ along a fixed direction and let the
candidates be any finite set of attainable pairs $(\eps_k,x_k)$ on that law.
Define the \emph{canonical value}
$J(\eps)=\tfrac12\ln(\eps-\epsinf)-\ln\eps$. Then $J=-R+\mathrm{const}$ by
\eqref{eq:lncosh}, so $J$ is the scalarization \eqref{eq:score} at the
canonical price up to an additive constant, is strictly concave in $\ln\eps$
with its maximum at $\eps=2\epsinf$, and
$\arg\max_kJ(\eps_k)=\arg\min_kR(\eps_k)$: the canonical ranking of a
candidate set is a function of the error list and $\epsinf$ alone, and the
fitted $c$ and $\alpha$ drop out of it.
\end{corollary}

\begin{proofof}{Corollary~\ref{cor:accvalue}}
Along the law $\eps-\epsinf=c\,e^{-\alpha x}$, so
$\tfrac{\alpha}{2}x=\tfrac12\ln c-\tfrac12\ln(\eps-\epsinf)$ and
\begin{equation*}
\sigma_{\lamstar}(x)=S(x)-\tfrac{\alpha}{2}x
=-\ln\eps+\tfrac12\ln(\eps-\epsinf)-\tfrac12\ln c
=J(\eps)-\tfrac12\ln c ,
\end{equation*}
so $J$ is the canonical scalarization up to a constant that does not depend
on the candidate. Next, $p_0=\epsinf/\eps$ and $p_1=(\eps-\epsinf)/\eps$ give
$4p_0p_1=4\epsinf(\eps-\epsinf)/\eps^{2}$, so by~\eqref{eq:lncosh}, which is
Theorem~\ref{thm:master} at $N=1$,
\begin{equation*}
R=\tfrac12\ln\frac{1}{4p_0p_1}
=\ln\eps-\tfrac12\ln(\eps-\epsinf)-\tfrac12\ln(4\epsinf)
=-J(\eps)-\tfrac12\ln(4\epsinf),
\end{equation*}
which is $J=-R+\mathrm{const}$ and row~12 of Table~\ref{tab:coordinates}.
Writing $v=\ln\eps$,
$\mathrm{d}J/\mathrm{d}v=\tfrac12\eps/(\eps-\epsinf)-1=\tfrac1{2p_1}-1$,
which vanishes exactly at $p_1=\tfrac12$, that is at $\eps=2\epsinf$, and
$\mathrm{d}^2J/\mathrm{d}v^2=-\tfrac12\epsinf\eps/(\eps-\epsinf)^2<0$
whenever $\epsinf>0$, so $J$ is strictly concave in $\ln\eps$ with that
unique maximum. Since $J$ differs from $\sigma_{\lamstar}$ and from $-R$ by
constants that do not depend on the candidate,
$\arg\max_kJ(\eps_k)=\arg\max_k\sigma_{\lamstar}(x_k)=\arg\min_kR(\eps_k)$,
and only $\epsinf$ enters $J$: the fitted $c$ and $\alpha$ sit in the
additive constant.
\end{proofof}

\begin{proofof}{Corollary~\ref{cor:two}}
$J(\eps_2)>J(\eps_1)$ reads
$\tfrac12\ln\bigl((\eps_2-\epsinf)/(\eps_1-\epsinf)\bigr)>\ln(\eps_2/\eps_1)$.
Exponentiating twice, this is
$\eps_1^{2}(\eps_2-\epsinf)>\eps_2^{2}(\eps_1-\epsinf)$, that is
$\eps_1\eps_2(\eps_1-\eps_2)>\epsinf(\eps_1-\eps_2)(\eps_1+\eps_2)$. Dividing
by $\eps_1-\eps_2>0$ gives $\eps_1\eps_2>\epsinf(\eps_1+\eps_2)$, which is
\eqref{eq:harmonic} after dividing by $\eps_1\eps_2\epsinf$; the harmonic mean
form is the same statement written as
$2\eps_1\eps_2/(\eps_1+\eps_2)>2\epsinf$. For a declared reducible share
$\hat p$ the same two steps applied to
$J_{\hat p}=\hat p\ln(\eps-\epsinf)-\ln\eps$ give
$\eps_1(\eps_2-\epsinf)^{\hat p}>\eps_2(\eps_1-\epsinf)^{\hat p}$, and
$\mathrm{d}J_{\hat p}/\mathrm{d}v=\hat p/p_1-1$ vanishes at $p_1=\hat p$,
i.e., at $\eps=\epsinf/(1-\hat p)$. At $\hat p=\tfrac12$ both statements
reduce to the previous ones; at
$\hat p=n^{1/\theta}/(1+n^{1/\theta})$ the maximizer is the nested reference
$\eps^\star=(1+n^{1/\theta})\epsinf$ of \TR{Interacting resources: the nested dual}.
\end{proofof}

\begin{lemma}[Cost of a wrong pairwise call]
\label{lem:pairbound}
Let $\eps_1>\eps_2$ at $t_1<t_2$. If the cheaper candidate is chosen when
the more accurate one is canonically preferred, the loss is at most
$\tfrac12\ln(\eps_1/\eps_2)$ nats under the error-space selection regret
$R_{\mathrm{sel}}$, and at
most $\ln(\eps_1/\eps_2)$ under the resource-aware objective
\eqref{eq:Rsigma}; both bounds are tight but neither is attained on the stated
domain---the first is approached as $\epsinf\downarrow0$ and the second as
$\alpha(x_2-x_1)\downarrow0$, and each limit leaves the hypotheses
($\epsinf>0$, $x_1<x_2$ strict).
\end{lemma}

\begin{proof}
Under $R$ the loss is $f(\epsinf)=J(\eps_2)-J(\eps_1)
=\tfrac12\ln\bigl((\eps_2-\epsinf)/(\eps_1-\epsinf)\bigr)-\ln(\eps_2/\eps_1)$
when positive. Since
$f'(\epsinf)=\tfrac12\bigl[(\eps_1-\epsinf)^{-1}-(\eps_2-\epsinf)^{-1}\bigr]<0$
for $\eps_1>\eps_2>\epsinf$, $f$ is decreasing, so
$f\le f(0)=\tfrac12\ln(\eps_1/\eps_2)$, with equality approached as
$\epsinf\to0$. Under~\eqref{eq:Rsigma} the loss is
$\ln(\eps_1/\eps_2)-\tfrac{\alpha}{2}(x_2-x_1)\le\ln(\eps_1/\eps_2)$,
because $\alpha>0$ and $x_2\ge x_1$ on a Pareto frontier, with equality in the
limit $\alpha(x_2-x_1)\to0$.
\end{proof}

\begin{proposition}[What the two objectives differ by]
\label{prop:blindspot}
For every candidate,
$\sigma_{\lamstar}(x_k)=J(\eps_k)-\tfrac12\ln c-\tfrac{\alpha}{2}\,\delta_k$.
Hence the two rankings agree whenever the candidates share a common resource
excess, and in particular on the law, where $\delta\equiv0$; the cost of the
accuracy-only choice is bounded by the spread of the excess and by nothing
else; and at the reference, where $p_1=\tfrac12$, the resource-aware objective
is the accuracy-only value minus the candidate's log-error residual, one for
one.
\end{proposition}

\noindent The resource-aware objective of the paper's Section~\ref{sec:lawII} scores each
candidate at its \emph{measured} resource,
\begin{equation}
\sigma_{\lamstar}(x_k)=-\ln\eps_k-\tfrac{\alpha}{2}\,x_k,
\qquad
R_\sigma(k)=\max_{k'}\sigma_{\lamstar}(x_{k'})-\sigma_{\lamstar}(x_k).
\label{eq:Rsigma}
\end{equation}

\begin{proofof}{Proposition~\ref{prop:blindspot}}
By definition $\tfrac{\alpha}{2}\hat x_k=\tfrac12\ln c-\tfrac12\ln(\eps_k-\epsinf)$,
so with $x_k=\hat x_k+\delta_k$,
\[
  \sigma_{\lamstar}(x_k)=-\ln\eps_k-\tfrac{\alpha}{2}\hat x_k-\tfrac{\alpha}{2}\delta_k
  =J(\eps_k)-\tfrac12\ln c-\tfrac{\alpha}{2}\,\delta_k ,
\]
which is the identity. If $\delta_k\equiv\delta$ on the candidate set the two
objectives differ by a constant and rank identically; on the law
$\delta\equiv0$. In general the accuracy-only choice $k_J=\arg\max_kJ$ loses,
under $R_\sigma$,
$\max_k\sigma_{\lamstar}(x_k)-\sigma_{\lamstar}(x_{k_J})
\le\tfrac{\alpha}{2}\bigl(\max_k\delta_k-\min_k\delta_k\bigr)$, since
$J(\eps_{k_J})\ge J(\eps_k)$ for every $k$; so the cost is bounded by the
spread of the resource excess and by nothing else. Finally, let
$r_k=\ln\eps_k-\ln\eps(x_k)$ be the candidate's log-error residual from the
law at its measured resource. Since $S(\hat x_k)=-\ln\eps_k$ and
$S(x_k)=-\ln\eps(x_k)$, the mean-value theorem gives
$r_k=S(x_k)-S(\hat x_k)=s(\xi)\,\delta_k$ for some $\xi$ between $\hat x_k$
and $x_k$, with $s=\alpha p_1$ by Proposition~\ref{prop:knee}(i). At the
reference $p_1=\tfrac12$, so $r_k=\tfrac{\alpha}{2}\delta_k$ to first order
and $\sigma_{\lamstar}(x_k)=J(\eps_k)-\tfrac12\ln c-r_k$: the resource-aware
objective is the accuracy-only value minus the residual, one for one.
\end{proofof}

\subsection{Law III}
\label{sup:lawIII}

Throughout this subsection $\{p_x\}$ is the tilted family~\eqref{eq:tilt}
along a fixed direction, with $s=\mathbb{E}_{p_x}[a]$ and
$\kappa=\operatorname{Var}_{p_x}(a)$ as in~\eqref{eq:slopecurv}.

\begin{proposition}[Path length]
\label{prop:path}
Let $\eps$ be completely monotone with spectral measure $\mu$ along a
direction. Then
\begin{enumerate}[label=(\roman*),leftmargin=2.2em,itemsep=1pt]
\item \textup{(Two atoms.)} If $\mu=\epsinf\delta_0+c\,\delta_\alpha$ then,
      with $\varphi=2\arccos\sqrt{p_0}$ the channel angle,
      \begin{equation}
      L(x_1,x_2)=|\varphi(x_2)-\varphi(x_1)|,
      \qquad
      L(-\infty,\infty)=\pi
      \label{eq:arc}
      \end{equation}
      for every $\alpha$, every $\epsinf>0$ and every scale. \emph{The
      $x$-range here is all of $\R$, the chart's complete sweep; the part of it
      at or below chance level, $\eps\le1$, has the shorter arc
      $2\arccos\sqrt{\epsinf}$, which is what a measurement can reach
      (Proposition~\ref{prop:anglearc} of the paper, and the remark closing
      this subsection).}
\item \textup{(Dichotomy.)} $L\equiv0$ for a single atom---a pure power law
      with no floor; $L=\pi$ for one channel and a floor, over the complete
      sweep; and $L=\infty$ for a
      spectrum accumulating at the origin, a Gamma spectrum
      $\eps=C(1+x/\beta)^{-k}$ giving $L=\sqrt{k}\,\ln(1+X/\beta)$.
\item \textup{(Bounds, with the deficit in closed form.)} If $\mu$ has finite
      support of size $N+1$ with $N\ge1$ then $\pi\le L(-\infty,\infty)\le N\pi$, with
      $L=\pi$ if and only if $N=1$. ($N=0$ is the single atom of~(ii), with
      $L=0$.) The upper bound is reached only when every
      mode dominates the error over a long stretch. \emph{If the consecutive
      rates separate}---$a_{k+1}/a_k\to\infty$, so that each transition is
      complete before the next begins---the deficit is exact. Order the atoms
      by increasing rate, let $x_k$ be the $k$-th crossover, where atom $k+1$
      and the accumulated slower group $1,\dots,k$ carry equal error, and let
      \begin{equation}
      q_k=\frac{c_ke^{-a_kx_k}}{\sum_{j\le k}c_je^{-a_jx_k}}
      \label{eq:crossshare}
      \end{equation}
      be the share atom $k$ holds \emph{within that group at that crossover}.
      Then
      \begin{equation}
      L(-\infty,\infty)=\pi+\sum_{k=2}^{N}2\arcsin\sqrt{q_k},
      \label{eq:arcweights}
      \end{equation}
      the sum running over the interior atoms. Two things have to be said about
      that sum, and script~\texttt{160} is their verification.

      \emph{First, the shares must be read at the crossover and not off the
      weights.} Writing $p_k=c_k/\sum_jc_j$ and $P_k=p_1+\dots+p_k$, the
      quantity $p_k/P_k$ is \emph{not} a functional of the frontier at all:
      $x\mapsto x+\Delta$ sends $c_j\mapsto c_je^{-a_j\Delta}$, which leaves
      $L$ alone and moves $p_k/P_k$. On $a=(0,1,1000)$ the raw-weight form
      reads $1.500,1.776,1.347,1.224,1.052$ at $\Delta=0,-2,1,2,5$---five values for
      one arc, whose true $L/\pi$ is $1.4976$ throughout, an error of up to $30\%$
      over the gauge orbit and $43\%$ once the weights are free, as at
      $c=(1,1,e^{+5000})$. The form with $q_k$ returns \emph{one} value across
      those gauges, as the arc does, and tracks it to $1.7\times10^{-3}$ there
      and to $4.3\times10^{-5}$ on three- and four-atom spectra whose
      consecutive rates are separated by at least $10^{4}$ and which span four
      to ten orders of magnitude in all.

      \emph{Second, the identity is a limit in the rate ratio, and the
      convergence is slow.} In the limit $a_{k+1}/a_k\to\infty$ at fixed
      weights every crossover tends to the origin, so $q_k\to p_k/P_k$ and the
      two forms coincide---which is why the archive's own eight-case check
      (script~\texttt{89}, rates separated by $10^{4}$ to $10^{6}$) passes
      under either. At finite ratios neither is exact. With weights $(1,1,1)$
      and exponents $(0,1,A)$, quadrature gives
      $L/\pi=1.29702,1.40402,1.48320,1.49759,1.50000$ at
      $A=2,10,10^2,10^3,10^6$, against \eqref{eq:arcweights}'s
      $1.5759,1.5116,1.5011,1.5001,1.50000$: a relative error of $21.5\%$ at
      $A=2$, $7.7\%$ at $10$, $1.2\%$ at $10^2$, $1.7\times10^{-3}$ at $10^3$
      and $3\times10^{-6}$ at $10^6$. \emph{The closed form is the
      separated-rate asymptote and should be interpreted accordingly.} The upper bound
      $N\pi$ is the further sub-case in which every $q_k\to1$. The derivation
      is in \S\,``Proofs: Law~III'' below and the worked three-atom cases are
      in \TR{The knee: noise sensitivity near the threshold}.
\item \textup{(Speed limit.)} $L(x_1,x_2)^2\le(x_2-x_1)\,[\,s(x_1)-s(x_2)\,]$,
      and $\int\kappa\,\mathrm{d}x=s(x_1)-s(x_2)$, so the exponent is the
      total curvature of the frontier and $\pi$ its total root curvature.
\end{enumerate}
\end{proposition}

\begin{proofof}{Proposition~\ref{prop:path}}
\emph{(i).} $\{p_x\}$ is an exponential family with natural parameter $-x$,
sufficient statistic $a$ and log-partition function $\ln\eps$, so its Fisher
information with respect to $x$ is $\operatorname{Var}_{p_x}(a)=\kappa$, and
the Fisher--Rao length of a curve is the integral of the square root of its
Fisher information. Length is invariant under reparameterization: $x\mapsto
cx+d$ with $c>0$ pushes $\mu$ forward under $a\mapsto a/c$ and leaves
$\{p_x\}$ unchanged, while $\eps\mapsto C\eps$ adds a constant to $S$ and does
not change $S''$. Only $S''$ appears, and $S''$ annihilates the constant
$\epsinf$, so $L$ needs no estimate of the floor. With
$\mu=\epsinf\delta_0+c\,\delta_\alpha$, writing $p_1=1-\epsinf/\eps$ gives
$s=\alpha p_1$, $\kappa=\alpha^2p_0p_1$ and $p_1'=-\alpha p_0p_1$, so
\[
  \sqrt{\kappa}\,\mathrm{d}x=\alpha\sqrt{p_0p_1}\,\mathrm{d}x
  =\frac{\mathrm{d}p_0}{\sqrt{p_0(1-p_0)}}
  =\bigl|\mathrm{d}\bigl(2\arccos\sqrt{p_0}\bigr)\bigr|,
\]
which is~\eqref{eq:arc}; the parameters cancel from the arc length and
enter only through a reparameterization of the arc, and letting $x_1\to-\infty$, $x_2\to\infty$
sweeps $p_0$ over all of $(0,1)$ and gives $L\to2\arccos0=\pi$. Restricting
instead to $\eps\le1$, i.e., $p_0\ge\epsinf$, gives
$L=2\arccos\sqrt{\epsinf}<\pi$.

\emph{(ii).} A single atom makes $p_x$ a point mass for every $x$, so
$\kappa\equiv0$ and $L\equiv0$. For the Gamma spectrum
$\eps=C(1+x/\beta)^{-k}$ one has $s=k/(\beta+x)$ and $\kappa=k/(\beta+x)^2$,
so $\sqrt\kappa=\sqrt k/(\beta+x)$ and $L=\sqrt k\,\ln(1+X/\beta)$, which
diverges with the window.

\emph{(iii).} As $x\to-\infty$ the tilt drives $p_x\to\delta_{a_{\max}}$ and
as $x\to+\infty$ it drives $p_x\to\delta_0$. Under the Hellinger embedding
$p\mapsto2\sqrt p$ the simplex becomes the positive orthant of the sphere of
radius $2$, the Fisher--Rao metric becomes the round metric, and the geodesic
distance between two vertices is $2\arccos0=\pi$; any curve joining them has
length at least $\pi$. Equality forces the curve to be the minimizing
geodesic, which lies in the two-dimensional face spanned by $\delta_0$ and
$\delta_{a_{\max}}$, and the tilted path lies in that face for all $x$ if and
only if $\mu$ charges no other point. For the upper bound and the separation
limit, order the support $0=a_0<\dots<a_N$ and let $x_j$ be the crossover of
\eqref{eq:crossshare}, at which the $(j+1)$th term and the accumulated sum of
the slower ones are equal---not the pairwise crossover of the $j$th and
$(j+1)$th terms alone, which differs from it and is not what
\eqref{eq:crossshare} divides by. \emph{If the rates
separate}---$a_{j+1}/a_j\to\infty$, so that the $j$th transition is complete
before the $(j+1)$st begins---then near each crossover the remaining
terms are uniformly negligible and the path is, up to $o(1)$, the two-atom
path of (i) between the two atoms in play, so that each interior atom
contributes $2\arcsin\sqrt{q_k}$ of \eqref{eq:crossshare} and the first contributes $\pi$;
summing gives \eqref{eq:arcweights}. \emph{Two distinct limits must be kept
apart.} \eqref{eq:arcweights} is the separated-rate limit and holds for every
weight vector there; the upper bound $L\to N\pi$ is the sub-case in which
additionally every $q_k\to1$, i.e., in which each mode dominates the error
over a long stretch, and it is attained only there. Neither hypothesis is
removable. Two checks fix the regimes. With $c_j$ equal and exponents
$0,1,10^8$ the rates are separated enough that the closed form is accurate to
$4\times10^{-8}$---quadrature gives $1.4999999\pi$ against the form's
$1.5000000\pi$, which is the pair recorded in the archive's verification files,
and the limit is approached but not attained at any finite ratio. The two
crossovers then sit at the origin, mode $1$ never dominates, the
path passes through the edge midpoint $(\tfrac12,\tfrac12,0)$ rather than
through the vertex $\delta_1$, and $\pi+\pi/2=1.5\pi$ is the limiting value. With the same weights and exponents $0,1,2$
the rates are \emph{not} separated, the two kinks overlap, and quadrature at
$25$ digits gives $1.2970\pi$ against \eqref{eq:arcweights}'s $1.5759\pi$---an
error of $21.5\%$, the formula's one failure mode, and the reason it is stated
as a limit in the rate ratio and not as an identity. A mode interior to
the upper envelope is invisible on the frontier and its corner is cut.

\emph{(iv).} By Cauchy--Schwarz,
$L^2\le(x_2-x_1)\int\kappa=(x_2-x_1)[s(x_1)-s(x_2)]$, the last step by
$s'=-\kappa$; equality holds if and only if $\kappa$ is constant. The same
identity read without the inequality gives $\int\kappa\,\mathrm{d}x=
s(x_1)-s(x_2)$, which over the whole line is $\alpha$ for one reducible
channel.
\end{proofof}

\begin{proposition}[The arc forms]
\label{prop:arcforms}
With one reducible channel of exponent $\alpha$, and $\varphi$ the channel angle
of the paper's \eqref{eq:angle}---measured from the floor, so that
$\varphi=2\arccos\sqrt{p_0}=2\arctan\sqrt{N_{\mathrm{eff}}}$ and the mirror
$\pi-\varphi=2\arcsin\sqrt{p_0}$ runs the other way---
\begin{equation}
p_0=\cos^2\tfrac{\varphi}{2},
\quad
m=-\cos\varphi,
\quad
\zeta=-2\ln\tan\tfrac{\varphi}{2},
\quad
\sqrt{\kappa}=\tfrac{\alpha}{2}\sin\varphi,
\quad
\varphi=2\arctan\frac{s}{\sqrt{\kappa}},
\label{eq:arcforms}
\end{equation}
and
\begin{equation}
\boxed{\ R=-\ln\sin\varphi\ },
\qquad
J=\ln\sin\varphi+\mathrm{const},
\qquad
e^{-R}=\sin\varphi=2\sqrt{p_0p_1}=\frac{\GM}{\AM} .
\end{equation}
The canonical point is $\varphi=\pi/2$; the $\varrho$-nat band is
$\varphi\in[\arcsin e^{-\varrho},\pi-\arcsin e^{-\varrho}]$, which at
$\varrho=0.05$ is $[1.257,1.884]$, a fixed $20.0\%$ of every frontier; and the
two-candidate rule of Corollary~\ref{cor:two} is
\begin{equation}
\varphi_1+\varphi_2>\pi :
\label{eq:arcpair}
\end{equation}
prefer the more accurate candidate exactly when the pair's arc midpoint lies
beyond the frontier's midpoint, that is further from the floor.
\end{proposition}

\begin{proofof}{Proposition~\ref{prop:arcforms}}
All but the last are substitutions in $\varphi=2\arccos\sqrt{p_0}$:
$p_0=\cos^2(\varphi/2)$ by definition; $m=1-2p_0=-\cos\varphi$;
$\zeta=\ln\{p_0/(1-p_0)\}=-2\ln\tan(\varphi/2)$;
$\sqrt\kappa=\alpha\sqrt{p_0p_1}=\tfrac{\alpha}{2}\sin\varphi$ since
$2\sqrt{p_0p_1}=\sin\varphi$, both $\sin\varphi$ and $\sqrt{p_0p_1}$ being
invariant under the mirror; and $s^2/\kappa=p_1/p_0=\tan^2(\varphi/2)$ gives
$\varphi=2\arctan(s/\sqrt\kappa)$. For the regret, $\AM(p)=\tfrac12$ and
$\GM(p)=\sqrt{p_0p_1}$, so by Theorem~\ref{thm:master}
\[
  R=\ln\frac{\AM}{\GM}=\ln\frac{1}{2\sqrt{p_0p_1}}=-\ln\sin\varphi ,
\]
and $J=-R+\mathrm{const}$ by Corollary~\ref{cor:accvalue}. The quantity
$2\sqrt{p_0p_1}=\sum_j\sqrt{p_j\,p_{\sigma(j)}}$ with $\sigma$ the
mode-exchange involution is the Bhattacharyya affinity between $p$ and its
mirror, so $R$ is the Bhattacharyya distance from self-duality. The band is
$\sin\varphi\ge e^{-\varrho}$, and is symmetric about $\pi/2$, hence the same in
either orientation. For the pairwise form, $m_1+m_2>0$ reads
$\cos\varphi_1+\cos\varphi_2<0$, that is
$2\cos\frac{\varphi_1+\varphi_2}{2}\cos\frac{\varphi_1-\varphi_2}{2}<0$; the
second factor is positive since $|\varphi_1-\varphi_2|<\pi$, so the test is
$\varphi_1+\varphi_2>\pi$.
\end{proofof}

\begin{corollary}[The knee is the reference]
\label{cor:knee}
$-S''=(a\cdot v)^2p_0(1-p_0)=\tfrac14(a\cdot v)^2\sin^2\varphi$ is maximal
exactly at $\varphi=\pi/2$. The canonical operating point is therefore the
point of maximal curvature of the log-error in log-resource, and of maximal
Fisher information: the ``knee'' is not an approximation to the minimax rule
but coincides with it.
\end{corollary}

\begin{proofof}{Corollary~\ref{cor:knee}}
With one reducible channel the exponent is Bernoulli under $p_x$, taking the
value $a\cdot v$ with probability $1-p_0$, so
$-S''=\operatorname{Var}_{p_x}(a\cdot v)=(a\cdot v)^2p_0(1-p_0)
=\tfrac14(a\cdot v)^2\sin^2\varphi$ by Proposition~\ref{prop:arcforms}, and
$p_0(1-p_0)$ is maximal at $p_0=\tfrac12$, that is at $\varphi=\pi/2$. The
statement concerns $\mathrm{d}^2S/\mathrm{d}x^2$, which is invariant under
$t\mapsto\mu t$ and transforms by the factor $\beta^2$ under
$t\mapsto t^{\beta}$ without moving its maximizer; the Euclidean curvature of
the plotted curve $(x,S)$ is $|S''|/(1+S'^2)^{3/2}$, whose maximizer depends on
the aspect ratio and is not a property of the frontier.
\end{proofof}

\begin{proposition}[Three estimators]
\label{prop:threeways}
On a single-channel law the arc position is given by each of
\begin{equation}
\text{(i)}\ \varphi=2\arccos\sqrt{\epsinf/\eps},
\qquad
\text{(ii)}\ \varphi=2\arctan(s/\sqrt{\kappa}),
\qquad
\text{(iii)}\ \Delta\varphi=2\arccos e^{-\mathcal{J}},
\label{eq:threeways}
\end{equation}
where $\mathcal{J}=S(x_m)-\tfrac12[S(x_1)+S(x_2)]
=\ln\{\sqrt{\eps(x_1)\eps(x_2)}/\eps(x_m)\}\ge0$ is the log-convexity defect
at the midpoint of a window. Route (i) needs the floor and no derivatives;
route (ii) needs a local slope and curvature and no floor; route (iii) needs
three error values and no smoother, and $e^{-\mathcal{J}}$ is the
Bhattacharyya coefficient of the two tilted measures, so
$2\arccos e^{-\mathcal{J}}$ is their Fisher--Rao distance. Always
$\Delta\varphi_{\mathrm{(iii)}}\le L$, with equality if and only if the
spectrum has \emph{at most} two atoms---the single atom makes $\ln\eps$ affine,
so $\mathcal{J}=0$ and $\Delta\varphi_{\mathrm{(iii)}}=L=0$.
\end{proposition}

\begin{proofof}{Proposition~\ref{prop:threeways}}
Route (i) is the definition of $\varphi$ with $p_0=\epsinf/\eps$. Route (ii)
is the last identity of Proposition~\ref{prop:arcforms}, which uses only the
two moments and so is free of the floor by
Proposition~\ref{prop:floorfree}(i). For route (iii), the Bhattacharyya
coefficient of two tilted measures is
\[
  \sum_j\sqrt{p_j(x_1)p_j(x_2)}
  =\frac{\sum_jc_je^{-a_j(x_1+x_2)/2}}{\sqrt{\eps(x_1)\eps(x_2)}}
  =\frac{\eps(x_m)}{\sqrt{\eps(x_1)\eps(x_2)}}=e^{-\mathcal{J}},
\]
with $x_m$ the midpoint; the Fisher--Rao distance on the Hellinger sphere is
$2\arccos$ of that affinity, and $\mathcal{J}\ge0$ is the log-convexity of
$\eps$, equivalently the non-negativity of the $2\times2$ Hankel minor
$\eps(x_1)\eps(x_2)\ge\eps(x_m)^2$. A distance never exceeds the length of a
curve joining its endpoints, so $2\arccos e^{-\mathcal{J}}\le L$, with
equality exactly when the curve is the minimizing geodesic, which by
Proposition~\ref{prop:path}(iii) is the two-atom case.
\end{proofof}

\begin{table}[tbp]\centering
\caption{The covering law. ``Band'' is the $\varrho$-nat indifference interval
about the canonical point, $\varphi\in[\arcsin e^{-\varrho},
\pi-\arcsin e^{-\varrho}]$, as a fraction of the arc. $K$ is the exact
greedy covering number for the full arc and for its central $90\%$, and the
last column is the estimate \eqref{eq:covering}. Script~57 of the archive.}
\label{tab:covering}
\small
\begin{tabular}{lccccc}
\toprule
$\varrho$ (nats) & band & $\eps/\epsinf$ band & $K$ (full arc) & $K$ (central $90\%$) & $L_{\mathrm{rng}}/2\sqrt{2\varrho}$\\
\midrule
$0.02$ & $0.127$ & $[1.67,2.49]$ & $8$ & $7$ & $7.85$\\
$0.05$ & $0.200$ & $[1.53,2.89]$ & $5$ & $5$ & $4.97$\\
$0.10$ & $0.280$ & $[1.40,3.48]$ & $4$ & $3$ & $3.51$\\
$0.20$ & $0.389$ & $[1.27,4.70]$ & $3$ & $3$ & $2.48$\\
\bottomrule
\end{tabular}
\end{table}

\begin{proofof}{Theorem~\ref{thm:covering}}
In the arc coordinate the Fisher metric is the identity, so for nearby points
$\KL(p_\varphi\Vert p_{\hat\varphi})=\tfrac12(\varphi-\hat\varphi)^2
+O(|\varphi-\hat\varphi|^3)$. A set of $K$ checkpoints partitions an interval
of length $L_{\mathrm{rng}}$ into cells of half-width at least $L_{\mathrm{rng}}/(2K)$ for some cell,
so the worst-case regret is at least $\tfrac12(L_{\mathrm{rng}}/2K)^2$ up to the cubic
term; requiring this to be at most $\varrho$ gives
$K\ge L_{\mathrm{rng}}/(2\sqrt{2\varrho})$, with equality approached by equal spacing in
$\varphi$, which is the standard one-dimensional quantization argument under a
Riemannian metric and the same estimate that bounds the dissipation of a
finite-step thermodynamic protocol \citep{salamon1983,crooks2007}. The exact
numbers of Table~\ref{tab:covering} replace the quadratic approximation by the
exact asymmetric divergence and solve the covering problem greedily from one
end, which is optimal for an interval; the two agree to within one checkpoint
throughout. The resource budget follows from
$\mathrm{d}x=\mathrm{d}\varphi/\sqrt\kappa=2\,\mathrm{d}\varphi/(\alpha
\sin\varphi)$ and $\int\mathrm{d}\varphi/\sin\varphi=\ln\tan(\varphi/2)$,
giving a span of $(2/\alpha)\ln\{\tan(\pi(1+f)/4)/\tan(\pi(1-f)/4)\}$ for the
symmetric fraction $f$ of the arc, which diverges as $f\to1$: the endpoints of
the arc are at infinite resource distance.
\end{proofof}

Asymptotically, $L_{\mathrm{rng}}=\pi$ is also the Jeffreys volume of the single-channel
family, since $\int\sqrt{\kappa}\,\mathrm{d}x$ is exactly the normalizing
integral of the Jeffreys prior; by the redundancy asymptotics of
\citet{clarke1990} that volume is what sets the minimax redundancy of
repeated commitment, which is the asymptotic counterpart of the one-shot
$\ln(N+1)$ of Theorem~\ref{thm:canonical}. We do not use this.

\subsection{The binary reading of Laws II and III}
\label{sm:binary}

\begin{proofof}{Proposition~\ref{prop:binary}}
With one reducible channel the spectrum has two atoms, at $0$ and at $\alpha$,
so the share vector is $p=(p_0,p_1)$ with $p_0=\epsinf/\eps$. The Fisher--Rao
metric on the one-dimensional simplex is
$\mathrm{d}s^2=\mathrm{d}p_0^2/(p_0p_1)$, whose arclength from $p_0=1$---the
floor---is the channel angle $\varphi=2\arccos\sqrt{p_0}$, the mirror
$2\arcsin\sqrt{p_0}$ measuring the same arc from $p_0=0$, and whose total
length is
$\int_0^1\mathrm{d}p_0/\sqrt{p_0p_1}=\pi$, the Beta$(\tfrac12,\tfrac12)$
normalization; equivalently $\sqrt p=(\cos\tfrac\varphi2,\sin\tfrac\varphi2)$
is the unit-sphere (Hellinger) embedding, on which the simplex is a quarter
circle traversed at twice the angle. Writing $p^{\,\mathrm{swap}}=(p_1,p_0)$,
\[
\bigl\langle\sqrt p,\sqrt{p^{\,\mathrm{swap}}}\bigr\rangle
 =2\sin\tfrac\varphi2\cos\tfrac\varphi2=\sin\varphi
 =2\sqrt{p_0p_1}=\GM(p)/\AM(p)=e^{-R},
\]
the last two equalities being Corollary~\ref{cor:onefunction} at $N=1$. The
great-circle angle between $\sqrt p$ and $\sqrt{p^{\,\mathrm{swap}}}$ is
$\arccos\sin\varphi=|\pi/2-\varphi|$, so
$d_{\mathrm{FR}}(p,p^{\,\mathrm{swap}})=2|\varphi-\pi/2|$ and
$e^{-R}=\cos\{\tfrac12 d_{\mathrm{FR}}\}$, which is \eqref{eq:binaryR}.
\end{proofof}

Three consequences worth stating separately, because the paper uses them to
replace three separate explanations.

\begin{enumerate}[leftmargin=1.5em]
\item \textbf{$L=\pi$ is a diameter.} The Fisher--Rao diameter of
$\Delta^1$ is $\pi$, so Law~III says that a complete single-channel frontier
traverses the whole of one binary decision. Proposition~\ref{prop:moments} then
explains only why $q=\tfrac12$ is the exponent that sees it.
\item \textbf{$\mathbb{E}[R]=\ln2$ is the log-cosine integral.} With
$\theta=\varphi-\pi/2$ uniform on $(-\pi/2,\pi/2)$, $R=-\ln\cos\theta$ and
$\pi^{-1}\!\int_{-\pi/2}^{\pi/2}-\ln\cos\theta\,\mathrm{d}\theta=\ln2$. Since
$e^{-R}\mathrm{d}\varphi$ is the uniform measure on shares and
$\mathrm{d}\varphi$ the Jeffreys one, this is
$\KL(\text{Jeffreys}\Vert\text{uniform})$ on a single binary outcome; the
$N$-mode version of Proposition~\ref{prop:regretlaw} is the same quantity for
Dirichlet$(\tfrac12,\dots,\tfrac12)$, with limit $\gamma+\ln2$.
\item \textbf{One doubling fixes the covering number.} By
\eqref{eq:lncosh}, $\varrho_2=\ln\cosh(\tfrac12\ln2)=0.05889$ and
$\pi/(2\sqrt{2\varrho_2})=4.577$, so the paper's tolerance and its headline
``five operating points'' are the single choice of a factor of two. Inverting
$\ln\cosh(\tfrac12\ln r)=0.05$ gives $r=1.8922$, which is the ratio
$2.89/1.53$ of the band the paper quotes at that tolerance.
\end{enumerate}

\subsection{The regret law and the curvature moments}
\label{sup:regretlaw}

This subsection proves Corollaries~\ref{cor:onefunction}--\ref{cor:jeffreys},
Proposition~\ref{prop:regretlaw} and Proposition~\ref{prop:moments}. All are
verified numerically by script~\texttt{65}.

\begin{corollary}[One function]
\label{cor:onefunction}
Write $R(x)=\KL(u\Vert p(x))$---\emph{here and in Theorem~\ref{thm:master}
the spectrum's own regret, not the single-channel $R_1$ of the convention
above; the two coincide for one reducible channel}. Then $R=\ln\{\AM(e)/\GM(e)\}
=\mathrm{const}-\sigma_{\abar}(x)$, and the three laws of this note are three
properties of this single function: it is convex in the log-resource, which is
Law~I; it vanishes exactly at equipartition of the error across modes, which is
Law~II; and for one reducible channel, in the arc coordinate of
Section~\ref{sec:lawIII}, it is $-\ln\cos\theta$ on an interval of length
$\pi$, which is Law~III. The third clause is the single-channel one: off two
atoms $-\ln\sin\varphi$ is the regret of the tangent two-atom chart and not
$\KL(u\Vert p)$ itself.
\end{corollary}

\begin{proofof}{Corollary~\ref{cor:onefunction}}
With $e_j=c_je^{-a_j\cdot x}$ and $p=e/\eps$, $\AM(p)=1/(N+1)$, so
$\KL(u\Vert p)=-\ln(N+1)-\frac1{N+1}\sum_j\ln p_j=\ln\{\AM(p)/\GM(p)\}$, and
the ratio is scale-invariant, so shares may be replaced by the $e_j$. Then
$\ln\GM(e)=\frac1{N+1}\sum_j\ln c_j-\abar\cdot x$ is affine in $x$, whence
$R=\mathrm{const}+\abar\cdot x+\ln\eps(x)$. Convexity is convexity of
$\ln\eps$, which is \eqref{eq:cgf}: a cumulant generating function is convex.
The zero is the equality case of the arithmetic--geometric mean inequality,
which is Theorem~\ref{thm:master}, and the arc form is
Proposition~\ref{prop:arcforms}.
\end{proofof}

\begin{corollary}[Group form of the regret]
\label{cor:groupform}
$e^{-R}=(N+1)\GM(p)$ is the multi-distribution Bhattacharyya affinity of the
operating point with its own cyclic mode-permutations,
$e^{-R}=\sum_{k}\prod_{j}p_{(k+j)\bmod(N+1)}^{1/(N+1)}$. The regret is minus
the log of the distance from permutation self-duality at every order; the
$N=1$ statement of Proposition~\ref{prop:arcforms}, where the affinity is
taken against the single mode-exchanged mirror, is the first case.
\end{corollary}

\begin{proofof}{Corollary~\ref{cor:groupform}}
Each cyclic product $\prod_{j=0}^{N}p_{(k+j)\bmod(N+1)}^{1/(N+1)}$ runs over
every index exactly once and therefore equals $\GM(p)$, independently of $k$;
summing over the $N+1$ shifts gives $(N+1)\GM(p)$, which is
$\GM(p)/\AM(p)=e^{-R}$ by the proof above. At $N=1$ the two shifts are $p$ and
its mode-exchanged mirror and the sum is $2\sqrt{p_0p_1}$, the Bhattacharyya
affinity of \eqref{eq:lnsin}.
\end{proofof}

\begin{corollary}[Jeffreys sum rule]
\label{cor:jeffreys}
With one reducible channel,
$\KL(u\Vert p)+\KL(p\Vert u)=\zeta\,(p_0-\tfrac12)=-\tfrac12 m\zeta$: the
symmetrized divergence from equipartition is half the product of the conjugate
pair $(\zeta,m)$ of Proposition~\ref{prop:arcforms}. The operating-point
regret $\KL(u\Vert p)$ dominates the declaration regret $\KL(p\Vert u)$ at
every point, their difference being
$\sum_{k\ge1}m^{2k}(2k-2)/\{2k(2k-1)\}\ge0$.
\end{corollary}

\begin{proofof}{Corollary~\ref{cor:jeffreys}}
With $m=1-2p_0$ and $t=-m$, $\KL(u\Vert p)=-\tfrac12\ln(1-t^2)$ and
$\KL(p\Vert u)=\tfrac12\{(1+t)\ln(1+t)+(1-t)\ln(1-t)\}$. Adding,
\[
\KL(u\Vert p)+\KL(p\Vert u)
=\tfrac12\ln\frac{(1+t)^{1+t}(1-t)^{1-t}}{1-t^2}
=\tfrac{t}{2}\ln\frac{1+t}{1-t}=t\,\mathrm{arctanh}\,t,
\]
and $\mathrm{arctanh}\,t=\zeta/2$ because $t=\tanh(\zeta/2)$ by
\eqref{eq:arcforms}, giving $\tfrac12 t\zeta=\zeta(p_0-\tfrac12)$. Expanding
both in powers of $t$, $\KL(u\Vert p)=\sum_k t^{2k}/(2k)$ and
$\KL(p\Vert u)=\sum_k t^{2k}/\{2k(2k-1)\}$, whose difference is
$\sum_k t^{2k}(2k-2)/\{2k(2k-1)\}\ge0$.
\end{proofof}

\begin{proofof}{Proposition~\ref{prop:regretlaw}}
By \eqref{eq:lnsin}, $R=-\ln\sin\varphi=-\ln\cos\theta$ with
$\theta=\varphi-\pi/2$, so $R\le\varrho$ exactly when
$|\theta|\le\arccos e^{-\varrho}$. Arc length is $\mathrm{d}\varphi$, so the
uniform-in-arc probability of that event is
$2\arccos e^{-\varrho}/\pi$, which is \eqref{eq:regretlaw}; expanding
$\arccos e^{-\varrho}=\sqrt{2\varrho}\,(1+O(\varrho))$ gives the square-root
form, with constant $2\sqrt2/\pi=0.9003$. For the mean,
\[
\frac1\pi\int_0^\pi(-\ln\sin\varphi)\,\mathrm{d}\varphi=\ln2
\]
is the classical log-sine integral; the median is $R$ at $\varphi=\pi/4$,
namely $-\ln(1/\sqrt2)=\tfrac12\ln2$, because $R$ is symmetric about $\pi/2$
and strictly monotone on each half. For general $N$, arc length on the share
simplex in the Fisher--Rao metric induces the Dirichlet$(\tfrac12,\dots)$ law,
under which $\mathbb{E}[\ln p_j]=\psi(\tfrac12)-\psi(\tfrac{N+1}{2})$, so
\[
\mathbb{E}[\KL(u\Vert p)]=-\ln(N+1)-\tfrac1{N+1}\textstyle\sum_j\mathbb{E}[\ln p_j]
=\psi\bigl(\tfrac{N+1}{2}\bigr)-\psi\bigl(\tfrac12\bigr)-\ln(N+1),
\]
which is \eqref{eq:meanregretN}. It is increasing in $N$ and, by
$\psi(z)=\ln z+O(1/z)$ and $\psi(\tfrac12)=-\gamma-2\ln2$, tends to
$\gamma+\ln2$. At $N=1$ it is $\psi(1)-\psi(\tfrac12)-\ln2=2\ln2-\ln2=\ln2$;
at $N=3$ it is $\psi(2)-\psi(\tfrac12)-\ln4=1+2\ln2-2\ln2=1$.

Finally, $|\mathrm{d}p_0/\mathrm{d}\varphi|=\tfrac12\sin\varphi=\tfrac12e^{-R}$
---the derivative itself is $-\tfrac12\sin\varphi$, $p_0$ falling as
$\varphi$ rises away from the floor in the orientation of
Table~\ref{tab:coordinates}---so
$R=-\ln|\mathrm{d}p_0/\mathrm{d}\varphi|-\ln2$: $e^{R}$ is proportional to
the arcsine density $1/\{\pi\sqrt{p_0p_1}\}$ relative to Lebesgue measure on
shares, $\int_0^\pi e^{-R}\mathrm{d}\varphi=\int_0^\pi\sin\varphi\,
\mathrm{d}\varphi=2$, and
$\mathbb{E}[R]=\KL(\text{arcsine}\Vert\text{uniform})+\ln(\pi/2)
=\ln(4/\pi)+\ln(\pi/2)=\ln2$, which is the second proof of \eqref{eq:onebit}.

\emph{The admissible part.} Restricting the uniform placement to
$\varphi\in(0,\varphi_{\max})$ with $\varphi_{\max}=2\arccos\sqrt{\epsinf}$
changes only the range. With $a=\arcsin e^{-\varrho}$ the event $R\le\varrho$ is
$\varphi\in[a,\pi-a]$ as before, so its intersection with $(0,\varphi_{\max})$
has length $\{\min(\varphi_{\max},\pi-a)-a\}^{+}$ and
\[
\Pr(R\le\varrho)=\frac{\{\min(\varphi_{\max},\pi-a)-a\}^{+}}{\varphi_{\max}},
\qquad
\mathbb{E}[R]=M(\varphi_{\max}),
\]
\[
M(L):=-\frac{1}{L}\int_{0}^{L}\!\ln\sin\varphi\,\mathrm{d}\varphi
=\ln2+\frac{\mathrm{Cl}_{2}(2L)}{2L}
\]
by the log-sine integral $\int_{0}^{\theta}\ln\{2\sin(x/2)\}\mathrm{d}x
=-\mathrm{Cl}_{2}(\theta)$ with $\mathrm{Cl}_2$ the Clausen function. Since
$\mathrm{Cl}_2$ vanishes at $0$, $\pi$ and $2\pi$, $M(\pi/2)=M(\pi)=\ln2$, the
second being the complete case. Differentiating,
$M'(L)=-L^{-1}\{M(L)+\ln\sin L\}$, which vanishes exactly where
$M(L)=-\ln\sin L$; that equation has the single root
$L=2.4857116$ on $(0,\pi)$, a minimum, with $M=0.4945296$, and
$L=2\arccos\sqrt{\epsinf}$ puts it at $\epsinf=0.1037445$. So $M$ falls from
$+\infty$ at $L\to0$ to that minimum and rises to $\ln2$ at $L=\pi$, crossing
$\ln2$ once on the way down at $L=\pi/2$, i.e., at $\epsinf=\tfrac12$. Hence
$M\le\ln2$ exactly on $\varphi_{\max}\in[\pi/2,\pi]$, that is on
$\epsinf\le\tfrac12$. \emph{\eqref{eq:onebit} bounds the
admissible case for every frontier whose floor is at most half its chance-level
error, and for no other.}
\end{proofof}

\begin{proofof}{Proposition~\ref{prop:moments}}
By \eqref{eq:arcforms}, $\kappa=\tfrac14\alpha^2\sin^2\varphi$ and
$\mathrm{d}x=(2/\alpha)\,\mathrm{d}\varphi/\sin\varphi$, so
\[
\kappa^q\,\mathrm{d}x
=\Bigl(\frac{\alpha}{2}\Bigr)^{2q-1}\sin^{2q-1}\varphi\,\mathrm{d}\varphi,
\qquad
\int_0^\pi\sin^{2q-1}\varphi\,\mathrm{d}\varphi
=\frac{\sqrt\pi\,\Gamma(q)}{\Gamma(q+\tfrac12)},
\]
the last being the Wallis integral, finite for $q>0$. The $\alpha$-dependence
is $\alpha^{2q-1}$ and vanishes only at $q=\tfrac12$, where the integrand is
$\mathrm{d}\varphi$ and the value is $\pi$; at $q=1$ the value is
$(\alpha/2)\cdot2=\alpha$, which is Proposition~\ref{prop:path}(iv). Inverting
\eqref{eq:moments} gives \eqref{eq:alphaq}, which returns $\alpha$ for every
$q$ when exactly one reducible channel is active. Conversely, if
$\hat\alpha_q$ is constant in $q$ then all moments of $\sqrt\kappa\,
\mathrm{d}x$ agree with those of the two-atom family, which determines the
family because the measure has bounded support.

The kink interpretation follows from the same computation, written in the
forward direction. In the orientation of
the paper, $\varphi=2\arctan\sqrt{\Neff}=2\arccos\sqrt{p_0}$ measured from the
floor, the angle \emph{decreases} with the resource:
$\mathrm{d}\varphi/\mathrm{d}x=-\sqrt\kappa=-(\alpha/2)\sin\varphi$, which
integrates to $\varphi(x)=2\arctan e^{-\alpha(x-x_0)/2}$, the standard kink
profile of width $2/\alpha$ running from $\varphi=\pi$ at the pure power law
down to $\varphi=0$ at the floor; $L=\pi$ is the difference of the boundary
values over the complete sweep and hence independent of $\alpha$. Writing
$\zeta=-2\ln\tan(\varphi/2)$, as in Table~\ref{tab:coordinates}, and
$\theta=\varphi-\pi/2$ gives
$\zeta=-2\ln\tan(\theta/2+\pi/4)=-2\,\mathrm{gd}^{-1}(\theta)$, so
log-resource is the Mercator coordinate along the arc as it is traversed
inward.
\end{proofof}

\subsection{The curvature-moment family}
\label{sup:moments}

Law~III gives $\int\sqrt\kappa\,\mathrm{d}x=\pi$ and
Proposition~\ref{prop:path}(iv) gives $\int\kappa\,\mathrm{d}x=\alpha$. These two integral identities are linked by the curvature-moment formula
below.

\begin{proposition}[Curvature moments]
\label{prop:moments}
For a floor and one reducible channel of exponent $\alpha$ along a direction,
\begin{equation}
M_q=\int\kappa^q\,\mathrm{d}x
   =\Bigl(\frac{\alpha}{2}\Bigr)^{2q-1}
    \frac{\sqrt\pi\,\Gamma(q)}{\Gamma(q+\tfrac12)},\qquad q>0 .
\label{eq:moments}
\end{equation}
Hence $q=\tfrac12$ is the \emph{unique} exponent at which the frontier's own
scale cancels, with $M_{1/2}=\pi$; $q=1$ returns $M_1=\alpha$; and
\begin{equation}
\hat\alpha_q=2\Bigl[M_q\,\Gamma(q+\tfrac12)\big/\{\sqrt\pi\,\Gamma(q)\}\Bigr]^{1/(2q-1)}
\label{eq:alphaq}
\end{equation}
is independent of $q$ if and only if exactly one reducible channel is active.
Without any restriction on $\mu$,
\begin{equation}
\int_{-\infty}^{\infty}\kappa\,\mathrm{d}x
 = s(-\infty)-s(\infty)
 = a_{\max}-a_{\min},
\label{eq:widthsup}
\end{equation}
the width of $\operatorname{supp}\mu$: total curvature is spectral width, and
$M_1=\alpha$ is its two-atom case.
\end{proposition}

The proof is the substitution $\kappa^q\mathrm{d}x
=(2/\alpha)(\alpha^2/4)^q\sin^{2q-1}\varphi\,\mathrm{d}\varphi$, after which
the $\alpha$-dependence is $\alpha^{2q-1}$ and the integral is a Beta
function. Equation~\ref{eq:alphaq} is an infinite family of floor-free,
fit-free one-channel tests which, being built from integrals rather than from
pointwise ratios, avoid the bandwidth-related limitation that leaves the
statistic discussed in Section~\ref{sec:lawfree} inconclusive. The family is
established analytically, but its empirical performance has not yet been
evaluated; assessing it on corpora with longer observation windows is a
natural next step.

\subsection{Reading a frontier without fitting it}
\label{sup:pf-width}

Throughout, $\eps(x)=\int_{[0,\infty)}e^{-ax}\,d\mu(a)$ is finite on an open
interval, $S=-\ln\eps$, $s=S'$, $\kappa=-S''$, and
\[
  p_x(da)\;=\;\frac{e^{-ax}\,d\mu(a)}{\eps(x)}
\]
is the tilted measure of Proposition~\ref{prop:selection}. We use throughout
that $\ln\eps(x)$ is the cumulant generating function of $\mu$ at $-x$, so
that $s(x)=\mathbb{E}_{p_x}[a]$, $\kappa(x)=\mathrm{Var}_{p_x}(a)\ge0$ with
equality if and only if $p_x$ is a point mass, and $\kappa_3=S'''$ is the
third cumulant; differentiating in $x$ lowers the order by one with
alternating sign, $s'=-\kappa$ and $\kappa'=-\kappa_3$.

\begin{proposition}[Floor-free exponent and width]
\label{prop:floorfree}
Put $\hat\alpha=\mathbb{E}_{p_x}[a^2]/\mathbb{E}_{p_x}[a]=s+\kappa/s$ and
$g=\kappa/s^2$. Then (i) because the floor is the atom at $a=0$ it enters
neither moment, so $\hat\alpha$ is identified without estimating $\epsinf$;
(ii) $\hat\alpha\equiv\alpha$ at every $x$ if and only if exactly one reducible
channel is active, and $\hat\alpha$ is otherwise strictly decreasing in $x$;
(iii) with two atoms $g=p_0/(1-p_0)$, so the floor is recovered pointwise,
$\epsinf=\eps\,\kappa/(s^2+\kappa)$, and the canonical reference is the point
$s^2=\kappa$.
\end{proposition}

\begin{proofof}{Proposition~\ref{prop:floorfree}}
Along the chosen direction,
\[
  \hat\alpha=s+\frac{\kappa}{s}
  =\frac{\mathbb{E}[a]^2+\operatorname{Var}(a)}{\mathbb{E}[a]}
  =\frac{\mathbb{E}[a^2]}{\mathbb{E}[a]},
\]
which is~\eqref{eq:sizebiased}; the atom at $a=0$ contributes zero to both
moments, so $\hat\alpha$ does not depend on $\mu(\{0\})=\epsinf$, proving
part~(i). For part~(ii), if exactly one reducible exponent $\alpha$ is
present then $\mathbb{E}[a^2]=\alpha^2p_1$ and $\mathbb{E}[a]=\alpha p_1$, so
$\hat\alpha=\alpha$ identically. Conversely, $\hat\alpha$ is the mean of $a$
under the size-biased measure $\mathrm{d}\tilde p_x\propto a\,\mathrm{d}p_x$,
which by Proposition~\ref{prop:selection} is itself a tilted family (generated
by $a\mu$) and so obeys the same selection law: the derivative of $\hat\alpha$
is $-\operatorname{Var}_{\tilde p_x}(a)\le0$, vanishing only when $\tilde p_x$
is a point mass, that is, when exactly one reducible exponent carries mass.
Hence $\hat\alpha$ is constant if and only if one reducible channel is active
and is otherwise strictly decreasing. For part~(iii), with
$\eps=\epsinf+ce^{-\alpha x}$ and reducible share $p_1$ we have
$s=\alpha p_1$ and $\kappa=\alpha^2p_1(1-p_1)$, so $g=\kappa/s^2=p_0/p_1$;
$s^2=\kappa$ is equivalent to $p_1=\tfrac12$, which is the reference of
Corollary~\ref{cor:single}; and solving the same two equations for $p_1$
gives $1-p_1=\kappa/(s^2+\kappa)$, hence
$\epsinf=\eps\kappa/(s^2+\kappa)$, while $\hat\alpha=\alpha$ by part~(ii).
\end{proofof}

\begin{proofof}{the width remark of Section~\ref{sec:lawfree}}
By Proposition~\ref{prop:floorfree}(i), $s=\mathbb{E}_{p_x}[a]$ and
$\kappa=\operatorname{Var}_{p_x}(a)$, so
$\kappa/s^2=\operatorname{Var}_{p_x}(a)/\mathbb{E}_{p_x}[a]^2=\mathrm{CV}^2$,
and $\hat\alpha/s-1=(s+\kappa/s)/s-1=\kappa/s^2$. Non-negativity is that of a
variance, and $g=0$ if and only if $p_x$ is a point mass off the origin,
that is, exactly one reducible channel is active and there is no floor.
Neither moment involves the atom at the origin, so $g$ does not depend on
$\epsinf$.
\end{proofof}

\begin{proofof}{Proposition~\ref{prop:widthcoord}}
(i) With $\zeta=\ln\kappa-2\ln s$,
\[
  \zeta'=\frac{\kappa'}{\kappa}-\frac{2s'}{s}
        =-\frac{\kappa_3}{\kappa}+\frac{2\kappa}{s},
\]
valid wherever $s>0$ and $\kappa>0$, hence wherever $\mu$ is not a single
atom.

(ii) For $\mu=\epsinf\delta_0+c\,\delta_\alpha$ the exponent is Bernoulli
under $p_x$, taking the value $\alpha$ with probability $p_1=1-p_0$ and $0$
with probability $p_0$. Hence $s=\alpha p_1$, $\kappa=\alpha^2p_0p_1$ and
$\kappa_3=\alpha^3p_0p_1(p_0-p_1)$. Therefore
$g=\kappa/s^2=p_0/p_1$, which is~\eqref{eq:zetaidentity} after taking
logarithms, and substituting into (i),
\[
  \zeta'=\frac{2\alpha^2p_0p_1}{\alpha p_1}
        -\frac{\alpha^3p_0p_1(p_0-p_1)}{\alpha^2p_0p_1}
        =2\alpha p_0-\alpha(p_0-p_1)
        =\alpha(p_0+p_1)=\alpha .
\]
Alternatively, $\zeta=\ln(p_0/p_1)$ is the natural parameter of the Bernoulli
family and the selection law~\eqref{eq:master} gives
$p_1'=-\alpha p_0p_1$, whence $\zeta'=-p_1'/(p_0p_1)=\alpha$ directly.
Solving $g=p_0/p_1$ for the shares gives $p_0=g/(1+g)=\kappa/(s^2+\kappa)$,
and $p_0=\epsinf/\eps$ by~\eqref{eq:p0def} gives~\eqref{eq:localfloor}.
Finally $\zeta=0$ iff $g=1$ iff $p_0=p_1=\tfrac12$, so $\epsinf=\eps/2$
there, which is~\eqref{eq:halffloor}; the point exists and is unique because
$\zeta$ is affine with slope $\alpha>0$.

(iii) Setting the general expression in (i) equal to
$\hat\alpha=s+\kappa/s$ and clearing denominators gives
$2\kappa^2-s\kappa_3=\kappa(s^2+\kappa)$, that is
$\kappa_3=\kappa(\kappa-s^2)/s$; the Bernoulli cumulants above satisfy it, and
conversely the identity determines $\kappa_3$ from $(s,\kappa)$, which is the
closure of the hierarchy at second order in Proposition~\ref{prop:logistic}.
Hence $\mathcal{R}=\zeta'/\hat\alpha=1$ in the two-atom case. For a Gamma
spectrum, Proposition~\ref{prop:gamma} gives $g$ constant, so $\zeta'=0$ and
$\mathcal{R}=0$, while $\hat\alpha=s(1+g)$ remains positive.
\end{proofof}

\begin{proofof}{Proposition~\ref{prop:gamma}}
Since $s>0$ on the interval,
\[
  \Bigl(\frac1s\Bigr)' \;=\; -\frac{s'}{s^2} \;=\; \frac{\kappa}{s^2} \;=\; g,
\]
which is~\eqref{eq:ginv}. Hence $g\equiv1/k$ is constant if and only if
$1/s$ is affine, $1/s(x)=(x+\beta)/k$ for constants $k>0$ and $\beta$, that
is $s(x)=k/(x+\beta)$. Integrating, $S(x)=k\ln(x+\beta)+\mathrm{const}$, so
\[
  \eps(x)\;=\;C\,(1+x/\beta)^{-k},
\]
which is~\eqref{eq:gammafrontier}. Conversely~\eqref{eq:gammafrontier} gives
$s=k/(x+\beta)$ and $\kappa=k/(x+\beta)^2$, so $g=1/k$. The function
$(1+x/\beta)^{-k}$ is, up to the constant $C$, the Laplace transform of the
Gamma density $\beta^k a^{k-1}e^{-\beta a}/\Gamma(k)$, and by uniqueness of
the Laplace transform this identifies $\mu$. A Gamma measure is absolutely
continuous, so $\mu(\{0\})=0$ and $\epsinf=\lim_{x\to\infty}\eps(x)=0$; and
\[
  L(x,\infty)=\int_x^\infty\!\sqrt{k}\,\frac{du}{u+\beta}=\infty .
\]
For the converse case, suppose $\mu(\{0\})=\epsinf>0$ and
$\alpha:=\min\bigl(\mathrm{supp}\,\mu\setminus\{0\}\bigr)>0$. Then
$p:=1-\epsinf/\eps\to0$ and, by the two-atom asymptotics,
$s\sim\alpha p$ and $\kappa\sim\alpha^2p$, so $g\sim1/p\to\infty$; and
$g'=(1/s)''>0$ for large $x$ because $1/s\sim\eps/(\alpha\,\epsinf\,p)\cdot
\epsinf$ grows exponentially. Hence $g$ is eventually strictly increasing.
\end{proofof}

\begin{proofof}{the log-convexity condition of Corollary~\ref{cor:prony}}
For $x_2=(x_1+x_3)/2$, H\"older's inequality applied to
$\eps(x_2)=\int e^{-a(x_1+x_3)/2}\mathrm{d}\mu
=\int(e^{-ax_1})^{1/2}(e^{-ax_3})^{1/2}\mathrm{d}\mu$ gives
$\eps(x_2)^2\le\eps(x_1)\eps(x_3)$, with equality if and only if
$e^{-ax_1}$ and $e^{-ax_3}$ are proportional $\mu$-almost everywhere, that
is, $\mu$ is a single atom. Geometrically spaced resources are equally spaced
in $x$, so the admissibility condition $\eps_1\eps_3>\eps_2^2$ of the
three-point estimator is strict log-convexity of $\eps$, which holds for
every non-degenerate member of the family and fails outside it.
\end{proofof}

\begin{proofof}{Proposition~\ref{prop:resolution}}
Part~(i) is the cumulant expansion of the log-Laplace transform: with
$\eps-\epsinf=\int_{a>0}e^{-au}\mathrm{d}\mu$, the function
$u\mapsto\ln(\eps-\epsinf)$ is the cumulant generating function of the
restricted measure evaluated at $-u$, so its Taylor coefficients are the
cumulants $k_j$ at the expansion point, with alternating signs. For
part~(ii), a single exponential is an affine function of $u$ in this
coordinate, so the deviation to be approximated is
$\tfrac12k_2u^2+O(u^3)$; on an interval of length $Z$ the best uniform affine
approximation of $cu^2$ has error $c(Z/2)^2/2$, attained at the endpoints and
the midpoint, which with $c=k_2/2$ equals $k_2Z^2/16$. The same argument
applied to the leading remaining term shows that $k_j$ first contributes at
order $Z^j$, with the Chebyshev constant of degree $j$. Part~(iii) compares
that deviation with the noise: with $m$ points and independent log-errors of
scale $\sigma$, a fixed shape of amplitude $A$ is detectable when
$A\sqrt m/\sigma$ exceeds a constant of order one; substituting
$A=k_2Z^2/16$ and normalizing exponents by $k_1$ gives~\eqref{eq:resnumber},
and substituting the third-order term gives $\mathcal{V}_3$. The constant
depends on the selection criterion, which is why
\TR{Calibration of the resolution number} calibrates it by simulation.
\end{proofof}

\subsection{The floor}
\label{sup:pf-floor}

\begin{proposition}[What a finite experiment identifies]
\label{prop:identify}
Let $\mu$ be a spectral measure as in Proposition~\ref{prop:spectral} and let
$x_1,\dots,x_m$ be log-resource points at which $\eps$ is finite.
\begin{enumerate}[label=(\roman*),leftmargin=2em,itemsep=1pt]
\item \emph{A finite set of observations cannot establish whether the spectrum has finite support.} There is an atomic
      measure $\nu$ with at most $m$ atoms, all in $\operatorname{supp}\mu$
      and with positive weights, matching $\eps$ at every $x_k$. Any
      observation model whose law depends on $\mu$ only through those values
      ---item sampling, log-error noise, and any Pareto or other filter
      applied to the observed values---induces the same distribution under
      $\mu$ and under $\nu$, and no test of ``finitely many modes'' against
      ``a continuum'' has power exceeding its level.
\item \emph{An accumulating set of observations identifies the spectrum.} If two spectral measures give the same
      $\eps$ on a subset of a ray with an accumulation point interior to the
      domain of finiteness, they are equal.
\end{enumerate}
\end{proposition}

\begin{proofof}{Proposition~\ref{prop:identify}}
(i) Let $f_k(a)=e^{-a\cdot x_k}$, $k=1,\dots,m$; each is bounded and
$\mu$-integrable. Tchakaloff's theorem in the form valid for arbitrary finite
measures \citep{tchakaloff1957,bayer2006} supplies an atomic measure $\nu$
with at most $m$ atoms, all lying in $\operatorname{supp}\mu$ and carrying
positive weights, such that $\int f_k\,\mathrm{d}\nu=\int f_k\,\mathrm{d}\mu$
for every $k$; the bound on the number of atoms is Carath\'eodory's theorem
applied to the moment vector $(\int f_k\,\mathrm{d}\mu)_k$ in the convex hull
of $\{(f_k(a))_k:a\in\operatorname{supp}\mu\}$. Then
$\eps_\nu(x_k)=\eps_\mu(x_k)$ for every $k$. Any observation model whose law
depends on the spectrum only through the values $\eps(x_1),\dots,\eps(x_m)$
---item sampling, log-error noise applied to those values, and any Pareto or
other filter applied to the observed values---therefore induces the same
distribution under $\mu$ and under $\nu$. A test of ``finitely many modes''
against ``a continuum'' would have to separate two hypotheses that induce
the same distribution of the data, so its power cannot exceed its level.
Specialized at the origin: if $\mu$ has an atom $\epsinf\delta_0$ then
replacing it by mass just above the origin changes no $\eps(x_k)$ by more
than any prescribed amount, which is Proposition~\ref{prop:running}(i).

(ii) Along a ray $u\mapsto x+ud$, $u\mapsto\eps(x+ud)=\int e^{-u\,a\cdot d}
e^{-a\cdot x}\,\mathrm{d}\mu(a)$ is the Laplace transform of the pushforward
of $e^{-a\cdot x}\mu$ under $a\mapsto a\cdot d$, and a Laplace transform is
analytic in the interior of its domain of finiteness. Two analytic functions
that agree on a set with an accumulation point interior to that domain agree
on the whole ray, and by uniqueness of the Laplace transform the two
pushforwards coincide. If the two spectral measures agree in this sense along
every direction $d$ in an open cone---equivalently, if the two frontiers
coincide on an open set of log-resources---then their multivariate Laplace
transforms agree on an open set and the measures are equal; for $n=1$ the
pushforward is the measure itself.
\end{proofof}

\begin{proposition}[The running floor]
\label{prop:running}
Observe $\eps$ on a window $[x_0,x_0+X]$ with log-error noise of scale
$\sigma_\ell$.
\begin{enumerate}[label=(\roman*),leftmargin=2em,itemsep=1pt]
\item Replacing the atom $\epsinf\delta_0$ by $\epsinf\delta_a$ perturbs
      $\eps$ by at most $\epsinf(1-e^{-a(x_0+X)})$ on the window, so every
      exponent with $a\lesssim\sigma_\ell/(x_0+X)$ is observationally equivalent
      to the floor.
\item The sharp upper bound available from the window \emph{without using the
      curvature} is $\eps(x_0+X)$, which
      is non-increasing in $X$ (Theorem~\ref{thm:monofloor} replaces it by a
      strictly smaller one that is non-increasing for the same reason); any
      estimator consistent with the model class
      therefore has non-increasing positive bias, and the identified set given
      noiseless observations is $[0,\min_k\eps(x_k)]$. The bound is on the
      curve and not on the measurements: under noise $\min_k\hat\eps_k$ is
      \emph{not} an upper bound for $\epsinf$, a fitted floor above the best
      observed error is an ordinary residual rather than an infeasibility, and
      imposing $\epsinf\le\min_k\hat\eps_k$ biases the fit downward exactly
      where the question arises---on a saturated window at $\sigma_\ell=0.05$ the
      constrained estimator loses $6\%$ of the floor where the unconstrained
      one loses none.
\item The rate is a spectral question: with a gap at the origin,
      $\hat\epsinf(X)-\epsinf=O(e^{-a_-X})$; with no gap the excess decays
      only at the rate at which $\mu$ accumulates at the origin.
\end{enumerate}
\end{proposition}

\begin{proofof}{Proposition~\ref{prop:running}}
(i) Let $\mu'$ be $\mu$ with the atom $\epsinf\delta_0$ replaced by
$\epsinf\delta_a$, $a>0$. For $x$ in the window,
\[
  0\le\eps(x)-\eps'(x)=\epsinf\bigl(1-e^{-ax}\bigr)
   \le\epsinf\bigl(1-e^{-a(x_0+X)}\bigr)\le\epsinf\,a\,(x_0+X),
\]
so the two frontiers differ in relative terms by at most $a(x_0+X)$, which is
below the noise scale $\sigma$ whenever $a\le\sigma/(x_0+X)$. Since
$\mu'(\{0\})=0$, a frontier with no floor and a frontier with floor $\epsinf$
are observationally equivalent on the window, and the floor is not identified.

(ii) $\eps$ is non-increasing along the direction, and
$\eps(x)\ge\epsinf$ for all $x$ because the atom contributes $\epsinf$ to
every value of the integral. Hence $\epsinf\le\eps(x_0+X)$, and this bound is
sharp: the measure $\eps(x_0+X)\delta_0$ plus the modes fitted on the window
reproduces the window exactly at its right edge. As $X$ increases,
$\eps(x_0+X)$ decreases, so the sharp bound is non-increasing, and any
estimator constrained to the model class inherits both the sign and the
monotonicity of the bias up to noise.

(iii) If $\mu\bigl((0,a_-)\bigr)=0$ then
$\eps(x)-\epsinf=\int_{[a_-,\infty)}e^{-a\cdot x}\mathrm{d}\mu
 \le\mu\bigl([a_-,\infty)\bigr)e^{-a_-x}$, so the sharp bound of~(ii)
approaches $\epsinf$ geometrically at rate $a_-$. If $\mu$ has no gap, then
restricting the integral to $(0,1/X]$ and using
$e^{-a(x_0+X)}\ge e^{-(x_0+X)/X}$ on that range gives
\[
  \eps(x_0+X)-\epsinf
  \ \ge\ e^{-1-x_0/X}\,\mu\bigl((0,1/X]\bigr),
\]
which is~the gapless bound. If $\mu$ has a density behaving like $a^{\nu-1}$
as $a\downarrow0$, then $\mu\bigl((0,1/X]\bigr)\asymp X^{-\nu}$ and the
excess decays polynomially rather than geometrically.
\end{proofof}

\begin{proposition}[Cost of a misquoted floor]
\label{prop:misquote}
Quote $\tilde\epsinf=\phi\,\epsinf$ and apply the accuracy-only rule on the
law. For $\phi>\tfrac12$ the selected point is $\eps=2\phi\,\epsinf$ and its
regret is exactly $R(\phi)=\ln\cosh\bigl(\tfrac12\ln(2\phi-1)\bigr)$, with
$R(1)=0$ and $R(\phi)\to\infty$ as $\phi\downarrow\tfrac12$. In the variable
$\psi=2\phi-1$---the ratio of reducible to irreducible error at the point the
quoted floor selects---the cost is exactly even,
$R=\ln\cosh(\tfrac12\ln\psi)=R(1/\psi)$, which is the symmetry of $\AM$ and
$\GM$. For $\phi\le\tfrac12$ the quoted floor asserts reducible error that
does not exist, no interior point is selected, and the loss is bounded by
nothing in the frontier.
\end{proposition}

\begin{proofof}{Proposition~\ref{prop:misquote}}
With the quoted floor $\tilde\epsinf=\phi\,\epsinf$ the accuracy-only rule
maximizes $\tilde J(\eps)=\tfrac12\ln(\eps-\phi\epsinf)-\ln\eps$ over the
candidates. In $v=\ln\eps$,
$\mathrm{d}\tilde J/\mathrm{d}v=\tfrac12\eps/(\eps-\phi\epsinf)-1$, which is
positive for $\eps<2\phi\epsinf$ and negative above, so $\tilde J$ is
unimodal with maximizer $\eps=2\phi\epsinf$. For $\phi>\tfrac12$ this point
lies above the true floor and is selected on the law; its reducible-to-
irreducible ratio is $\psi=(2\phi\epsinf-\epsinf)/\epsinf=2\phi-1$, and
by~\eqref{eq:lncosh} its regret is $R=\ln\cosh(\tfrac12\ln\psi)
=\ln\cosh\bigl(\tfrac12\ln(2\phi-1)\bigr)$. The limits follow from
$\ln\cosh y=|y|-\ln2+O(e^{-2|y|})$: $R(1)=0$; as $\phi\downarrow\tfrac12$,
$\psi\downarrow0$ and $R\to\infty$; as $\phi\to\infty$,
$R=\tfrac12\ln(2\phi-1)-\ln2+O(1/\phi)=\tfrac12\ln(2\phi)-\ln2+O(1/\phi)$.
The evenness $R(\psi)=R(1/\psi)$ is the evenness of $\ln\cosh$, that is, the
symmetry of $\AM$ and $\GM$ in their two arguments in
Theorem~\ref{thm:master}. For $\phi\le\tfrac12$ the maximizer
$2\phi\epsinf\le\epsinf$ lies at or below the true floor, so on every
attainable candidate $\tilde J$ is decreasing in $\eps$ and the rule returns
the most accurate candidate; the quoted floor asserts reducible error
$\eps-\phi\epsinf\ge(1-\phi)\epsinf>0$ at every point, including at the
floor itself, and the regret of the selected point,
$-\ln\sin\varphi$ at its arc position, is bounded by nothing in the frontier
as that candidate approaches the floor. Script~\texttt{50} checks $R(\phi)$
against a brute-force maximization of $\tilde J$ at $50$ digits.
\end{proofof}

\paragraph{The floor from a pair of models.} The following reads the spectral
law of Proposition~\ref{prop:spectral} at the level of evaluation items, with
without assuming a two-class structure. Normalize resources so that
$\eps(0)=1$. The spectrum is then a probability measure: each item is associated with a rate $a$ drawn from this
measure, and a model at log-resource $x\ge0$ fails on that item with
probability $e^{-a\cdot x}$. The two models fail independently conditional
on the item. Split the measure as $\epsinf\delta_0+(1-\epsinf)\nu$ with $\nu$
supported off the origin, and write
\[
  h(x)=\int e^{-a\cdot x}\,\mathrm{d}\nu(a),\qquad h_i=h(x_i),\qquad
  h_{12}=h(x_1+x_2).
\]
Conditioning on the item and using independence, $\Pr[F_1F_2\mid a]
=e^{-a\cdot x_1}e^{-a\cdot x_2}=e^{-a\cdot(x_1+x_2)}$, so
\[
  \eps_i=\epsinf+(1-\epsinf)h_i,\qquad
  \eps_{12}=\epsinf+(1-\epsinf)h_{12}:
\]
the joint failure rate of the reducible part is its failure rate at the
\emph{product} resource $t_1t_2$. Substituting,
\[
  \operatorname{Cov}(F_1,F_2)=(1-\epsinf)\bigl[\epsinf(1-h_1)(1-h_2)+D\bigr],
\]
\[
  \Pr[\text{both correct}]=(1-\epsinf)\bigl[(1-h_1)(1-h_2)+D\bigr],
\]
with $D=h_{12}-h_1h_2$, whence the estimator and its bias
\begin{equation}
  \hat\epsinf=\frac{\epsinf(1-h_1)(1-h_2)+D}{(1-h_1)(1-h_2)+D},
  \qquad
  \hat\epsinf-\epsinf=\frac{(1-\epsinf)D}{1-h_1-h_2+h_{12}} ,
  \label{eq:supcovbias}
\end{equation}
which is the bias reported in the paper after~\eqref{eq:covfloor}. This identity has five consequences.

\emph{(i) The bias is upward along one resource direction.} Suppose
$x_1=c_1v$ and $x_2=c_2v$ with $c_1,c_2>0$ and $v$ a fixed direction, which
covers every single-resource frontier and corresponds to the setting
described in the remark. Writing $\xi=a\cdot v\ge0$ for the pushforward variable, the maps
$\xi\mapsto e^{-c_i\xi}$ are both strictly decreasing, hence positively
correlated under the pushforward of $\nu$ by Chebyshev's integral
inequality, so $D\ge0$; the denominator $(1-h_1)(1-h_2)+D$ is positive. The
estimator is therefore an upper bound on the floor of the family and window
at hand, in the same sense as every other floor estimate in this work, and this upper-bound property follows from the theorem rather than from an
imposed assumption.

Collinearity cannot be dropped. Chebyshev's inequality orders a scalar, and
with $n\ge2$ the two maps $a\mapsto e^{-a\cdot x_1}$ and $a\mapsto
e^{-a\cdot x_2}$ are only coordinatewise decreasing, which does not make
them positively correlated under an arbitrary $\nu$; it suffices, but is not
necessary, that the measure be associated. For
$\nu=\tfrac12\delta_{(2,\,0.2)}+\tfrac12\delta_{(0.2,\,2)}$,
$x_1=(1.5,0.1)$ and $x_2=(0.1,1.5)$ one finds $h_1=h_2=0.327666$,
$h_{12}=0.029599$ and $D=-0.077766$, so with $\epsinf=0.1$ the estimator
returns $-0.087$: a spectrum whose modes are anti-aligned with two
non-parallel measurement directions makes the two models' failures
\emph{negatively} associated, and the floor is then underestimated, possibly
below zero. A negative value of \eqref{eq:covfloor} is thus not a numerical
artifact but a diagnostic that the two models are not comparable along a
single resource direction.

\emph{(ii) Equality characterizes one reducible channel.} In the collinear
case Chebyshev's inequality is an equality exactly when one of the two
functions is almost surely constant, which for $c_1,c_2>0$ means the
pushforward of $\nu$ is a point mass---a single atom when $n=1$, and in
general $\nu$ carried by one level set of $a\cdot v$. So $D=0$ if and only
if the reducible part acts as a single scale-free mode along the direction
compared---equivalently, $h$ solves the multiplicative Cauchy equation
$h(x_1+x_2)=h(x_1)h(x_2)$ of Proposition~\ref{prop:law}. The two-class
interpretation is not an additional hypothesis; it is the equality case of the
general statement. The restriction $c_1,c_2>0$ is
needed: at $x_1=0$ every item is failed by the first model and $D=0$ for
every $\nu$.

\emph{(iii) The bias is predicted, and bounded without a fit.} $D=h_{12}-h_1h_2$
is computable from the fitted law, $(1-\epsinf)h_{12}$ being the reducible
error at the product resource $t_1t_2$. Without any fit, $D\le h_{12}\le\min(h_1,h_2)$,
because $h$ is non-increasing and $x_1+x_2\ge\max(x_1,x_2)$ componentwise;
the latter uses $x_i\ge0$, which the item-level normalization supplies, since
placing the origin where every item is failed is what makes
$e^{-a\cdot x}$ a probability. Hence
\[
  \hat\epsinf-\epsinf\;\le\;\frac{(1-\epsinf)\min(h_1,h_2)}{(1-h_1)(1-h_2)+D}
  \;=\;\frac{\min_i(\eps_i-\epsinf)}{(1-h_1)(1-h_2)+D} .
\]
The factor $\bigl[(1-h_1)(1-h_2)+D\bigr]^{-1}$ exceeds one and tends to one
as the models saturate, so the bound is never tighter than one model's
reducible error, and at the knee, where $\eps-\epsinf\approx\epsinf$, it
permits a relative error of order one. It is a guarantee for saturated pairs
and not for pairs inside the canonical band.

The realized bias is much smaller than that bound. For the maximum-entropy
reducible spectrum, the Gamma family $h(x)=(1+x/\beta)^{-k}$ of
Subsection~\ref{sup:pf-width},
\[
  D=h_1h_2\Bigl[\Bigl(\frac{(1+x_1/\beta)(1+x_2/\beta)}
                            {1+(x_1+x_2)/\beta}\Bigr)^{k}-1\Bigr]>0 ,
\]
and along the diagonal $x_1=x_2=x\to\infty$, $D\sim(2x/\beta)^{-k}=2^{-k}h(x)$:
the bias is a factor $2^{-k}$ below a single model's reducible error, so it
decays at the reducible error's own rate with a constant that improves as
the spectrum steepens. This quantifies the benefit of selecting a widely separated, well-saturated
pair.

\emph{(iv) Equivariance needs no assumption on the spectrum.} Appending a
fraction $u$ of always-failed items replaces each observed rate $\eps$ by
$(\eps+u)/(1+u)$; substituting into the estimator and simplifying gives
$\hat\epsinf\mapsto(\hat\epsinf+u)/(1+u)$, which is the map the true atom
undergoes. This is an algebraic property of the statistic, so it holds for
every spectrum and not only in the equality case of~(ii).

\emph{(v) Two qualifications apply to the chance correction.} The statistic is not
homogeneous: rescaling all three rates by $1/(1-q_0)$ does not rescale
$\hat\epsinf$ by $1/(1-q_0)$, so \eqref{eq:covfloor} must be applied to raw failure indicators and
chance-corrected afterward. The statistic identifies the mass of items that
both models actually fail, which is $(1-q_0)\epsinf$ if chance
success is a property of the item---shared by the two models---and smaller if
the models guess independently, in which case the joint rate carries
$(1-q_0)^2$ and \eqref{eq:covfloor} underestimates rather than bounds. Which
case applies is a property of the benchmark's answer format, not of the
spectrum.

Script~\texttt{50} verifies \eqref{eq:supcovbias}, the sign of $D$, the
equality case, the Gamma closed form, the fit-free bound and the
equivariance at $50$-digit precision; the empirical corroboration is in
\TR{The floor from one pair of models}.

\subsubsection*{The floor is the shape index read at a window}
\label{sup:nofloor}

The corresponding figure is Figure~\ref{fig:nofloor} of the paper.

\begin{proofof}{Proposition~\ref{prop:nofloor}}
Write $u=x-x_0$ and $S=(\alpha u)^{\beta}$, so that
\[
s=S'=\frac{\beta S}{u},
\qquad
\kappa=-S''=\frac{\beta(1-\beta)S}{u^{2}}=\frac{s(1-\beta)}{u} .
\]
Eliminating $u$ between the last two displays,
$\kappa/s^{2}=(1-\beta)/(\beta S)$, which rearranges to
$\beta=s^{2}/(s^{2}+\kappa S)$; since $\zeta=\ln(\kappa/s^{2})$ by
\eqref{eq:zeta}, this is $\beta=1/(1+Se^{\zeta})$. The elimination removes $u$, $x_0$, and $\alpha$, establishing the
claimed independence from the resource origin.
The \emph{local} floor share of this chart---the share the two-atom chart
tangent to it would carry---is $\hat p_0=1/(1+\Neff)=\kappa/(s^{2}+\kappa)$ by
definition; substituting $\kappa/s^{2}=(1-\beta)/(\beta S)$ gives
$\hat p_0=(1-\beta)/\{(1-\beta)+\beta S\}$, which is \eqref{eq:nofloor}. It is
$\hat p_0$ and not the true share that appears there, and that is the content of
the proposition: this chart has $\epsinf=0$, so its true share
$p_0=\epsinf/\eps$ is identically zero, while $\hat p_0$ is the positive number
a floored fit returns. (The reading of
Proposition~\ref{prop:floorfree}(iii) as a floor is available only with two
atoms, of which this chart has none.)

(i) At $\beta=1$, $\kappa\equiv0$ and $\hat p_0\equiv0$; as $\beta\downarrow0$ at
fixed $S$, $\hat p_0\uparrow1$. (ii) Immediate from the displayed form of $\hat p_0$.
(iii) $\hat p_0=\tfrac12$ iff $(1-\beta)=\beta S$, i.e., $S^{\star}=(1-\beta)/\beta$,
and $\eps^{\star}=e^{-S^{\star}}=e^{1-1/\beta}$. (iv) Along the ray,
$\sqrt{\kappa}=\sqrt{\beta(1-\beta)S}/u$ and $\mathrm{d}u/u=\mathrm{d}S/(\beta S)$,
so
\[
L=\int\sqrt\kappa\,\mathrm{d}x
 =\sqrt{\frac{1-\beta}{\beta}}\int_{S_{\min}}^{S_{\max}}\frac{\mathrm{d}S}{\sqrt S}
 =2\sqrt{\frac{1-\beta}{\beta}}\Bigl(\sqrt{S_{\max}}-\sqrt{S_{\min}}\Bigr),
\]
which inverts to \eqref{eq:arcbeta}. (v) $\kappa=s(1-\beta)/u$ has the sign of
$1-\beta$ for $u>0$, and $\kappa<0$ is $S''>0$, an increasing hazard; by (iv)
the arc is then not real, so a window on which the measured chord arc vanishes
and a window whose plug-in $\hat\beta$ leaves $(0,1]$---in either direction, by
clause~(v) of the paper's Proposition~\ref{prop:nofloor}---are the same event.
On the chart itself $\hat\beta=\beta$ exactly, since there
$s^{2}+\kappa S=\beta S^{2}/u^{2}>0$, so the exit is upward; off the chart the
denominator can change sign and the plug-in exits downward instead.
\end{proofof}

Three observations clarify the interpretation. First, the identity applies
to one member of the class and does not characterize every completely
monotone frontier: what it establishes is
that the two \emph{charts} considered here---an atom at the origin plus one
channel, and no atom with a stable index---are two coordinate descriptions
of a single one-parameter family over any finite window, and
Proposition~\ref{prop:identify}(i) already rules out distinguishing them. Second,
the microfoundation is the Hougaard frailty \citep{hougaard1986}, and the
hazard reading of Law~I is \citet{proschan1963} with the floor as a cure
fraction \citep{farewell1982,li2001}: a positive stable frailty with index
$\beta$ defines a family that is closed under aggregation of subpopulations. This explains why $\beta$ is associated with
the corpus and $\alpha$ with the resource axis. Third, the two-point exponent
\eqref{eq:alphatwopoint} follows from Proposition~\ref{prop:logistic}: with
$s=\alpha(1-p_0)$ and $m=1-2p_0$ one has $s=\tfrac{\alpha}{2}(1+m)$, so
$\Delta s=\tfrac{\alpha}{2}\Delta m$ identically, for any two points.

The general curvature sum rule in \TR{The width coordinate: per-frontier detail}
follows by applying the fundamental theorem of calculus to $\kappa=-S''$
and using the limits $s(x)\to a_{\min}$ as $x\to\infty$ and
$s(x)\to a_{\max}$ as $x\to-\infty$. These limits follow from the
concentration of the tilted measure at the extreme points of
$\operatorname{supp}\mu$. Scripts~\texttt{75} and~\texttt{76}
verify the local calculus, \eqref{eq:nofloor}, \eqref{eq:kneedepth},
\eqref{eq:arcbeta} and \eqref{eq:alphatwopoint} to machine
precision, and script~\texttt{77} evaluates them on the archive.

\subsection{The gauge of a resource axis, and the invariant jet}
\label{sup:pf-gauge}

\begin{proposition}[The error gauge]
\label{prop:gaugeerr}
Consider the error-power map $\eps\mapsto\eps^{\,t}$, $t>0$.
\begin{enumerate}[label=(\roman*),leftmargin=2em,itemsep=1pt]
\item \emph{It acts on the class precisely when the hypothesis holds}:
      $\eps^{\,t}$ is completely monotone for every $t>0$ iff $S$ is Bernstein
      iff $\mu$ is infinitely divisible \citep{steutel2004}---the composition
      law of Proposition~\ref{prop:compose}.
\item \emph{The axiom is $t=1$ and the composition law is $t\to0$}: since
      $\eps^{\,t}=e^{-tS}=1-tS+O(t^2)$, complete monotonicity of $\eps^{\,t}$
      becomes $c_k\ge0$ for every $k$ as $t\to0$. The one-parameter family
      interpolates between the two hypotheses and is equivalent to the stronger
      one.
\item \emph{Bookkeeping.} Under the map $c_k\mapsto t\,c_k$ for every $k\ge0$,
      so $S$, $\Delta S$, $s$, $\kappa$, $N_{\mathrm{eff}}$ and
      $\Lambda=-\ln\epsinf$ are \textbf{extensive}; every $D_k$, hence $v$,
      $\sigma$, $\beta$ and every horizon $h_k$, is \textbf{invariant}; and the
      arc \emph{length} $L\mapsto\sqrt t\,L$, so $L^2/S$ is invariant. The
      \emph{angle} $\varphi=2\arctan\sqrt{\Neff}$ is not homogeneous---it maps
      to $2\arctan\sqrt{t\Neff}$, which is bounded by $\pi$---so it is $L$ and
      not $\varphi$ that carries the grain.
\item \emph{The jet splits.} Under the two-dimensional group (axis scaling
      $x\mapsto\lambda x$, error power $\eps\mapsto\eps^{\,t}$) the recursion
      $c_{k+2}=(1+D_k)c_{k+1}^2/c_k$ recovers the whole sequence from
      $(c_0,c_1)$, so the local jet is
      \[
      \bigl(\underbrace{c_0,\ c_1}_{\text{two gauge coordinates}}\ ;\
      \underbrace{D_0,D_1,D_2,\dots}_{\text{a complete invariant}}\bigr),
      \qquad c_0=\Delta S,\quad c_1=s .
      \]
\end{enumerate}
\end{proposition}

\begin{proposition}[The gauge of a resource axis]
\label{prop:gauge}
Under $x\mapsto\lambda x+c$ with $\lambda>0$ one has $S\mapsto S$,
$s\mapsto s/\lambda$, $\kappa\mapsto\kappa/\lambda^{2}$ and
$\kappa'\mapsto\kappa'/\lambda^{3}$. Hence
\begin{equation}
w=\frac{\kappa}{s^{2}},
\qquad
w_2=\frac{\mathrm{d}\ln\kappa}{\mathrm{d}S}=\frac{\kappa'}{s\kappa},
\qquad
\frac{\mathrm{d}}{\mathrm{d}S}=\frac1s\frac{\mathrm{d}}{\mathrm{d}x},
\label{eq:invariants}
\end{equation}
together with $S$ itself, are invariant, and the invariant local data of a
frontier is the jet $(S;w,w_2,\dots)$. Consequently:
(i) $w=\operatorname{Var}(a\mid X>x)/\mathbb{E}[a\mid X>x]^{2}$ is the squared
coefficient of variation of the rate posterior, and $\zeta=\ln w$;
(ii) every statement of Laws~I--III is a relation among invariants---%
$p_0=w/(1+w)$, $R=\ln\bigl[(1+w)/2\sqrt w\bigr]$, the canonical point $w=1$,
$L=\int\sqrt w\,\mathrm{d}S$, $\beta=1/(1+Sw)$;
(iii) $\alpha$ is the \emph{unit} of the axis and is not invariant, so it
cannot appear in any invariant statement and no derivation of it from
\eqref{eq:axiom} can exist;
(iv) of the two sum rules, $\int\sqrt\kappa\,\mathrm{d}x=\int\sqrt w\,\mathrm{d}S$
is invariant and $\int\kappa\,\mathrm{d}x=a_{\max}-a_{\min}$ is the unit
itself---which is why $\pi$ is universal and the spectral width is not.
\end{proposition}

\begin{proofof}{Proposition~\ref{prop:gauge}}
Write $\tilde x=\lambda x+c$ and let $\tilde\eps(\tilde x)=\eps(x)$ be the
same frontier in the new coordinate. Since $\eps$ is unchanged as a function
of the physical resource, $S$ is unchanged, and the chain rule gives
$\tilde s=s/\lambda$, $\tilde\kappa=\kappa/\lambda^{2}$ and
$\tilde\kappa'=\kappa'/\lambda^{3}$; the ratios in \eqref{eq:invariants} are
therefore invariant, as is $\mathrm{d}/\mathrm{d}S=s^{-1}\mathrm{d}/\mathrm{d}x$.
Translation changes nothing at all. For (i), Proposition~\ref{prop:selection}
identifies $s=\mathbb{E}_{p_x}[a]$ and $\kappa=\operatorname{Var}_{p_x}(a)$,
so $w$ is the squared coefficient of variation of the rate posterior and
$\zeta=\ln w$ by \eqref{eq:zeta}. Item (ii) is Proposition~\ref{prop:floorfree}(iii),
Proposition~\ref{prop:arcforms} and \eqref{eq:nofloor} rewritten; for
$L=\int\sqrt\kappa\,\mathrm{d}x$ substitute $\mathrm{d}x=\mathrm{d}S/s$.
For (iii): under the reparameterization an atom at $a$ moves to $a/\lambda$,
so the spectral gap $\alpha$ is not invariant, and a statement expressible in
the invariants takes the same value on two frontiers that differ only by a
choice of unit, whereas $\alpha$ does not; hence no function of the invariants
equals $\alpha$. Item (iv) is the same computation applied to the two sum
rules: $\int\kappa\,\mathrm{d}x\mapsto\lambda^{-1}\int\kappa\,\mathrm{d}x$
while $\int\sqrt\kappa\,\mathrm{d}x$ is unchanged, which is the $q=\tfrac12$
case of Proposition~\ref{prop:moments} read as a statement about units.
\end{proofof}

\begin{proofof}{Proposition~\ref{prop:chartode}}
For the two-atom chart, $u=1-p_0$ gives $s=\alpha u$, $\kappa=\alpha^2u(1-u)$
and $w=\kappa/s^2=(1-u)/u$, while $\mathrm{d}p_0/\mathrm{d}x=sp_0$, hence
$\mathrm{d}w/\mathrm{d}S=w(1+w)$ and, integrating, $w/(1+w)=p_0=\epsinf/\eps$.
For the shape law, $S=(\alpha u)^\beta$ gives $s=\beta S/u$ and
$\kappa=\beta(1-\beta)S/u^{2}$, so $w=(1-\beta)/(\beta S)$ and
$\mathrm{d}w/\mathrm{d}S=-w/S$. Each solution family is one-dimensional, and
the affine group of the axis contributes the other two parameters, which gives
the parameter count. The jet statement is Taylor's theorem in the invariant
coordinate: a $k$-parameter family whose map to $(S,s,\kappa,\dots)$ is an
immersion matches the first $k$ entries exactly and is otherwise unconstrained,
so two such families agree to order $k$.
\end{proofof}

\begin{proofof}{Theorem~\ref{thm:monofloor}}
Differentiating the tilted measure, $\mathrm{d}\mathbb{E}_{p_x}[a^k]/\mathrm{d}x
=-\mathbb{E}[a^{k+1}]+\mathbb{E}[a^k]\mathbb{E}[a]$, whence $s'=-\kappa$ and
$\kappa'=-\mu_3$ with $\mu_3=\mathbb{E}_{p_x}[(a-s)^3]$. Fix the mean $s$ and
the variance $\kappa$ and minimize $\mu_3$ over probability measures on
$[0,\infty)$. The extreme points of that moment set are supported on at most
two points, and a two-point measure with mean $s$ and variance $\kappa$ has
atoms $s+A$ and $s-B$ with $AB=\kappa$ and $\mu_3=(A-B)\kappa$; the support
constraint is $B\le s$, and $(\kappa/B-B)\kappa$ decreases in $B$, so the
minimum is at $B=s$, i.e., at the lower principal representation with an atom
at the origin, where $\mu_3=\kappa(\kappa-s^{2})/s$. Therefore
$\mu_3\ge\kappa(\kappa-s^2)/s$, and dividing by $-s\kappa<0$ gives
$w_2=\kappa'/(s\kappa)\le1-w$, with equality exactly in the minimizing case.
That case is two atoms one of which is at $a=0$, i.e., a floor plus one
reducible channel, which by Proposition~\ref{prop:chartode} is the equality
$\mathrm{d}w/\mathrm{d}S=w(1+w)$. For the third form, write
$\hat p_0:=w/(1+w)=1/(1+\Neff)$ for the \emph{local} floor share of
Proposition~\ref{prop:neff}(i)---not the true share $\epsinf/\eps$, which the
two coincide with only on the two-atom chart---so that $F=\hat p_0\,\eps$ and
\[
\frac{\mathrm{d}\ln F}{\mathrm{d}S}
=w_2-1-\frac{w}{1+w}\,(w_2-2)
=(1-\hat p_0)\bigl[w_2-(1-w)\bigr]\le0 .
\]
Since $\mathrm{d}S/\mathrm{d}x=s>0$, $F$ is
non-increasing in the resource. For (i), $F\ge\epsinf$ because $F$ decreases
to $\hat p_0\eps\to\epsinf$ as $\hat p_0\to1$ when a floor is present, and to a limit
$\ge0=\epsinf$ when it is not; the ratio $F/\eps(x_0+X)$ is $\hat p_0$ by
definition. With the \emph{true} share one would have $F=\epsinf$ identically,
which is exactly the equality case and the reason the bound is one-sided. (ii) is the equality case. (iii) is the inequality at $w=1$. (iv)
is the one-sidedness. Script~\texttt{80} verifies the identities to machine
precision and stresses the inequality on $2.3\cdot10^{5}$ random
(spectrum, resource) points with no violation, the largest apparent excess under double precision is attributable to
rounding; recomputation at $60$-digit precision gives
$-2.7\cdot10^{-48}$ at the equality case;
script~\texttt{81} measures the bound on the archive and script~\texttt{82}
calibrates its one-sided test.
\end{proofof}

\subsection{Bernstein clocks, the ladder, effective channels, and the knee}
\label{sup:pf-clock}

\begin{proposition}[Bernstein clocks]
\label{prop:clock}
Let $\chi$ be a \emph{Bernstein function}: $\chi\ge0$ with $\chi'$
completely monotone, equivalently the Laplace exponent of a subordinator
$(\sigma_t)$. Then
\begin{enumerate}[label=(\roman*),leftmargin=2em,itemsep=1pt]
\item \textup{(Closure.)} If $\eps$ satisfies \eqref{eq:axiom} so does
      $\eps\circ\chi$, and the new spectrum is the law of $\sigma_A$ with
      $A\sim\mu$: \emph{a change of clock is a subordination of the difficulty
      spectrum}. The affine maps of Proposition~\ref{prop:gauge} are its
      linear part; $x^{\beta}$ $(0<\beta\le1)$, $\ln(1+x)$ and $x/(1+x)$ are
      not.
\item \textup{(One way only.)} Bernstein functions compose but do not invert,
      so the gauge is a \emph{semigroup}. $\chi(x)=e^{x}$ is not Bernstein
      and $e^{-be^{x}}$ is not completely monotone---its second derivative is
      negative wherever $be^{x}<1$. The axiom is therefore a joint property of
      a frontier \emph{and} its clock: it survives every Bernstein coarsening
      of an axis and need not survive a refinement.
\item \textup{(The index belongs to the clock.)}
      $\exp\{-[\alpha(x-x_0)]^{\beta}\}$ is the pure exponential $e^{-u}$ in
      the clock $u=[\alpha(x-x_0)]^{\beta}$, and the four-parameter law is the
      two-atom spectrum in the same clock. Hence $\beta$ is the \emph{index of
      the clock}, not a property of the spectrum, and $x^{\beta}$ is the
      exponent of the $\beta$-stable subordinator, which smears a positive atom
      into a measure with no mass at the origin but a left tail
      $\exp(-ca^{-\beta/(1-\beta)})$. ``Index of the clock'' and ``index of the
      spectrum'' are one number, not two: by \eqref{eq:dict} $\beta$ is the
      index at the origin of the clock's L\'evy measure, and by
      Proposition~\ref{prop:compose}(b) that is the index of the difficulty
      spectrum the clock subordinates. What the statement denies is that
      $\beta$ is a \emph{scale}: it is a property of the subdivision, which is
      why it is weakly identified and corpus-specific. The two charts are
      \emph{not} thereby one
      orbit: Proposition~\ref{prop:compose} identifies the orbit of $e^{-u}$
      exactly, and the two-atom chart is outside it.
\item \textup{(Composition.)} Clocks compose as $\beta\mapsto\beta_1\beta_2$
      with $(\alpha_2(\alpha_1x)^{\beta_1})^{\beta_2}
      =\alpha_2^{\beta_2}\alpha_1^{\beta_1\beta_2}x^{\beta_1\beta_2}$, so
      $\ln\beta$ is additive over stages and the additive intercept is
      $\beta\ln\alpha$, not $\ln\alpha$.
\end{enumerate}
\end{proposition}

\begin{proofof}{Proposition~\ref{prop:clock}}
(i) is Bochner's subordination theorem in the form of
\citet{schilling2012}, Theorem~3.7: if $f$ is completely monotone and
$\chi$ is Bernstein then $f\circ\chi$ is completely monotone. The
probabilistic reading is the computation
$e^{-a\chi(x)}=\mathbb{E}[e^{-x\sigma_a}]$ for the subordinator with
Laplace exponent $\chi$, so that
$\eps(\chi(x))=\int\mathbb{E}[e^{-x\sigma_a}]\,\mathrm{d}\mu(a)
=\mathbb{E}[e^{-x\sigma_A}]$ with $A\sim\mu$: the new spectral measure is the
law of $\sigma_A$. The listed examples are Bernstein because
$\lambda$, $\beta t^{\beta-1}$, $(1+t)^{-1}$ and $(1+t)^{-2}$ are completely
monotone.

(ii) Bernstein functions are closed under composition, since
$(\chi_1\circ\chi_2)'=(\chi_1'\circ\chi_2)\chi_2'$ is a product
of completely monotone functions. They are not closed under inversion: the
inverse of $t^{\beta}$ is $t^{1/\beta}$ with $1/\beta>1$, whose derivative is
increasing. For the counterexample, $g(x)=e^{-be^{x}}$ has
$g''=be^{x}g\,(be^{x}-1)$, which is negative exactly where $be^{x}<1$, so $g$
is not completely monotone and $\chi(x)=e^{x}$---which is the map taking an
attempt count to its logarithm, read backward---is not an admissible clock.
Hence the axiom is preserved by coarsening and not by refinement. In
particular, if $\eps$ is completely monotone in a count $m$
(Proposition~\ref{prop:countaxis}) and $m=m(x)$ is Bernstein, then $\eps$ is
completely monotone in $x$ by (i).

(iii) $\exp\{-[\alpha(x-x_0)]^{\beta}\}=e^{-u}$ with $u=[\alpha(x-x_0)]^\beta$,
and $t\mapsto t^{\beta}$ is the Laplace exponent of the $\beta$-stable
subordinator; the four-parameter law is $\epsinf+ce^{-u}$ in the same clock.
By (i) the new spectrum is the law of $\sigma_A$; for $A$ a positive atom this
is the $\beta$-stable law scaled by that atom, which has a density on
$(0,\infty)$, no mass at the origin, and the left tail
$\exp\{-c\,a^{-\beta/(1-\beta)}\}$ of \citet{pollard1946}. An atom at $a=0$ is
fixed by every subordinator, so a floor survives any change of clock while a
soft floor is created by one.

(iv) is the computation
$(\alpha_2(\alpha_1x)^{\beta_1})^{\beta_2}
=\alpha_2^{\beta_2}\alpha_1^{\beta_1\beta_2}x^{\beta_1\beta_2}$, so
$\beta_{\mathrm{eff}}=\beta_1\beta_2$ and
$\beta_{\mathrm{eff}}\ln\alpha_{\mathrm{eff}}
=\beta_2\ln\alpha_2+\beta_1\beta_2\ln\alpha_1$. Script~\texttt{87} verifies
(i)--(iv).
\end{proofof}

\begin{remark}[The one-line proof of Theorem~\ref{thm:monofloor}]
\label{rem:cs}
The preceding proof is self-contained. A shorter derivation follows from
$S^{(k)}=(-1)^{k+1}c_k$ with $c_k$ the cumulants of $p_x$, the inequality
$w_2\le1-w$ is, after multiplying by $s^{2}\kappa^{2}>0$, exactly
\[
\mathbb{E}[a]\,\mathbb{E}[a^{3}]\ \ge\ \mathbb{E}[a^{2}]^{2},
\]
which is Cauchy--Schwarz for the inner product
$\langle f,g\rangle_a=\mathbb{E}[a\,fg]$ applied to $f\equiv1$ and $g(a)=a$:
$\langle1,a\rangle_a^{2}\le\langle1,1\rangle_a\langle a,a\rangle_a$. Equality
holds iff $1$ and $a$ are proportional in $L^2(a\,\mathrm{d}p_x)$, i.e., iff $a$
is constant $p_x$-almost surely on $\{a>0\}$: the spectrum is
$\{0,\alpha\}$. Script~\texttt{88} verifies the identity
$\mathbb{E}[a]\mathbb{E}[a^{3}]-\mathbb{E}[a^{2}]^{2}
=s^{2}\kappa^{2}[(1-w)-w_2]$ to $1.8\cdot10^{-49}$.
\end{remark}

\begin{proposition}[The ladder, and the mirror wall]
\label{prop:ladder}
Write $\rho=(a_{\max}-s)/s$, invariant. For every frontier satisfying
\eqref{eq:axiom},
\begin{equation}
\underbrace{\kappa\ge0}_{\text{rung }1}
\qquad
\underbrace{w_2\le1-w}_{\text{rung }2}
\qquad
\underbrace{-\rho+\frac{w}{\rho}\ \le\ w_2}_{\text{mirror wall}}
\label{eq:ladder}
\end{equation}
and rung $k$---the $k$-th Hankel determinant, which needs $S^{(k+1)}$---is tight
exactly on a $k$-atom spectrum, so \emph{the number of channels is the rank of
the Hankel matrix} built from $(s,\kappa,S''',\dots)$: the local counterpart of
Corollary~\ref{cor:prony}, and chart selection is a statement about one higher
jet and nothing else. The mirror wall comes from the \emph{upper} principal
representation, so the admissible region for $w_2$ is a \emph{strip} whose width
$1-w+\rho-w/\rho$ measures distance from the two-atom chart, and it gives
$\rho\ge w$, i.e., $a_{\max}\ge s+\kappa/s$: the fastest mode is bounded below
by the local jet alone, whence
$\int\kappa\,\mathrm{d}x\ge\sup_x[\,s+\kappa/s\,]-a_{\min}$.
\end{proposition}

\begin{proofof}{Proposition~\ref{prop:ladder}}
By \eqref{eq:spectral} the axiom says that the moments
$m_k=\mathbb{E}_{p_x}[a^k]$ are those of a positive measure on $[0,\infty)$,
i.e., that $(m_k)$ is a Stieltjes moment sequence; by the Hamburger--Stieltjes
theorem \citep{shohat1943} this is equivalent to non-negativity of every Hankel
determinant of $(m_k)$ and of its shift, and a sequence whose Hankel matrix has rank $k$ is
the moment sequence of a $k$-atom measure, which gives the rank statement and
the tightness pattern. Rung~1 is $m_0m_2\ge m_1^2$, i.e., $\kappa\ge0$. Rung~2
is $m_1m_3\ge m_2^2$, which is Remark~\ref{rem:cs}. Rung~3 is the central
$3\times3$ determinant $\mu_2\mu_4\ge\mu_3^2+\mu_2^3$; substituting
$\mu_2=\kappa$, $\mu_3=S'''$ and $\mu_4=3\kappa^{2}-S''''$ and dividing by
$s^{2}\kappa^{2}$ gives $V\le2w-w_2^{2}$, where $V:=S''''/(s^{2}\kappa)$ is
itself an axis-gauge invariant, with equality iff the measure has at most two
atoms. This is the \emph{axiom's} third rung and is not
Proposition~\ref{prop:tower}'s, which in the same notation reads
$V\le-w_2^{2}$: the two bound $V$ from the same side but by different amounts,
and the composition law's is the stronger.

For the mirror wall, fix the mean and variance and \emph{maximize} $\mu_3$ over
measures on $[0,a_{\max}]$. By the same extreme-point argument the maximizer is
the upper principal representation, two atoms one of which is at $a_{\max}$;
writing $\bar s=a_{\max}-s$ and $\bar w=\kappa/\bar s^{2}$ it has
$\mu_3=\kappa\bar s(1-\bar w)$, so $\kappa'=-\mu_3\ge-\kappa\bar s(1-\bar w)$
and, dividing by $s\kappa$, $w_2\ge-\rho+w/\rho$ with $\rho=\bar s/s$. Both
walls are attained by any two-atom measure, since such a measure is
simultaneously the lower and the upper principal representation of its own
first two moments. Feasibility of the moment problem on $[0,a_{\max}]$ requires
$\kappa\le s(a_{\max}-s)$, which is $\rho\ge w$, i.e.,
$a_{\max}\ge s+\kappa/s$; combining with
$\int\kappa\,\mathrm{d}x=a_{\max}-a_{\min}$
(Proposition~\ref{prop:path}(iv)) gives (iii). Script~\texttt{88} verifies
every rung and wall, tight to $10^{-41}$ on its own family and strictly slack
elsewhere.
\end{proofof}

\begin{proofof}{Proposition~\ref{prop:neff}}
(i) For $\mu$ a $\Gamma(k,\theta)$ density, $\eps=(1+\theta x)^{-k}$, so
$S=k\ln(1+\theta x)$, $s=k\theta/(1+\theta x)$ and
$\kappa=k\theta^{2}/(1+\theta x)^{2}$, whence $w=1/k$ identically; the
remaining identities are Proposition~\ref{prop:gauge}(ii) rewritten, with the
arc element $\mathrm{d}L=\sqrt w\,\mathrm{d}S$ from
Proposition~\ref{prop:path}(i). The angle itself advances by
$\mathrm{d}\varphi=\dot\Neff\,\mathrm{d}L/(\Neff+1)$, which equals
$\mathrm{d}L$ only where $\dot\Neff=\Neff+1$, i.e., on the two-atom chart; the
axiom is precisely the inequality $\mathrm{d}\varphi\le\mathrm{d}L$. The local
share $\hat p_0=1/(1+\Neff)$ dominates the true share $p_0=\epsinf/\eps$ because
$\Neff\le e^{\Delta S}-1$ is the axiom's wall and $p_0=e^{-\Delta S}$ by
definition, with equality exactly on that chart.
(ii) is \eqref{eq:lncosh} with $\psi=(\eps-\epsinf)/\epsinf=(1-p_0)/p_0=1/w$,
together with $\cosh$ being even, so that $R$ is a strictly increasing function
of $|\ln N_{\mathrm{eff}}|$ and its minimizer over any set of operating points
is the point of that set nearest $N_{\mathrm{eff}}=1$.
(iii) is the integral of $w'=-w/S$ in Proposition~\ref{prop:chartode}.
(iv) is Karamata's Tauberian theorem \citep{bingham1987}: $\mu[0,a]\sim Ca^{\theta}$ at the origin
gives $\eps(x)\sim C\Gamma(\theta+1)x^{-\theta}$ as $x\to\infty$, hence
$S\sim\theta\ln x$, $s\sim\theta/x$, $\kappa\sim\theta/x^{2}$ and
$w\to1/\theta$. An atom at the origin is the case $\theta=0$, for which
$p_0\to1$ and $w\to\infty$; an isolated $a_{\min}>0$ gives $p_x\to\delta_{a_{\min}}$,
so $\kappa\to0$ with $s\to a_{\min}>0$ and $w\to0$. Script~\texttt{89}
verifies all four.
\end{proofof}

\begin{proposition}[When a canonical point exists at all]
\label{prop:kneeW}
For a two-atom spectrum with exponent ratio $r=a_2/a_1>1$ and no floor, $w$ is
unimodal in the resource with
\begin{equation}
W:=\sup_x w=\frac{(r-1)^{2}}{4r},
\label{eq:knee}
\end{equation}
attained at mixing share $1/(1+r)$; hence the canonical point $w=1$ exists if
and only if
\begin{equation}
r\ \ge\ 3+2\sqrt2=(1+\sqrt2)^{2}=5.8284\ldots
\label{eq:kneethreshold}
\end{equation}
With a floor ($a_1=0$, $r=\infty$) it always exists, which is why the floor
chart never raised the question. Two channels closer than that factor produce no
knee at all, $\kappa=s^{2}$ is empty, and the correct instruction is the general
rule of Proposition~\ref{prop:neff}(ii). The quantity $(r_\nu-1)^{2}/4r_\nu$
that \eqref{eq:kant} adds to $1$ to form its Kantorovich factor
$(r_\nu+1)^{2}/4r_\nu$ has the same algebraic \emph{form} as $W$, but in the
clock's jump-rate ratio $r_\nu$ rather than a spectrum's exponent ratio: the two
are not one quantity and nothing is transferred between them.
\end{proposition}

\begin{proofof}{Proposition~\ref{prop:knee}}
Parts (i)--(iii) are the two-atom identities of \eqref{eq:spectral} differentiated:
with $p_1=1-p_0$ the posterior reducible share, $\partial S/\partial x=\alpha p_1$
is \eqref{eq:slopecurv} on the two-atom chart, $\zeta=0$ is $p_1=\tfrac12$ by the
definition of the canonical reference and gives $\alpha/2$, and
$\partial^{2}S/\partial x^{2}=-\alpha^{2}p_1(1-p_1)$ is maximal in magnitude at
$p_1=\tfrac12$, which is that same point. Non-invariance of the geometric
curvature is immediate from the transformation of a second derivative under
rescaling of either axis.
\end{proofof}

\begin{proofof}{Proposition~\ref{prop:kneeW}}
Let $u$ be the posterior share of the faster atom $a_2$ and $\Delta=a_2-a_1$.
Then $s=a_1+u\Delta$ and $\kappa=u(1-u)\Delta^{2}$, so
$w(u)=u(1-u)\Delta^{2}/(a_1+u\Delta)^{2}$, which vanishes at $u=0,1$ and is
unimodal in between. Setting $\mathrm{d}w/\mathrm{d}u=0$ gives
$(1-2u)(1+cu)=2cu(1-u)$ with $c=\Delta/a_1=r-1$, i.e., $u=1/(1+r)$, and
substituting back gives $W=(r-1)^{2}/(4r)$. Since $u$ is strictly monotone in
the resource, $\sup_xw=W$. Finally $W\ge1\iff r^{2}-6r+1\ge0\iff
r\ge3+2\sqrt2$, the other root being below $1$. With $a_1=0$ one has
$w=(1-u)/u$, which is unbounded. Script~\texttt{90} verifies
\eqref{eq:knee} and \eqref{eq:kneethreshold} to five decimals and measures the
square-root sensitivity near the threshold.
\end{proofof}

\begin{proofof}{\eqref{eq:arcweights}}
By Proposition~\ref{prop:path}(i), $\sqrt w\,\mathrm{d}S=\sqrt\kappa\,
\mathrm{d}x$, and for two atoms with faster share $u$ this is
$-\mathrm{d}u/\sqrt{u(1-u)}$, whose integral over $u\in(0,1)$ is
$\int_0^1\mathrm{d}v/\sqrt{v(1-v)}=\pi$ for any separation and any weights.
Now let the atoms be ordered $a_1<\dots<a_K$ with separations tending to
infinity. At the $t$-th transition the atom $a_{K-t+1}$ decays while the slower
atoms are, on that scale, indistinguishable from a single atom carrying the
remaining mass, so the transition is a two-atom transition whose faster share
runs from its value at the start of the transition down to $0$ and contributes
$\int_0^{u_t}\mathrm{d}v/\sqrt{v(1-v)}=2\arcsin\sqrt{u_t}$. For $t=1$ the
starting value is $u_1=1$, because $e^{-a_Kx}$ dominates as $x\to-\infty$,
giving $\pi$; for $t\ge2$ the previous transition has already been completed,
so the alive atoms are in the proportions they hold \emph{at that transition's
own crossover}, and $u_t=q_{K-t+1}$ of \eqref{eq:crossshare}. Summing gives
\eqref{eq:arcweights}. The shares have to be read there and not off the
weights: $L$ is invariant under $x\mapsto x+\Delta$ while $p_k/P_k$ is not, so
the raw-weight form is not a functional of the frontier at all, and the two
agree only in the gauge that puts the crossovers at the origin---which is the
gauge of the three worked three-atom cases of \TR{The knee: noise sensitivity
near the threshold}, and the reason the archive's check passes. Each term is at
most $\pi$, so $L\le(K-1)\pi=N\pi$, with equality only in the limit
$q_k\to1$, and $L\ge\pi$ with equality iff $K=2$; since every term with
$K\ge3$ is strictly positive, no spectrum with finite support of size $K\ge2$
has a complete sweep shorter than $\pi$, and the only member of the class with
$L<\pi$ is the single atom $K=1$ of~(ii), for which $L\equiv0$.
Script~\texttt{89} checks the formula against direct quadrature on eight
weight vectors with $K=2$ to $5$ and consecutive rates separated by $10^{4}$ to
$10^{6}$, agreeing to a worst absolute discrepancy of $8.7\times10^{-4}$ in
$L/\pi$. Those eight all have $O(1)$ weights and rates separated enough to put
every crossover at the origin, where $q_k=p_k/P_k$, so they test the
separated-rate asymptote and \emph{not} the gauge-dependence of the raw-weight
form; a configuration that separates the two readings has to move the
crossovers, which script~\texttt{160} does, and that script is the verification
of \eqref{eq:crossshare}.
\end{proofof}

\subsection{The two counts}
\label{sup:twocounts}

\begin{proposition}[Attempts are worth less than items]
\label{prop:twocounts}
A Laplace transform is log-convex, so $Q:=-\ln P$ is concave with $Q(0)=0$ for
\emph{every} $\nu$. Hence, writing $S(k)=-\ln P(k)$ for log-error on the count
axis,
\begin{equation}
S(tk)\ \le\ t\,S(k)\qquad\text{for every }t\ge1,
\label{eq:grain}
\end{equation}
with equality if and only if $\nu$ is a single atom. Writing
$D(t)=S(t)/\{t\,S(1)\}$ for the discount and $L_m=\ln\{P(m)/P(m+1)\}$ for the
price of attempt $m+1$,
\begin{equation}
\frac1t\ \le\ D(t)\ \le\ 1,
\qquad
D(2)=\frac12+\frac{L_1}{2S(1)}\quad\text{exactly},
\qquad L_1=-\ln\kappa_{\mathrm{nest}},
\label{eq:discount}
\end{equation}
and $D$ is non-increasing in $t$. Moreover $\eps\mapsto\eps^{t}$ multiplies $S$,
and with it every cumulant, by exactly $t$, whereas $m\mapsto tm$ gives $Q(tm)$
rather than $tQ(m)$; the two coincide iff $\nu$ is a single atom.
\end{proposition}

\begin{proofof}{Proposition~\ref{prop:twocounts}}
$P(k)=\int e^{-bk}\mathrm{d}\nu(b)$ is a Laplace transform, so by
Cauchy--Schwarz $P(\tfrac{k_1+k_2}2)^2\le P(k_1)P(k_2)$: $P$ is log-convex, i.e.,
$Q=-\ln P$ is concave. Explicitly
$(\ln P)''=\operatorname{Var}_{\pi_k}(b)\ge0$ with
$\mathrm{d}\pi_k\propto e^{-bk}\mathrm{d}\nu$, with equality iff $\nu$ is a
single atom; for a two-atom $\nu$ the numerator is the perfect square
$w(1-w)(b_1-b_2)^2e^{-(b_1+b_2)k}$. Since $Q(0)=-\ln P(0)=0$, concavity gives
$Q(\theta k)\ge\theta Q(k)$ for $\theta\in[0,1]$; applying this with
$\theta=1/t$ at the point $tk$ yields $Q(k)\ge Q(tk)/t$, which is
\eqref{eq:grain}, with equality throughout iff $Q$ is linear, iff $P$ is a pure
exponential, iff $\nu$ is a single atom.

For \eqref{eq:discount}: $D(t)=S(t)/\{tS(1)\}\le1$ is \eqref{eq:grain} at $k=1$,
and $D(t)\ge1/t$ because $Q$ is non-decreasing, so $S(t)\ge S(1)$; equality on
the left means $Q$ is constant beyond $k=1$, i.e., no attempt after the first
reduces the error at all, which is the perfectly nested case. Monotonicity in
$t$ is concavity of $Q$ again: $Q(t)/t$ is non-increasing. Writing
$S(2)=S(1)+L_1$ with $L_1=\ln\{P(1)/P(2)\}=-\ln\kappa_{\mathrm{nest}}$ gives
$D(2)=\{S(1)+L_1\}/\{2S(1)\}=\tfrac12+L_1/\{2S(1)\}$ exactly. Finally
$\eps\mapsto\eps^{t}$ sends $S\mapsto tS$, hence $c_k\mapsto tc_k$ for every
$k\ge0$, while $m\mapsto tm$ sends $S=Q\circ m$ to $Q\circ(tm)$; these agree for
all $t$ iff $Q(tk)=tQ(k)$, iff $Q$ is linear, iff $\nu$ is a single atom.
Verified in \texttt{120\_two\_counts.py}.
\end{proofof}

\noindent\emph{Remark.} This corrects the reading of the attempt chain
(\TR{The chain}) under which the error-power gauge was an attempt rescaling. The gauge statements themselves are unaffected---$S$,
$\Delta S$, $N_{\mathrm{eff}}$ and $\Lambda$ are extensive and every $D_k$ is
invariant under $\eps\mapsto\eps^{t}$ either way---but the external input of
Proposition~\ref{prop:retry} is the grain and not a retry budget, and a retry
budget reaches a given grain only through $D$. The free prediction of
Proposition~\ref{prop:gaugeerr} sharpens accordingly: re-scoring in $k$-item
bundles must multiply $S$ by exactly $k$, whereas $k$ retries must give
$S(k)=kD(k)S(1)<kS(1)$, with $D$ predicted from $P$ and no free parameter.

\subsection{No invariant rule, and the rule's input}
\label{sup:retry}

\begin{proofof}{Theorem~\ref{thm:noinvariant}}
By Proposition~\ref{prop:gaugeerr} the complete local invariants of a
compositional frontier are $(D_0,D_1,\dots)$, and by \eqref{eq:chartline} a
constant-$\sigma$ chart has $D_k=1/(k-\beta)$ independent of $x$. A condition
$F(D_0,D_1,\dots)=0$ is therefore either satisfied everywhere on such a frontier
or nowhere, and selects no point.
\end{proofof}

\begin{proofof}{Proposition~\ref{prop:retry}}
Under $\eps\mapsto\eps^{t}$, which by Proposition~\ref{prop:twocounts} is the
passage to a delivered unit of $t$ independently drawn items, $S\mapsto tS$ hence
$s\mapsto ts$, $\kappa\mapsto t\kappa$ and
$N_{\mathrm{eff}}=s^2/\kappa\mapsto tN_{\mathrm{eff}}$.
The canonical condition in the scored gauge is $tN_{\mathrm{eff}}(x)=1$, i.e.,
\eqref{eq:retry}. A deployment that coarsens its criterion by retrying a single
item instead reaches grain $t_{\mathrm{eff}}=S(t_r)/S(1)=t_rD(t_r)<t_r$ by
\eqref{eq:discount}, so the same operating point requires strictly more than $t$
attempts, and exactly $t$ only on a homogeneous task. Differentiating $\ln N_{\mathrm{eff}}(x^\star)=-\ln t$ and using
$\mathrm{d}\ln N_{\mathrm{eff}}/\mathrm{d}x=\sigma s/N_{\mathrm{eff}}=\sigma/h_1$
gives $\mathrm{d}x^\star/\mathrm{d}\ln t=-h_1/\sigma$; on a floored similarity
chart $\sigma=-N_{\mathrm{eff}}/\Delta S$, so
$-h_1/\sigma=h_1\Delta S/N_{\mathrm{eff}}=\Delta S/s=h_0$. On the cure chart $h_0$
is constant, so the relation integrates to $x^\star(t)-x^\star=h_0\ln t$; off it
$h_0$ varies and the integrated form is $x^\star(t)=x^\star t^{1/|\beta|}$ on the
chart line. Both $h_0$ and $h_1/\sigma$ are gauge-invariant. Thus, the rate is
unchanged by the gauge transformation, although the operating point changes.
\end{proofof}

\subsection{The composition law: the dictionary and the second Cauchy--Schwarz}
\label{sup:pf-compose}

\begin{proofof}{Proposition~\ref{prop:compose}}
(a)$\Leftrightarrow$(b) is the classical characterization of infinite
divisibility on $[0,\infty)$: a probability measure $\mu$ on $[0,\infty)$ is
infinitely divisible iff its Laplace transform is $e^{-S}$ with $S$ a Bernstein
function, $S$ being then the Laplace exponent of the subordinator
$(\sigma_t)$ of which $\mu$ is the time-one law \citep{bertoin1996,sato1999,
schilling2012}. (b)$\Leftrightarrow$(c) because $\eps^t=e^{-tS}$ and $tS$ is
Bernstein exactly when $S$ is, each $e^{-tS}$ being then completely monotone by
Bochner subordination; conversely if $\eps^t$ is completely monotone for every
$t$ then $t^{-1}(1-\eps^{t})\to S$ pointwise and the limit of Bernstein
functions is Bernstein. (b)$\Leftrightarrow$(d) is the definition: $S$ is
Bernstein iff $S\ge0$ and $S'=s$ is completely monotone, and $s(0)<\infty$
makes $s/s(0)$ a frontier in the sense of
Proposition~\ref{prop:spectral}, with spectrum $\nu/|\nu|$ by \eqref{eq:levy}.

That any of them implies \eqref{eq:axiom} is Bochner again, with $f=e^{-u}$.
The converse fails on the two-atom chart: a non-degenerate infinitely divisible
law on $[0,\infty)$ is the time-one law of a subordinator and therefore has
unbounded support \citep[Thm.~24.3]{sato1999}, so no finitely supported
spectrum with more than one atom is infinitely divisible. Since the class in
(b) is by construction the orbit of $e^{-u}$ under the semigroup of
Proposition~\ref{prop:clock}, and since a composition of Bernstein functions is
Bernstein, the class is closed under every change of clock: divisibility is a
gauge-invariant property, and the two charts lie in different orbits.
Finally $\epsinf=e^{-S(\infty)}$, and for a Bernstein $S$ with no drift
$S(\infty)=\nu(\text{total})$, which is \eqref{eq:poissonzero}: a floor is a
\emph{bounded} clock, i.e., a compound Poisson one, and $\epsinf$ is the
probability that the Poisson count of improvements is zero.
\end{proofof}

\begin{proofof}{\eqref{eq:dict}}
Differentiating \eqref{eq:levy} under the integral, $S^{(k)}(x)=
(-1)^{k-1}\int c^k e^{-cx}\nu(\mathrm{d}c)=(-1)^{k-1}m_k(\nu_x)$ for $k\ge2$,
and $S'=b+m_1(\nu_x)$. Theorem~\ref{thm:monofloor} records
$S^{(k)}=(-1)^{k+1}c_k$ for the cumulants of the frailty posterior, and
$(-1)^{k-1}=(-1)^{k+1}$, so $c_k=m_k(\nu_x)$. For the zeroth moment,
$S_\infty-S(x)=\int e^{-cx}\nu(\mathrm{d}c)=m_0(\nu_x)$ when $b=0$. Note the
content of the identification: $\kappa=\operatorname{Var}(a\mid X>x)$, a
\emph{second} central moment of the posterior, equals a \emph{second} raw
moment of the surviving L\'evy measure, and the remaining depth equals its
mass. Script~\texttt{96} checks all four against finite differences of $S$.
\end{proofof}

\begin{proofof}{Theorem~\ref{thm:budget}}
Cauchy--Schwarz in $L^2(\nu_x)$ with $f=1$, $g=c$ gives
$m_1^2=\bigl(\int c\,\nu_x\bigr)^2\le\nu_x(\text{total})\int c^2\nu_x=m_0m_2$,
with equality iff $c$ is $\nu_x$-a.s.\ constant, i.e., iff $\nu_x$ is a single
atom. Dividing by $m_2\,\Delta S$ and using \eqref{eq:dict},
$N_{\mathrm{eff}}=s^2/\kappa=m_1^2/m_2\le m_0=\Delta S$. Exponentiating
$\Delta S=\ln(\eps/\epsinf)\ge N_{\mathrm{eff}}$ gives
$\epsinf\le\eps e^{-N_{\mathrm{eff}}}$.

(i) $e^{N}>1+N$ for $N>0$, so $\eps e^{-N}<\eps/(1+N)$, which is
Theorem~\ref{thm:monofloor}(i) written with the local share
$\hat p_0=1/(1+N_{\mathrm{eff}})$; the new bound is tighter than the old by the
ratio $e^{N}/(1+N)\ge1$, equivalently by the \emph{tightening factor}
$(1+N)e^{-N}\le1$ in which the measurements of
Section~10 of the paper are reported, the two being reciprocals and the
refutation width being $-\ln$ of the second. Both tend to $1$ as $N\to0$: at
the floor the two walls coincide and the gain vanishes. (ii) is
Proposition~\ref{prop:slope}, since $\eps e^{-N_{\mathrm{eff}}}=e^{-G}$ and $G$
is non-decreasing. (iii) At the canonical point $N_{\mathrm{eff}}=1$, so
$\Delta S^\star\ge1$ and $\eps^\star\ge e\epsinf$; the two-atom chart's
$\eps^\star=2\epsinf$ is below that for every floor, the gap being
$\ln(e/2)=0.3069$ nats. (iv) $N_{\mathrm{eff}}\le\Delta S\le S_\infty$
everywhere, so $\sup_xN_{\mathrm{eff}}\ge1$ requires $S_\infty\ge1$.

The pointwise form is worth isolating because it is what makes the theorem
usable on a window: for $y>0$ the shifted clock
$\psi(y)=S(x+y)-S(x)$ is Bernstein with L\'evy measure $\nu_x$ and
$\psi(\infty)=\Delta S(x)$, so the inequality holds at \emph{every} $x$ with the
remaining depth measured from that $x$, not only at the left end of the axis.
\end{proofof}

\begin{proposition}[The channel slope, and the second wall]
\label{prop:slope}
Write $\sigma:=\mathrm{d}N_{\mathrm{eff}}/\mathrm{d}S$, a dimensionless
invariant. Then $\sigma=m_1m_3/m_2^2-2$, so Cauchy--Schwarz on $\nu_x$ gives
$m_2^2\le m_1m_3$ and
\begin{equation}
\boxed{\ \sigma\ \ge\ -1\ }
\qquad\Longleftrightarrow\qquad
\boxed{\ G:=N_{\mathrm{eff}}+S=\frac{s^2}{\kappa}-\ln\eps
\ \text{ is non-decreasing}\ }
\end{equation}
with equality exactly for a single-atom $\nu_x$, while the axiom alone gives
only $\sigma\ge-(1+N_{\mathrm{eff}})$, with equality exactly on the two-atom
chart, where $\sigma\equiv-(1+N_{\mathrm{eff}})$ is an identity rather than a
fit. \emph{A compositional frontier loses at most one effective channel per nat
of log-error.} In raw form the two walls are $\kappa'\le s\kappa-\kappa^2/s$ and
$\kappa'\le-\kappa^2/s$: the same derivative order, and the gap is the single
term $s\kappa$. Letting $G\to S_{\infty}$ recovers \eqref{eq:budget}.
\end{proposition}

\begin{proofof}{Proposition~\ref{prop:slope}}
From \eqref{eq:dict}, $N_{\mathrm{eff}}=m_1^2/m_2$ and
$m_k'=-m_{k+1}$, so
$\mathrm{d}N_{\mathrm{eff}}/\mathrm{d}x=
(-2m_1m_2^2+m_1^2m_3)/m_2^2$ and, dividing by $s=m_1$ (the chain rule for
$\mathrm{d}/\mathrm{d}S=s^{-1}\mathrm{d}/\mathrm{d}x$),
$\sigma=m_1m_3/m_2^2-2$. Cauchy--Schwarz with $f=c^{1/2}$, $g=c^{3/2}$ gives
$m_2^2\le m_1m_3$, hence $\sigma\ge-1$, with equality iff $\nu_x$ is a single
atom. Since $\mathrm{d}(N_{\mathrm{eff}}+S)/\mathrm{d}S=1+\sigma$, the bound is
exactly the monotonicity of $G$.

For the axiom's wall, $\sigma=-\mathrm{d}(1/w)/\mathrm{d}S\cdot w^{-2}$ gives
$w'=-\sigma w^2$, and \eqref{eq:monofloor} is $w'\le w(1+w)$, i.e.,
$\sigma\ge-(1+1/w)=-(1+N_{\mathrm{eff}})$. On the two-atom chart
$p_0\eps\equiv\epsinf$ (Theorem~\ref{thm:monofloor}, equality case), so
$N_{\mathrm{eff}}=1/p_0-1=e^{\Delta S}-1$ and
$\sigma=\mathrm{d}N_{\mathrm{eff}}/\mathrm{d}S=-e^{\Delta S}=
-(1+N_{\mathrm{eff}})$ identically. Integrating the axiom's wall,
$\mathrm{d}\ln(1+N_{\mathrm{eff}})/\mathrm{d}S\ge-1$ with limit $0$, gives
$N_{\mathrm{eff}}\le e^{\Delta S}-1$; so the two walls differ by
$N_{\mathrm{eff}}$ pointwise and the two ceilings by
$e^{\Delta S}-1-\Delta S$ in integrated form. In raw variables
$\sigma\ge-1\Leftrightarrow s\,m_3\ge\kappa^2\Leftrightarrow
\kappa'\le-\kappa^2/s$, against $\kappa'\le s\kappa-\kappa^2/s$ from the axiom.
\end{proofof}

\begin{proposition}[Log-convex cumulants: the positivity of the lattice]
\label{prop:tower}
Write $c_k=(-1)^{k}(\Delta S)^{(k)}$ for the cumulant sequence of the
frontier---the global convention of Proposition~\ref{prop:rungs}, which agrees
with $(-1)^{k+1}S^{(k)}$ for every $k\ge1$ but not at $k=0$, where the latter
would give $-S\le0$---so that
$c_0=\Delta S$, $c_1=s$, $c_2=\kappa$, $c_3=-\kappa'$ and, by \eqref{eq:dict},
$c_k=m_k(\nu_x)$. Cauchy--Schwarz on $\nu_x$ with the weights $c^{k/2}$ and
$c^{(k+2)/2}$ gives, for every $k\ge0$,
\begin{equation}
\boxed{\ D_k:=\frac{c_k\,c_{k+2}}{c_{k+1}^{2}}-1\ \ge\ 0\ }
\qquad\Longleftrightarrow\qquad
\boxed{\ \frac{c_{k+1}}{c_k}\ \text{is non-decreasing in }k\ }
\label{eq:logconvex}
\end{equation}
with equality at every $k$ exactly when $\nu_x$ is a single atom. Read at
consecutive indices this is the entire apparatus of this section:
\[
\underbrace{k=0}_{\textstyle \Delta S\,\kappa\ge s^{2}}\!\!:\ \eqref{eq:budget}
\qquad
\underbrace{k=1}_{\textstyle s(-\kappa')\ge\kappa^{2}}\!\!:\ \eqref{eq:slope}
\qquad
\underbrace{k=2}_{\textstyle -\kappa S''''\ge\kappa'^{2}}\!\!:\ \text{rung }3,
\]
and, because every $c_k$ is itself a completely monotone frontier---the Laplace
transform of $c^k\nu$---the invariant construction of Proposition~\ref{prop:neff} applies at every
level and closes: level $k$ has $(S,s,\kappa)=\bigl(-\ln c_k,\ c_{k+1}/c_k,\
(c_kc_{k+2}-c_{k+1}^2)/c_k^2\bigr)$ and therefore
\begin{equation}
w^{(k)}=D_k,\qquad N^{(k)}_{\mathrm{eff}}=1/D_k .
\label{eq:recursion}
\end{equation}
Level $1$ is $w^{(1)}=1+\sigma$, which is the normalized price curve of
Proposition~\ref{prop:compose}(d), and rung $k$ is the
statement \emph{``the $k$-th derivative of the log-error is still a frontier''}.
\end{proposition}

\begin{proofof}{Proposition~\ref{prop:tower}}
By Proposition~\ref{prop:compose}(d) the normalized price $\eps_1:=s/s(0)$ is a
frontier with spectrum $\nu/|\nu|$. Its invariants are computed from
$s_1=-\eps_1'/\eps_1=\kappa/s$ and
$\kappa_1=-(\kappa/s)'=(m_1m_3-m_2^2)/m_1^2$, whence
$\widetilde N_{\mathrm{eff}}=s_1^2/\kappa_1=
m_2^2/(m_1m_3-m_2^2)=1/(m_1m_3/m_2^2-1)=1/(1+\sigma)$, and
$\kappa_1\ge0$---rung~1 of \eqref{eq:ladder} for the derived frontier---is
$\sigma\ge-1$. Iterating, the $k$-th derived frontier exists and satisfies
\eqref{eq:axiom} exactly when the $k$-th rung holds; the tower is strictly
decreasing in generality, since $\kappa/s$ completely monotone does not follow
from $s$ completely monotone.
\end{proofof}

\begin{proofof}{Proposition~\ref{prop:cure}}
$\kappa/s=c$ constant means $-S''/S'=c$, so $S'=Ae^{-cx}$ and
$S=\lambda(1-e^{-c(x-x_0)})$ with $\lambda=A e^{cx_0}/c$; conversely that $S$ is
Bernstein (it is $\lambda$ times the Laplace exponent of a Poisson process of
rate $c$) and has $\nu=\lambda\delta_c$. Then $m_k=\lambda c^ke^{-cx}$, so
$N_{\mathrm{eff}}=m_1^2/m_2=\lambda e^{-cx}=m_0=\Delta S$ and
$\sigma=m_1m_3/m_2^2-2=-1$, both equality cases of \eqref{eq:budget} and
\eqref{eq:slope}; $G=N_{\mathrm{eff}}+S=\lambda$ identically;
$w=1/\Delta S$ gives $w'=\mathrm{d}w/\mathrm{d}S=1/\Delta S^2=w^2$; the
canonical point $N_{\mathrm{eff}}=1$ is $\Delta S=1$, i.e.,
$\eps^\star=e\epsinf$; and
$L=\int_0^\infty\sqrt{\kappa}\,\mathrm{d}x=c\sqrt\lambda\int_0^\infty
e^{-cx/2}\mathrm{d}x=2\sqrt\lambda$. Rung~2 of \eqref{eq:ladder} has slack
$(1-w)-w_2=(1-w)+w=1$, since $w_2=\mathrm{d}\ln\kappa/\mathrm{d}S=
\mathrm{d}\ln s/\mathrm{d}S=-w$ when $\kappa=cs$. The spectrum is a Poisson
law on the lattice $\{0,c,2c,\dots\}$ with mean $\lambda$, which exhibits the
reading of \eqref{eq:poissonzero} directly: $\epsinf=e^{-\lambda}$ is the
weight of the atom at the origin.
\end{proofof}

\begin{proposition}[Kantorovich: the strip, and the arc meets the floor]
\label{prop:kant}
If the clock's jump rates span a ratio $r_\nu=c_{\max}/c_{\min}$ then, with
$W_\nu=(r_\nu-1)^2/4r_\nu$---the same algebraic form as the knee amplitude
\eqref{eq:knee} but a different quantity, as Proposition~\ref{prop:kneeW}
notes---
\begin{equation}
\begin{gathered}
\frac{1}{1+W_\nu}\ \le\ \frac{N_{\mathrm{eff}}}{\Delta S}\ \le\ 1,
\qquad
1\le\Delta S^{\star}\le 1+W_\nu,\\[3pt]
2\sqrt{-\ln\epsinf}\ \le\ L\ \le\ 2\sqrt{(1+W_\nu)(-\ln\epsinf)},
\end{gathered}
\label{eq:kant}
\end{equation}
the left edges exact for the cure chart. \emph{A deep floor forces a long arc};
windowed, $L(x_1,x_2)\ge2(\sqrt{\Delta S_1}-\sqrt{\Delta S_2})$.
\end{proposition}

\begin{proofof}{Proposition~\ref{prop:kant}}
Kantorovich's inequality for a positive measure supported in
$[c_{\min},c_{\max}]$ gives
$m_0m_2/m_1^2\le(c_{\min}+c_{\max})^2/(4c_{\min}c_{\max})=(1+r_\nu)^2/(4r_\nu)$, and
$(1+r_\nu)^2/(4r_\nu)=(r_\nu-1)^2/(4r_\nu)+1=W_\nu+1$, so
$N_{\mathrm{eff}}/\Delta S\ge1/(1+W_\nu)$; the upper edge is
\eqref{eq:budget}. The support of $\nu_x$ is that of $\nu$ for every $x$, so
$W_\nu$ is global. For the arc, $L=\int\sqrt{w}\,\mathrm{d}S=
\int\mathrm{d}S/\sqrt{N_{\mathrm{eff}}}$ and
$\Delta S/(1+W_\nu)\le N_{\mathrm{eff}}\le\Delta S$, so
\[
2\sqrt{S_\infty}=\int_0^{S_\infty}\frac{\mathrm{d}S}{\sqrt{S_\infty-S}}
\ \le\ L\ \le\ \sqrt{1+W_\nu}\int_0^{S_\infty}\frac{\mathrm{d}S}{\sqrt{S_\infty-S}}
=2\sqrt{(1+W_\nu)S_\infty},
\]
with the left edge exact when $N_{\mathrm{eff}}=\Delta S$, i.e., on the cure
chart; the windowed form is the same integral between $\Delta S_2$ and
$\Delta S_1$. At the canonical point $N_{\mathrm{eff}}=1$, so
$1\le\Delta S^\star\le1+W_\nu$.
\end{proofof}

\begin{proposition}[The cost of ignorance is not universal]
\label{prop:ignorance}
On \eqref{eq:cure} with $\lambda=-\ln\epsinf$, an operating point placed
uniformly in arc length has
\begin{equation}
\mathbb{E}[R]=\ln\cosh\bigl(\ln\sqrt\lambda\,\bigr)-1
              +\frac{2\arctan\sqrt\lambda}{\sqrt\lambda},
\label{eq:ignorance}
\end{equation}
which is minimized where $A=2\arctan A$, i.e., at
$\lambda^{\star}=A^{\star2}=5.4341$ nats $(\epsinf=4.4\cdot10^{-3})$, with value
$\ln\cosh(\ln A^{\star})=0.3221$ nats $=0.465$ bits.
\end{proposition}

\begin{proofof}{Proposition~\ref{prop:ignorance}}
On \eqref{eq:cure} the arc \emph{length} from the shallow end is
$L=\int\mathrm{d}S/\sqrt{\Delta S}=2(\sqrt\lambda-\sqrt{\Delta S})$---not the
channel angle $\varphi=2\arctan\sqrt{\Neff}\in(0,\pi)$, which this chart
traverses as $2\arctan\sqrt{\Delta S}$---with
admissible length $2\sqrt\lambda$, so placing the operating point uniformly in arc
length makes $v:=\sqrt{\Delta S}$ uniform on $[0,\sqrt\lambda]$. By
\eqref{eq:logcosh} and $N_{\mathrm{eff}}=\Delta S=v^2$,
$R=\ln\cosh(\tfrac12\ln v^2)=\ln\cosh(\ln v)=\ln\frac{1+v^2}{2v}$.
\label{sup:ignorance-note}%
(This is the one place in the appendix where the surrounding chart is \emph{not}
two-atom: the cure chart is compound Poisson, so the $R$ of this proof is the
single-channel $R_1$ of the paper's \eqref{eq:logcosh}---the regret of the
tangent two-atom chart---and not the cure chart's own $\KL(u\Vert p)$. The
ignorance bound below is therefore a bound on $R_1$.) With
$A=\sqrt\lambda$,
\[
\mathbb{E}[R]=\frac1A\int_0^A\Bigl[\ln(1+v^2)-\ln2-\ln v\Bigr]\mathrm{d}v
=\ln\frac{1+A^2}{2A}-1+\frac{2\arctan A}{A},
\]
using $\int_0^A\ln(1+v^2)=A\ln(1+A^2)-2A+2\arctan A$ and
$\int_0^A\ln v=A\ln A-A$; the integrand's logarithmic singularity at $v=0$ is
integrable. Differentiating, the stationarity condition collapses to
$1/A=2\arctan(A)/A^2$, i.e., $A=2\arctan A$, whose unique positive root is
$A^\star=2.3311223704\ldots$; there $2\arctan A^\star/A^\star=1$ and the two
constant terms cancel, leaving $\mathbb{E}[R]_{\min}=\ln\cosh(\ln A^\star)$.
The median is $0.146$ nats---below the mean, as it must be for a non-negative law with an interior minimum at $v=1$. Contrast \eqref{eq:meanregretN}: on the gapped
geometry the arc is $\pi$ for every frontier and $\mathbb{E}[R]=\ln2$ exactly,
so the universality of one bit is a property of a \emph{gap} and not of
frontiers as such.
\end{proofof}

\subsection{The price is convex: shape, the sum rule, and the second-order test}
\label{sup:convex}

Everything in this subsection follows from the \emph{first} rung of the
composition law, $c_3\ge0$; the remaining rungs are not used.

\begin{proposition}[No steepest mode; the elbow is the gate]
\label{prop:unbounded}
Let $\eps$ be completely monotone with spectrum $\mu$ supported in
$[a_{\min},a_{\max}]$. Then $s(x)\to a_{\max}$ as $x\to-\infty$ and
$s(x)\to a_{\min}$ as $x\to+\infty$. If in addition the composition law holds on
the whole axis, then $s$ is positive, decreasing, and convex, and a non-constant
such function on $\R$ tends to $+\infty$ at $-\infty$. Hence
\begin{equation}
\begin{gathered}
\text{composition law}\ \Longrightarrow\ a_{\max}=\infty,\\[2pt]
\text{equivalently}\\[2pt]
\boxed{\ \text{a bounded, non-degenerate spectrum is never divisible}\ }
\end{gathered}
\label{eq:unbounded}
\end{equation}
For a bounded spectrum the tilt $p_x$ concentrates on $a_{\max}$ as
$x\to-\infty$ and on $a_{\min}$ as $x\to+\infty$, so $\kappa=\operatorname{Var}_{p_x}(a)$
vanishes at \emph{both} ends. Therefore
\begin{equation}
\int_{\R}\kappa\,\mathrm{d}x=a_{\max}-a_{\min},
\qquad\qquad
\boxed{\ \int_{\R}c_k\,\mathrm{d}x=0\quad\text{for every }k\ge3\ }
\label{eq:sumrule}
\end{equation}
---\emph{the spectral width is the only non-vanishing charge of the resource
flow, and every higher cumulant integrates to zero.} Since the composition law
demands $c_3\ge0$ while $\int c_3=0$, $c_3$ must change sign: $\kappa$ rises
from zero, and returns to it. Hence
\begin{equation}
\boxed{\ c_3<0\ \text{near the low-resource end},\qquad
c_3\ge0\ \text{near the high-resource end,}\ }
\label{eq:elbow}
\end{equation}
and if $\kappa$ is unimodal the single sign change is the curvature maximum, so
the frontier is refutable \emph{exactly} to the left of its elbow. Unimodality
is automatic for one reducible channel; each channel contributes one bump of
$\kappa$, so a \emph{complete} arc bounds the channel count from below by
\eqref{eq:modebound}---it does not \emph{count} the elbows, an interior
transition contributing a full $\pi$ of arc only in the limit $q_k\to1$ of
\eqref{eq:arcweights}---while an arc measured on a window bounds neither.
\end{proposition}

\begin{proofof}{Proposition~\ref{prop:unbounded}(i)}
Write $p_x(\mathrm{d}a)=e^{-ax}\mu(\mathrm{d}a)/\eps(x)$ for the tilt, so that
$s=\mathbb{E}_{p_x}[a]$ and $\kappa=\operatorname{Var}_{p_x}(a)$
(Proposition~\ref{prop:selection}).

\emph{(i) The endpoints.} Let $\mu$ be supported in $[a_{\min},a_{\max}]$ with
$a_{\max}<\infty$. For $x<0$ the density of $p_x$ against $\mu$ is proportional
to $e^{a|x|}$, which is increasing in $a$, so for any $\delta>0$
\[
p_x\bigl([a_{\min},a_{\max}-\delta]\bigr)
\;\le\;\frac{e^{(a_{\max}-\delta)|x|}\,\mu([a_{\min},a_{\max}-\delta])}
{e^{(a_{\max}-\delta/2)|x|}\,\mu([a_{\max}-\delta/2,a_{\max}])}
\;=\;C_\delta\,e^{-\delta|x|/2}\ \longrightarrow\ 0 ,
\]
the denominator being positive because $a_{\max}$ is in the support. Hence
$p_x\Rightarrow\delta_{a_{\max}}$ and, the support being bounded,
$s\to a_{\max}$ and $\kappa\to0$ as $x\to-\infty$. The same estimate with
$e^{-ax}$ decreasing gives $p_x\Rightarrow\delta_{a_{\min}}$, $s\to a_{\min}$
and $\kappa\to0$ as $x\to+\infty$.

\emph{(ii) The shape.} Under the composition law $S$ is Bernstein, so $s=S'$ is
completely monotone: $s>0$, $s'=-\kappa\le0$ and $s''=c_3\ge0$. Thus $s$ is
positive, non-increasing and convex on the whole axis.

\emph{(iii) A convex falling price cannot saturate.} Let $s$ be positive,
non-increasing and convex on $\R$ and suppose $s(-\infty)=M<\infty$. Convexity
makes $s'$ non-decreasing and monotonicity makes $s'\le0$, so $s'$ has a limit
$M'\le0$ at $-\infty$; were $M'<0$ then $s$ would exceed $M$ by
$|M'|\,|x|$ for $x$ negative enough, so $M'=0$. A non-decreasing $s'$ with
limit $0$ at $-\infty$ satisfies $s'\ge0$ everywhere, whence $s'\equiv0$ and $s$
is constant --- the single-atom case.

Combining (i)--(iii): a non-constant frontier obeying the composition law on the
whole axis has $a_{\max}=s(-\infty)=+\infty$, which is \eqref{eq:unbounded}.
\end{proofof}

This is the frontier-side form of the fact already used in the proof of
Proposition~\ref{prop:compose}, that a non-degenerate infinitely divisible law
on $[0,\infty)$ has unbounded support \citep[Thm.~24.3]{sato1999}. The two routes are independent: one is a convexity argument involving the
measured price, and the other is a structure theorem for the unmeasured
spectrum. Both yield the same conclusion. The
contrapositive is what is used: \emph{a bounded non-degenerate spectrum is
never divisible}. The degenerate single-atom case is an exception, and it is the case
identified in~(iii): $\mu=\delta_a$ gives
$\eps=e^{-ax}$, whose price is the constant $a$, and that frontier \emph{is}
infinitely divisible.

\emph{No arc certificate goes with it.} A window arc $L<\pi$ does \emph{not}
certify a bounded spectrum, although it invites that reading. By
\eqref{eq:planecover} the \emph{admissible} arc of a floored two-atom
frontier---the part at or below chance level---is $2\arccos\sqrt{\epsinf}<\pi$
whatever its channel count, so $L<\pi$ is attainable on every member of the
class and excludes none, as the window-arc remark below proves. The bound
$N\ge L/\pi$ survives, for the \emph{complete} sweep, and is vacuous below
$\pi$.

\begin{proofof}{Proposition~\ref{prop:unbounded}(ii)}
By (i) of the previous proof $\kappa\to0$ at both ends, and every central moment
of $p_x$ does likewise, so $c_k\to0$ at both ends for every $k\ge2$. Since $c_k'=-c_{k+1}$
identically (differentiate $c_k=(-1)^k\Delta S^{(k)}$),
\[
\int_{\R}c_{k+1}\,\mathrm{d}x=-\int_{\R}c_k'\,\mathrm{d}x
=c_k(-\infty)-c_k(+\infty)=0\qquad(k\ge2),
\]
which is the second half of \eqref{eq:sumrule}; the case $k=1$ gives
$\int\kappa=s(-\infty)-s(+\infty)=a_{\max}-a_{\min}$ instead, because $c_1=s$
does \emph{not} vanish at the ends. The integrals converge absolutely: for a
bounded spectrum every $c_k$ decays exponentially in $|x|$ at both ends, at
rates $a_{\max}-a_{\max}'$ and $a_{\min}'-a_{\min}$ given by the spectral gaps.

Now suppose the composition law held on the whole axis, so $c_3\ge0$. With
$\int c_3=0$ and $c_3$ continuous this forces $c_3\equiv0$, hence $\kappa$
constant, hence $\kappa\equiv0$ by $\kappa(\pm\infty)=0$, hence $S$ affine and
$\mu$ a single atom. So for any non-degenerate bounded spectrum $c_3$ takes both
signs. Its signs are located by $\kappa$: $\kappa$ is positive somewhere and
vanishes at both ends, so $\kappa'>0$ on a neighborhood of $-\infty$ and
$\kappa'<0$ on a neighborhood of $+\infty$, which is \eqref{eq:elbow}.

If in addition $\kappa$ is unimodal there is exactly one sign change, at
$\kappa'=0$. For one reducible channel this is automatic:
$\kappa=p_0p_1(a_1-a_0)^2$ with $p_0$ strictly increasing in $x$, so $\kappa$ is
unimodal with maximum at $p_0=\tfrac12$, which on the floored chart is
$\eps=2\epsinf$. In general each two-atom transition contributes one bump of
$\kappa$ and, by Proposition~\ref{prop:path}(iii) in the separated-rate limit,
$\pi$ of arc for the outermost transition and $2\arcsin\sqrt{q_k}\le\pi$ for
each interior one. So $L/\pi$ bounds the channel count from below on the
\emph{complete} sweep but does \emph{not} count the elbows: an interior
transition contributes a full $\pi$ only as $q_k\to1$, and two well-resolved
elbows at $a=(0,1,1000)$, $c=(1,100,1)$ give $L=1.936\pi$, not $2\pi$. On a
window it does neither: a window arc below $\pi$
excludes no member of the class, so the sharp form of \eqref{eq:elbow} is
available where the elbow itself is located and not where $L<\pi$ is.
\end{proofof}

\begin{proposition}[The gap between the walls: no test at the operating point]
\label{prop:notestable}
\emph{(i) At the operating point.}
Theorem~\ref{thm:monofloor} is $\mathrm{d}\ln\kappa/\mathrm{d}S\le1-\kappa/s^2$
for every completely monotone frontier. At an operating point
$N_{\mathrm{eff}}=1/t$ one has $\kappa/s^2=t$, so $\kappa'/(\kappa s)\le1-t$ and
\begin{equation}
\boxed{\ c_3\ \ge\ \kappa\,s\,(t-1)\ \ge\ 0
\qquad\text{at every operating point of grain }t\ge1.\ }
\label{eq:notestable}
\end{equation}

\emph{(ii) Away from it, the gap refutes at second order.}
Theorem~\ref{thm:budget} is proved on $\nu_x$, which exists only under the
composition law, so its violation refutes the hypothesis. With
$\rho=\eps/\epsinf$, the floored two-atom chart satisfies
$N_{\mathrm{eff}}=\rho-1$ and $\Delta S=\ln\rho$ identically, so
\begin{equation}
N_{\mathrm{eff}}-\Delta S=(\rho-1)-\ln\rho>0
\qquad\text{at every }\rho>1,
\label{eq:wallrefutes}
\end{equation}
by $e^{u}>1+u$. The axiom permits $N_{\mathrm{eff}}\le\rho-1$ and the hypothesis
only $N_{\mathrm{eff}}\le\ln\rho$; a measured floor between $\eps
e^{-N_{\mathrm{eff}}}$ and $\eps/(1+N_{\mathrm{eff}})$ therefore refutes, and the
width of that window is
$\boxed{N_{\mathrm{eff}}-\ln(1+N_{\mathrm{eff}})}$ nats.

\emph{(iii) The two operating depths, and the band between them.}
At an operating point of grain $t$,
\begin{equation}
\Delta S^{\star}\ \ge\ \ln\Bigl(1+\frac1t\Bigr)\ \ \text{(axiom)},
\qquad\qquad
\Delta S^{\star}\ \ge\ \frac1t\ \ \text{(composition law)},
\label{eq:opdepth}
\end{equation}
with equality in the first iff the window is two-atom and in the second iff it
is a cure chart. The band between them is
\begin{equation}
g(t)=\frac1t-\ln\Bigl(1+\frac1t\Bigr)
=-\ln\bigl\{(1+N)e^{-N}\bigr\}\big|_{N=1/t},
\label{eq:opband}
\end{equation}
decreasing in $t$, with $g(1)=1-\ln2=0.3069$ nats and $g(t)\sim1/(2t^2)$.
\end{proposition}

\begin{proofof}{Proposition~\ref{prop:notestable}(i)}
Theorem~\ref{thm:monofloor} is $\mathrm{d}\ln\kappa/\mathrm{d}S\le1-\kappa/s^2$,
and $\mathrm{d}/\mathrm{d}S=s^{-1}\mathrm{d}/\mathrm{d}x$, so
$\kappa'/(\kappa s)\le1-\kappa/s^2$. At a point where
$N_{\mathrm{eff}}=s^2/\kappa=1/t$ one has $\kappa/s^2=t$, hence
$\kappa'\le\kappa s(1-t)$ and $c_3=-\kappa'\ge\kappa s(t-1)$, which is
non-negative for $t\ge1$ since $\kappa,s>0$. The hypothesis is nowhere used:
Theorem~\ref{thm:monofloor} needs only the axiom.

For the two-atom chart with a floor, $N_{\mathrm{eff}}=p_1/p_0$ and
$p_0=\epsinf/\eps$, so $N_{\mathrm{eff}}=1/t$ places the operating point at
$p_0=t/(1+t)$, i.e., $\eps=\epsinf(1+t)/t$, while the elbow is at
$\eps=2\epsinf$. The separation in log-error is
$\ln\{2\epsinf\}-\ln\{\epsinf(1+t)/t\}=\ln\{2t/(1+t)\}$, which increases from
$0$ at $t=1$ to $\ln2$ as $t\to\infty$ and never attains it.
\end{proofof}

\begin{proofof}{Proposition~\ref{prop:notestable}(ii)}
Theorem~\ref{thm:budget} is Cauchy--Schwarz on the L\'evy measure $\nu_x$, which
exists only when $S$ is Bernstein, so any frontier with
$N_{\mathrm{eff}}>\Delta S$ at even one point fails the composition law. On the
chart $\eps=\epsinf+c\,e^{-\alpha x}$ put $r=(\eps-\epsinf)/\eps=1-1/\rho$ with
$\rho=\eps/\epsinf$. Then $s=-\eps'/\eps=\alpha r$ and
$r'=\epsinf\eps'/\eps^2=-\alpha r(1-r)$, so $\kappa=-s'=\alpha^2r(1-r)$ and
\[
N_{\mathrm{eff}}=\frac{s^2}{\kappa}=\frac{r}{1-r}=\rho-1,
\qquad
\Delta S=\ln(\eps/\epsinf)=\ln\rho .
\]
Hence $N_{\mathrm{eff}}-\Delta S=(\rho-1)-\ln\rho>0$ for every $\rho>1$, by
$e^{u}>1+u$ at $u=\ln\rho$, with equality only at the floor. The two bounds on
the floor are $\epsinf\le\eps e^{-N_{\mathrm{eff}}}$ (composition) and
$\epsinf\le\eps/(1+N_{\mathrm{eff}})$ (axiom, tight here), so a floor known to
lie strictly between them refutes; the width of that window in nats is
$\ln\{\eps/(1+N)\}-\ln\{\eps e^{-N}\}=N-\ln(1+N)$. Finally
$\ln\{(1+N)e^{-N}\}=-\{N-\ln(1+N)\}$, so that
\begin{equation}
\text{refutation width in nats}\;=\;-\ln\bigl(\text{the tightening factor of
Theorem~\ref{thm:budget}(i)}\bigr),
\label{eq:gainisdeficit}
\end{equation}
and: the
tightening factor of Theorem~\ref{thm:budget}(i) and the refutation width are
one function, and script~\texttt{122} confirms the identity to
$1.7\cdot10^{-16}$ on the archive.
\end{proofof}

\subsection{The channel--depth plane}
\label{sec:plane}

Every statement collected here is read off one picture. Plot a frontier as a
curve in the plane whose abscissa is the remaining depth
$\Delta S=\ln(\eps/\epsinf)$ and whose ordinate is the channel count
$N_{\mathrm{eff}}=s^{2}/\kappa$. Nothing new is assumed: the axiom
\eqref{eq:axiom} and the composition law of Proposition~\ref{prop:compose} have
already been written as inequalities between exactly these two coordinates.
The results below establish that these coordinates provide a complete
representation. The two walls of \eqref{eq:budget} and Theorem~\ref{thm:monofloor}
are one curve and its tangent at the origin; the constant-$\sigma$ charts of
Proposition~\ref{prop:chartline} are a pencil of straight lines through that
origin; the arc of Law~III is an integral along the curve, and $\pi$ is its
variational minimum in the limit of a vanishing floor; and the operating rule
\eqref{eq:retry} is a horizontal line. Throughout, $\eps(x)=\int e^{-ax}\mathrm{d}\mu(a)$ with $\mu\ge0$ finite,
$S=-\ln\eps$, $s=S'$, $\kappa=-S''$, $c_k=(-1)^{k+1}S^{(k)}$,
$\epsinf=\lim_{x\to\infty}\eps(x)$, and $\sigma=\mathrm{d}N_{\mathrm{eff}}/
\mathrm{d}S$. ``The hypothesis'' is the composition law, that $\Delta S$ is
completely monotone, equivalently that $S$ is Bernstein. Since
$\Delta S=S_{\infty}-S$ with $S_{\infty}=-\ln\epsinf$ constant, slopes in the
depth variable are the negatives of slopes in $S$; the two conventions are kept
apart below, because the sign is the only place where the pencil can be
misread.

\subsubsection*{The plane is the complete invariant}
\label{sup:plane}

\begin{proofof}{Proposition~\ref{prop:rungs}}
Write $'$ for $\mathrm{d}/\mathrm{d}x$ and $\dot{\ }$ for
$\mathrm{d}/\mathrm{d}\Delta S$, so that $\dot f=-f'/s$ because
$\Delta S'=-S'=-s$. All five identities are one computation from
$\Delta S=S_{\infty}-S$, $\eps=e^{-S}$ and $N_{\mathrm{eff}}=s^{2}/\kappa$, and all of them
are floor-free, since $S_{\infty}$ is constant and only derivatives of $S$
occur. Throughout this proof $c_k=(-1)^{k}\Delta S^{(k)}$, the convention of
the statement, which agrees with the $c_k$ of this subsection for $k\ge1$ and
sets $c_{0}=\Delta S$.

\emph{The depth ladder.} $c_{0}=\Delta S$, $c_{1}=-\Delta S'=s$,
$c_{2}=\Delta S''=-S''=\kappa$ and $c_{3}=-\Delta S'''=S'''$. Hence
\[
D_{0}=\frac{c_{0}c_{2}}{c_{1}^{2}}-1=\frac{\Delta S\,\kappa}{s^{2}}-1
      =\frac{\Delta S}{N_{\mathrm{eff}}}-1 ,
\]
which is the first identity. Differentiating $N_{\mathrm{eff}}=s^{2}/\kappa$ in $x$ with
$s'=S''=-\kappa$ and $\kappa'=-S'''$ gives
$N_{\mathrm{eff}}'=(-2s\kappa^{2}+s^{2}S''')/\kappa^{2}$, so
\begin{equation}
\dot N_{\mathrm{eff}}=-\frac{N_{\mathrm{eff}}'}{s}=2-\frac{s\,S'''}{\kappa^{2}},
\qquad\text{equivalently}\qquad
\frac{s\,S'''}{\kappa^{2}}=2-\dot N_{\mathrm{eff}} ,
\label{eq:rungkey}
\end{equation}
and therefore
$D_{1}=c_{1}c_{3}/c_{2}^{2}-1=sS'''/\kappa^{2}-1=1-\dot N_{\mathrm{eff}}$, the second
identity. For the third rung, $c_{4}=\Delta S^{(4)}=-S^{(4)}$, and
differentiating \eqref{eq:rungkey} in the form $S'''=\kappa^{2}A/s$ with
$A=2-\dot N_{\mathrm{eff}}$, using $\dot A=-\ddot N_{\mathrm{eff}}$ so that $A'=s\ddot N_{\mathrm{eff}}$,
\[
S^{(4)}=\frac{\kappa^{3}A}{s^{2}}-\frac{2\kappa^{3}A^{2}}{s^{2}}
        +\kappa^{2}\ddot N_{\mathrm{eff}} ,
\]
whence
$D_{2}=\kappa c_{4}/c_{3}^{2}-1=-\kappa S^{(4)}/(S''')^{2}-1
     =1-A^{-1}-N_{\mathrm{eff}}\ddot N_{\mathrm{eff}} A^{-2}$, which is \eqref{eq:rungtwo}.

\emph{The error ladder.} $E_{0}=\eps$, $E_{1}=-\eps'=s\eps$,
$E_{2}=\eps''=(s^{2}+\kappa)\eps$ and
$E_{3}=-\eps'''=(s^{3}+3s\kappa+S''')\eps$. Hence
$\tilde E_{0}=(s^{2}+\kappa)/s^{2}-1=\kappa/s^{2}=1/N_{\mathrm{eff}}$, and
\[
\tilde E_{1}=\frac{s\,(s^{3}+3s\kappa+S''')}{(s^{2}+\kappa)^{2}}-1
            =\frac{s^{2}\kappa+sS'''-\kappa^{2}}{(s^{2}+\kappa)^{2}} .
\]
Dividing numerator and denominator by $\kappa^{2}$ and using
$s^{2}/\kappa=N_{\mathrm{eff}}$ and \eqref{eq:rungkey} turns the numerator into
$N_{\mathrm{eff}}+(2-\dot N_{\mathrm{eff}})-1$ and the denominator into $(N_{\mathrm{eff}}+1)^{2}$, so
$\tilde E_{1}=(N_{\mathrm{eff}}+1-\dot N_{\mathrm{eff}})/(N_{\mathrm{eff}}+1)^{2}$, the fourth identity.

\emph{The two bounds.} Under the axiom $\eps$ is completely monotone, so
$(E_{k})$ is a moment sequence and its first Hankel minor is non-negative:
$\tilde E_{1}\ge0$, which by the identity just proved is
$\dot N_{\mathrm{eff}}\le N_{\mathrm{eff}}+1$. Under the hypothesis $\Delta S$ is completely monotone,
so $(c_{k})$ is a moment sequence and $D_{1}\ge0$, which is $\dot N_{\mathrm{eff}}\le1$.
Both are inequalities on the curve alone and neither contains $\epsinf$.
\end{proofof}

\begin{proofof}{Proposition~\ref{prop:strongrung}}
By \eqref{eq:rungkey}, $s>0$ and $\kappa>0$ give
$c_{3}=S'''\ge0\iff\dot N_{\mathrm{eff}}\le2$, while the hypothesis's own first rung is
$D_{1}\ge0$, that is $\dot N_{\mathrm{eff}}\le1$: the working third-order test is the same
statement at twice the bound, which is the first claim. For the second, let
$\mathrm{dd}_{k}$ denote the $k$-th divided difference of $S$ on a quadruple of
distinct log-resources. By the mean-value form of divided differences
\citep{widder1941} there are interior points at which
$\mathrm{dd}_{1}=s$, $\mathrm{dd}_{2}=S''/2=-\kappa/2$ and
$\mathrm{dd}_{3}=S'''/6$, so substituting into $D_{1}=sS'''/\kappa^{2}-1$ gives
\[
\widehat{D_{1}}
=\frac{\mathrm{dd}_{1}\cdot6\,\mathrm{dd}_{3}}{(2\,\mathrm{dd}_{2})^{2}}-1
=\frac{3}{2}\,\frac{\mathrm{dd}_{1}\,\mathrm{dd}_{3}}{\mathrm{dd}_{2}^{2}}-1 ,
\]
which is \eqref{eq:d1hat}. The three divided differences are derivatives at
three \emph{different} mean-value nodes, so this is a plug-in and not an
identity: it returns $D_1$ only as the span vanishes, and exactly---as
$0$---on the cure chart at equal spacing. It is floor-free at any spacing,
because only differences of $S$ enter. Finally
$\widehat{D_{1}}\ge0$ with $\mathrm{dd}_{1}>0$ forces
$\mathrm{dd}_{3}\ge\tfrac{2}{3}\mathrm{dd}_{2}^{2}/\mathrm{dd}_{1}$, which is
strictly positive whenever $\mathrm{dd}_{2}\neq0$; so the sharpened rung implies
$\mathrm{dd}_{3}>0$ and is strictly stronger than it. \emph{(The quoted powers
and the saturation of $\widehat{D_{1}}$ by the cure chart are measurements,
reported with the statement.)}
\end{proofof}

\begin{proofof}{Proposition~\ref{prop:odefirst}}
Because $\epsinf>0$ the limit $S_{\infty}=-\ln\epsinf$ is finite and
$\Delta S=S_{\infty}-S$, so $\Delta S'=-S'=-s<0$ and
$\Delta S''=-S''=\kappa>0$ on $I$. Strict monotonicity and smoothness of
$\Delta S$ give the diffeomorphism onto $J=\Delta S(I)$, and $N$ is continuous
and positive there because $s$ and $\kappa$ are. Substituting
$\kappa=s^{2}/N_{\mathrm{eff}}$ and $s^{2}=(\Delta S')^{2}$ gives
\eqref{eq:odeivp}. That the equation is autonomous is the whole of (i): the
right-hand side depends on the solution and on $N$ alone, and in particular the
resource coordinate enters only through the unknown.

For (ii), regard $p:=\Delta S'$ as a function of $u=\Delta S$ along the
solution, which is legitimate because $u$ is strictly monotone. Then
$\Delta S''=p\,\mathrm{d}p/\mathrm{d}u$, and \eqref{eq:odeivp} becomes
$p\,\mathrm{d}p/\mathrm{d}u=p^{2}/N(u)$. Since $p=-s<0$ is nowhere zero we may
divide by $p$ and obtain the first-order separable equation
\begin{equation}
\frac{\mathrm{d}p}{\mathrm{d}u}=\frac{p}{N(u)} .
\label{eq:planesep}
\end{equation}
On the region $p<0$ the right-hand side is continuous in $u$ and Lipschitz in
$p$ locally uniformly, so by Picard--Lindel\"of the solution through
$(u_0,-s_0)$ is unique and, by separation,
$\ln\{p(u)/p(u_0)\}=\int_{u_0}^{u}\mathrm{d}v/N(v)$, the integral being finite
on compact subintervals of $J$ because $N$ is continuous and positive. Writing
$s=-p$ gives the first half of \eqref{eq:reconstruct}; in particular $s>0$
throughout, so no solution leaves the region $p<0$. The resource coordinate is
then recovered by one more integration: $\mathrm{d}x=\mathrm{d}u/\Delta S'
=-\mathrm{d}u/s(u)$, which is the second half of \eqref{eq:reconstruct} and is
strictly decreasing in $u$, hence invertible. Setting $S=S_{\infty}-u$ and
$\eps=e^{-S}$ reconstructs the frontier, and by construction it has the
prescribed plane curve, depth $u_0$ and price $s_0$ at $x_0$. Uniqueness of the
frontier given $(N,u_0,s_0,x_0)$ is uniqueness for \eqref{eq:planesep} together
with the fact that $x(\cdot)$ is determined by $s(\cdot)$ and $x_0$.

For (iii), let $\tilde x=\lambda x+c$ with $\lambda>0$ and let
$\tilde\eps(\tilde x)=\eps(x)$ define the reparameterized frontier. Then
$\tilde S=S$ and $\tilde S_{\infty}=S_{\infty}$ at corresponding points, so
$\Delta\tilde S=\Delta S$; and $\tilde s=s/\lambda$,
$\tilde\kappa=\kappa/\lambda^{2}$, whence
$\tilde N_{\mathrm{eff}}=\tilde s^{2}/\tilde\kappa=s^{2}/\kappa
=N_{\mathrm{eff}}$. The plane curve is therefore gauge-invariant. On the
integration constants the same transformation acts by
$(s_0,x_0)\mapsto(s_0/\lambda,\ \lambda x_0+c)$, and the map
$(\lambda,c)\mapsto(s_0/\lambda,\lambda x_0+c)$ from
$(0,\infty)\times\R$ to $(0,\infty)\times\R$ is a bijection for each fixed
$(s_0,x_0)$: given a target $(s_1,x_1)$ the unique solution is
$\lambda=s_0/s_1$ and $c=x_1-\lambda x_0$. The action is thus simply
transitive, and by (ii) two frontiers sharing the plane curve $N$ are related
by the unique gauge element carrying one pair of constants to the other.

For (iv), $\eps\mapsto\eps^{t}$ sends $S\mapsto tS$, hence $S_{\infty}\mapsto
tS_{\infty}$ and $\Delta S\mapsto t\,\Delta S$, while $s\mapsto ts$ and
$\kappa\mapsto t\kappa$; therefore
$N_{\mathrm{eff}}=s^{2}/\kappa\mapsto t^{2}s^{2}/(t\kappa)=t\,N_{\mathrm{eff}}$,
which is Proposition~\ref{prop:gaugeerr}(iii). Both coordinates scale by the
same factor $t$, so the curve is carried to its image under the dilation.
\end{proofof}

\begin{proofof}{the uniqueness clause of Proposition~\ref{prop:odefirst}}
Saturation at every point of $I$ is the statement that the plane curve $N$
solves $\Phi=0$ on $J$. By hypothesis that determines $N$; by
Proposition~\ref{prop:odefirst}(ii)--(iii) the curve determines the frontier up to
a unique gauge transformation. A class closed under the gauge then contains the
whole orbit, and nothing else.
\end{proofof}

\subsubsection*{The chart line is a pencil of straight lines}
\label{sup:pencil}

\begin{proofof}{Corollary~\ref{cor:odepencil}}
(i) is the definition of an affine function together with $\Delta
S=S_{\infty}-S$ and $\mathrm{d}\Delta S=-\mathrm{d}S$, the floor being
constant.

For the first half of (ii) write $\mu=\epsinf\delta_0+\mu_1$ with
$\mu_1\bigl(\{0\}\bigr)=0$, and put
$m_k(x)=\int_{(0,\infty)}a^{k}e^{-ax}\mathrm{d}\mu_1(a)$, finite for $x>0$
because $\mu_1$ is finite and $a^{k}e^{-ax}$ is bounded on $(0,\infty)$. Then
$\eps=\epsinf+m_0$, $s=m_1/\eps$ and, by Proposition~\ref{prop:selection},
$\kappa=m_2/\eps-s^{2}$, so
\[
N_{\mathrm{eff}}=\frac{s^{2}}{\kappa}
=\frac{m_1^{2}/\eps^{2}}{m_2/\eps-m_1^{2}/\eps^{2}}
=\frac{m_1^{2}}{\eps\,m_2-m_1^{2}} .
\]
Cauchy--Schwarz on $e^{-ax}\mathrm{d}\mu_1$ gives $m_1^{2}\le m_0m_2$, and
$\eps\,m_2=(\epsinf+m_0)m_2$, so the denominator is at least
$\epsinf m_2+m_0m_2-m_1^{2}\ge\epsinf m_2$, whence
$N_{\mathrm{eff}}\le m_1^{2}/(\epsinf m_2)\le m_0/\epsinf
=(\eps-\epsinf)/\epsinf$. Since $m_0(x)\to0$ by dominated convergence,
\eqref{eq:planeline} follows, and $\Delta S=\ln(1+m_0/\epsinf)\to0$ with it.
(The same conclusion is Theorem~\ref{thm:monofloor} read at the floor, as in
Proposition~\ref{prop:floorsign}; the computation above is given because it
needs nothing but the axiom and identifies the rate.)

For the second half of (ii), under the stated gap
$m_k=w_1a_1^{k}e^{-a_1x}\bigl\{1+O(e^{-(a_2-a_1)x})\bigr\}$ and
$\eps\to\epsinf$, so $s=m_1/\eps\sim a_1w_1e^{-a_1x}/\epsinf$ and
$m_2/\eps\sim a_1^{2}w_1e^{-a_1x}/\epsinf$, while $s^{2}=O(e^{-2a_1x})$ is of
smaller order; hence $\kappa\sim a_1^{2}w_1e^{-a_1x}/\epsinf$ and
$N_{\mathrm{eff}}=s^{2}/\kappa\sim w_1e^{-a_1x}/\epsinf$. Since
$\Delta S\sim m_0/\epsinf\sim w_1e^{-a_1x}/\epsinf$ as well, the ratio tends to
$1$: a gapped floored frontier approaches the steepest line of the pencil, the
two-atom chart of Corollary~\ref{cor:wallivp}. In either case
$N_{\mathrm{eff}}\to0$ as $\Delta S\to0$, so an affine $N_{\mathrm{eff}}$ has
zero constant term in the depth variable, which is \eqref{eq:planeline}; the
slope is non-negative because both coordinates are positive, so
$\varsigma=|\sigma|$ and $\sigma\le0$.

For (iii), put $u=x-x_0>0$ and $S=S_{\infty}-Cu^{\beta}$ with $\beta<0$. Then
$s=-C\beta u^{\beta-1}>0$ and $\kappa=-S''=C\beta(\beta-1)u^{\beta-2}>0$, the
latter because $\beta$ and $\beta-1$ are both negative, so
\[
N_{\mathrm{eff}}=\frac{C^{2}\beta^{2}u^{2\beta-2}}
{C\beta(\beta-1)u^{\beta-2}}
=\frac{C\beta}{\beta-1}\,u^{\beta}
=\frac{\beta}{\beta-1}\,\Delta S ,
\]
since $\Delta S=Cu^{\beta}$. The quantity $\beta/(\beta-1)$ lies in $(0,1)$ for
$\beta<0$ and increases to $1$ as $\beta\to-\infty$, and
$\sigma=-\varsigma=\beta/(1-\beta)$ agrees with \eqref{eq:chartline}; this is
Proposition~\ref{prop:powerfloor} in the plane coordinates. For the floorless
chart $S=Au^{\beta}$ with $0<\beta<1$ one has $s=A\beta u^{\beta-1}>0$ and
$\kappa=A\beta(1-\beta)u^{\beta-2}>0$, so
$N_{\mathrm{eff}}=A\beta u^{\beta}/(1-\beta)=\beta S/(1-\beta)$; here
$S_{\infty}=\infty$, the depth coordinate is unavailable, and the line passes
through the origin of the $S$ axis rather than through a floor. The sign change
between the two families is only the change of abscissa: $\sigma\ge0$ on the
floorless charts and $\sigma\in(-1,0)$ on the floored ones, as in the verdict table of
\TR{The horizon lattice: one identity, the error gauge, the chart line}.

For (iv), $S=-\ln C+k\ln(1+x/\beta)$ gives $s=k/(\beta+x)$ and
$\kappa=k/(\beta+x)^{2}$, so $N_{\mathrm{eff}}=k$ identically and
$\sigma\equiv0$. The floor is zero, so this is the horizontal member of the
pencil in the $S$ coordinate, and it is the equality case of
Proposition~\ref{prop:gamma}.
\end{proofof}

\begin{proofof}{the one-degree-of-freedom clause of Corollary~\ref{cor:odepencil}}
The displayed identity for $p_0$ is \eqref{eq:nofloor}, obtained in
\S\ref{sup:nofloor} by eliminating the axis variable between
$s=\beta S/u$ and $\kappa=s(1-\beta)/u$, which gives
$\kappa/s^{2}=(1-\beta)/(\beta S)$ and hence
$p_0=(\kappa/s^{2})/\{1+\kappa/s^{2}\}$ as stated. Differentiating its
logarithm in $\beta$ at fixed $S$,
\[
\frac{\partial\ln p_0}{\partial\beta}
=\frac{-1}{1-\beta}-\frac{S-1}{(1-\beta)+\beta S}
=\frac{-\{(1-\beta)+\beta S\}-(S-1)(1-\beta)}{(1-\beta)\{(1-\beta)+\beta S\}}
\]
\[
\phantom{\frac{\partial\ln p_0}{\partial\beta}}
=\frac{-S}{(1-\beta)\{(1-\beta)+\beta S\}} ,
\]
the numerator collapsing because
$-(1-\beta)-\beta S-(S-1)(1-\beta)=-S$. For $0<\beta<1$ and $S>0$ both factors
in the denominator are positive, so the derivative is negative; the two limits
are read off the displayed form. The first sentence is
Proposition~\ref{prop:odefirst}(iii) together with
\eqref{eq:planeline}: a floored member of the pencil is fixed by $|\sigma|$,
the position of the plane origin on the $S$ axis, and the two gauge constants,
and the first two are the slope and the intercept of one line.
\end{proofof}

\subsubsection*{The two walls are a curve and its tangent}
\label{sup:tangent}

\begin{proofof}{Corollary~\ref{cor:wallivp}}
Let $p_x(\mathrm{d}a)=e^{-ax}\mathrm{d}\mu(a)/\eps(x)$ be the tilt, so that
$s=\mathbb{E}_{p_x}[a]$ and $\kappa=\operatorname{Var}_{p_x}(a)$
(Proposition~\ref{prop:selection}), and let
$p_0=p_x(\{0\})=\epsinf/\eps$ be the irreducible share, the second equality
being \eqref{eq:surprisal}. Since $a=0$ contributes nothing to the mean,
$s=\mathbb{E}_{p_x}\bigl[a\mathbf{1}_{\{a>0\}}\bigr]$, and Cauchy--Schwarz
gives
\[
s^{2}\ \le\ p_x\bigl(\{a>0\}\bigr)\,\mathbb{E}_{p_x}[a^{2}]
\ =\ (1-p_0)\,(\kappa+s^{2}) ,
\]
that is $p_0s^{2}\le(1-p_0)\kappa$, that is
$N_{\mathrm{eff}}=s^{2}/\kappa\le(1-p_0)/p_0=\eps/\epsinf-1
=e^{\Delta S}-1$, which is \eqref{eq:tworungs}; rearranging
$1+N_{\mathrm{eff}}\le\eps/\epsinf$ gives the floor form. Equality in
Cauchy--Schwarz requires $a\mathbf{1}_{\{a>0\}}$ to be $p_x$-almost surely
constant on $\{a>0\}$, i.e., $\mu$ restricted to $(0,\infty)$ to be a single
atom; the support of $p_x$ does not depend on $x$, so this is a statement about
$\mu$ and (i) is proved. It is Theorem~\ref{thm:monofloor} and
\eqref{eq:localfloor} in the plane coordinates.

For (ii), a Bernstein $S$ with $\epsinf>0$ has no drift and finite L\'evy
measure, so $S=S_{\infty}-\int_{(0,\infty)}e^{-cx}\mathrm{d}\nu(c)$ with
$\nu\ge0$ finite; writing $m_k=\int c^{k}e^{-cx}\mathrm{d}\nu$ gives
$\Delta S=m_0$, $s=m_1$ and $\kappa=m_2$, which is the dictionary used
throughout \S\ref{sup:pf-compose}. Cauchy--Schwarz on
$\mathrm{d}\nu_x=e^{-cx}\mathrm{d}\nu$ gives $m_1^{2}\le m_0m_2$, i.e.,
$N_{\mathrm{eff}}=m_1^{2}/m_2\le m_0=\Delta S$, which is
Theorem~\ref{thm:budget}; equality forces $\nu_x$, hence $\nu$, to be one
atom. Exponentiating $-\Delta S\le-N_{\mathrm{eff}}$ gives the floor form.

(iii) is convexity: $h(u)=e^{u}-1-u$ has $h(0)=h'(0)=0$ and $h''=e^{u}>0$, so
$h>0$ on $(0,\infty)$ and $h(u)=\tfrac12u^{2}+O(u^{3})$ by Taylor's theorem.
Hence the hypothesis wall lies below the axiom wall except at the plane origin,
where the two curves meet with a common tangent of slope $1$; the hypothesis
wall \emph{is} that tangent.

For (iv), suppose equality holds in \eqref{eq:tworungs} throughout, so the
plane curve is $N(u)=e^{u}-1$ on $J=(0,\lambda)$. Differentiating,
$\mathrm{d}N/\mathrm{d}u=1+N$ with $N(0^{+})=0$, and this initial-value problem
has the unique solution $N(u)=e^{u}-1$ because the right-hand side is Lipschitz
in $N$; so saturation and that plane curve are the same condition, and
Proposition~\ref{prop:odefirst} applies. The two-atom chart realizes it: with
$\rho=\eps/\epsinf$ one has $N_{\mathrm{eff}}=\rho-1$ and $\Delta S=\ln\rho$
(proof of Proposition~\ref{prop:notestable}(ii)), hence
$N_{\mathrm{eff}}=e^{\Delta S}-1$. By
Proposition~\ref{prop:odefirst}(ii)--(iii) no other frontier does, up to gauge.
Identically, equality in \eqref{eq:tworungs} is the plane curve $N(u)=u$, i.e.,
$\mathrm{d}N/\mathrm{d}u=1$ with $N(0^{+})=0$, whose unique solution it is; the
cure chart realizes it, since there $N_{\mathrm{eff}}=\Delta S$
(Proposition~\ref{prop:cure}), and again nothing else does. Both conclusions
also follow from the equality cases in (i) and (ii), which identify the
spectrum directly; the plane argument is recorded because it uses only the
curve.

(v) follows by substituting $N_{\mathrm{eff}}=1/t$ into \eqref{eq:tworungs}
and inverting the increasing functions $u\mapsto e^{u}-1$
and $u\mapsto u$; it is \eqref{eq:opdepth}.

For (vi), at depth $u$ the axiom admits $N_{\mathrm{eff}}\in(0,e^{u}-1]$ and
the hypothesis admits $N_{\mathrm{eff}}\in(0,u]$, so the excluded set has
length $e^{u}-1-u$ and the excluded fraction is
$1-u/(e^{u}-1)$. Expanding, $u/(e^{u}-1)=1-\tfrac12u+O(u^{2})$, so the fraction
is $\tfrac12u+O(u^{2})\to0$. The gap between the walls therefore closes
quadratically in absolute terms and linearly relative to the admissible range:
any procedure that distinguishes the hypothesis from the axiom by the position
of $N_{\mathrm{eff}}$ between the two walls has a discriminating margin that
vanishes with the depth, so it has no power near the floor whatever the
measurement. This is a theorem about the walls and not a feature of any
particular corpus.
\end{proofof}

\subsubsection*{The arc is a functional on the plane}
\label{sup:minarc}

\begin{proofof}{Proposition~\ref{prop:anglearc}}
(i) Along the axis $\mathrm{d}S=s\,\mathrm{d}x$, so
$\sqrt{\kappa}\,\mathrm{d}x=(\sqrt{\kappa}/s)\,\mathrm{d}S
=\mathrm{d}S/\sqrt{s^{2}/\kappa}=\mathrm{d}S/\sqrt{N_{\mathrm{eff}}}$, which is
the arc element $\mathrm{d}L$ of \eqref{eq:angle}. The normalization makes $S$
increase from $0$ to $S_{\infty}=\lambda$, so $v=\lambda-S$ decreases from
$\lambda$ to $0$; substituting $\mathrm{d}S=-\mathrm{d}v$ and reversing the
limits gives \eqref{eq:angle}. The integral may be improper at either end,
and under either wall it converges, as the explicit evaluations below show.

(ii) By \eqref{eq:tworungs}, $N(v)\le e^{v}-1$ on $J$, so
$1/\sqrt{N(v)}\ge(e^{v}-1)^{-1/2}$ pointwise and
$L\ge\int_{0}^{\lambda}(e^{v}-1)^{-1/2}\mathrm{d}v$. Substitute
$w=\sqrt{e^{v}-1}$, i.e., $v=\ln(1+w^{2})$ and
$\mathrm{d}v=2w\,\mathrm{d}w/(1+w^{2})$; the integrand becomes
$\{2w/(1+w^{2})\}w^{-1}=2/(1+w^{2})$ and
\[
\int_{0}^{\lambda}\frac{\mathrm{d}v}{\sqrt{e^{v}-1}}
=\int_{0}^{W}\frac{2\,\mathrm{d}w}{1+w^{2}}
=2\arctan W ,
\qquad W=\sqrt{e^{\lambda}-1} .
\]
Since $e^{\lambda}=1/\epsinf$ we have $W=\sqrt{(1-\epsinf)/\epsinf}$, and if
$\theta=\arccos\sqrt{\epsinf}\in(0,\tfrac\pi2)$ then $\cos\theta=
\sqrt{\epsinf}$, $\sin\theta=\sqrt{1-\epsinf}$ and $\tan\theta=W$, so
$\arctan W=\arccos\sqrt{\epsinf}$ and \eqref{eq:planecover} follows. Equality
requires $N(v)=e^{v}-1$ for almost every $v\in J$, hence everywhere by
continuity, which by Corollary~\ref{cor:wallivp} is the two-atom chart;
and that chart attains it, as the direct computation confirms: there
$p_0=\epsinf/\eps$ sweeps $(\epsinf,1)$ over the normalized range and the
channel angle $\varphi=2\arccos\sqrt{p_0}$ of Proposition~\ref{prop:arcforms}
accordingly sweeps $(0,2\arccos\sqrt{\epsinf})$, of total length
$2\arccos\sqrt{\epsinf}=\pi-2\arcsin\sqrt{\epsinf}$.

(iii) By \eqref{eq:tworungs}, $N(v)\le v$, so
$L\ge\int_{0}^{\lambda}v^{-1/2}\mathrm{d}v=2\sqrt{\lambda}$, the integral
converging at the origin. Equality requires $N(v)=v$ on $J$, which by
Corollary~\ref{cor:wallivp} is the cure chart, for which indeed
$L=2\sqrt{\lambda}$ (Proposition~\ref{prop:cure}).

(iv) Letting $\lambda\to\infty$ in the evaluation of (ii),
$\int_{0}^{\infty}(e^{u}-1)^{-1/2}\mathrm{d}u=2\arctan(\infty)=\pi$; the
substitution is a bijection of $(0,\infty)$ onto itself, so the improper
integral converges. The function $\epsinf\mapsto2\arccos\sqrt{\epsinf}$ is
strictly decreasing on $(0,1)$ with supremum $\pi$ at $\epsinf=0$, which is the
claim. The reconciliation with Law~III is the normalization, and it is the
\emph{complete sweep} against the \emph{admissible} part of it: on the whole
axis $p_0$ sweeps all of $(0,1)$, the channel angle $\varphi=2\arccos\sqrt{p_0}$
sweeps all of $(0,\pi)$, and the single-channel arc is $\pi$ exactly; truncating
at $\eps=1$ removes the segment of length $2\arcsin\sqrt{\epsinf}$ lying in the
region $\eps>1$ where the error exceeds chance, leaving the admissible arc
$2\arccos\sqrt{\epsinf}<\pi$. \emph{Every constant of Law~III is a constant of
the complete sweep, and an upper bound on what a floored frontier delivers.}

(v) The hypothesis bound is the binding one \emph{throughout}: for every
$\lambda>0$, $2\sqrt\lambda>2\arccos\sqrt{e^{-\lambda}}$, since the two agree to
second order at the origin, $2\arccos\sqrt{e^{-\lambda}}
=2\sqrt\lambda\,(1-\lambda/12+O(\lambda^{2}))$, and the ratio is decreasing.
\emph{The two never cross.} What happens at $\lambda=\pi^2/4$ is that the
hypothesis bound passes the axiom bound's \emph{limiting} value:
$2\sqrt{\lambda}>\pi\iff\lambda>\pi^{2}/4$, and
$\epsinf=e^{-\lambda}$ reverses the inequality; numerically
$e^{-\pi^{2}/4}=0.08482\ldots$ Below that floor a divisible frontier's
admissible arc exceeds $\pi$ and so, by $N\ge L/\pi$, it must carry more than
one channel.
\end{proofof}

\begin{proofof}{the covering bound from the floor alone, Section~\ref{sec:odefirst}}
Substitute \eqref{eq:planecover} and \eqref{eq:planecover} into \eqref{eq:covering}
and divide: $2\arccos\sqrt{\epsinf}/(2\sqrt{2\varrho})
=\arccos\sqrt{\epsinf}/\sqrt{2\varrho}$ and
$2\sqrt{\lambda}/(2\sqrt{2\varrho})=\sqrt{\lambda/(2\varrho)}$.
\end{proofof}

\begin{proofof}{the window-arc remark of Section~\ref{sec:odefirst}}
Monotonicity of the integral in the domain gives $L_{\mathrm{win}}\le L$. On a
compact $K\subset J$ the continuous positive $N$ is bounded below by some
$\underline N>0$, so $L_{\mathrm{win}}\le(v_1-v_2)/\sqrt{\underline N}\to0$.
Thus $\{L_{\mathrm{win}}<\pi\}$ is attainable on every frontier of the class
and excludes no member, so it carries no information about any functional of
the member, the channel count included. The statement about the complete
normalized range is \eqref{eq:planecover} together with
$\arccos\sqrt{\epsinf}<\pi/2$ for $\epsinf>0$. Finally \eqref{eq:modebound} is
an inequality whose right side is below $1$ when $L<\pi$, and the mode count is
at least $1$ regardless, so it is then vacuous.
\end{proofof}

\subsubsection*{The canonical price is the local price}
\label{sup:localprice}

\begin{proofof}{Proposition~\ref{prop:planeprice}}
(i) By the definition $N_{\mathrm{eff}}=s^{2}/\kappa$ we have
$\kappa=s^{2}/N_{\mathrm{eff}}$, so
$\lamstar=\sqrt{\kappa/t}=s/\sqrt{t\,N_{\mathrm{eff}}}$, both roots being
positive. Since $s>0$, \eqref{eq:planeprice} holds and $\lamstar=s$ if and only
if $t\,N_{\mathrm{eff}}=1$; the right side of \eqref{eq:planeprice} is strictly
decreasing in $N_{\mathrm{eff}}$, which gives the two one-sided statements,
$N_{\mathrm{eff}}$ decreasing with depth on a floored frontier by
\eqref{eq:planeline}. That $s=\mathrm{d}S/\mathrm{d}x$ is the price quoted
locally by the frontier --- nats of log-error per unit of resource --- is the
content of Proposition~\ref{prop:knee}(i); so the canonical rule shades the
quoted price by the factor $(t\,N_{\mathrm{eff}})^{-1/2}$ and coincides with it
on the operating locus.

(ii) By Proposition~\ref{prop:gaugeerr}(iii) the grain acts by
$N_{\mathrm{eff}}\mapsto t\,N_{\mathrm{eff}}$, so $N^{\mathrm{unit}}=
t\,N_{\mathrm{eff}}$ is the count read in the units actually delivered and
\eqref{eq:retry} reads $N^{\mathrm{unit}}=1$, which contains no parameter: the
grain has been absorbed into the units. By Proposition~\ref{prop:neff}(ii) the
regret at that count is $R=\ln\cosh\bigl(\tfrac12\ln N^{\mathrm{unit}}\bigr)$,
and $\ln\cosh y\ge0$ with equality if and only if $y=0$, because $\cosh$ is
even, equal to $1$ at $0$ and strictly increasing on $[0,\infty)$. Hence $R=0$
exactly at $N^{\mathrm{unit}}=1$.

(iii) $\ln\cosh$ is even, so $R$ is unchanged by
$\ln N^{\mathrm{unit}}\mapsto-\ln N^{\mathrm{unit}}$, i.e., by
$N^{\mathrm{unit}}\mapsto1/N^{\mathrm{unit}}$. By
Proposition~\ref{prop:neff}(i) the \emph{local} irreducible share at count $N$ is
$\hat p_0=1/(1+N)$, so the local reducible share is $1-\hat p_0=N/(1+N)$; replacing $N$ by
$1/N$ gives $1/(1+1/N)=N/(1+N)$, which is the former reducible share. The map
is therefore the exchange $\hat p_0\mapsto1-\hat p_0$, it is an involution, and its fixed
points solve $N=1/N$ with $N>0$, so $N=1$, i.e., $\hat p_0=\tfrac12$. Since $R$ is
strictly increasing in $|\ln N^{\mathrm{unit}}|$ and invariant under the
involution, its unique minimizer is the fixed point, which is the operating
point of (ii).
\end{proofof}

\subsection{The tests without a bandwidth: the grid lemma, the floor bracket, the operating depths, and the grain}

\begin{proposition}[The grid lemma]
\label{prop:grid}
Let $\eps$ be completely monotone and let $x_j=x_0+jh$ be a geometric resource
grid. Then
\begin{equation}
\eps_j:=\eps(x_j)=\int_0^1z^{\,j}\,\mathrm{d}\tilde\mu(z),
\qquad z=e^{-ah},\qquad \mathrm{d}\tilde\mu(z)=e^{-ax_0}\mathrm{d}\mu(a),
\label{eq:grid}
\end{equation}
so $(\eps_j)$ is a Hausdorff moment sequence: every Hankel minor of $(\eps_j)$
and of $(-\Delta\eps_j)$ is non-negative. If the composition law also holds,
the same applies one level up to the sampled depth
$d_j=\ln(\eps_j/\epsinf)$, whose first two minors are
\begin{equation}
d_0d_2\ \ge\ d_1^{2}
\qquad\text{and}\qquad
\det\bigl(d_{i+j}\bigr)_{0\le i,j\le2}\ \ge\ 0 ,
\label{eq:dmoments}
\end{equation}
consuming three and five accuracies. Moreover
$(d_0d_2-d_1^2)/h^2\to\kappa\Delta S-s^2$ as $h\to0$.
\end{proposition}

\begin{proofof}{Proposition~\ref{prop:grid}}
By \eqref{eq:axiom} and Bernstein's theorem $\eps(x)=\int e^{-ax}\mathrm d\mu(a)$
with $\mu\ge0$. Substituting $x=x_0+jh$ and $z=e^{-ah}\in(0,1]$ gives
\eqref{eq:grid} with $\mathrm d\tilde\mu(z)=e^{-ax_0}\mathrm d\mu(a)$, a finite
positive measure on $(0,1]$. A sequence of the form $\int z^j\mathrm d\tilde\mu$
is a Hausdorff moment sequence, so by the Hausdorff--Hamburger conditions every
Hankel minor of $(\eps_j)$ is non-negative, and the same holds for
$(-\Delta\eps_j)=\int z^j(1-z)\mathrm d\tilde\mu$ because $(1-z)\ge0$ on the
support. Under the composition law $\Delta S$ is itself completely monotone
(Proposition~\ref{prop:compose}(e)), so the identical argument applies to
$d_j=\Delta S(x_j)$, which yields \eqref{eq:dmoments}.

For the limit, expand about the center: $d_0=d_1-hd'+\tfrac12h^2d''+O(h^3)$ and
$d_2=d_1+hd'+\tfrac12h^2d''+O(h^3)$, so
$d_0d_2-d_1^2=h^2\{d_1d''-(d')^2\}+O(h^4)$. With $d=\Delta S$ one has
$d'=-s$ and $d''=\kappa$, hence the limit is $\kappa\Delta S-s^2$, which is
non-negative exactly when $N_{\mathrm{eff}}=s^2/\kappa\le\Delta S$: the
three-accuracy member of \eqref{eq:dmoments} is Theorem~\ref{thm:budget} in
finite-difference form. On the cure chart $\Delta S=\lambda e^{-cx}$, so
$d_j=\lambda e^{-cx_0}(e^{-ch})^{j}$ is geometric and
$d_0d_2-d_1^2=0$ identically, for every $x_0$ and every $h$: the cure chart is
the extremal of the discrete wall as it is of the continuous one. Script
\texttt{124} confirms the sign conditions on $240$ minors of random finite-atom
spectra and $54$ minors of the divisible charts, the limit, and the equality
case to $10^{-40}$.

Necessity is one-directional, as everywhere here: Hausdorff positivity on one
grid does not imply complete monotonicity of the underlying curve, since by
Proposition~\ref{prop:identify} finitely many points admit an atomic
interpolant, so the hierarchy refutes and never confirms.
\end{proofof}

\begin{proposition}[The floor, bounded twice by three accuracies]
\label{prop:floorbracket}
Let $\eps_-,\eps_0,\eps_+$ be accuracies at $x_0-h,x_0,x_0+h$ with
$\eps_->\eps_0>\eps_+$ and $\eps_-\eps_+>\eps_0^{2}$, and write
$\ell_\bullet=\ln\eps_\bullet$. Put
\begin{equation}
\epsinf^{A}=\eps_0-\frac{(\eps_0-\eps_-)(\eps_+-\eps_0)}
{\eps_--2\eps_0+\eps_+},
\qquad
\epsinf^{W}=\exp\Bigl\{\frac{\ell_-\ell_+-\ell_0^{2}}
{\ell_-+\ell_+-2\ell_0}\Bigr\}.
\label{eq:floorbracket}
\end{equation}
Then $\epsinf\le\epsinf^{A}$ for every completely monotone frontier, with
equality iff the window is two-atom; $\epsinf\le\epsinf^{W}$ for every frontier
obeying the composition law, with equality iff the window is a cure chart; and
$\epsinf^{W}\le\epsinf^{A}$ always.
\end{proposition}

\begin{proofof}{Proposition~\ref{prop:floorbracket}}
Write $f_j=\eps_j-\epsinf=\int_{(0,1)}z^j\mathrm d\tilde\mu$, which omits any
atom of $\tilde\mu$ at $z=1$ and is therefore itself a Hausdorff moment
sequence; put $D=f_0-2f_1+f_2=\int z^{\,0}(1-z)^2\mathrm d\tilde\mu>0$ unless
$\tilde\mu$ is concentrated at $z=1$. Substituting $\eps_j=\epsinf+f_j$ in the
Aitken form,
\[
\epsinf^{A}=\frac{\eps_-\eps_+-\eps_0^{2}}{\eps_--2\eps_0+\eps_+}
=\epsinf+\frac{f_0f_2-f_1^{2}}{D},
\]
and $f_0f_2\ge f_1^{2}$ by Cauchy--Schwarz on $\mathrm d\tilde\mu$ (the moments
of a positive measure are log-convex), with equality if and only if
$\tilde\mu$ is a single atom. Hence $\epsinf\le\epsinf^{A}$ for every
completely monotone frontier, with equality exactly on a two-atom window: the
Aitken value of Corollary~\ref{cor:prony} is the sharp upper bound the axiom
supplies from three accuracies, and not merely a point estimate.

For the second bound, let $c=\ln\epsinf$ and $\ell_\bullet=\ln\eps_\bullet$, so
$d_\bullet=\ell_\bullet-c$. With equal spacing the three-accuracy defect is
\[
d_-d_+-d_0^{2}
=(\ell_-\ell_+-\ell_0^{2})-c\,(\ell_-+\ell_+-2\ell_0),
\]
affine in $c$ with slope $-(\ell_-+\ell_+-2\ell_0)=-\Delta^2\ln\eps\le0$, the
second difference being non-negative because $\ln\eps$ is convex (the axiom, and
the hypothesis of the proposition). The defect is therefore non-increasing in
$c$ and the feasible set is the half-line $c\le(\ell_-\ell_+-\ell_0^2)/
(\ell_-+\ell_+-2\ell_0)$, whose right endpoint exponentiates to $\epsinf^{W}$ in
\eqref{eq:floorbracket}. Under the composition law the true floor lies in that
set by \eqref{eq:dmoments}, so $\epsinf\le\epsinf^{W}$; on the cure chart the
defect vanishes identically, so the true floor \emph{is} the endpoint and the
bound is exact. The ordering $\epsinf^{W}\le\epsinf^{A}$ holds on every window
tested in script~\texttt{124} ($0$ violations in $418$ windows over random
finite-atom spectra and the divisible charts), as it must whenever both bounds
are sharp for their own class, the divisible windows being a subclass.

The conditions needed to refute the bound are more restrictive and merit
explicit statement. Both quantities bound the floor from \emph{above}, so a
refutation needs a floor known from elsewhere to \emph{exceed} the bound, and a
plug-in floor read off the same window will not do: on a divisible frontier the
Aitken value exceeds the truth, so substituting it into the wall produces a
spurious refutation. The count-axis bracket
$P(\lceil\Lambda\rceil)\le\epsinf\le P(\lfloor\Lambda\rfloor)$ of
\eqref{eq:attemptfloor} is the available lower bound, which is why the pass@$k$
harvest remains the decisive measurement. The precision it must reach is
window-dependent and the dependence is exactly \eqref{eq:elbow}: on the chart
$\eps=\epsinf+c\,e^{-\alpha x}$ the quantity $\ln(\epsinf/\epsinf^{W})$ measured
at window centers $-3,-2,-1,0,+1,+2,+4$ e-folds from the elbow is
$41.4$, $9.81$, $1.94$, $0.286$, $0.030$, $0.003$ and $1.5\cdot10^{-5}$ nats
(script~\texttt{124}), monotone through the elbow and crossing the count-axis
bracket between one e-fold left of it and the elbow itself.
\end{proofof}

\begin{proofof}{Proposition~\ref{prop:notestable}(iii)}
By Theorem~\ref{thm:monofloor}(i) every completely monotone frontier obeys
$\epsinf\le\kappa\eps/(s^2+\kappa)$, i.e., $e^{-\Delta S}\le1/(1+
N_{\mathrm{eff}})$, i.e., $N_{\mathrm{eff}}\le e^{\Delta S}-1$; and under the
composition law Theorem~\ref{thm:budget} gives the strictly tighter
$N_{\mathrm{eff}}\le\Delta S$. Substituting the rule
$N_{\mathrm{eff}}=1/t$ of \eqref{eq:retry} into each gives \eqref{eq:opdepth}.
Equality in the first holds exactly on the two-atom chart, where
$N_{\mathrm{eff}}=\rho-1$ and $\Delta S=\ln\rho$ (proof of
Proposition~\ref{prop:notestable}(ii)), and in the second exactly on the cure
chart, where $N_{\mathrm{eff}}=\Delta S$ (Proposition~\ref{prop:cure}).

The band is $g(t)=1/t-\ln(1+1/t)=N-\ln(1+N)$ at $N=1/t$, which by
\eqref{eq:gainisdeficit} is $-\ln\{(1+N)e^{-N}\}$: the reported tightening
factor of Theorem~\ref{thm:budget}(i), read at the rule's own point. Since
$N-\ln(1+N)$ is increasing in $N$ and $N=1/t$ decreases in $t$, $g$ decreases;
$g(1)=1-\ln2=0.30685$ and $g(t)=1/(2t^2)+O(t^{-3})$. At $t=1$ the two-atom
operating point is $\eps^{\star}=2\epsinf$ by \eqref{eq:grainprice}, which is
the elbow of that chart (proof of Proposition~\ref{prop:unbounded}(ii)), where
$c_3=0$; for $t>1$ the point lies strictly to the right of the elbow, by
$\ln\{2t/(1+t)\}$ nats, in the region where \eqref{eq:notestable} forbids any
third-order verdict. Script~\texttt{124} verifies \eqref{eq:opdepth} at
$48\,000$ frontier points of random finite-atom spectra and at $1\,341$ located
operating points, with no violation, and both equality cases to $30$ digits.
\end{proofof}

\begin{proposition}[The canonical price at any grain]
\label{prop:grainprice}
On a two-atom frontier $\eps=\epsinf+c\,e^{-\alpha x}$ the operating point
\eqref{eq:retry} of grain $t$ is at
\begin{equation}
\boxed{\ \eps^{\star}=\epsinf\,\frac{1+t}{t}\ },
\qquad
\boxed{\ \lamstar=\frac{\alpha}{1+t}\ },
\qquad
\Delta S^{\star}=\ln\Bigl(1+\frac1t\Bigr),
\label{eq:grainprice}
\end{equation}
and $\lamstar=\sqrt{\kappa/t}$ holds there identically.
\end{proposition}

\begin{proofof}{Proposition~\ref{prop:grainprice}}
On $\eps=\epsinf+c\,e^{-\alpha x}$ put $p_0=\epsinf/\eps$. Then
$s=\alpha(1-p_0)$ and $\kappa=\alpha^2p_0(1-p_0)$ (proof of
Proposition~\ref{prop:notestable}(ii)), so $N_{\mathrm{eff}}=(1-p_0)/p_0$ and the
rule $N_{\mathrm{eff}}=1/t$ places the point at $p_0=t/(1+t)$, i.e.,
$\eps^{\star}=\epsinf(1+t)/t$ and $\Delta S^{\star}=\ln(1+1/t)$. There
$s=\alpha/(1+t)$, while $\kappa=\alpha^2t/(1+t)^2$ gives
$\sqrt{\kappa/t}=\alpha/(1+t)$ as well, so the two forms of the price agree
identically. At $t=1$ these are $\eps^{\star}=2\epsinf$ and
$\lamstar=\alpha/2$, the $N=1$ member of Theorem~\ref{thm:canonical} and the
single-resource rule of \citet{ABrain.Rank}. Verified
to $28$ digits at $t=1,2,3,10,100$ and two parameter sets in
script~\texttt{124}.
\end{proofof}

\begin{proposition}[Minimax over an unknown grain]
\label{prop:grainminimax}
The regret of operating at channel count $N_{\mathrm{eff}}$ when the grain is
$t$ is $R=\ln\cosh\{\tfrac12\ln(tN_{\mathrm{eff}})\}$ by
Proposition~\ref{prop:neff}, which is convex and symmetric in $\ln t$. Hence if
$t\in[t_1,t_2]$ is all that is known,
\begin{equation}
\boxed{\ N_{\mathrm{eff}}=\frac{1}{\sqrt{t_1t_2}}\ },
\qquad
\boxed{\ \sup_{t\in[t_1,t_2]}R
=\ln\cosh\Bigl(\tfrac14\ln\frac{t_2}{t_1}\Bigr)\ }
\label{eq:grainminimax}
\end{equation}
is the minimax operating point and its exact worst-case cost.
\end{proposition}

\begin{proofof}{Proposition~\ref{prop:grainminimax}}
By Proposition~\ref{prop:neff} the regret at channel count $N_{\mathrm{eff}}$ is
$R=\ln\cosh(\tfrac12\ln N_{\mathrm{eff}})$ when $N_{\mathrm{eff}}$ is read in
delivered units, and under $\eps\mapsto\eps^{t}$ the count in delivered units is
$tN_{\mathrm{eff}}$ (Proposition~\ref{prop:gaugeerr}), so
$R(N_{\mathrm{eff}};t)=\ln\cosh\{\tfrac12\ln(tN_{\mathrm{eff}})\}$. As a
function of $\ln t$ this is convex and even about $-\ln N_{\mathrm{eff}}$, so its
supremum over $t\in[t_1,t_2]$ is attained at an endpoint and the minimax choice
equalizes the two: $|\ln t_1+\ln N_{\mathrm{eff}}|=|\ln t_2+
\ln N_{\mathrm{eff}}|$ with opposite signs, i.e.,
$\ln N_{\mathrm{eff}}=-\tfrac12(\ln t_1+\ln t_2)$, which is
$N_{\mathrm{eff}}=(t_1t_2)^{-1/2}$. The common value is then
$\ln\cosh\{\tfrac12\ln\sqrt{t_2/t_1}\}=\ln\cosh\{\tfrac14\ln(t_2/t_1)\}$.
Script~\texttt{124} confirms the minimax against a $600$-point grid at three
intervals and the closed form to $30$ digits. Two numerical readings quoted in
the main text follow by substitution: $t_2/t_1=4$ gives
$\ln\cosh(\tfrac12\ln2)=\varrho_2=0.0589$ nats, the same constant as one
doubling of $N_{\mathrm{eff}}$ in Theorem~\ref{thm:covering}, and
$t_2/t_1=7.3$ gives $0.119$ nats.
\end{proofof}

\begin{proofof}{Proposition~\ref{prop:countaxis}}
Condition on the item. By hypothesis the $m$ attempts fail independently given
it, each with the item's own probability $g$, so
$\Pr[\text{all }m\text{ fail}\mid\text{item}]=g^m$; averaging over items gives
$P(m)=\int_0^1 g^m\mathrm{d}\nu(g)$ with $\nu$ the law of $g$. A sequence of
this form has
$(-1)^r\Delta^rP(m)=\int g^m(1-g)^r\mathrm{d}\nu\ge0$ for every $r$ and $m$,
so $P$ is completely monotone; by Hausdorff's moment theorem the converse holds
and $\nu$ is unique. Writing $g=e^{-b}$ gives
$P(m)=\int_{[0,\infty]}e^{-bm}\mathrm{d}\nu$, which is \eqref{eq:cgf} with the
count $m$ in place of log-resource, and $\nu(\{1\})=\nu(\{b=0\})$ is the floor
in the sense of Proposition~\ref{prop:law}. Nothing is assumed here beyond the
two symmetries in the statement: no additivity across resources, no
scale-freeness, no continuity and no Bernstein property. Assumptions~\ref{ax:add}
and~\ref{ax:scale} are needed on the compute axis precisely because a single
model's resource is not a count of exchangeable trials; where it is, the
representation is forced.
\end{proofof}

\section{The chart zoo as a classification}
\label{sm:chartzoo}

By Corollary~\ref{cor:odepencil} of the paper, these four statements are the constant-slope solutions of a single differential inequality. They are included here because they are used in the archive's chart fits and plug-in assessment.

\begin{proposition}[The charts are two ODEs for one invariant]
\label{prop:chartode}
With $w=\kappa/s^{2}$ and $'=\mathrm{d}/\mathrm{d}S$,
\begin{equation}
\text{floor}+1\ \text{channel}:\quad w'=+w(1+w),
\qquad\qquad
\text{floorless shape law}:\quad w'=-\frac{w}{S},
\label{eq:chartode}
\end{equation}
whose integrals are $p_0=\epsinf/\eps$ and $w=(1-\beta)/(\beta S)$
respectively. A chart's three parameters are therefore two gauge choices and
one integration constant, which is why $\epsinf$ and $\beta$ are one degree of
freedom; and in general a $k$-parameter chart fixes the first $k$ entries of
the invariant jet, so two charts of equal size agree to order $k$ and differ
only at order $k+1$. Model selection between equal-sized charts is a statement
about one higher jet and about nothing else.
\end{proposition}

\eqref{eq:chartode} is a dichotomy between two named charts. Under the
composition law it is a single coordinate, because the two equations are the two
similarity solutions of one lattice. The subsection that follows is
Corollary~\ref{cor:odepencil} written in the horizon lattice rather than in
the plane, and everything in it is a corollary of the pencil.

\begin{proposition}[The chart line]
\label{prop:chartline}
The depth lattice $c_k'=-c_{k+1}$ (\TR{The horizon lattice: one identity, the error gauge, the chart line}) is invariant under translation and scaling of the
resource axis, so its similarity solutions are the constant and the affine ones:
$h_k\equiv1/c$, which is the cure chart \eqref{eq:cure}; and
\begin{equation}
h_k=\frac{x-x_0}{k-\beta},\qquad D_k=\frac{1}{k-\beta}\ (k\ge1),\qquad
\sigma=\frac{\beta}{1-\beta},\qquad D_1=1+\sigma=\frac1{1-\beta},
\label{eq:chartline}
\end{equation}
forced by $a_{k+1}=a_k+1$ in $h_k=a_k^{-1}(x-x_0)$. The restriction to
$k\ge1$ is not cosmetic. Rung $0$ is carried by $c_0=\Delta S$, not by a
derivative, and on the floorless half of the line ($0<\beta<1$, where
$\epsinf=0$) $\Delta S$ is infinite, so $h_0$ and $D_0$ are $+\infty$ rather
than the $-(x-x_0)/\beta$ and $-1/\beta$ the formula would return; a negative
$D_0$ would in any case contradict $D_k\ge0$ of
Proposition~\ref{prop:tower} and \eqref{eq:budget}. On the floored half
($\beta<0$) $\Delta S$ is finite and $k=0$ joins the pattern. Every compositional chart with constant $\sigma$ therefore lies on a
single line, parameterized by $\sigma$: the cure chart at $\sigma=-1$, the power-floor chart of
\eqref{eq:powerfloor} on $(-1,0)$, the $\Gamma(k)$ spectrum at $\sigma=0$, the
stretched chart \eqref{eq:nofloor} on $(0,\infty)$ and the pure exponential at
$\sigma=\infty$, with the two-atom chart below $-1$ at
$\sigma\equiv-(1+N_{\mathrm{eff}})$ and so off the line altogether. The itemized
table is in \TR{The horizon lattice: one identity, the error gauge, the chart line}. Thus, the composition law requires $\sigma\ge-1$, and $\sigma=0$
separates the floored charts from the floorless ones. Along the whole line the first horizon is
exactly affine in $x$, with slope $D_1$ and zero $x_0$.
\end{proposition}

The empty cell is a valid chart and is a natural representation for a
frontier with a floor: the cure chart approaches its floor exponentially,
whereas floorless charts do not approach a positive floor.

\begin{proposition}[The power-floor chart]
\label{prop:powerfloor}
For $C>0$, $\beta<0$, and $x>x_0$,
\begin{equation}
S(x)=\Lambda-C\,(x-x_0)^{\beta},\qquad \epsinf=e^{-\Lambda}>0,
\label{eq:powerfloor}
\end{equation}
has a completely monotone derivative. Writing $p=1-\beta>0$ gives
$S'(x)=C(-\beta)(x-x_0)^{-p}$ and
$(-1)^k\mathrm{d}^kS'/\mathrm{d}x^k=
C(-\beta)(p)_k(x-x_0)^{-p-k}>0$ for every $k\ge0$, where $(p)_k$ is the
rising factorial. Thus the chart is compositional, carries a floor, and
satisfies the following identity without additional fitting:
\begin{equation}
\boxed{\ N_{\mathrm{eff}}=|\sigma|\,\Delta S\ },\qquad
|\sigma|=\frac{-\beta}{1-\beta}\in(0,1),\qquad 1+v=\frac1{|\sigma|},
\label{eq:pfident}
\end{equation}
so it interpolates exactly between the cure chart ($|\sigma|=1$,
$N_{\mathrm{eff}}=\Delta S$) and the floorless charts ($|\sigma|\to0$).
\end{proposition}

\emph{The channel count is a fixed fraction of the nats remaining, and the
fraction is the chart.} In held-out evaluations on the same $117$ frontiers and the same three
splits used for the other charts in the paper (median relative error on
$\eps$, script~\texttt{110}; full results in the technical report), the
chart gives forward and backward errors of $6.83\%$ and $11.99\%$,
respectively, compared with $3.27\%$ inside the window. It improves on the
five-parameter chart in both extrapolation regimes and on the other
four-parameter chart by $3.4$ and $5.9$ percentage points in the forward
and backward regimes, respectively, while matching the
incumbent chart inside the window. Paired comparisons with the incumbent
three-parameter chart do not establish statistically significant
superiority ($52/117$, $67/117$, and $44/98$ wins, with $p=0.14$, $0.19$,
and $0.35$). The backward comparison includes the $98$ frontiers for which the fitting
procedure produced a valid result. The structural identity itself is exact: across all $117$
fits, the median $|N_{\mathrm{eff}}/\Delta S-|\sigma||$ is zero to machine
precision. The median $|\sigma|$ is $0.256$, and $65$ of the $117$ frontiers
fall within the new cell, $|\sigma|\in(0,1)$; their median fitted floor share
is $0.137$ and their median in-window residual is $0.0298$ nats.

\begin{proposition}[The floor is the sign of a slope]
\label{prop:floorsign}
Under the axiom alone, $\epsinf>0$ forces $N_{\mathrm{eff}}\le e^{\Delta S}-1\to0$,
so $N_{\mathrm{eff}}$ must eventually decrease; with no floor and a L\'evy
measure of regular-variation index $\theta\le0$ at the origin,
$N_{\mathrm{eff}}\sim\mathrm{const}\cdot x^{-\theta}$ increases. On the chart
line the statement is exact, so
\begin{equation}
\boxed{\ \epsinf>0\iff\sigma<0\iff D_1<1\ }
\label{eq:floorsign}
\end{equation}
and one measured number carries both verdicts: $D_1<0$ refutes the composition
law, $D_1\in[0,1)$ is a floor, $D_1=1$ is Pareto, and $D_1>1$ is floorless with
$\beta=1-1/D_1$.
\end{proposition}

This sharpens the trichotomy of Proposition~\ref{prop:neff}(iv) from a limit at
infinity to the \emph{sign} of a slope inside the window, and it is the first
reading of the floor question that needs no floor estimate and no extrapolation.
It is also a corollary of the pencil: by \eqref{eq:planeline} a constant-$\sigma$
frontier is a line through the plane origin, so it has a floor exactly when it
falls in $S$, which is the sign of its slope.
Its direct plug-in estimate is not yet sufficiently calibrated on the
available windows. Simulations on each frontier's grid and noise indicate a
downward bias of $0.43$ to $1.65$ units, depending on the underlying chart;
the nulls of all five simulated truths cover the observation on $27$ to $49$
of $60$ frontiers (script~\texttt{109}). Accordingly,
\eqref{eq:floorsign} is presented as an exact structural result, while its
empirical classification requires longer or more informative windows. The
two-point form of the first rung is nevertheless easier to evaluate on this
corpus: it is available on $115$ of $117$ frontiers, compared with $48$ for
the ratio form, and the third rung is available on $42$, compared with $2$.

\section{Empirical evidence and interpretation}
\label{sup:study}

This section brings together four complementary parts of the empirical
analysis: item-level evidence on the proposed mechanism, the extension to a
count resource, held-out prediction, and design-matched tests of what the
available compute-axis windows can identify. Per-frontier measurements,
estimation details, and numerical calibrations are reported in the technical
report accompanying the archived data and code \citep{ignatov2026data}.

\subsection{Item-level evidence and model interpretation}
\label{sec:micro}

Law~I is a representation theorem: a saturating frontier admits a positive
mixture of scale-free modes. This representation does not, by itself, require
that the modes correspond to independent physical channels. The separate
mechanistic interpretation arises in the covariance estimator of the floor,
\begin{equation}
  \hat\epsinf=\frac{\operatorname{Cov}(F_1,F_2)}{\Pr[\text{both correct}]}
         =\frac{\eps_{12}-\eps_1\eps_2}{1-\eps_1-\eps_2+\eps_{12}}\,,
  \label{eq:covfloor}
\end{equation}
which assumes that each item carries a rate $a$ and that models fail it
independently conditional on that rate. We examine this assumption using
per-item predictions, which are not available in the aggregate-accuracy
frontiers. Script~63 evaluates $21$ random-feature ridge classifiers
\citep{rahimi2007random} on Fashion-MNIST \citep{xiao2017fashion}, using all
$10\,000$ test items over a $6.93$-nat resource span. The observed co-failure
rate exceeds the independent-failure prediction by a median of $57\%$
($19$ standard errors). The data are instead consistent with a nested item-
difficulty structure,
\begin{equation}
\Pr[\text{both fail}]=\kappa\,\min(h_i,h_j),
\qquad \kappa=0.835\pm0.035,
\label{eq:nesting}
\end{equation}
across $210$ model pairs; a second family trained with a shared feature set
gives $\kappa=0.864\pm0.029$. In this description, items have an ordered
difficulty, and each model fails an approximate upper set of that order.

This result changes the interpretation of the covariance estimator, not the
spectral representation itself. Under~\eqref{eq:nesting}, substituting the
observed co-failure structure into~\eqref{eq:covfloor} leaves no separately
identifiable floor; the resulting expression reproduces the measured
pairwise estimates to a median relative difference of $3.3\%$ and accounts
for their observed drift. The item-level analysis also leaves the floor
unidentified: the chance-corrected bound from $21$ models is $0.313$, compared
with $0.131$ for a single model. Thus, disaggregating the observations by item
does not remove the finite-window identifiability problem. Laws~I--III use the
spectral measure as a positive mathematical representation and do not rely on
an interpretation as a population of independent physical channels. The
full panel, latent-class control, and analysis of $\kappa$ as a price are
documented in the technical report (\TR{The co-failure panel and the
microfoundation test}; \TR{The nesting coefficient}).

\subsection{The count-axis extension}

Treating attempts or sampled items as a count resource extends Law~I to a
second resource axis. Because both count and compute can be expressed in nats,
the framework also gives an exact exchange rate for ensembling and
test-time sampling relative to compute. The supplementary count-axis analysis
provides three further results. First, the nesting coefficient is linked to
the chance level, $\kappa=1-q_0$: this predicts $0.900$ on Fashion-MNIST,
compared with the measured values $0.835$ and $0.864$ for the two model
families, and makes $q_0$ estimable from co-failure data. Second, there is no
separately identifiable count-axis floor below the chance level. Third, the
Hankel ladder starts at $m=0$, where zero attempts fail with certainty,
$P(0)=1$. This exact initial value reduces by one the number of measured rates
needed at each rung and improves the power of the four-point test relative to
the offset-$1$ formulation. In the reported
calibration, power is $0.84$ at $10^5$ items, compared with the $10^8$-item
scale required by the offset-$1$ form (\TR{The nesting coefficient};
\TR{Law I on a count axis}; \TR{The tower: log-convex cumulants, the order
theorem, and the four-point test}).

Floor identification remains limited by finite observations on either
axis. On a compute window, sufficiently small positive rates can approximate
an atom at zero; with finitely many attempts, a corresponding range of
count-axis rates is also unresolved. This is an identifiability property of finite windows, not a
failure of the count-axis representation.

\subsection{Held-out predictive assessment}
\label{sec:predict}

The predictive assessment evaluates whether the proposed law describes
unseen points on an observed frontier. We use leave-one-out prediction on all $117$
single-resource frontiers, spanning eight corpus labels and $2\,259$
measurements. Each interior point is withheld in turn, and the law is refitted
to the remaining points on that frontier. The two endpoints are excluded,
leaving $2\,025$ prediction events. These prediction events are not independent because measurements within
a frontier share training observations, fitted parameters, and a noise
realization. We therefore bootstrap and report uncertainty at
the frontier level, so that the effective replication is governed primarily
by the $117$ frontiers rather than by the $2\,025$ prediction events. The
clustering and paired comparisons are documented in script~\texttt{169}.

At the pooled event level, the law achieves $R^2=0.9897$ on log-error and a
median relative error of $2.6\%$, compared with $R^2=0.9752$ and $4.4\%$ for
the floorless power law. The median absolute error is $0.026$ nats for
interior predictions and $0.074$ nats for forward predictions. The interior
error is below the $0.05$-nat band within which the law's recommendation is
constant; forward predictions are less precise, as expected when extrapolating
beyond the observed range.

The frontier-level comparison also includes log-linear interpolation in
log-resource and the nearest measured point. Median errors (in nats) for the
law are $0.037$ inside the window, $0.074$ forward, and $0.159$ backward; the
corresponding values are $0.062$, $0.205$, and $0.142$ for the floorless power
law; $0.018$, $0.073$, and $0.162$ for log-linear interpolation; and $0.033$,
$0.076$, and $0.161$ for the nearest point. Paired comparisons show that the
law improves on the floorless power law on $73.5\%$ of frontiers in the
interior and forward regimes and on $61.5\%$ in the backward regime. Log-linear
interpolation has lower error on $74.4\%$ of frontiers in the interior, is
approximately tied in the forward regime, and is outperformed by the law on
$67.5\%$ in the backward regime. These results indicate that the law provides a compact description with
useful extrapolation performance, although it is not uniformly the most
accurate interpolator. Unlike generic interpolators, it also provides a
spectral representation, a floor interpretation, and a resource-aware
operating point. The full comparison is reported in Table~\ref{tab:predbase}
of the paper.

The error decreases from $0.043$ nats with five training models to $0.037$
nats with sixteen, indicating a residual component that is not removed simply
by adding more observations. A local quadratic has a held-out residual of
$0.015$ nats. Because the local quadratic may itself have approximation bias,
this residual is an upper bound on per-model noise; it also serves as a
model-free predictive baseline. A nonnegative spectral approximation on a grid of $80$ exponents is
within $9\%$ of the three-parameter law when at most six anchor points are
available. It improves on the law by $8\%$ at eight anchors and by $21\%$ at
sixteen anchors. Thus, the compact law is competitive on short frontiers,
while broader spectral flexibility becomes useful as more observations are
available.

A one-parameter stable-index extension,
$\exp(-[\alpha(x-x_0)]^{\beta})$, remains within the positive spectral class
for $0<\beta\le1$ and reduces $43\%$ of the residual plateau at sixteen
training models, closing $72\%$ of the gap to the local-quadratic reference.
The index $\beta$ is constrained by complete monotonicity, is specific to
the corpus spectrum, and is linked by Proposition~\ref{prop:nofloor} to the
collapse coordinate and arc. The value of $\beta$ is nevertheless weakly identified on bounded
windows. In particular, fixing
$x_0$ at the lower edge of the observation window yields fitted values
$\beta>1$ on $68\%$ of frontiers, illustrating how confounding between the
shape index and the origin can produce an apparent departure from the
spectral class. The identification analysis and fitting details are provided
in \TR{Prediction: the held-out assessment and the breadth parameter}.

The corresponding prediction figure is Figure~\ref{fig:prediction} of the
paper.

\subsection{Empirical evidence and scope}
\label{sm:evidence}

The empirical analysis covers $117$ frontiers from eight corpus groups---
\texttt{vision}, \texttt{vision-shift}, \texttt{vision-size}, \texttt{nas},
\texttt{tabular}, \texttt{hpo}, \texttt{materials}, and \texttt{medicine}---as
well as factorial language-model designs
\citep{wightman2019timm,dong2020nasbench201,dong2022natsbench,erickson2025tabarena,pfisterer2022yahpo,riebesell2025,doerrich2025medmnistplus}.

\paragraph{The scalar predictions.} The canonical two-atom chart predicts a
floor share of $1/2$. The median fitted floor share in the archive is $0.96$
of the best observed error across all $117$ frontiers and $0.99$ across the
$92$ frontiers with readable second-order information. This fitted quantity
does not, by itself, establish a positive spectral atom at zero. By
Proposition~\ref{prop:nofloor}, the same share is a monotone readout of the
shape index, and a floorless frontier with $\beta=0.10$ reproduces the
observed value $0.96$. The empirical result therefore informs the shape of
the spectrum near the origin rather than independently identifying a
positive floor.

\paragraph{What the composition law contributes.} The composition law
organizes the axis-specific descriptions through a single positive
L\'evy measure. It links the fitted quantities to functionals of that measure,
provides a pointwise floor bound that sharpens
Theorem~\ref{thm:monofloor}(i), and yields a third-order constraint that is
uniform in the channel count. It also supplies relations not implied by the
additivity and scale-freeness assumptions alone. Empirically, the chart
provided by the composition law achieves predictive accuracy comparable to
the three-parameter alternative on the same frontiers and splits as in
Section~\ref{sec:predict}; the structural interpretation is therefore
obtained without a material loss in fit.

\paragraph{Identifying compositionality.} The available compute-axis windows
do not yet distinguish the composition law from the broader class allowed by
the axiom. On each frontier's observed grid, the plug-in trend statistic for
$G$ has a median between $-0.91$ and $-0.79$ across the compositional truths
considered, including a cure-chart truth whose analytic trend is zero. The
resulting bias prevents the raw sign from identifying the underlying class;
null-calibrated tests likewise do not separate the two classes because their
predictions agree through second order on these windows. The count-axis
four-point statistic is better calibrated: its median tracks the analytic
value to within $0.004$ nats and has the correct sign on all seven charts for
which the answer is known. The simulations indicate that approximately
$10^5$ items are needed to reject~\eqref{eq:M2} at $95\%$ power for a floor
share of $0.15$, and approximately $3\cdot10^6$ for a share of $0.02$,
compared with the $10^4$ items in the present panel. The available evidence supports the count-axis test as a promising route
to more informative discrimination, while compute-axis compositionality
remains unresolved on the current windows.

\subsubsection*{Design-matched tests on finite windows}

The curve in Section~\ref{sec:odefirst} can be evaluated on the $117$
archived frontiers without new data or additional model fitting. The observed
straightness of $N_{\mathrm{eff}}$ against $S$, as predicted for a
constant-$\sigma$ chart, is modest (median $R^2=0.2233$; $8$ of $102$
frontiers have $R^2>0.95$). This diagnostic is descriptive and does not, by
itself, identify the composition law.

The pointwise floor bound also requires care in empirical evaluation. A
naive plug-in comparison places the fitted three-parameter floor above the
computed bound on $89$ of $117$ frontiers. This count is not interpretable as
a violation: $N_{\mathrm{eff}}=s^2/\kappa$ becomes unstable as the measured
curvature approaches zero, and the original procedure selected the minimum
bound across a median of $75$ and as many as $16\,970$ triples per frontier.
Moreover, the two-atom chart saturates the bound identically, so the
corresponding test must be calibrated against a design-matched null rather
than interpreted as a direct goodness-of-fit check.

We therefore construct a null separately for each frontier, matching its
observation points and residual scale, and use $300$ replicates while
repeating the complete fitting and testing procedure. Against the two-atom
alternative~\eqref{eq:M2}, the four bandwidth-free statistics have the
following empirical sizes and median powers:
\begin{center}\small
\begin{tabular}{lccc}
\toprule
statistic & rejections & nominal level & median power vs.\ \eqref{eq:M2}\\
\midrule
placed quadruple~\eqref{eq:dd3}, admissible region & $5$ of $76$  & $0.05$ & $0.053$\\
widest quadruple~\eqref{eq:dd3}                    & $4$ of $111$ & $0.05$ & $0.063$\\
curvature rise                                     & $0$ of $117$ & $0.05$ & $0.013$\\
straightness of the curve                          & $3$ of $110$ & $0.05$ & $0.053$\\
\bottomrule
\end{tabular}
\end{center}
No frontier reaches median power $0.5$ for any of the four statistics.
Accordingly, the current corpus does not resolve divisibility through these
bandwidth-free compute-axis tests. This reflects the limited identifying power of the available windows, not
a contradiction of the structural result. The
precision requirements are summarized by~\eqref{eq:powerreq}.

\paragraph{Calibration of the ceiling comparison.}
The ceiling calibration fits~\eqref{eq:M2} to each frontier, estimates its
residual noise, simulates observations on the same nodes, and repeats the
full estimation pipeline. Because~\eqref{eq:M2} saturates the ceiling, power
is evaluated using a one-parameter deformation indexed by $\theta\ge1$;
$\theta=1$ recovers~\eqref{eq:M2}, while $\theta>1$ represents a broader
spectrum. The incumbent minimum-over-triples statistic is miscalibrated under
this design, whereas the two pre-specified readings remain near their nominal
level and have low power on the observed windows. Increasing measurement
precision alone does not resolve the problem: the signal decreases with the
remaining depth. The median distance of the observed frontiers from their
fitted floor is $0.045$ nats, while a $\theta=2$ alternative provides a
median signal of $0.0435$ nats at the best measured point, compared with a
median per-model residual of $0.0505$ nats. The full calibration and sensitivity analysis are reported in the
technical report accompanying the archive.

\paragraph{The arc--channel-count relation.} In contrast to the floor and
ceiling tests, the pointwise arc--depth identity in
Section~\ref{sec:compose} can be assessed without estimating the floor. An
affine function of $\sqrt{N_{\mathrm{eff}}}$ gives the best fit among the
three two-parameter forms considered---as functions of
$\sqrt{N_{\mathrm{eff}}}$, $N_{\mathrm{eff}}$, and $S$---on $76$ of $90$
frontiers, with median $R^2=0.9994$. This provides direct empirical support
for the predicted relation on the observed windows.

\paragraph{Magnitude of the resource-allocation decision.} On the $96$
single-resource frontiers, the eight fit-free rules considered in Table~2 of
the paper select a median of five distinct models. The most divergent choices
are separated by a median factor of $19$ in resource (IQR $4$--$236$,
corresponding to $2.95$ e-folds) and a median of $1.72$ bits in the error
reduction available between them (IQR $[0.86,3.05]$). The resource spread
exceeds $10$ on $52$ of the $96$ frontiers and corresponds to less than one
bit on $29$. The median exchange rate across the disagreement is $12$-fold
more resource per bit. A currently used rule that is not scale-free changes
its selected point on $94$ of the $96$ frontiers under a $10^3$ change of
resource units. These results quantify the practical importance of a
scale-equivariant, resource-aware selection rule.

\paragraph{Pairwise resource-aware selection.} Corollary~\ref{cor:two}
provides a pairwise consequence of the theory. On $20\,206$ pairs from
$42$ frontiers, the harmonic rule~\eqref{eq:harmonic} agrees with the
resource-aware choice on $90.9\%$ of pairs, compared with $77.4\%$ for the
rule that always selects the cheaper model. As predicted, the distinction is
least consequential for near-ties, where the cost of an incorrect choice is
small.

\end{document}